\documentclass[10pt,twocolumn,letterpaper]{article}
\usepackage{brdfusion}
\usepackage[dvipsnames]{xcolor}

\usepackage{bm}
\usepackage{makecell}
\usepackage{multirow}
\usepackage{array}
\usepackage{colortbl}
\usepackage{algorithm}
\usepackage{algpseudocode}
\makeatletter
\AtBeginDocument{%
    \providecommand*{\theHALG@line}{\thealgorithm.\arabic{ALG@line}}%
}
\makeatother
\usepackage{microtype}

\newcommand\method{{BRDFusion~}}

\newcolumntype{L}[1]{>{\raggedright\let\newline\\\arraybackslash\hspace{0pt}}m{#1}}
\newcolumntype{C}[1]{>{\centering\let\newline\\\arraybackslash\hspace{0pt}}m{#1}}
\newcolumntype{R}[1]{>{\raggedleft\let\newline\\\arraybackslash\hspace{0pt}}m{#1}}

\newcommand{\ignorethis}[1]{}

\makeatletter
\DeclareRobustCommand\onedot{\futurelet\@let@token\@onedot}
\def\@onedot{\ifx\@let@token.\else.\null\fi\xspace}

\def\ie{\emph{i.e}\onedot}

\makeatother

\definecolor{citecolor}{RGB}{34,139,34}
\definecolor{mydarkblue}{rgb}{0,0.08,1}
\definecolor{mydarkgreen}{rgb}{0.02,0.6,0.02}
\definecolor{mydarkred}{rgb}{0.8,0.02,0.02}
\definecolor{mydarkorange}{rgb}{0.40,0.2,0.02}
\definecolor{mypurple}{RGB}{111,0,255}
\definecolor{myred}{rgb}{1.0,0.0,0.0}
\definecolor{mygold}{rgb}{0.75,0.6,0.12}
\definecolor{myblue}{rgb}{0,0.2,0.8}
\definecolor{mydarkgray}{rgb}{0.66,0.66,0.66}

\newcommand{\bbR}{{\mathbb{R}}}

\newcommand{\bX}{\mathbf{X}}

\newcommand{\bx}{\mathbf{x}}

\newcommand{\bg}{\mathbf{g}}
\newcommand{\bS}{\mathbf{S}}
\newcommand{\bs}{\mathbf{s}}
\newcommand{\bM}{\mathbf{M}}

\newcommand{\bN}{\mathbf{N}}
\newcommand{\bn}{\mathbf{n}}

\newcommand{\bq}{\mathbf{q}}
\newcommand{\bR}{\mathbf{R}}

\newcommand{\bp}{\mathbf{p}}

\newcommand{\bD}{\mathbf{D}}

\newcommand{\bP}{\mathbf{P}}
\newcommand{\bC}{\mathbf{C}}
\newcommand{\bA}{\mathbf{A}}

\newcommand{\bc}{\mathbf{c}}

\newcommand{\bL}{\mathbf{L}}

\newcommand{\ba}{\mathbf{a}}
\newcommand{\bE}{\mathbf{E}}

\newcommand{\bH}{\mathbf{H}}

\newcommand{\bmu}{\boldsymbol{\mu}}

\newcommand{\bomega}{\boldsymbol{\omega}}

\newcommand{\SE}[1]{\mathbb{SE}(#1)}

\newcommand{\enc}{\mathcal{E}}
\newcommand{\dec}{\mathcal{D}}

\newcommand{\diffusionModelF}{\mathbf{F}_{\theta}^{\text{fwd}}}
\newcommand{\diffusionModelI}{\mathbf{F}_{\theta}^{\text{inv}}}

\newcommand{\diffusionTime}{\tau}
\newcommand{\latent}{\mathbf{z}}
\newcommand{\sdedit}{\textsc{SDEdit}}
\newcommand{\cond}{\mathbf{G}}
\newcommand{\noise}{\boldsymbol{\epsilon}}

\newcommand{\bO}{\mathbf{O}}

\newcommand{\Cpbr}{\bC_{\text{pbr}}}
\newcommand{\Cldr}{\bC_{\text{ldr}}}
\newcommand{\Cgen}{\bC_{\text{gen}}}  
\newcommand{\Cfin}{\bC_{\text{fin}}}
\newcommand{\Cgt}{\bC_{\text{gt}}} 

\newcommand{\loss}{\mathcal{L}}
\newcommand{\weight}{\lambda}

 \definecolor{linkblue}{rgb}{0.21,0.49,0.74}
\usepackage[pagebackref,breaklinks,colorlinks,allcolors=linkblue]{hyperref}
\hypersetup{
  pdftitle={BRDFusion: Physics Meets Generation for Urban Scene Inverse Rendering},
  pdfauthor={Yi-Ruei Liu, Jie-Ying Lee, Zheng-Hui Huang, Yu-Lun Liu, Chih-Hao Lin},
  pdfsubject={}, pdfkeywords={}, pdfcreator={LaTeX}
}

\title{BRDFusion: Physics Meets Generation for Urban Scene Inverse Rendering}

\author{
Yi-Ruei Liu$^{1}$ \quad Jie-Ying Lee$^{1}$ \quad Zheng-Hui Huang$^{3}$ \quad Yu-Lun Liu$^{1\dagger}$ \quad Chih-Hao Lin$^{2\dagger}$\\[2pt]
{\small $^1$National Yang Ming Chiao Tung University \quad $^2$University of Illinois Urbana-Champaign \quad $^3$National Taiwan University}\\[2pt]
{\small \url{https://shigon255.github.io/brdfusion-page/}}
}

\begin{document}
\twocolumn[
\maketitle
\begin{center}
\centering
\vspace{-3mm}
\includegraphics[width=\linewidth]{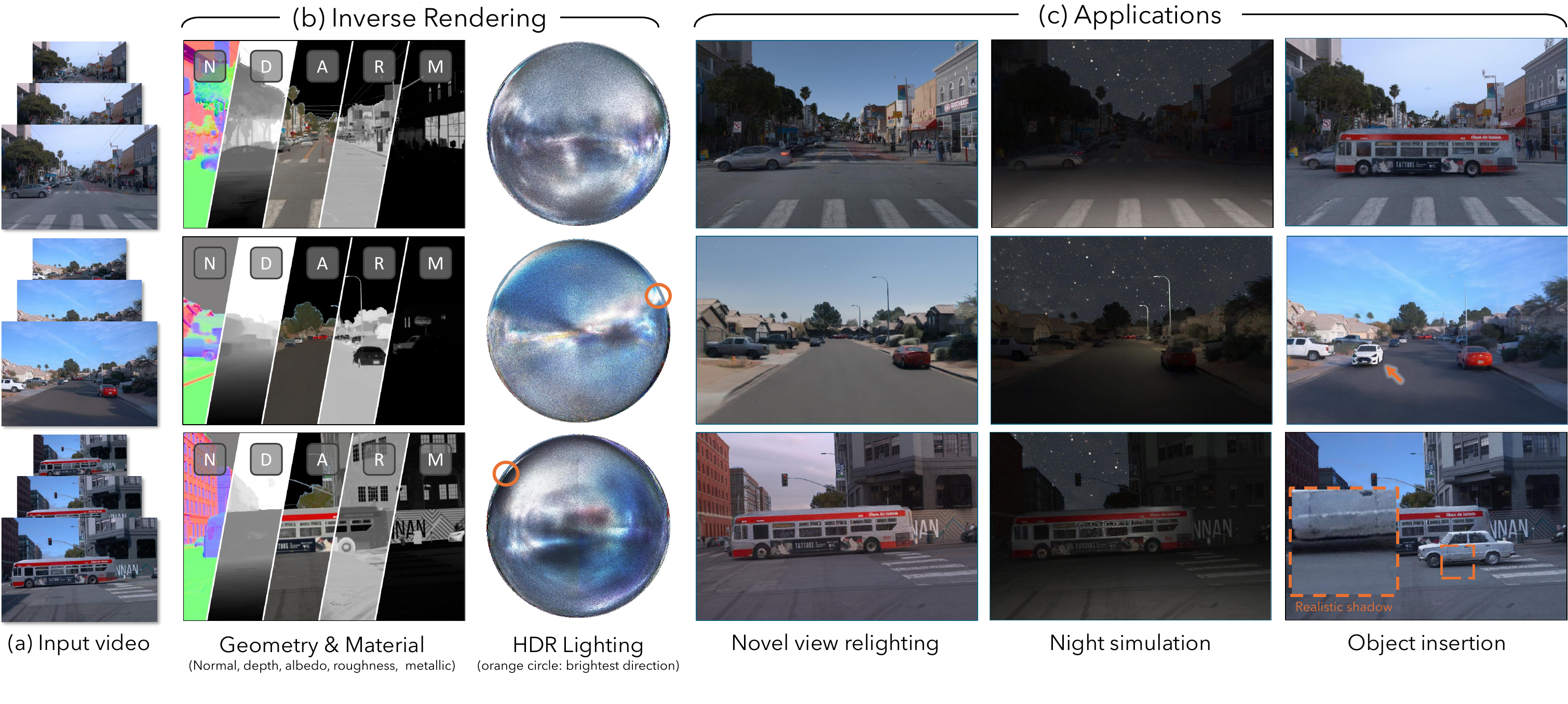}
\vspace{-9mm}
\captionsetup{type=figure}\caption{
\textbf{Inverse rendering and re-rendering of urban driving videos.}
Given (a) an input driving video, BRDFusion recovers (b) geometry (normal N, depth D), materials (albedo A, roughness R, metallic M), and an HDR environment map (the orange circle marks its brightest direction). (c) The recovered scene is re-rendered from novel views under new lighting, at night with added lights, and with inserted objects that cast shadows (\textcolor{orange}{\textbf{orange}} boxes).
}
\label{fig:teaser}
\vspace{-3mm}

\end{center}
]
\footnotetext[2]{Equal advising.}

\begin{abstract}
  Inverse rendering recovers the geometry, materials, and lighting of urban scenes from driving videos, so that simulators can relight them.
  Physically-based rendering (PBR) controls lighting explicitly, but inherits the artifacts of imperfect reconstructions.
  Generative video models render realistic frames, but cannot place explicit lights, such as headlights, and are inconsistent across clips.
  Our key insight is that these failures are complementary: a PBR rendering of a persistent 3D scene can anchor a generative model, which then only needs to remove artifacts rather than invent the lighting.
  BRDFusion applies this anchoring in both directions.
  In forward rendering, a video diffusion model refines the noised PBR video, so the lighting comes from physics and the realism from generation.
  In inverse rendering, the same refinement turns material maps rendered from the 3D scene into clean, temporally consistent priors, and PBR recovers the HDR lighting.
  On synthetic scenes, BRDFusion improves novel-view relighting PSNR by 2.8 dB over the strongest baseline. On real driving videos, it relights scenes, adds local lights, and inserts dynamic objects. Project page: \url{https://shigon255.github.io/brdfusion-page/}.
  \vspace{-0.3em}
\end{abstract}
\vspace{-1.0em}
\section{Introduction}
\label{sec:intro}
\vspace{-0.5em}

Reconstructing and editing large-scale urban environments from driving videos with photorealism and controllable lighting is crucial for autonomous driving simulation (Fig.~\ref{fig:teaser}), AR/VR, and content creation.
This requires solving the highly ill-posed \textbf{inverse rendering} problem, which decomposes a scene into its 3D geometry, material properties, and environment lighting.
It is especially challenging for urban scenes, since the input videos are sparse and capture only parts of the scene.

\begin{figure}[t]
    \centering
    \includegraphics[width=\linewidth]{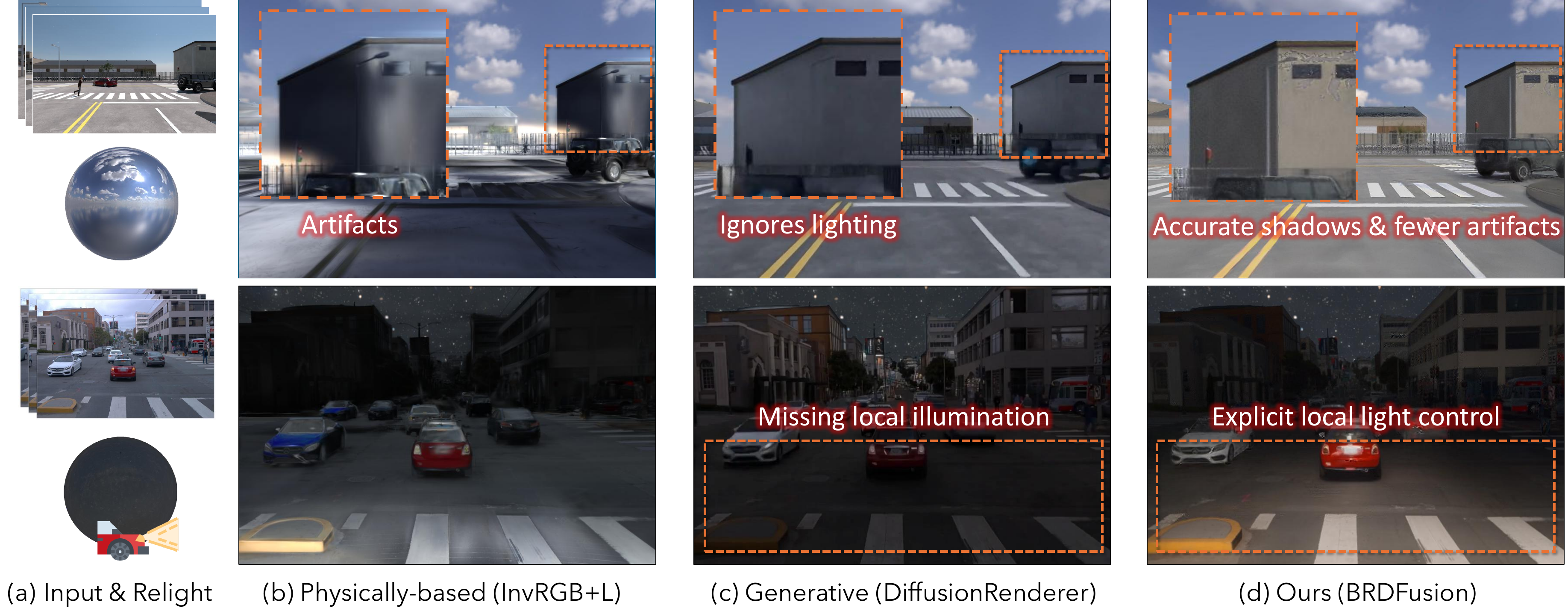}
    \vspace{-6mm}
    \caption{
    \textbf{Existing paradigms vs.\ BRDFusion.}
    Under the target lighting \textbf{(a)}, \textbf{(b)} a physically-based method (InvRGB+L~\cite{invrgbl}) follows the lighting but shows reconstruction artifacts, and \textbf{(c)} a generative model (DiffusionRenderer~\cite{diffusionrenderer}) produces realistic images but does not follow the target lighting or the car lights.
    \textbf{(d)} BRDFusion produces realistic images that follow both.
    }
    \label{fig:motivation}
    \vspace{-6mm}
\end{figure}

Existing methods trade physical correctness for visual realism, or the reverse.
Physically-based inverse rendering methods such as InvRGB+L~\cite{invrgbl} represent viewpoints, lighting, and scene composition explicitly, so each can be edited.
However, the optimization is ambiguous, so they are prone to wrong decompositions and rendering artifacts (Fig.~\ref{fig:motivation} (b)).
Another line of work uses generative models pretrained on large-scale data~\cite{diffusionrenderer, unirelight} to render clean and realistic frames, but these models are inconsistent over long sequences and cannot control local illumination (e.g., vehicle headlights), which is essential for simulating scenarios such as nighttime driving (Fig.~\ref{fig:motivation} (c)).
Recent hybrid methods~\cite{LightSim, TRON, LightCrafter, autoweather4d} combine the two, but either do not natively support local lights~\cite{LightSim} or dynamic urban scenes~\cite{TRON}, or rely on temporally inconsistent material maps that a generative model predicts clip by clip~\cite{LightCrafter,autoweather4d}.

\begin{figure*}[t]
    \centering
    \includegraphics[width=1.0\textwidth]{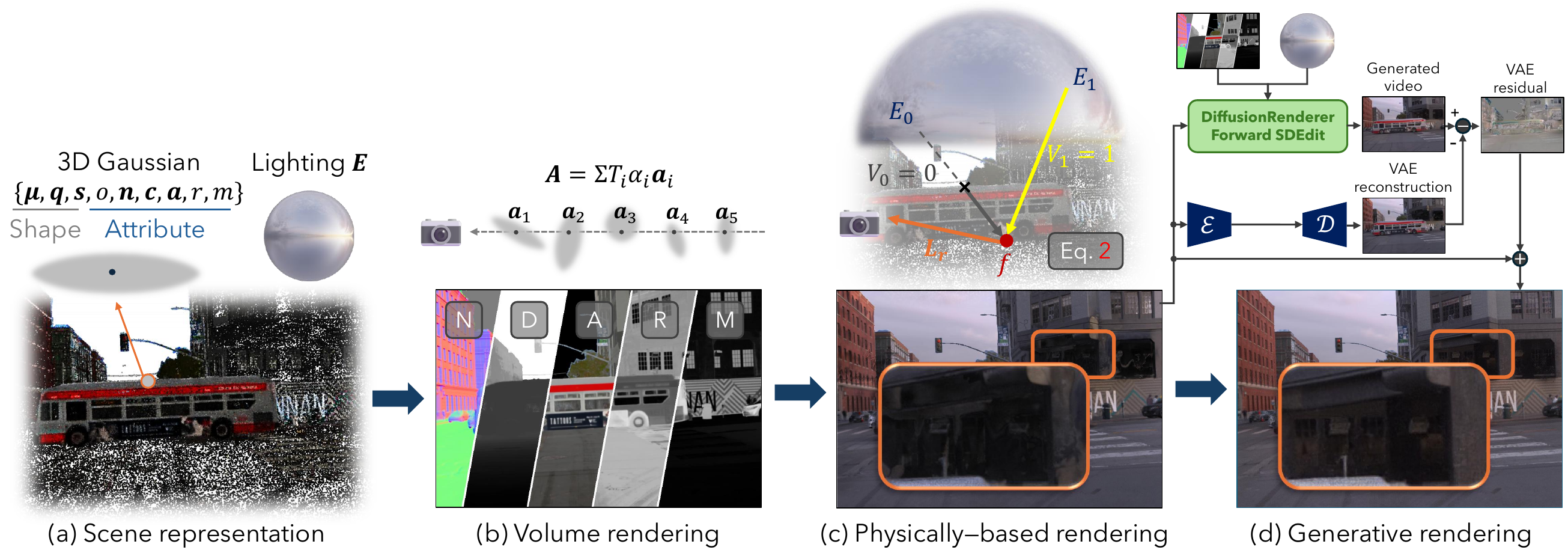}
    \caption{
    \textbf{Overview of the hybrid forward rendering pipeline.}
    \textbf{(a)} The scene is represented as 3D Gaussians and HDR environment lighting. 
    \textbf{(b)} Rasterizing the 3D attributes yields 2D G-buffers, but no light transport.
    \textbf{(c)} Hence, physically-based rendering (PBR) estimates the visibility $V$, queries the lighting $\bE$ along sampled directions, and computes direct illumination (Eq.~\ref{eq:pbr}). The result obeys light physics but shows reconstruction artifacts. \textbf{(d)} To remove them, the DiffusionRenderer~\cite{diffusionrenderer} forward model refines the PBR result with SDEdit~\cite{sdedit}. To preserve details, the VAE residual is added back to the PBR result instead of using the generated video directly. The final rendering keeps the PBR lighting without the artifacts (\textcolor{orange}{\textbf{orange}} insets).
    }
    \label{fig:rendering}
    \vspace{-5mm}
\end{figure*}

Our key insight is that the two paradigms fail in complementary ways. PBR follows the target lighting but inherits reconstruction artifacts, whereas a generative model produces clean results but does not reliably follow the target lighting, especially local lights.
A generative model that starts from a PBR rendering of a persistent 3D scene therefore only needs to remove artifacts, while the lighting and its consistency come from physics.
\textbf{BRDFusion} builds on this insight (Fig.~\ref{fig:motivation} (d)): It represents geometry and materials as 3D Gaussians~\cite{3dgs} and lighting as an HDR environment map.
Its PBR pass computes direct illumination on G-buffers rasterized from the Gaussians, and a pretrained video diffusion model~\cite{diffusionrenderer} refines the noised PBR video, following SDEdit~\cite{sdedit} (Fig.~\ref{fig:rendering}).
The same anchoring serves inverse rendering, where generative priors reduce ambiguity but are inconsistent across clips. BRDFusion refines material maps rendered from its 3D reconstruction with the generative inverse model, which yields clean and temporally consistent supervision, and then recovers the lighting with PBR (Fig.~\ref{fig:training}).

We summarize our main contributions as follows:

\begin{table}[t]
    \centering
    \scriptsize
    \vspace{0mm}
    \caption{\textbf{Capabilities of related methods.} A checkmark denotes native support. \emph{Dynamic}: scenes with moving objects. \emph{NVS}: novel-view synthesis.
    \emph{Consistent Material}: materials stored in a persistent 3D representation, hence consistent across frames and views.
    \emph{HDR Env.}: relighting directly from a single HDR environment map. 
    $^\dagger$Baseline composed for this paper.
    }
    \vspace{-2mm}
    \label{tab:capability_comparison}
    \setlength{\tabcolsep}{2pt}
    \renewcommand{\arraystretch}{1.0}
    \resizebox{\linewidth}{!}{%
    \begin{tabular}{@{}lccccccc@{}}
    \toprule
    Method & Urban & Dynamic & NVS & Relight. & \makecell{Consistent\\Material} & \makecell{Local\\Light} & \makecell{HDR\\Env.} \\
    \midrule
    \multicolumn{8}{@{}l}{\textit{Physically-based}} \\
    IRGS~\cite{IRGS}
        &  &  & $\checkmark$ & $\checkmark$ & $\checkmark$ &  & $\checkmark$ \\
    SRUG~\cite{srug}
        & $\checkmark$ &  & $\checkmark$ & $\checkmark$ & $\checkmark$ &  &  \\
    FEGR~\cite{FEGR}
        & $\checkmark$ &  & $\checkmark$ & $\checkmark$ & $\checkmark$ &  & $\checkmark$ \\
    UrbanIR~\cite{urbanir}
        & $\checkmark$ &  & $\checkmark$ & $\checkmark$ & $\checkmark$ & $\checkmark$ &  \\
    InvRGB+L~\cite{invrgbl}
        & $\checkmark$ & $\checkmark$ & $\checkmark$ & $\checkmark$ & $\checkmark$ & $\checkmark$ &  \\
    \midrule
    \multicolumn{8}{@{}l}{\textit{Generative}} \\
    DiffusionRenderer (DR)~\cite{diffusionrenderer}
        & $\checkmark$ & $\checkmark$ &  & $\checkmark$ &  &  & $\checkmark$ \\
    Gen3C~\cite{Gen3C}+DR~\cite{diffusionrenderer}$^\dagger$
        & $\checkmark$ & $\checkmark$ & $\checkmark$ & $\checkmark$ &  &  & $\checkmark$ \\
    \midrule
    \multicolumn{8}{@{}l}{\textit{Hybrid}} \\
    LightSim~\cite{LightSim}
        & $\checkmark$ & $\checkmark$ & $\checkmark$ & $\checkmark$ & $\checkmark$ &  & $\checkmark$ \\
    TRON~\cite{TRON}
        &  &  & $\checkmark$ & $\checkmark$ & $\checkmark$ &  & $\checkmark$ \\
    LightCrafter~\cite{LightCrafter}
        & $\checkmark$ & $\checkmark$ &  & $\checkmark$ &  & $\checkmark$ & $\checkmark$ \\
    AutoWeather4D~\cite{autoweather4d}
        & $\checkmark$ & $\checkmark$ &  & $\checkmark$ &  & $\checkmark$ & $\checkmark$ \\
    \midrule
    \textbf{BRDFusion (Ours)}
        & $\checkmark$ & $\checkmark$ & $\checkmark$ & $\checkmark$ & $\checkmark$ & $\checkmark$ & $\checkmark$ \\
    \bottomrule
    \end{tabular}%
    }
    \vspace{-4mm}
\end{table} 
\begin{itemize}
\item A hybrid forward renderer for dynamic urban scenes that anchors a video diffusion model on PBR renderings of a 3D Gaussian scene, combining the controllable lighting of PBR with the realism of generation.
\item A multi-stage inverse rendering pipeline in which generative priors regularize geometry and materials, generative refinement of material maps rendered from the 3D scene makes these priors temporally consistent, and PBR recovers the HDR lighting.
\item State-of-the-art novel-view relighting on synthetic scenes (2.8 dB PSNR over the strongest baseline), together with relighting, night simulation with local lights, and dynamic object insertion on real driving videos.

\end{itemize}
\vspace{-0.5em}
\section{Related Work}
\label{sec:related}
\vspace{-0.5em}

\paragraph{\textbf{Physically-Based Inverse Rendering in Urban Scenes.}}
Recovering geometry, material, and lighting from images is a long-standing, ill-posed problem~\cite{shape_illumination_reflectance_from_shading,mit_intrinsic_images_dataset,intrinsic_images_in_the_wild}.
Differentiable rendering with NeRFs~\cite{nerf}, meshes~\cite{nvdiffrec,nvdiffrecmc}, Monte Carlo light transport~\cite{pbrt4ed}, and 3D Gaussian Splatting (3DGS)~\cite{3dgs} enables the joint optimization of these properties~\cite{NerFactor,nerv,physg,TensoIR,neilf,neilf++,pbr_nerf,GaussianShader,relightable3dgs,gsir,GIR,3dgs_dr,svg_ir,chen2025gigs,IRGS,VisibilitySplat,LumiMotion,SGS_Intrinsic,BEAM,ye2024progressiveradiancedistillationinverse}.
Outdoor scenes add unbounded extents, HDR sunlight, and sparse viewpoints, which have been tackled with parametric sky models~\cite{hosek2012sky}, learning-based decomposition~\cite{deep_outdoor_illumination_estimation,deep_sky_modeling_for_single_image_outdoor_lighting_estimation,InverseRenderNet,self_supervised_outdoor_scene_relighting,DeRenderNet,multiview_relighting_geometry_aware_network,outcast,RENI}, and NeRF- or 3DGS-based relighting~\cite{nerf_osr,neural_light_field_estimation_for_street_scenes,neulighting,complementary_intrinsics,the_skys_the_limit,LumiGauss,rosgs,GaRe}.
Dynamic Gaussian scene graphs~\cite{street_gaussians,chen2025omnire} reconstruct driving scenes under the captured illumination, while lighting-aware urban methods decompose them with sky domes~\cite{FEGR}, SDF shadows~\cite{sol_nerf}, or single-camera~\cite{urbanir} and LiDAR-assisted optimization~\cite{invrgbl}, and others simulate or reconstruct varying lighting, weather, and nighttime~\cite{climatenerf,pam_tod,solo_nighttime,npsim,hybrid_rendering_ad,nighttime_pbr_gs,srug}.
Like these methods, we optimize a physically-based scene graph, but instead of relying mainly on \emph{physics}, we couple it with generative models in both inverse and forward rendering.

\vspace{-1.0em}
\paragraph{\textbf{Generative and Diffusion Priors for Inverse Rendering.}}

Learned priors are essential to resolve the ambiguity of inverse rendering. Feed-forward networks offer fast initialization~\cite{midas,MVInverse}, while large-scale diffusion models~\cite{ldm} provide richer priors for geometry~\cite{Marigold,geowizard,StableNormal,lotus}, materials~\cite{intrinsic_image_diffusion,intrinsicanything,idarb,matfusion,DreamMat,material_anything,materialfusion}, and lighting~\cite{diffusionlight,diffusionlight_turbo,luxdit,DiLightNet,ICLight,dpi_illumination,ir_hgp}, coupled to optimization via score distillation~\cite{dreamfusion,materialfusion}, direct prediction~\cite{neural_lightrig,unirenderer,DNF-Intrinsic,rgbx}, or iterative updates~\cite{instruct_nerf2nerf,reconfusion,srug}. These priors are predicted per image or clip and can be temporally inconsistent over long driving videos. Our generative refinement instead re-derives them from material maps rendered by our 3D scene, yielding \emph{temporally consistent} supervision.

\vspace{-1.0em}
\paragraph{\textbf{Generative Rendering and Relighting.}}

Neural renderers~\cite{deferred_neural_rendering,epe} and generative models~\cite{IllumiNeRF,NeuralGaffer} can render and relight scenes without physically-based inverse rendering. Among them, DiffusionRenderer (DR)~\cite{diffusionrenderer} and related models~\cite{rgbx,rgbx-next,beisswenger2025framediffuser,unirelight,HorizonRelight,GeoRelight,Relit-LiVE,LightSwitch,light-x,light-a-video,UniLumos,tc-light,Lumen} build on large image and video diffusion models~\cite{ldm,stable_video_diffusion} to render and relight images and videos directly, but are typically conditioned on global lighting such as an environment map or text. Recent works further add explicit light control in images~\cite{lightlab}, videos~\cite{relightmaster,livelight}, and static, mostly indoor 3D scenes~\cite{synclight,gr3en,luxremix,lightbridge}. We also use DR for realism, but whereas these models \emph{synthesize} lighting effects, we \emph{simulate} them with PBR in a persistent 3D scene, so newly added lights cast consistent, ray-traced shadows.

\vspace{-1.0em}
\paragraph{\textbf{Generative Fixers and Driving Simulation.}}
Diffusion-based fixers remove artifacts of neural reconstructions as pseudo-view supervision or render-time enhancers~\cite{difix3d,artifixer,gs_enhancer,genfusion,diffusion_harmonizer,realitybridge}, and driving simulators extend reconstructions to novel trajectories, conditions, and inserted objects~\cite{streetcrafter,recondreamer,cosmos_transfer1,chatsim,dipir,r3d2,spectralsplat}. However, they either keep the \emph{captured} illumination or change lighting only globally through generative models. In contrast, our generative rendering pass uses a relighting model and starts from a PBR rendering under the \emph{target} lighting, so it simulates both global and local lighting changes, such as nighttime with headlights and streetlights.

\vspace{-1.0em}
\paragraph{\textbf{Comparison with Prior Works.}}

Tab.~\ref{tab:capability_comparison} summarizes these differences.
Physically-based inverse rendering methods~\cite{IRGS,FEGR,urbanir,invrgbl,srug} store materials in a persistent 3D representation but either target static scenes~\cite{IRGS,FEGR,urbanir,srug} or use sun-sky models instead of HDR environment maps~\cite{urbanir,invrgbl}, while generative methods~\cite{diffusionrenderer,Gen3C} produce high-fidelity results without explicit local light control.
Closest to ours are four hybrid approaches: LightSim~\cite{LightSim} refines PBR renderings with a 2D U-Net~\cite{UNet}, and TRON~\cite{TRON} feeds ray-traced renderings of 3D Gaussians into a neural renderer, but neither natively supports local lights, and TRON reconstructs only static, non-urban scenes. LightCrafter~\cite{LightCrafter} refines PBR renderings with a video diffusion model, and AutoWeather4D (AW4D)~\cite{autoweather4d} adds PBR-shaded local lights to driving videos relit by DR. Both rely on per-clip \emph{2D} material maps predicted by DR, which can be temporally inconsistent. AW4D also refines its output with an edge-conditioned video model, which can alter scene textures, and it merges local and global lighting by per-pixel blending rather than in a single light-transport model.
BRDFusion instead stores materials on \emph{3D} Gaussians and simulates both global and local lighting with PBR in dynamic urban scenes.

\vspace{-0.5em}
\section{Method}
\label{sec:method}
\vspace{-0.5em}

\paragraph{\textbf{Problem Setting.}}
Given $F$ posed video frames $\{(\bP^t, \Cgt^{t})\}_{t=1}^F$ with camera poses $\bP^t \in \SE{3}$ and images $\Cgt^{t} \in \bbR^{H \times W \times 3}$, captured in an outdoor scene under fixed lighting, together with sparse LiDAR points and 3D bounding-box tracks of moving objects, our goal is to recover a \emph{relightable} scene representation that supports novel-view synthesis, relighting, and object insertion.

\subsection{Relightable Scene Representation}
\label{sec:representation}
\vspace{-0.5em}
 
An explicit 3D scene supplies what a 2D generative model lacks, namely geometry for ray-traced shadows, materials that stay consistent across frames, and lighting that can be edited. BRDFusion therefore represents the scene with 3D Gaussians~\cite{3dgs, chen2025omnire}.
Each Gaussian $\bg = (o, \bmu, \bq, \bs, \bn, \bc, \ba, r, m)$ encodes geometry, appearance, and material: opacity $o \in (0, 1)$, 3D mean $\bmu \in \bbR^3$, rotation quaternion $\bq \in \bbR^4$, scale $\bs \in \bbR^3$, normal $\bn \in \bbR^3$, view-dependent color $\bc \in \bbR^3$, albedo $\ba \in \bbR^3$, roughness $r \in \bbR$, and metallic $m \in \bbR$.
Following OmniRe~\cite{chen2025omnire}, we model moving objects with a scene graph built from the 3D bounding-box tracks. Each node holds Gaussians in its local frame and is placed in the world at each time $t$ by a rigid transformation for vehicles and by OmniRe's deformable human nodes for pedestrians.
The lighting is an HDR environment map $\bE$, stored as a $512 \times 1024 \times 3$ image, which returns the incoming radiance $\bE(\bomega) \in \bbR^3$ from direction $\bomega$.

\begin{figure*}[t]
    \centering
    \includegraphics[width=1.0\textwidth]{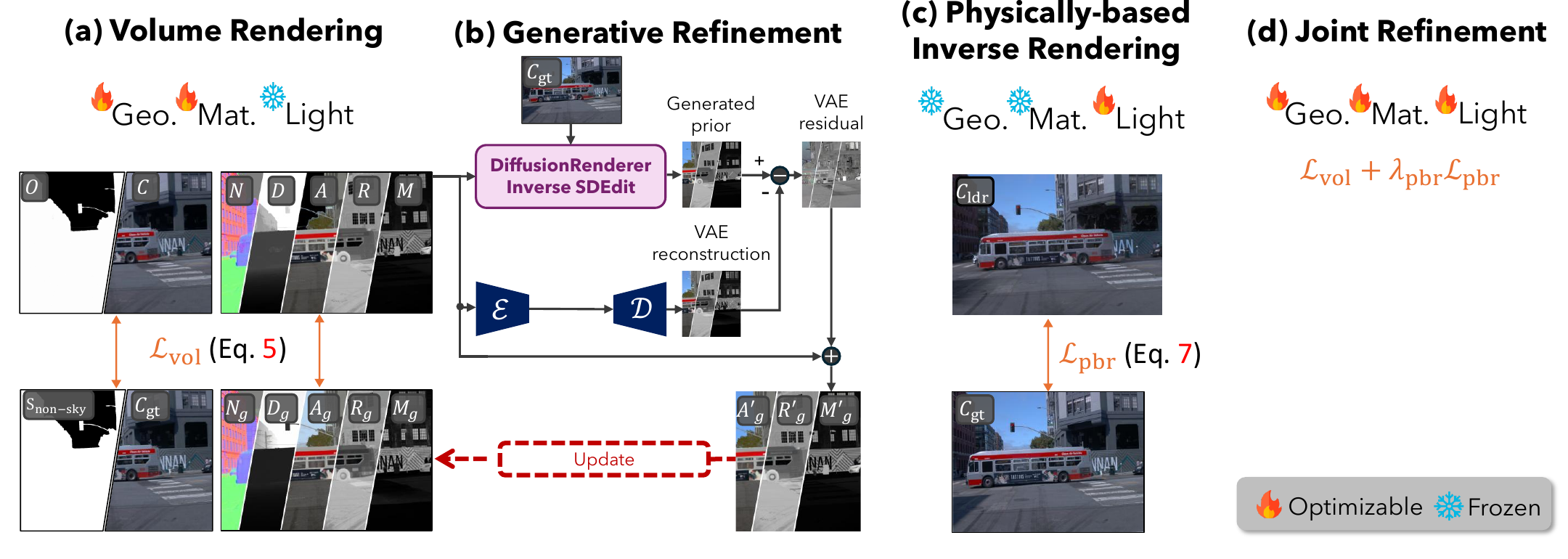}
    \caption{\textbf{Inverse rendering with physical and generative models.} \textbf{(a)} We first freeze the lighting and optimize geometry and material with volume rendering, supervised by the input frames, sky masks from SegFormer~\cite{segformer}, and G-buffer priors from DiffusionRenderer~\cite{diffusionrenderer} (Eq.~\ref{eq:loss_volume}). Since the material priors are temporally inconsistent, this initial reconstruction contains material artifacts. However, because the reconstruction is in 3D, its rendered material maps are temporally consistent, though with artifacts. \textbf{(b)} Therefore, we refine these consistent but imperfect material maps with the DiffusionRenderer inverse model using SDEdit~\cite{sdedit}. We add the VAE residual back to the rendered material maps to preserve details. We then use the refined maps as new priors to rerun stage \textbf{(a)}, resulting in a cleaner reconstruction. \textbf{(c)} With clean geometry and material, we then estimate the lighting with PBR (Eq.~\ref{eq:loss_pbr}) while keeping geometry and material frozen to reduce ambiguity in the optimization. \textbf{(d)} Finally, we jointly refine all scene properties to improve the overall quality.
    }
    \label{fig:training}
    \vspace{-5mm}
\end{figure*}
 
\subsection{Forward Rendering with PBR Anchoring}
\label{sec:forward_render}
\vspace{-0.5em}
Fig.~\ref{fig:rendering} shows the forward rendering pipeline, which turns the representation into an image through three passes: a \emph{volume rendering} pass that rasterizes scene attributes (Fig.~\ref{fig:rendering} (b)), a \emph{PBR} pass that simulates physical light transport (Fig.~\ref{fig:rendering} (c)), and a \emph{generative rendering} pass that removes the reconstruction artifacts (Fig.~\ref{fig:rendering} (d)). Each pass addresses a specific shortcoming of the previous one.

\vspace{-1.0em}
\paragraph{\textbf{Volume Rendering Pass.}}
At time $t$, the Gaussians of all scene-graph nodes are transformed into a shared world frame, sorted by depth for the given camera, and splatted to image space.
The color at pixel $\bp \in \bbR^2$ is alpha-blended~\cite{3dgs} over the depth-sorted Gaussians:
$\bC = \sum_{i}T_i\alpha_i\bc_i$, where $T_i = \prod_{j=1}^{i-1}(1-\alpha_j)$, $\alpha_i=o_ig_i(\bp)$, and $g_i(\bp) = \exp\!\left(-\tfrac{1}{2}(\bp - \bmu'_i)^\top\mathbf{\Sigma}_i^{-1}(\bp - \bmu'_i)\right)$, with $\bmu'_i$ and $\mathbf{\Sigma}_i$ the 2D mean and covariance projected from $\bmu_i$, $\bq_i$, and $\bs_i$.
The same compositing yields the opacity $\bO = \sum_i T_i\alpha_i$, depth $\bD = \sum_i T_i\alpha_i d_i$ (with $d_i$ the depth of Gaussian $i$), normal $\bN = \sum_i T_i\alpha_i\bn_i$, albedo $\bA = \sum_i T_i\alpha_i\ba_i$, roughness $\bR = \sum_i T_i\alpha_i r_i$, and metallic $\bM = \sum_i T_i\alpha_i m_i$.
This pass models no light transport, hence no shadows or reflections.

\vspace{-1.0em}
\paragraph{\textbf{Physically-Based Rendering Pass.}}
To simulate the light transport, this pass approximates the rendering equation~\cite{rendering_equation}:
{\small
\begin{equation}\label{eq:rendering}
    \bL_\text{o}(\bx,\bomega_\text{o})=\bL_\text{e}(\bx,\bomega_\text{o})+\int_{\Omega}\bL_\text{i}(\bx,\bomega_\text{i})f(\bx,\bomega_\text{i},\bomega_\text{o})(\bomega_\text{i}\cdot\bN(\bx))\,d\bomega_\text{i},
\end{equation}
}
where $\bL_\text{o}$ is the outgoing radiance at surface point $\bx$ in direction $\bomega_\text{o}$, $\bL_\text{e}$ the emission, $\bL_\text{i}$ the incident radiance from direction $\bomega_\text{i}$, $f$ the BRDF, and $\bN(\bx)$ the normal at $\bx$, queried from the normal map $\bN$.
BRDFusion models no surface emission and estimates the integral by Monte Carlo integration~\cite{distributed_ray_tracing}.
For each pixel, the surface point $\bx$ is unprojected from the depth $\bD$, and $N_r$ incident directions $\bomega_k$ are drawn by importance sampling the environment map~\cite{pbrt4ed}, with densities $p_k$.
The incident radiance is $\bL_\text{i}(\bx, \bomega_k) = \bE(\bomega_k)\,V(\bx, \bomega_k)$, where the visibility $V = 1 - \sum_i T_i\alpha_i \in [0,1]$ accumulates the Gaussians along the ray and is computed with 3D Gaussian ray tracing~\cite{3dgrt}.
The BRDF $f$ is the Cook-Torrance model~\cite{cooktorrancebrdf}, which captures both diffuse and specular reflection (details in the supplementary material).
The per-pixel HDR radiance composites the reflected radiance $\bL_\text{r}$ with the environment radiance, weighted by the opacity $\bO$~\cite{chen2025omnire}:
{\small
\begin{equation}\label{eq:pbr}
\begin{aligned}
 \Cpbr &= (1 - \bO)\,\bE(-\bomega_\text{o}) + \bL_\text{r}, \\
 \bL_\text{r} &= \sum_{k=1}^{N_r}\frac{\bE(\bomega_k)\,V(\bx, \bomega_k)\,f(\bx, \bomega_k, \bomega_\text{o})\,(\bomega_k \cdot \bN(\bx))}{N_r\,p_k}.
\end{aligned}
\end{equation}
}
Local lights such as headlights are point or spot lights whose direct contributions are added to $\bL_\text{r}$, with shadow rays traced against the Gaussians (details in the supplementary material).
Since the input videos are LDR, the HDR radiance is tone-mapped to $\Cldr = \Cpbr^{1/\gamma}$ with $\gamma=2.2$~\cite{inverse_rendering_complex_indoor_scenes, FEGR}.

\vspace{-1.0em}
\paragraph{\textbf{Generative Rendering Pass.}}
The PBR pass follows the target lighting, but its output inherits the artifacts of the reconstruction. To remove them, BRDFusion uses the pretrained \emph{forward} model of DR~\cite{diffusionrenderer}, a video diffusion model trained on large-scale data that renders realistic frames from G-buffers and an environment map. Unlike image models~\cite{rgbx}, it processes all frames of a clip jointly. It consists of a VAE encoder $\enc$, a decoder $\dec$, and a latent denoiser $\diffusionModelF$.

To retain the structure and lighting computed by PBR, the generation starts from the PBR video $\Cldr^{1:F}$ (the tone-mapped frames $1$ to $F$) instead of pure noise, following SDEdit~\cite{sdedit}. The encoded PBR video is noised to step $\diffusionTime$, denoised by $\diffusionModelF$ under the condition $\cond$, and decoded,
\begin{equation}\label{eq:sdedit}
\begin{split}
    \Cgen^{1:F} &= \sdedit(\Cldr^{1:F} \mid \diffusionModelF, \cond, \diffusionTime) \\
    &= \dec(\texttt{denoise}(\latent_{\diffusionTime} \mid \diffusionModelF, \cond, \diffusionTime)), \\
    \latent_{\diffusionTime} &= \beta_{\diffusionTime}\enc(\Cldr^{1:F}) + \sigma_{\diffusionTime}\noise,
\end{split}
\end{equation}
where $\Cgen^{1:F}$ is the generated video, $\noise$ is Gaussian noise, and
$\beta_{\diffusionTime}, \sigma_{\diffusionTime}$ follow
EDM~\cite{EDM}. The step $\diffusionTime$ trades fidelity for realism. A higher noise level removes more artifacts but may drift from the PBR lighting (values in the supplementary material). The condition $\cond = \{\enc(\bE), \enc(\bN^{1:F}), \enc(\bD^{1:F}), \enc(\bA^{1:F}), \enc(\bR^{1:F}), \enc(\bM^{1:F})\}$ consists of the encoded target environment map and rasterized G-buffer videos.

Since the VAE is lossy and blurs fine details, the final result $\Cfin^{1:F}$ adds the residual between the generated video and the VAE-reconstructed PBR video back to the PBR video:
\begin{equation}\label{eq:vae_residual}
\begin{split}
    \Cfin^{1:F} &= \Cldr^{1:F} + \bigl(\Cgen^{1:F} - \dec(\enc(\Cldr^{1:F}))\bigr).
\end{split}
\end{equation}
Eq.~\ref{eq:vae_residual} keeps the fine details of the PBR video that the VAE cannot reconstruct, at the cost of also keeping part of its Monte Carlo noise (Sec.~\ref{sec:ablation}).
Sequences longer than the 57-frame window of DR are processed as overlapping clips whose outputs are averaged (sliding-window averaging).

\subsection{Hybrid Inverse Rendering}
\label{sec:inverse_render}
\vspace{-0.5em}
Recovering geometry, materials, and lighting jointly from the input video is highly ill-posed.
BRDFusion therefore recovers them in four stages (Fig.~\ref{fig:training}). A one-stage optimization with equal iterations performs worse (Sec.~\ref{sec:ablation}).

\begin{figure*}[t]
    \centering\setlength{\tabcolsep}{2pt}
    \newcommand{\figfivelabel}[1]{\raisebox{0.5in}[0pt][0pt]{\rotatebox[origin=c]{90}{#1}}}
    \resizebox{1.0\textwidth}{!}{%
    \begin{tabular}{@{}lcccccc@{}}

     & Novel-View Synthesis & + Relighting & Novel-View Synthesis & + Relighting & Novel-View Synthesis & + Relighting \\[0.2em]

    \figfivelabel{GT \& Lighting}&
    \frame{\includegraphics[width=1.5in]{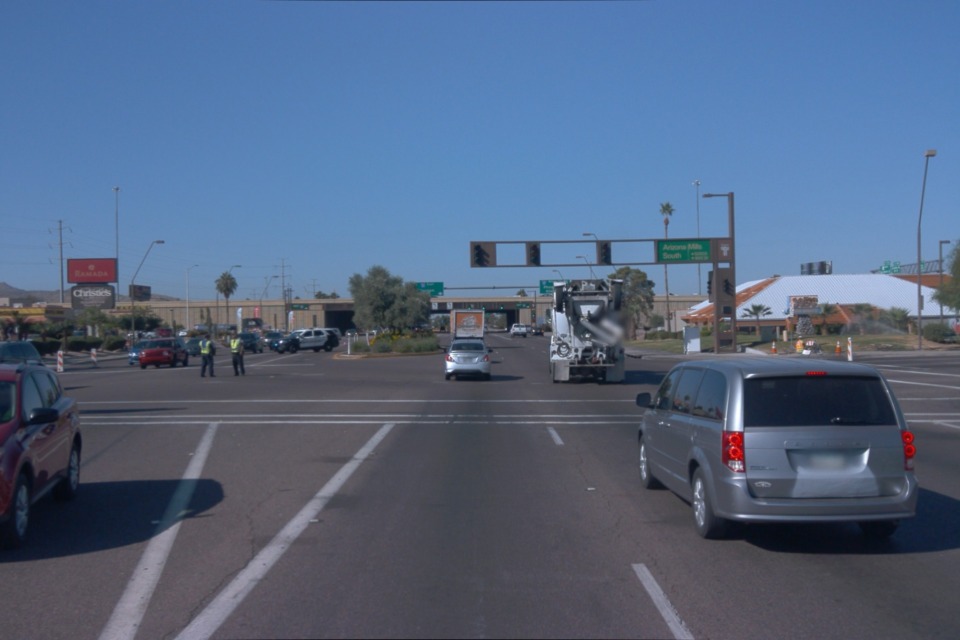}}&
    \includegraphics[width=1.5in]{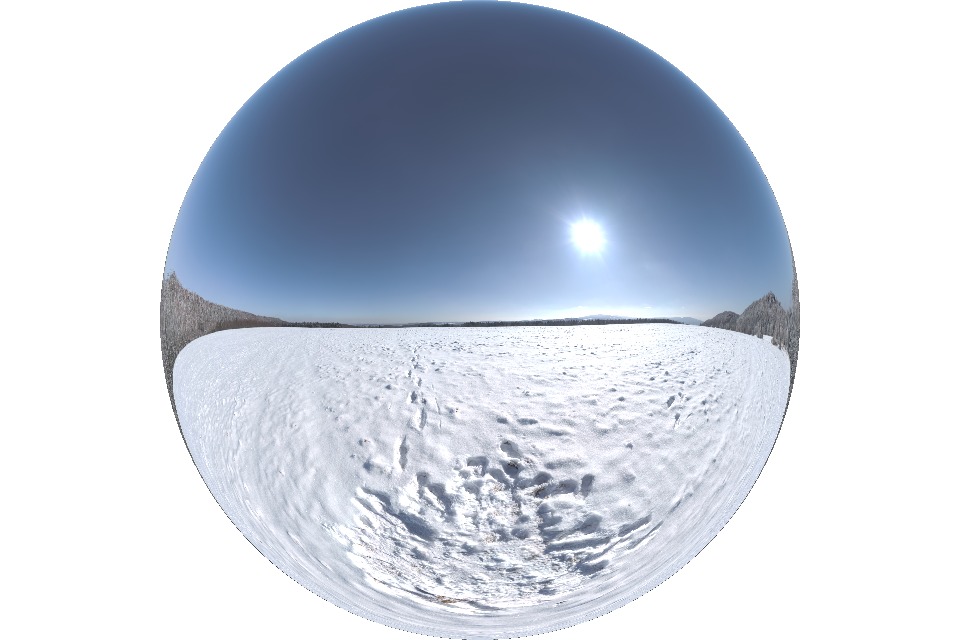}&
    \frame{\includegraphics[width=1.5in]{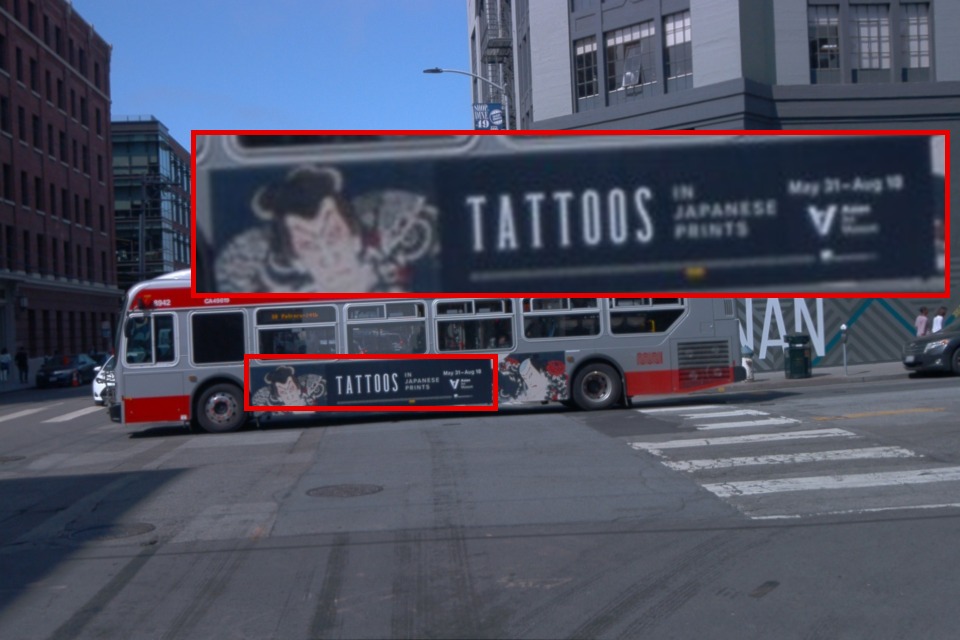}}&
    \includegraphics[width=1.5in]{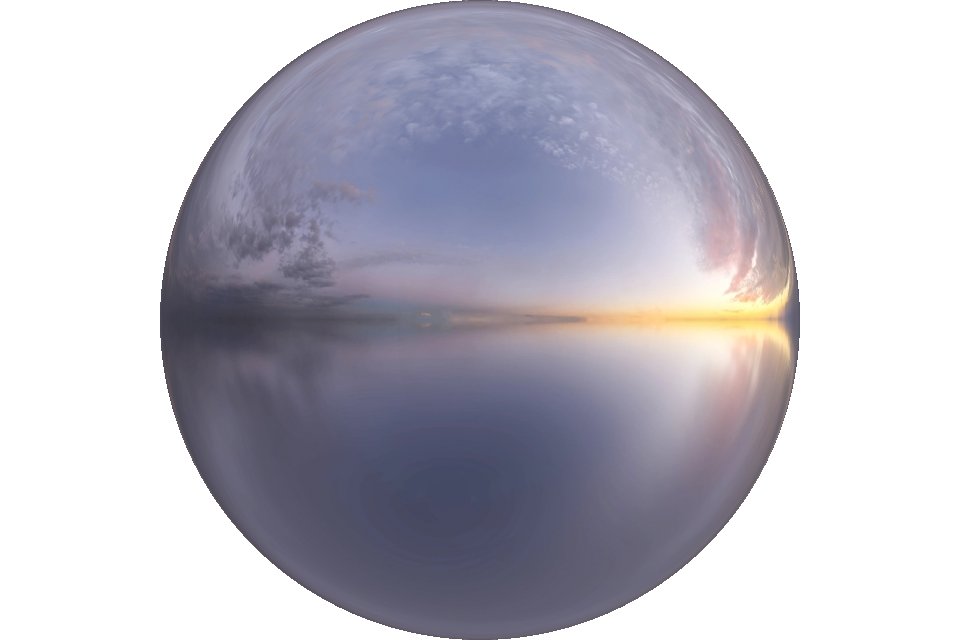}&
    \frame{\includegraphics[width=1.5in]{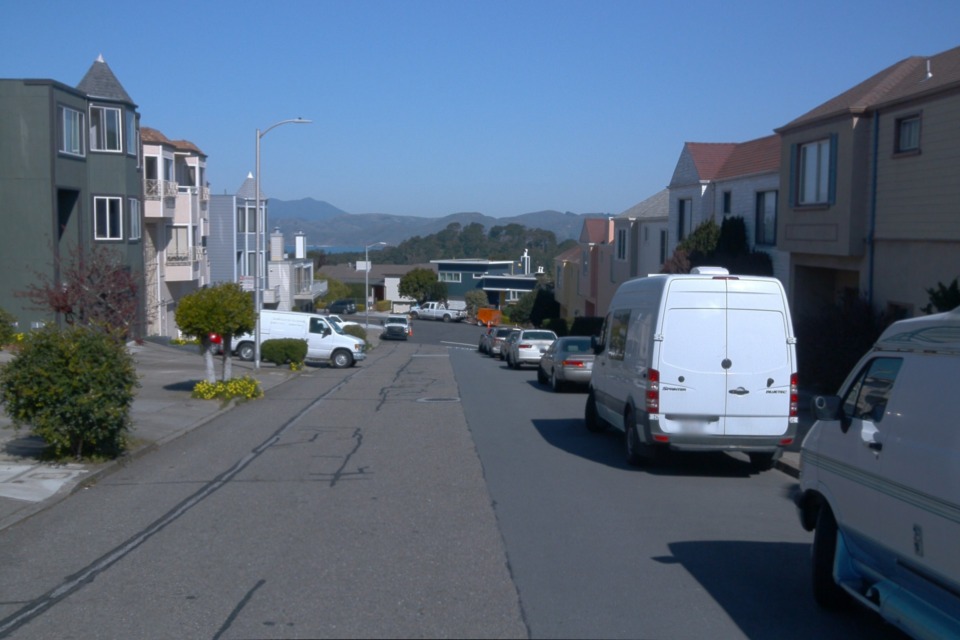}}&
    \includegraphics[width=1.5in]{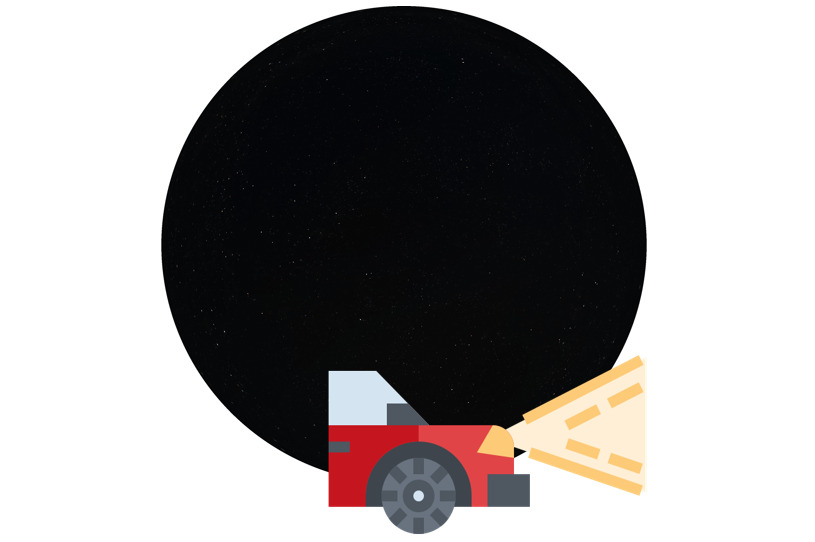} \\

    \figfivelabel{InvRGB+L~\cite{invrgbl}}&
    \frame{\includegraphics[width=1.5in]{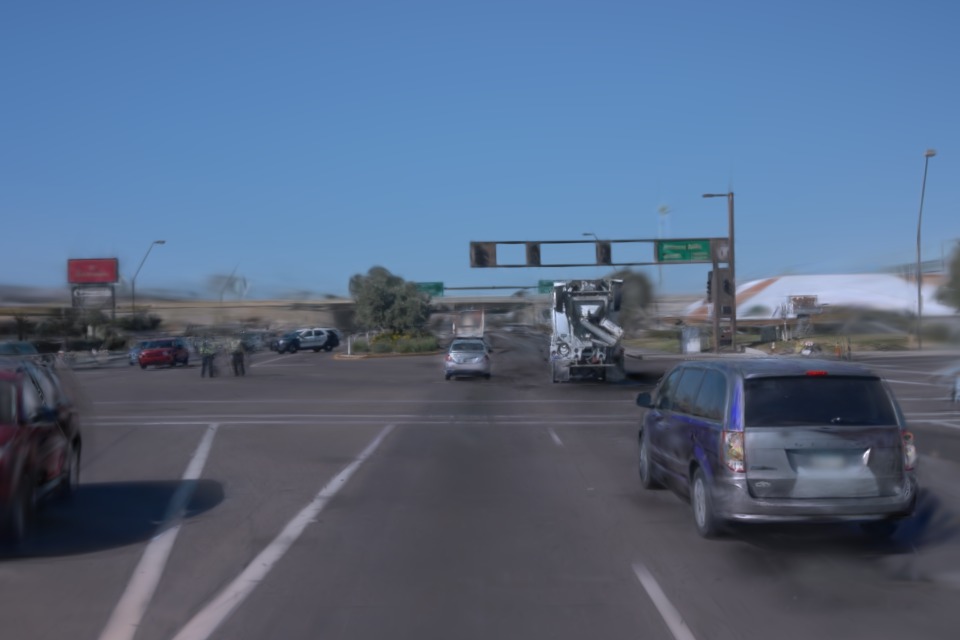}}&
    \frame{\includegraphics[width=1.5in]{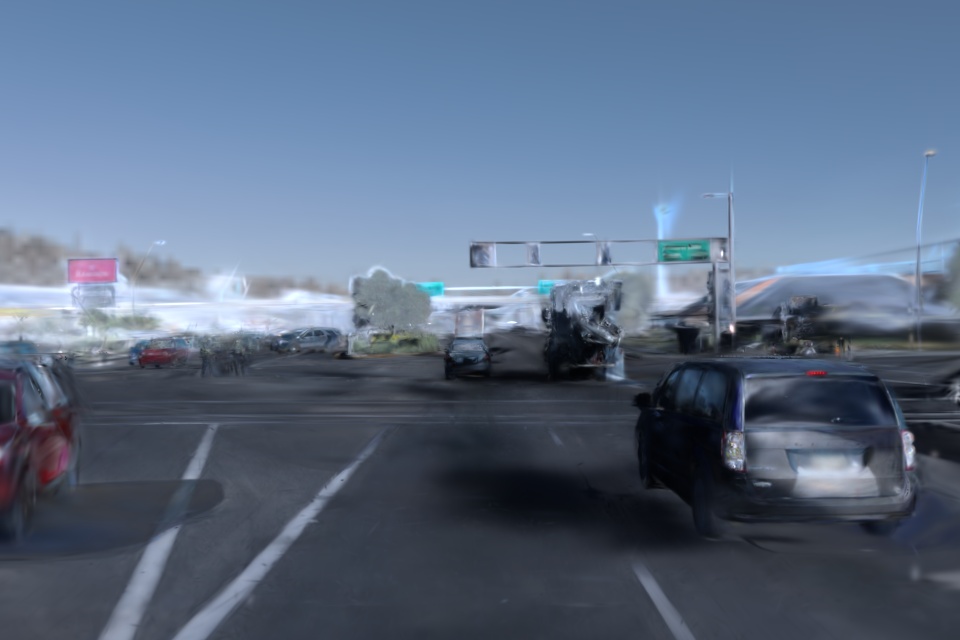}}&
    \frame{\includegraphics[width=1.5in]{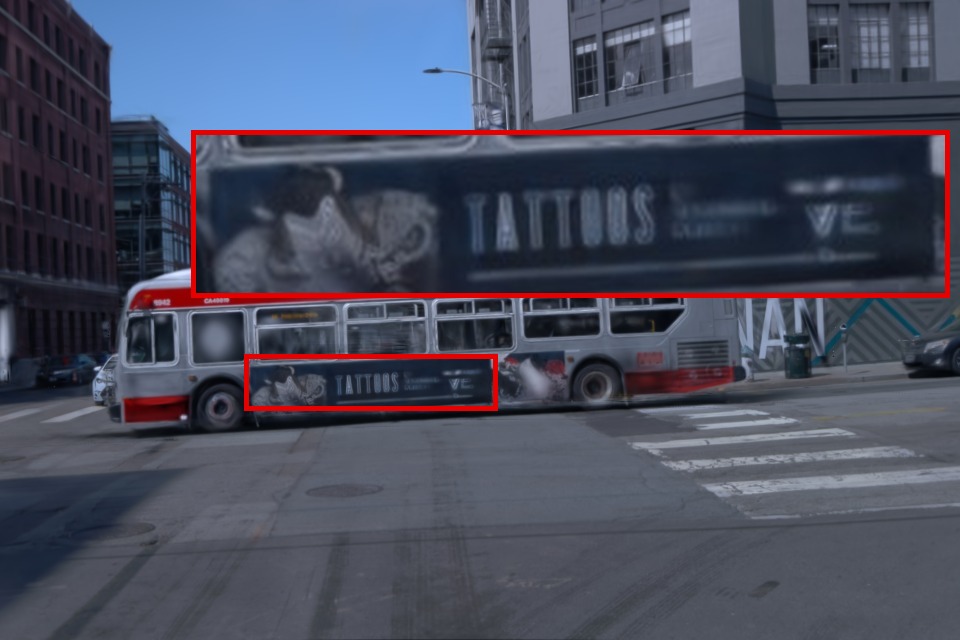}}&
    \frame{\includegraphics[width=1.5in]{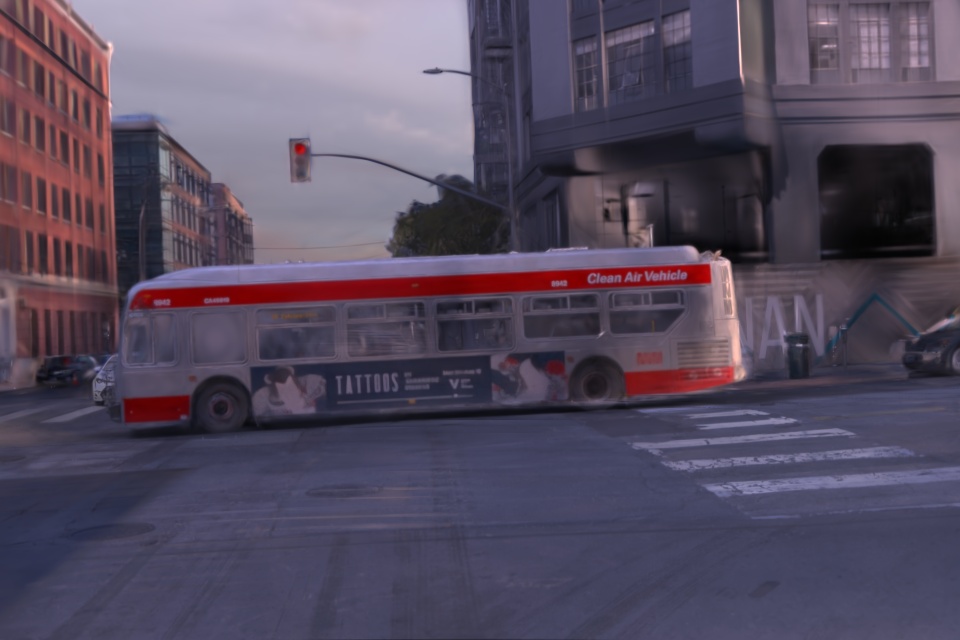}}&
    \frame{\includegraphics[width=1.5in]{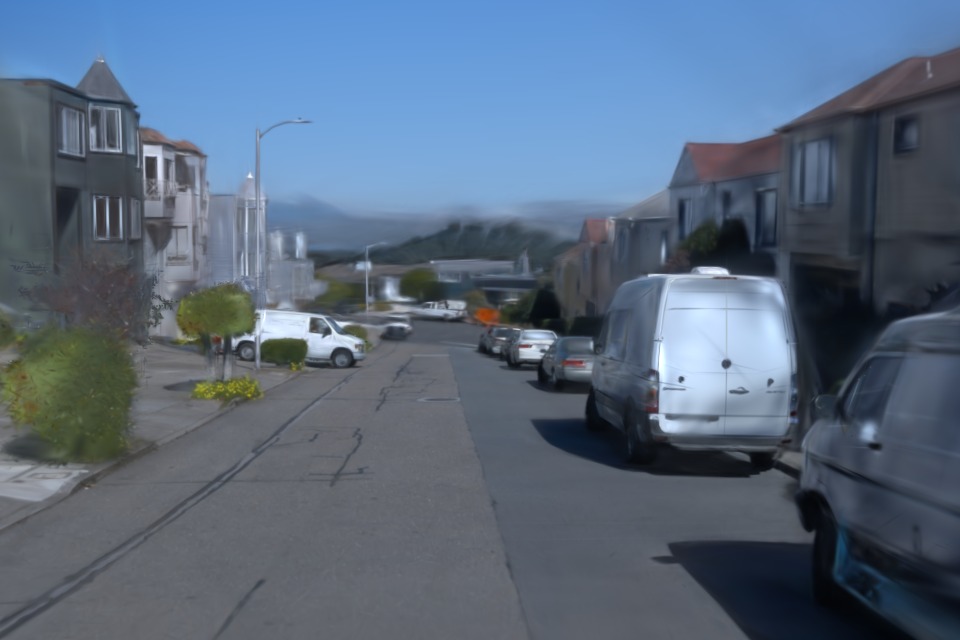}}&
    \frame{\includegraphics[width=1.5in]{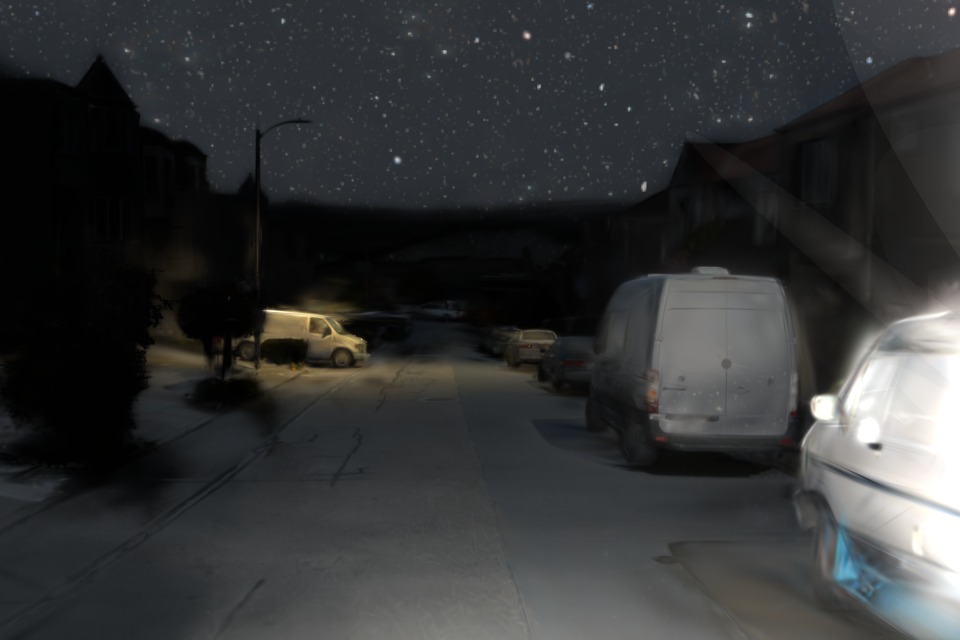}} \\

    \figfivelabel{\begin{tabular}{@{}c@{}}Gen3C~\cite{Gen3C}\\+DR~\cite{diffusionrenderer}\end{tabular}}&
    \frame{\includegraphics[width=1.5in]{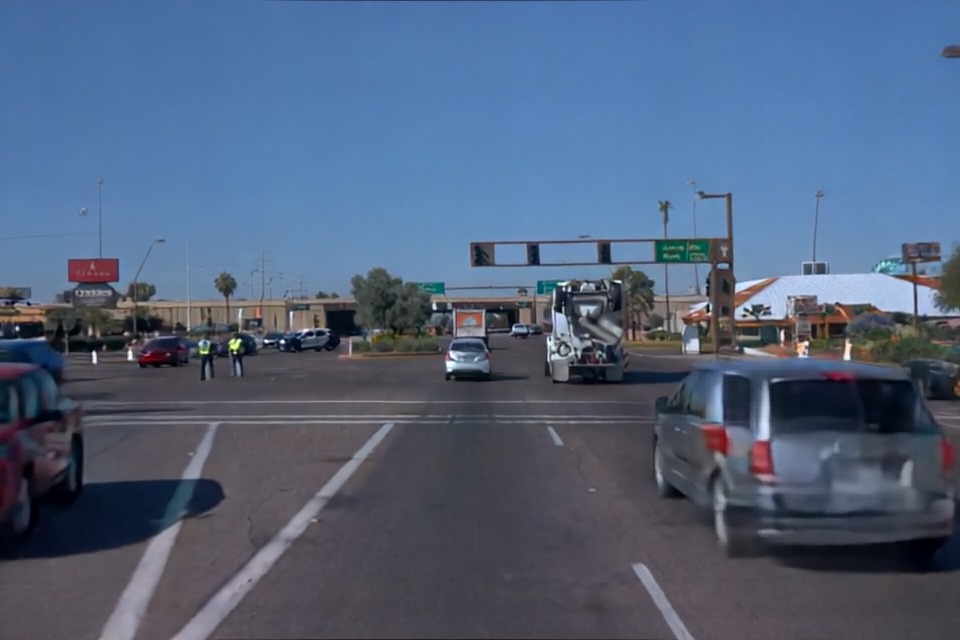}}&
    \frame{\includegraphics[width=1.5in]{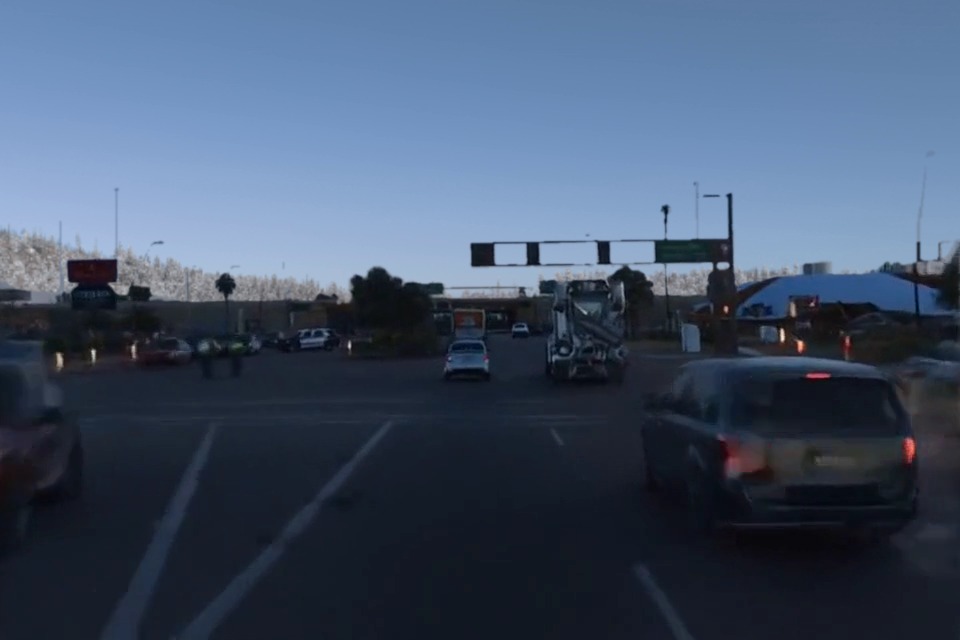}}&
    \frame{\includegraphics[width=1.5in]{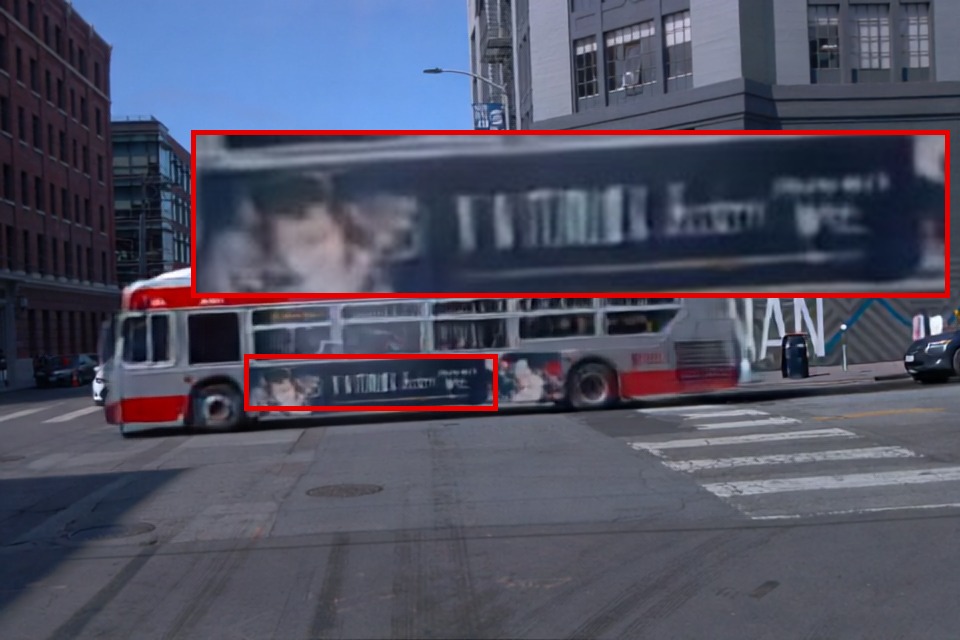}}&
    \frame{\includegraphics[width=1.5in]{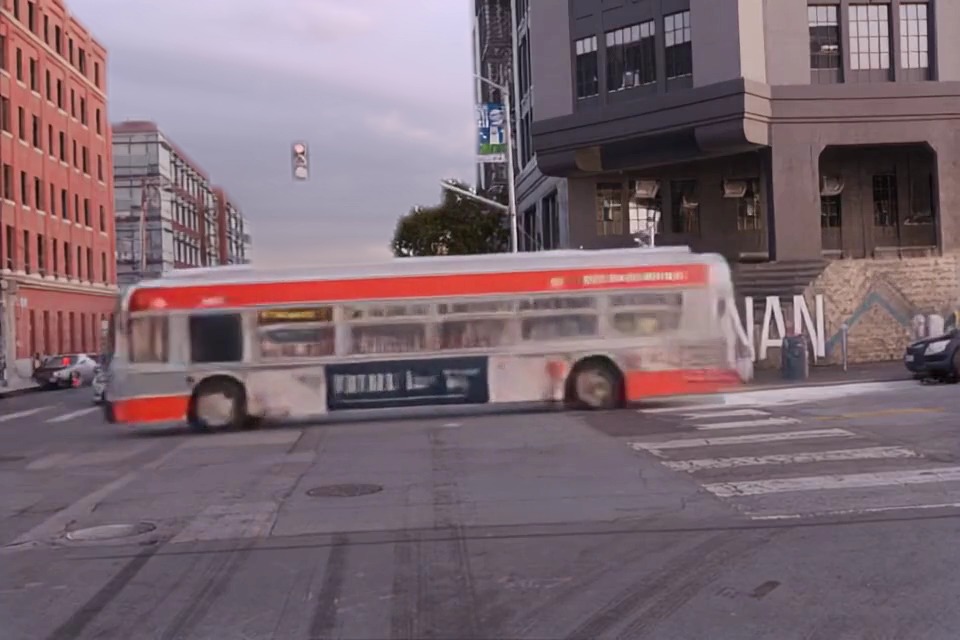}}&
    \frame{\includegraphics[width=1.5in]{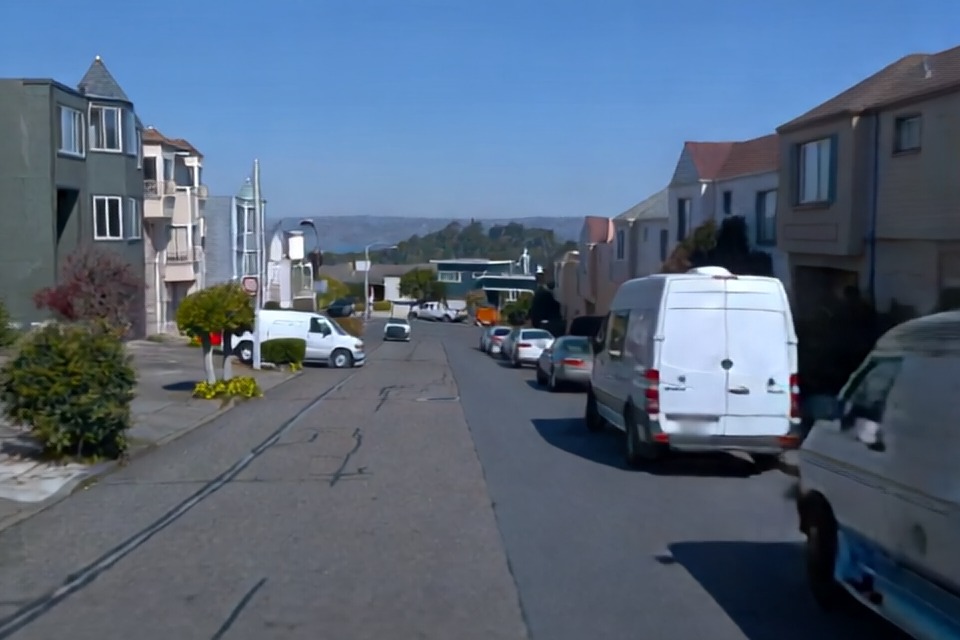}}&
    \frame{\includegraphics[width=1.5in]{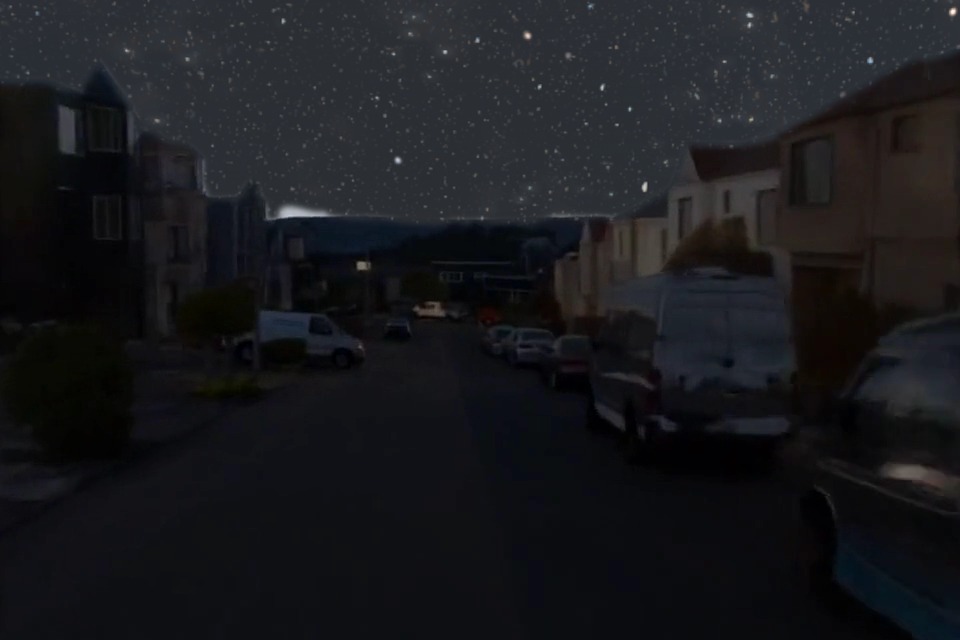}} \\

    \figfivelabel{AW4D~\cite{autoweather4d}}&
    \frame{\includegraphics[width=1.5in]{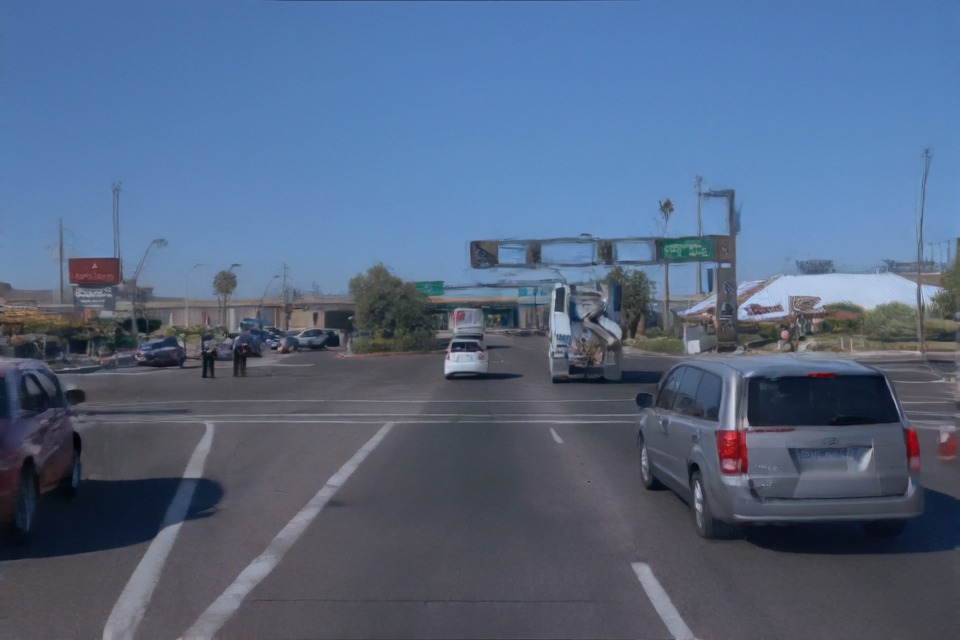}}&
    \frame{\includegraphics[width=1.5in]{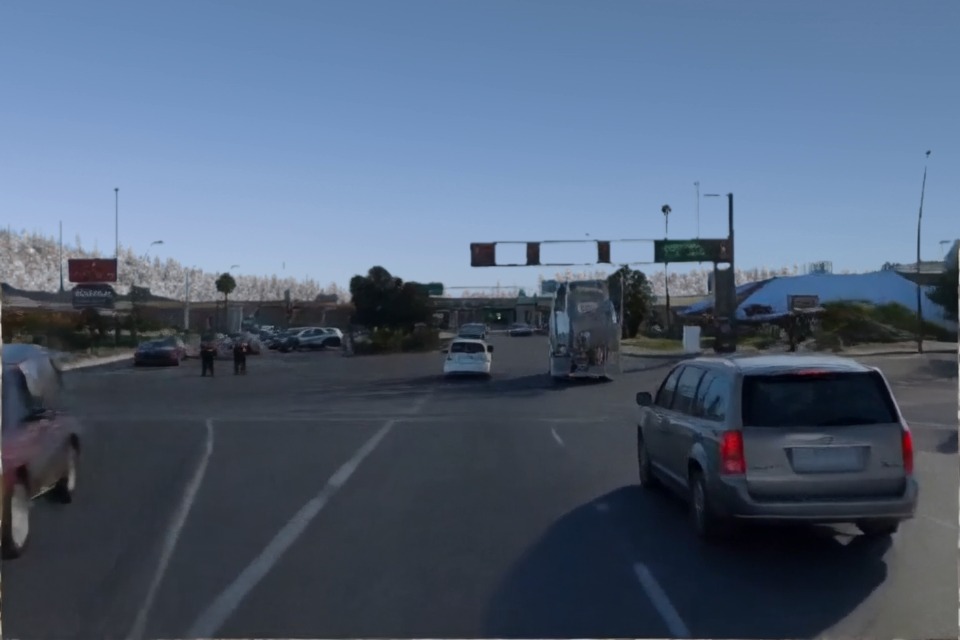}}&
    \frame{\includegraphics[width=1.5in]{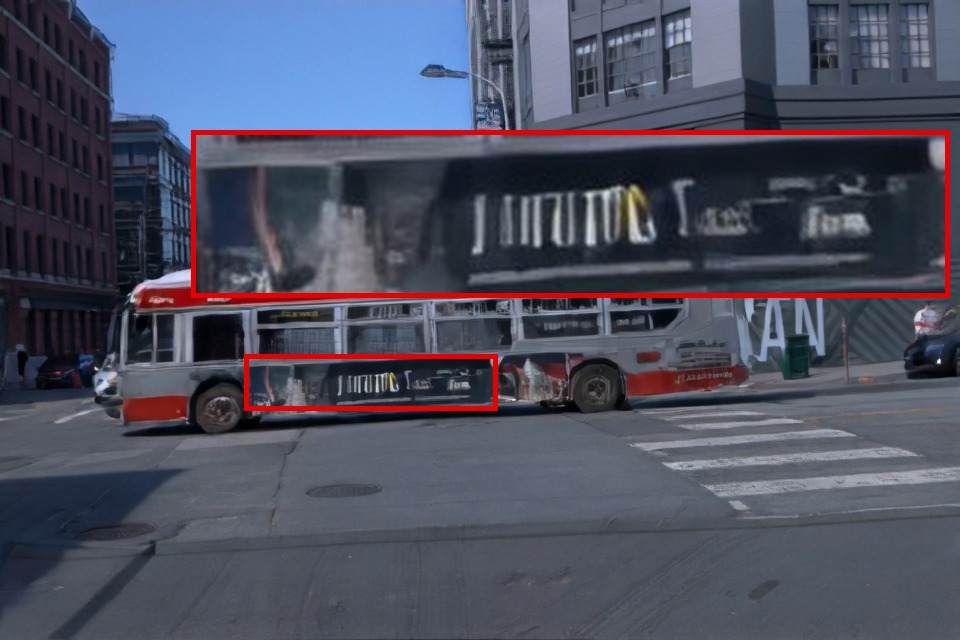}}&
    \frame{\includegraphics[width=1.5in]{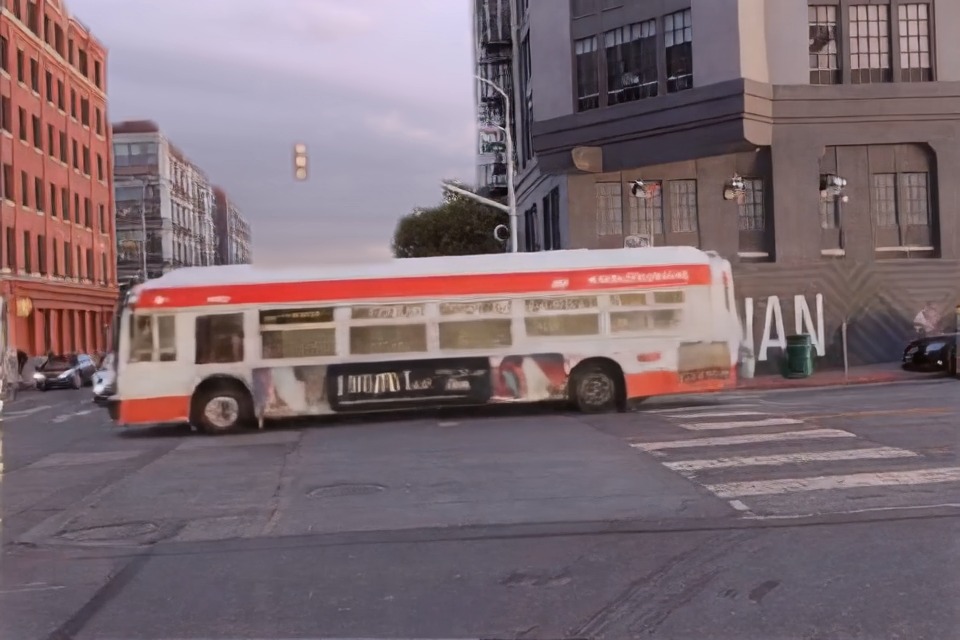}}&
    \frame{\includegraphics[width=1.5in]{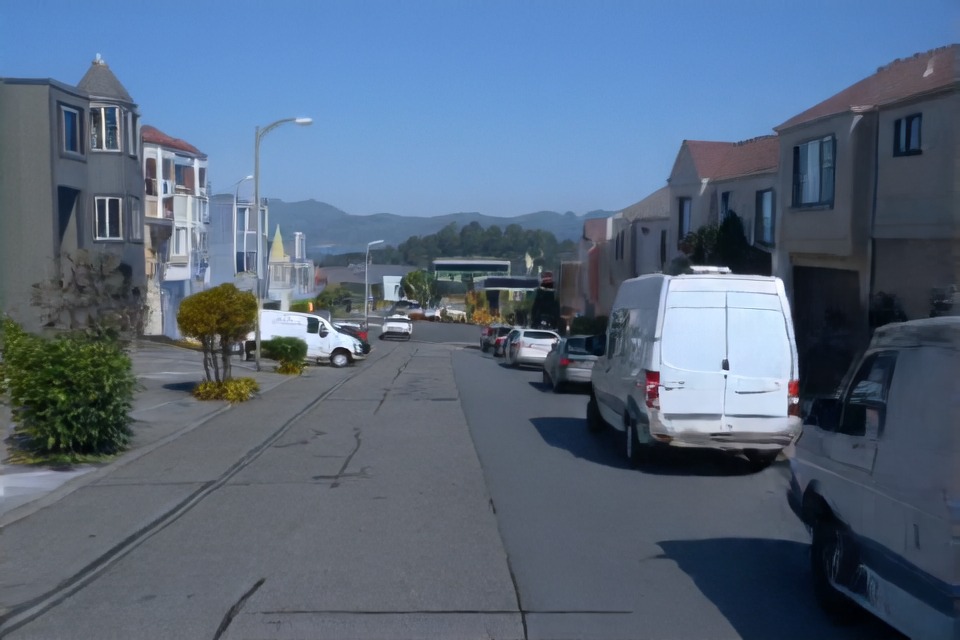}}&
    \frame{\includegraphics[width=1.5in]{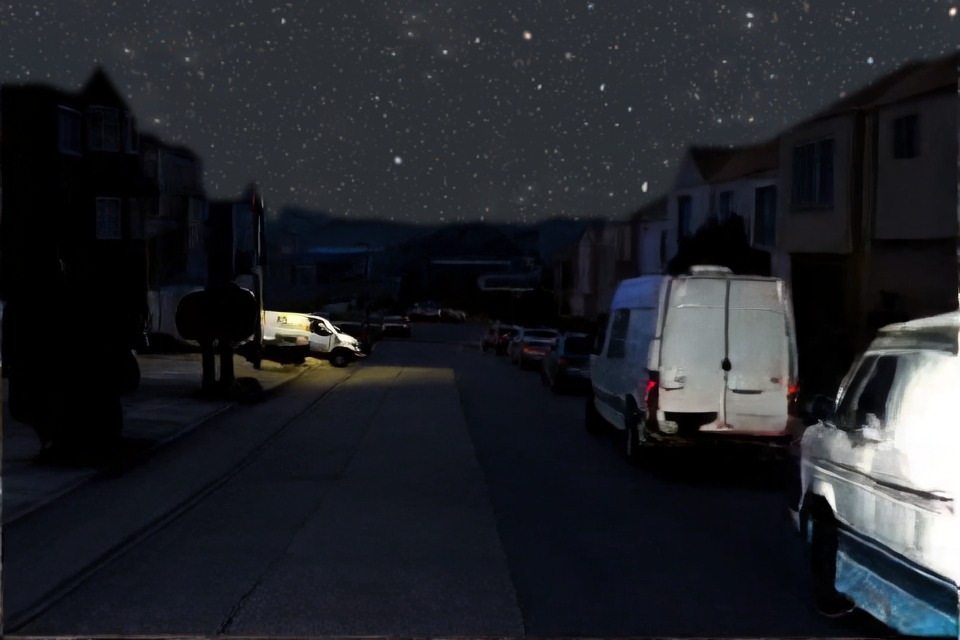}} \\

    \figfivelabel{Ours}&
    \frame{\includegraphics[width=1.5in]{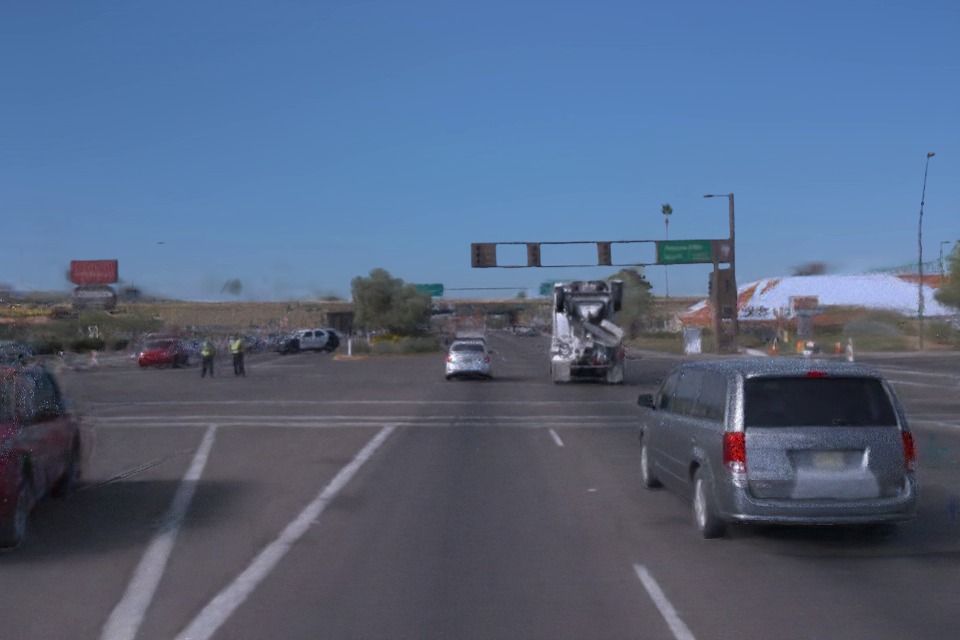}}&
    \frame{\includegraphics[width=1.5in]{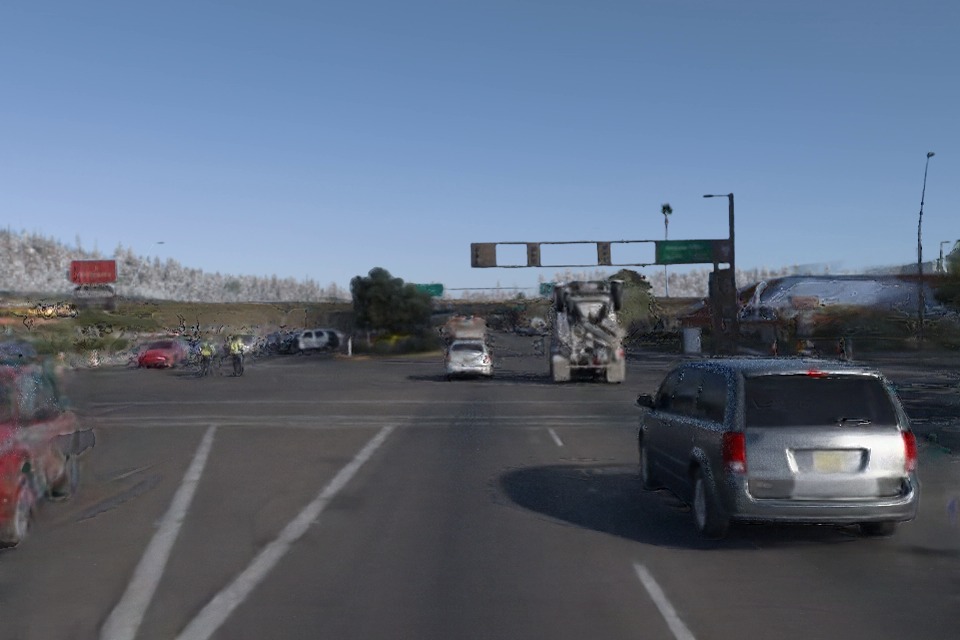}}&
    \frame{\includegraphics[width=1.5in]{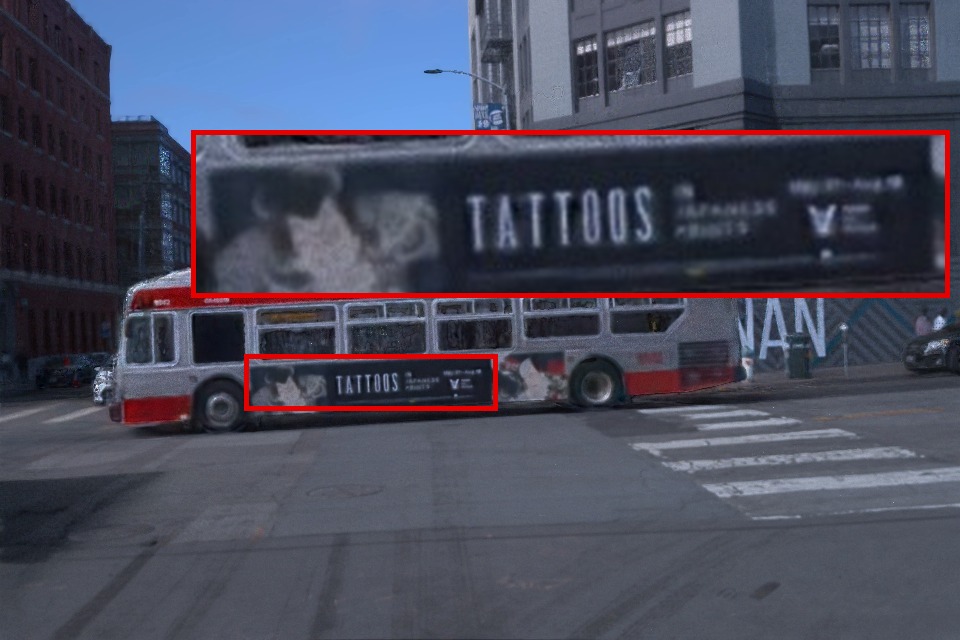}}&
    \frame{\includegraphics[width=1.5in]{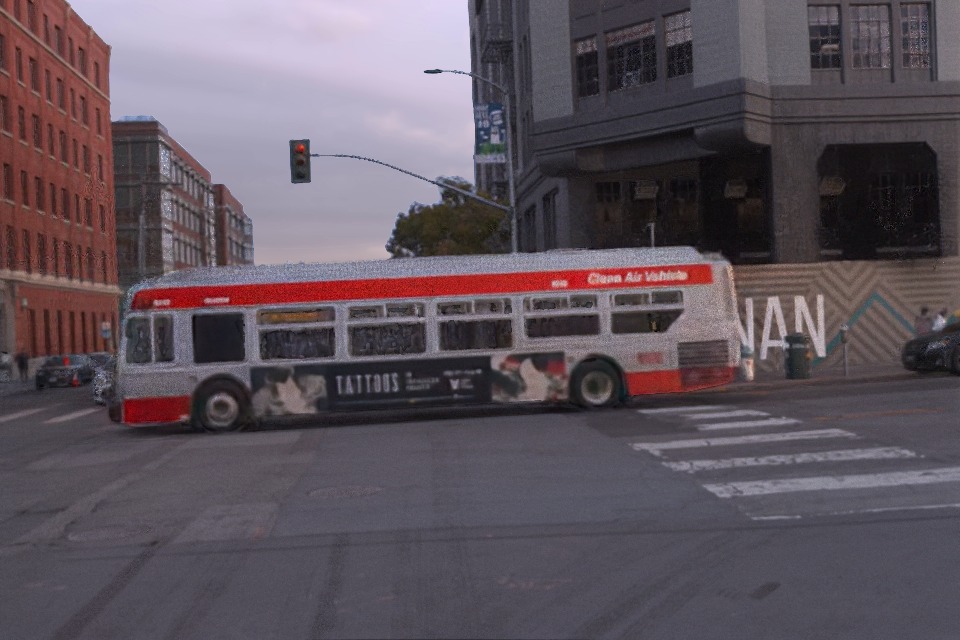}}&
    \frame{\includegraphics[width=1.5in]{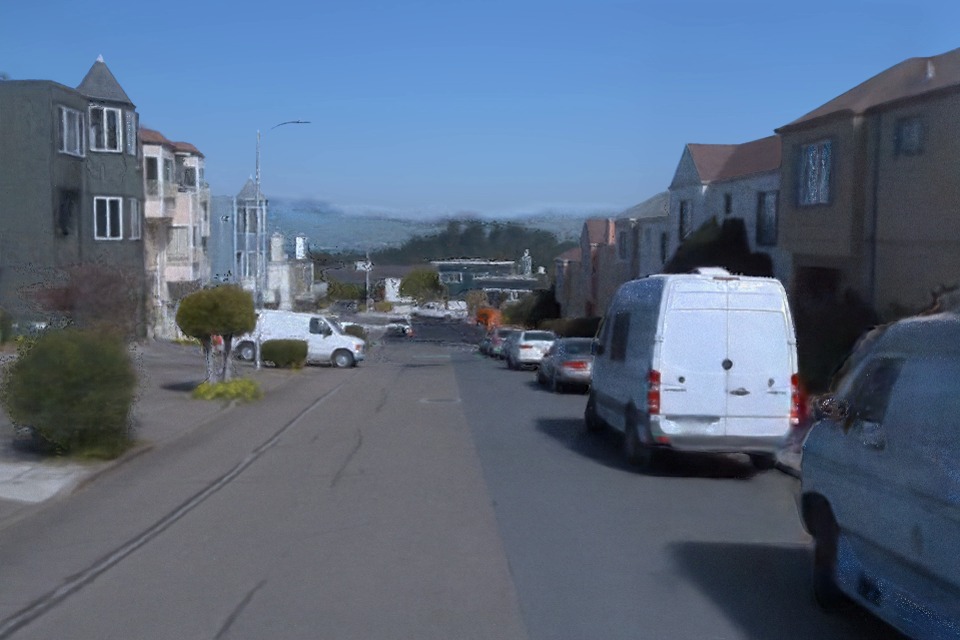}}&
    \frame{\includegraphics[width=1.5in]{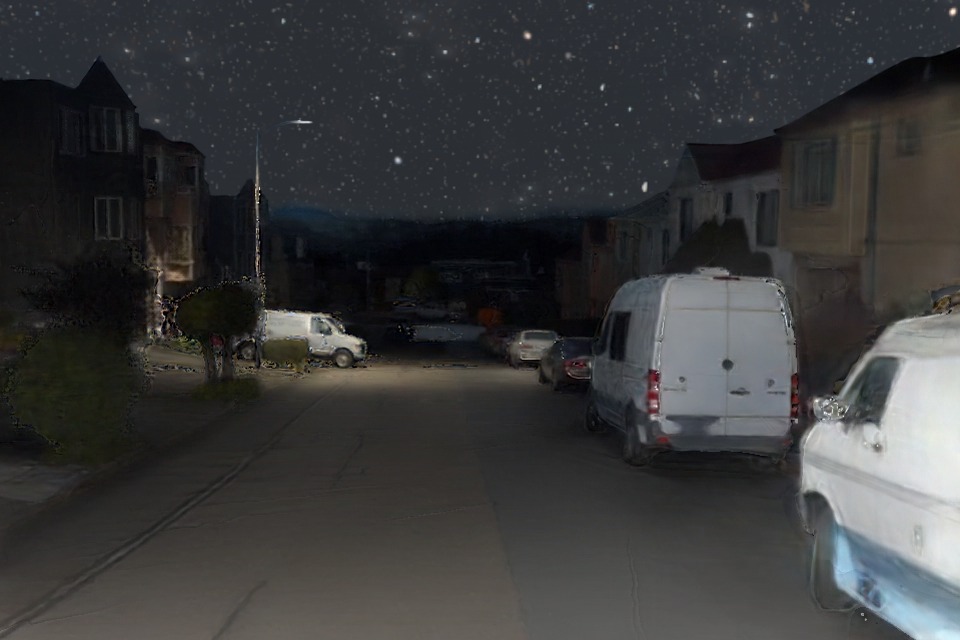}} \\
    \end{tabular}%
    }
    \caption{\textbf{Novel-view synthesis and relighting on WOD.}
    Each scene shows a novel view and its relighting with the lighting in the top row; the right scene is relit at night with headlights and a streetlight placed in 3D.
    InvRGB+L~\cite{invrgbl} shows geometric artifacts.
    Gen3C~\cite{Gen3C}+DR~\cite{diffusionrenderer} does not reproduce the shadows and the local lighting.
    AW4D~\cite{autoweather4d} blends local and global lighting per pixel, which leaves the road under the headlights dark.
    \textcolor{red}{\textbf{Red}} boxes: Gen3C+DR and AW4D alter fine details such as the text on the bus.
    BRDFusion keeps the scene content and follows both the global and the local lighting.
    For Gen3C+DR and AW4D, the sky is alpha-blended from the target environment map.
    }
    \vspace{-5mm}
    \label{fig:qual_relighting}
\end{figure*}

\begin{figure*}[t]
    \centering\setlength{\tabcolsep}{3pt}
    \resizebox{1.0\textwidth}{!}{%
    \begin{tabular}{@{}lccccc@{}}

     & Novel-View Synthesis & Normal & Albedo & Roughness & Metallic \\[0.2em]

    \raisebox{2.5\normalbaselineskip}[0pt][0pt]{\rotatebox[origin=c]{90}{InvRGB+L~\cite{invrgbl}}}& \frame{\includegraphics[width=1.5in,height=1in]{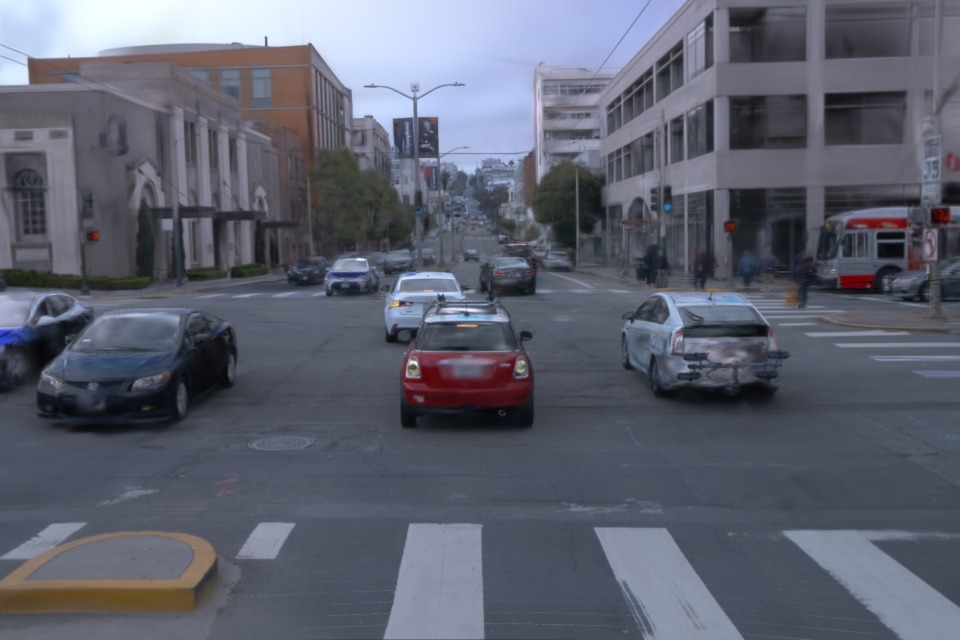}}& \frame{\includegraphics[width=1.5in,height=1in]{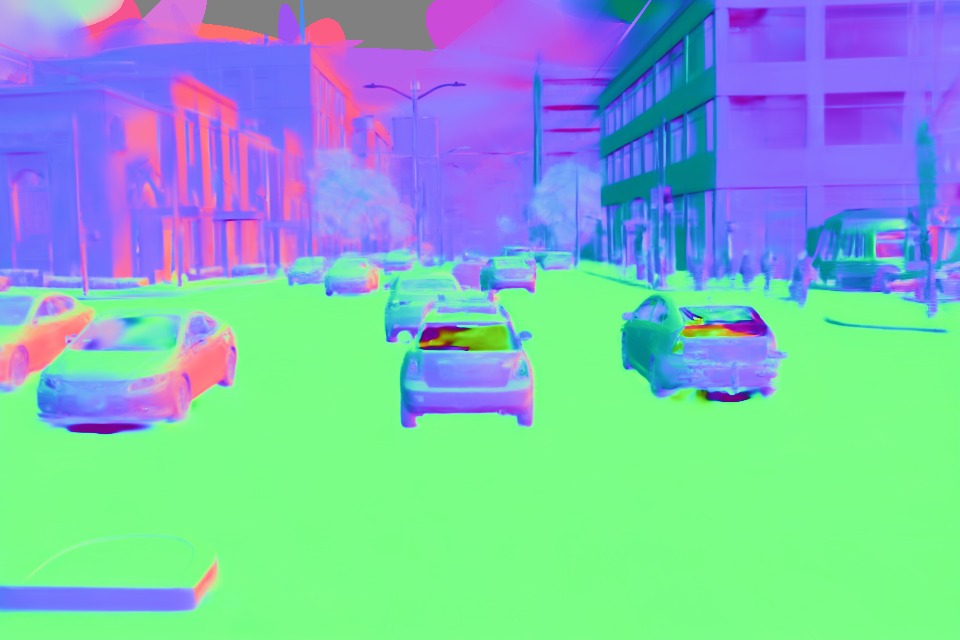}}& \frame{\includegraphics[width=1.5in,height=1in]{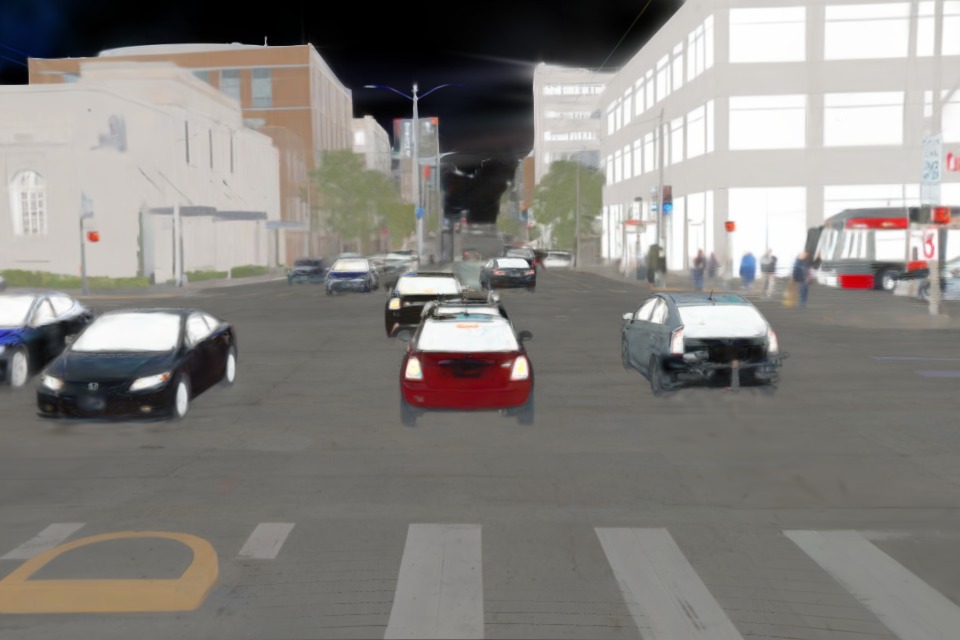}}& \frame{\includegraphics[width=1.5in,height=1in]{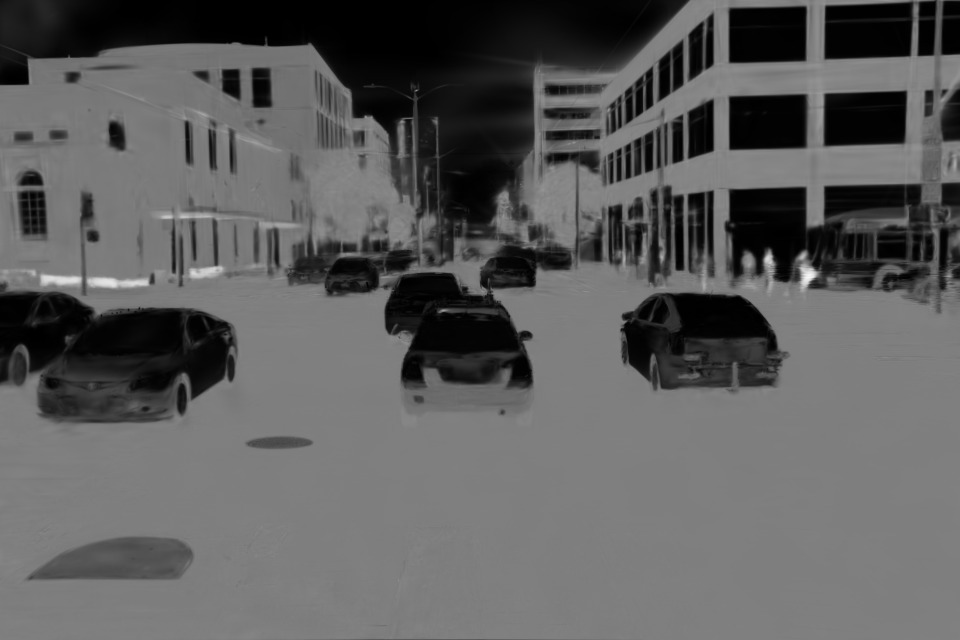}} & \frame{\includegraphics[width=1.5in,height=1in]{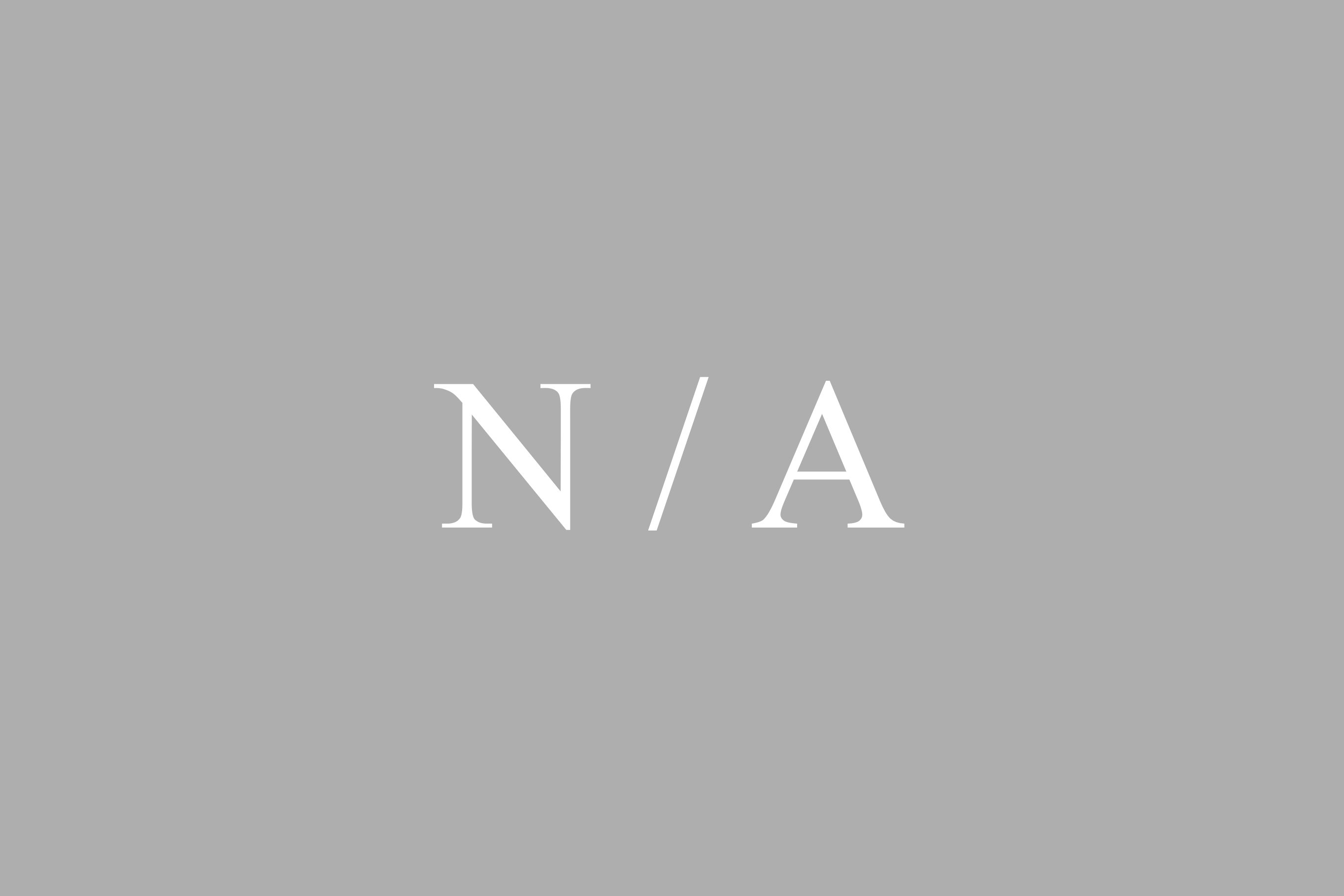}}  \\

    \raisebox{2.5\normalbaselineskip}[0pt][0pt]{\rotatebox[origin=c]{90}{\begin{tabular}{@{}c@{}}Gen3C~\cite{Gen3C}\\+DR~\cite{diffusionrenderer}\end{tabular}}}& \frame{\includegraphics[width=1.5in,height=1in]{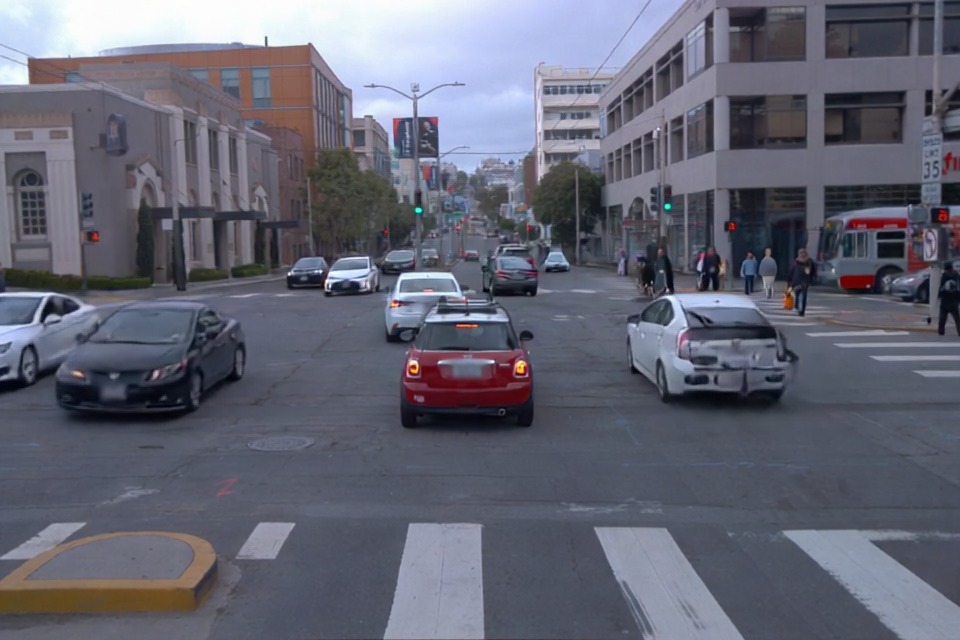}}& \frame{\includegraphics[width=1.5in,height=1in]{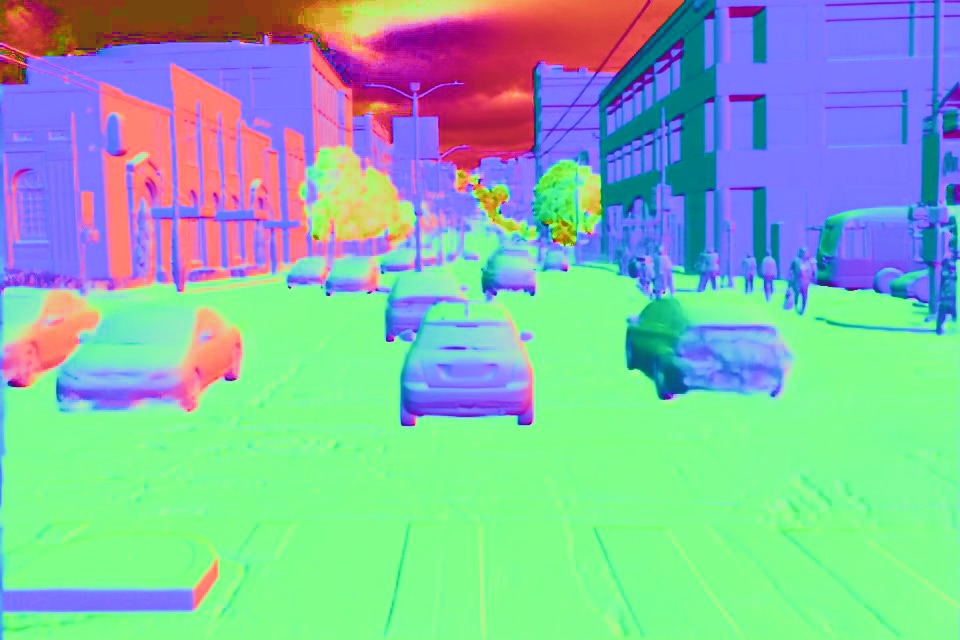}}& \frame{\includegraphics[width=1.5in,height=1in]{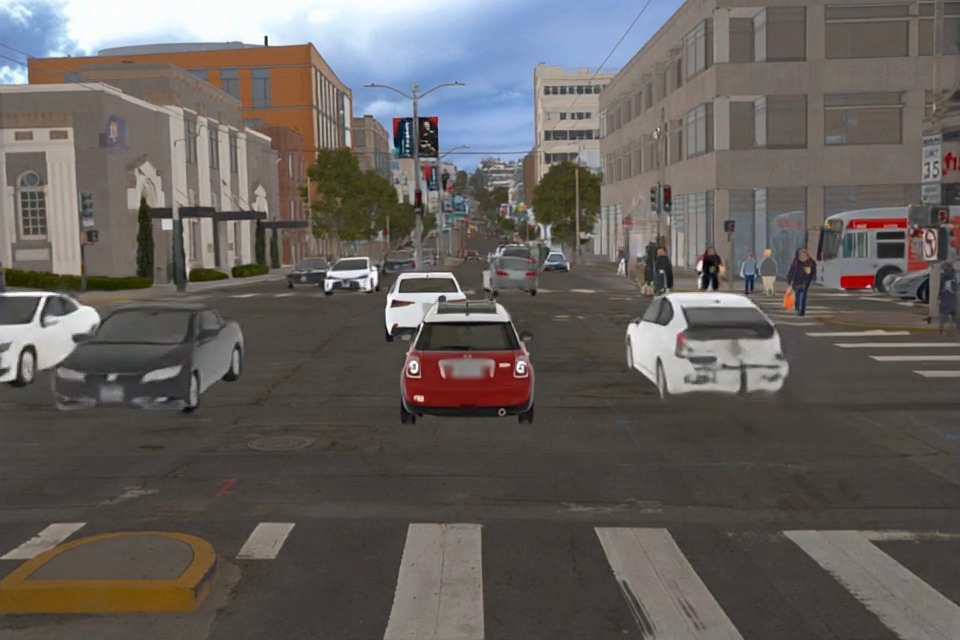}}& \frame{\includegraphics[width=1.5in,height=1in]{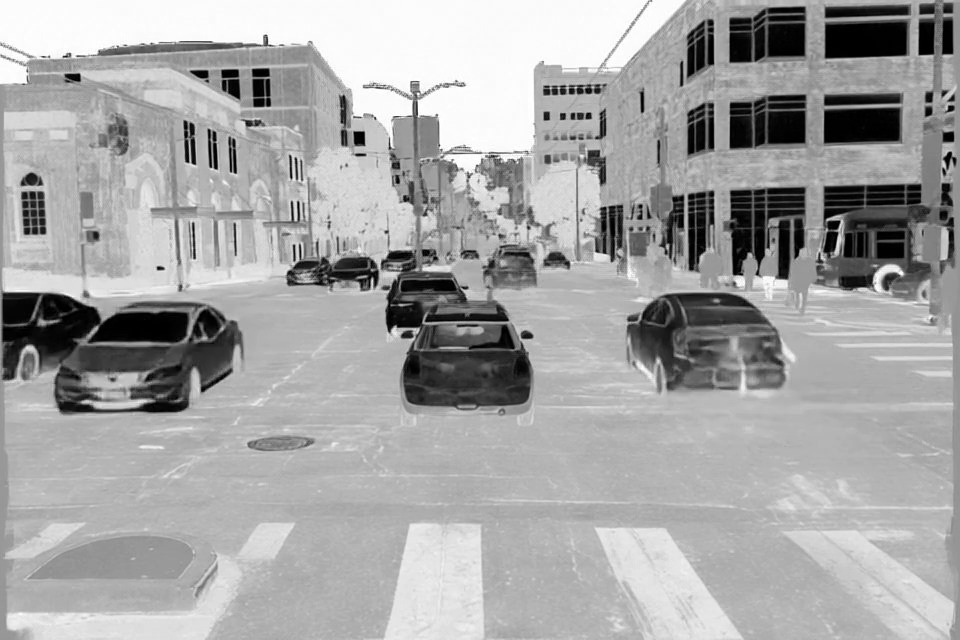}} & \frame{\includegraphics[width=1.5in,height=1in]{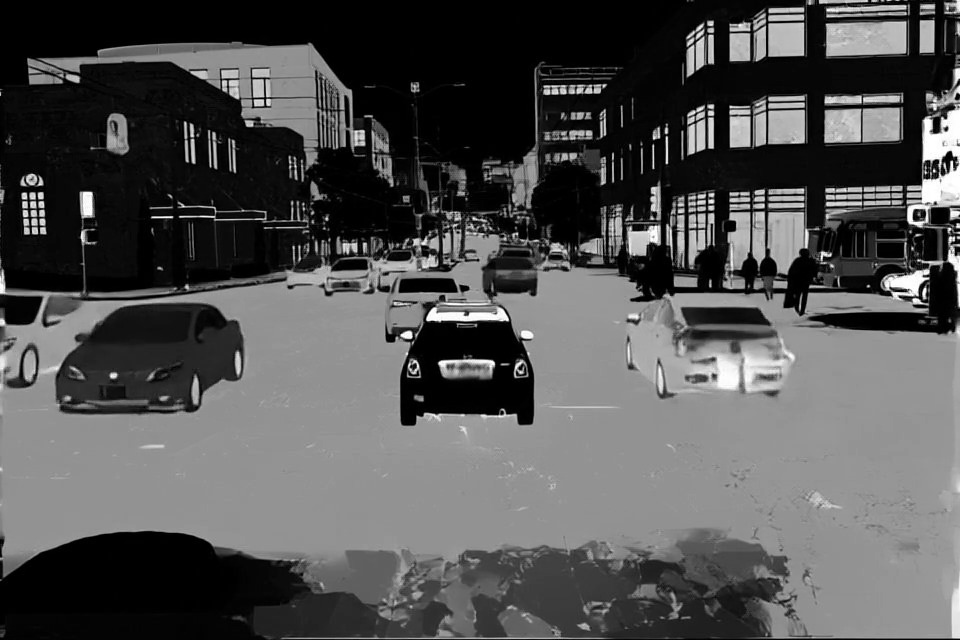}} \\

        \raisebox{2.5\normalbaselineskip}[0pt][0pt]{\rotatebox[origin=c]{90}{AW4D~\cite{autoweather4d}}}& \frame{\includegraphics[width=1.5in,height=1in]{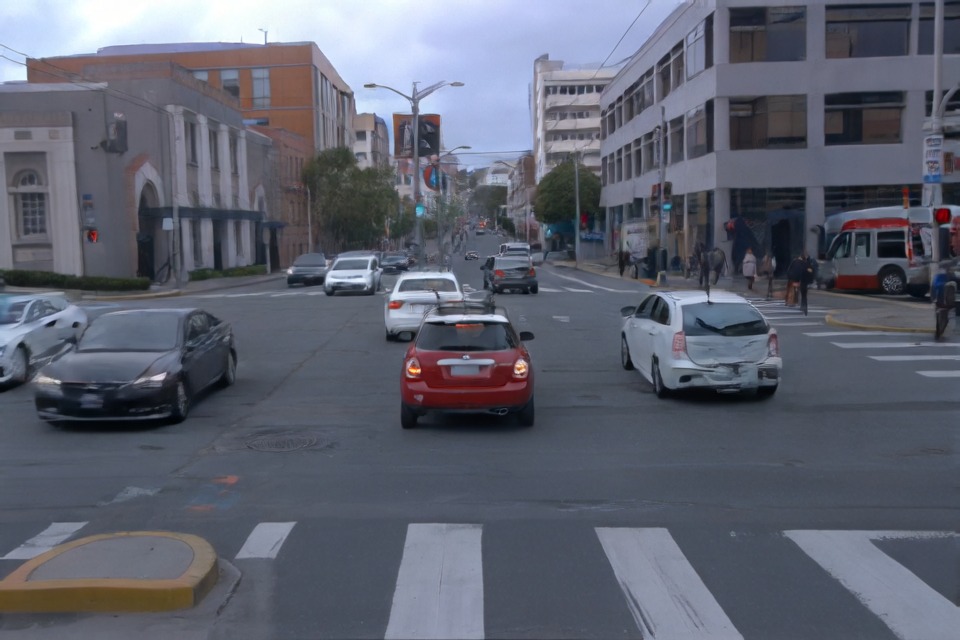}}& \frame{\includegraphics[width=1.5in,height=1in]{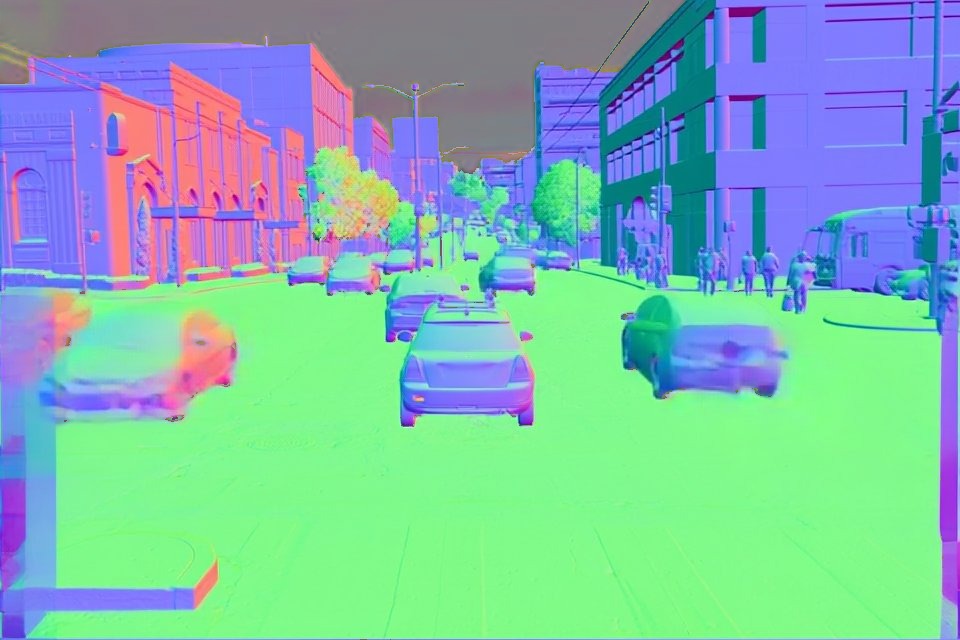}}& \frame{\includegraphics[width=1.5in,height=1in]{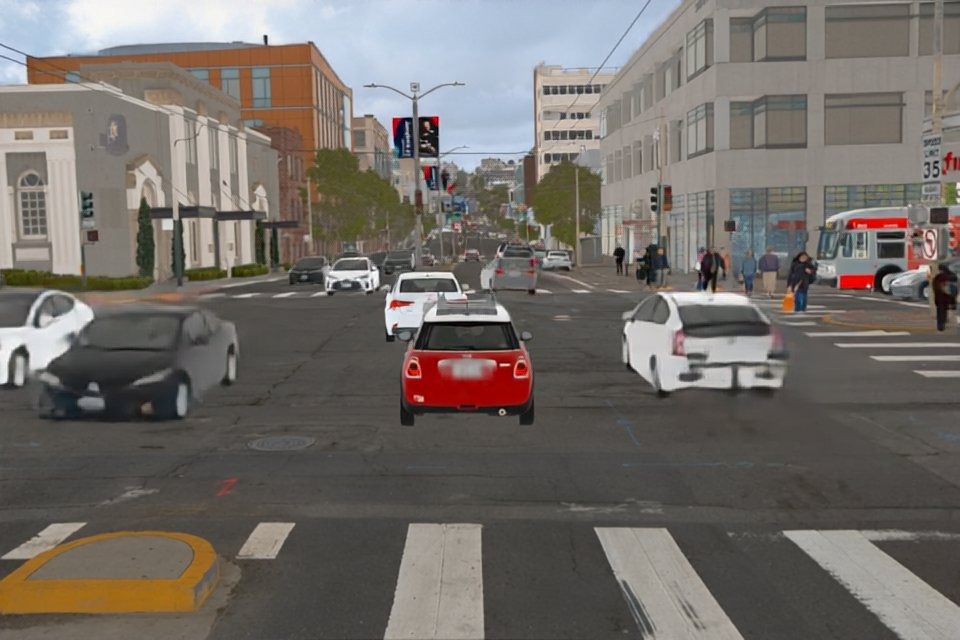}}& \frame{\includegraphics[width=1.5in,height=1in]{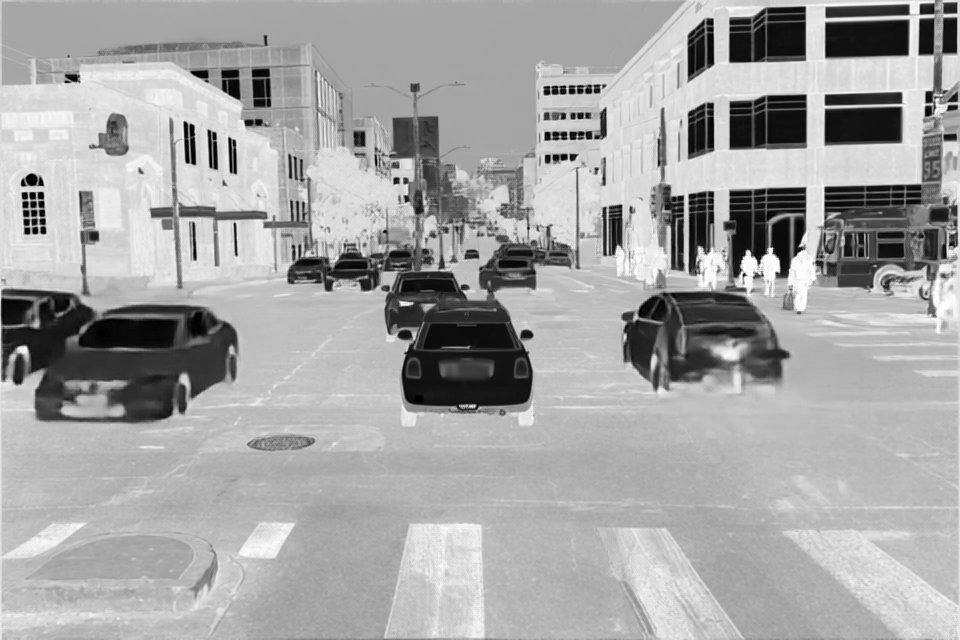}} & \frame{\includegraphics[width=1.5in,height=1in]{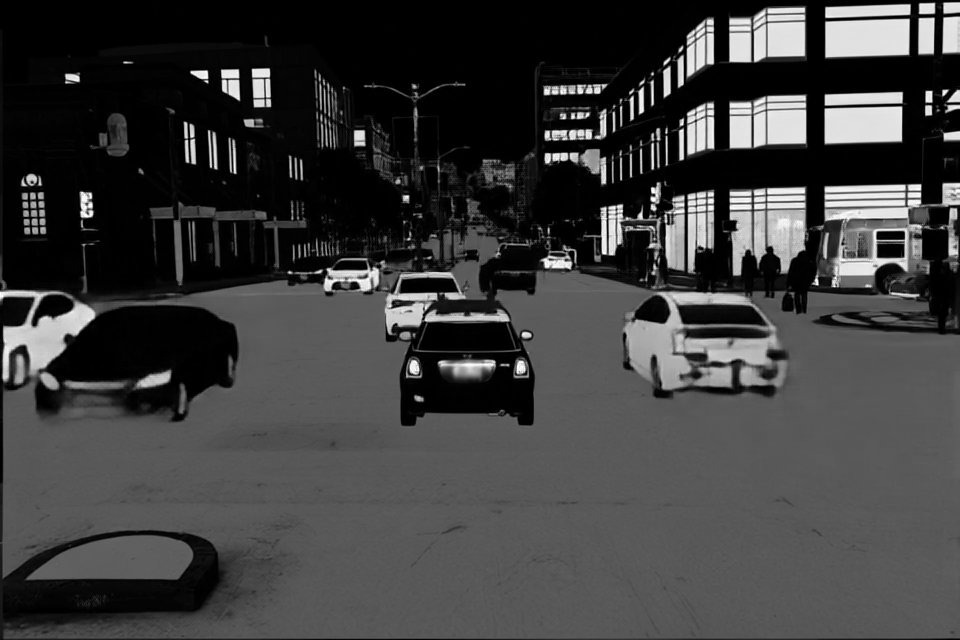}} \\

    \raisebox{2.5\normalbaselineskip}[0pt][0pt]{\rotatebox[origin=c]{90}{Ours}}& \frame{\includegraphics[width=1.5in,height=1in]{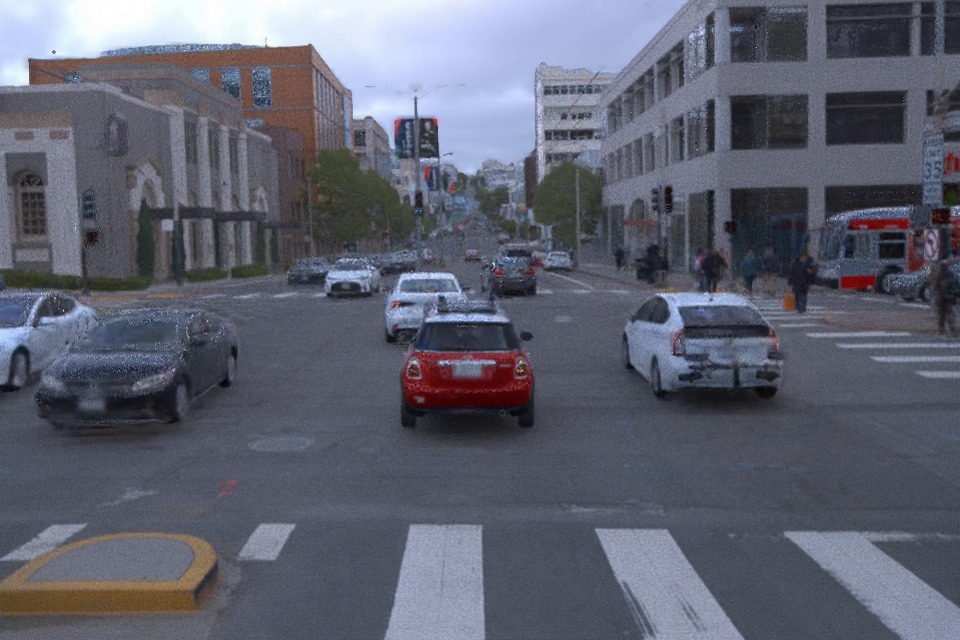}}& \frame{\includegraphics[width=1.5in,height=1in]{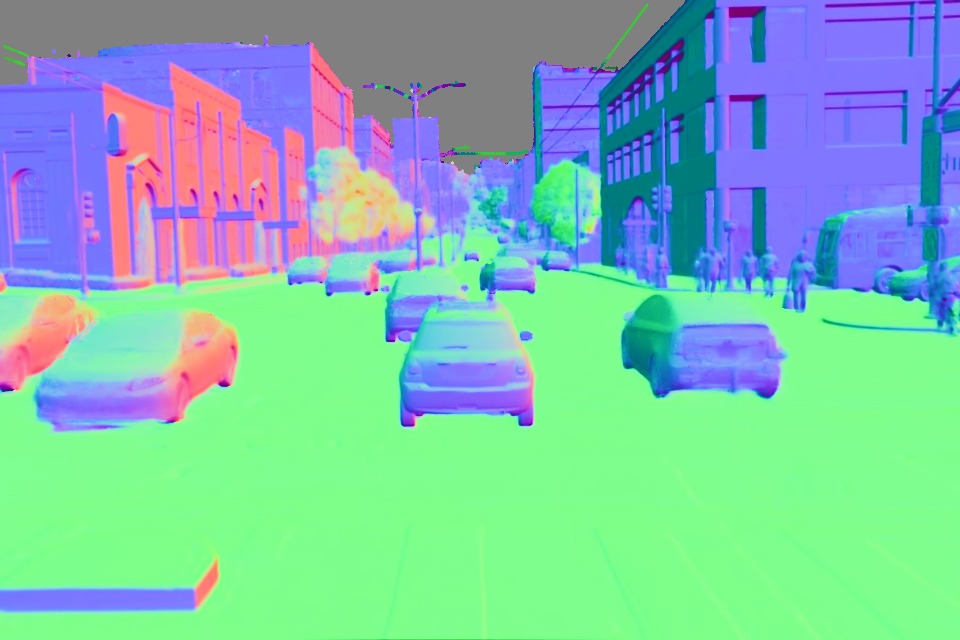}}& \frame{\includegraphics[width=1.5in,height=1in]{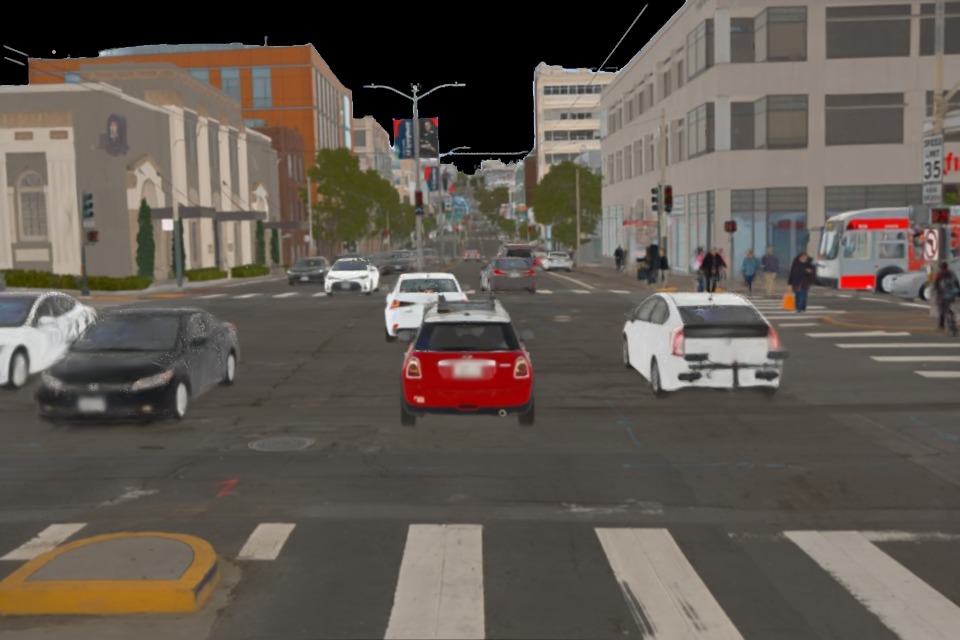}}& \frame{\includegraphics[width=1.5in,height=1in]{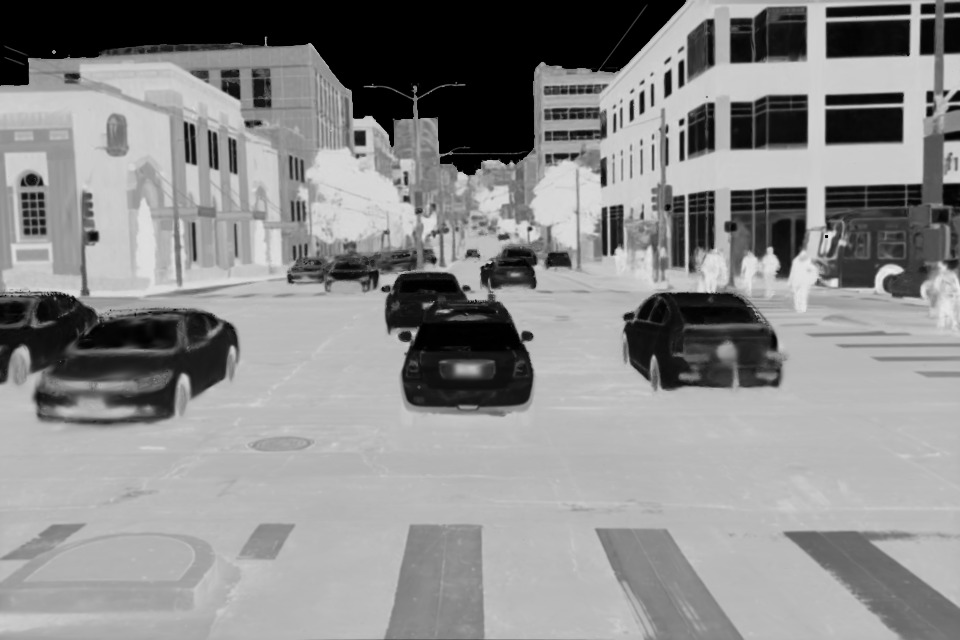}} & \frame{\includegraphics[width=1.5in,height=1in]{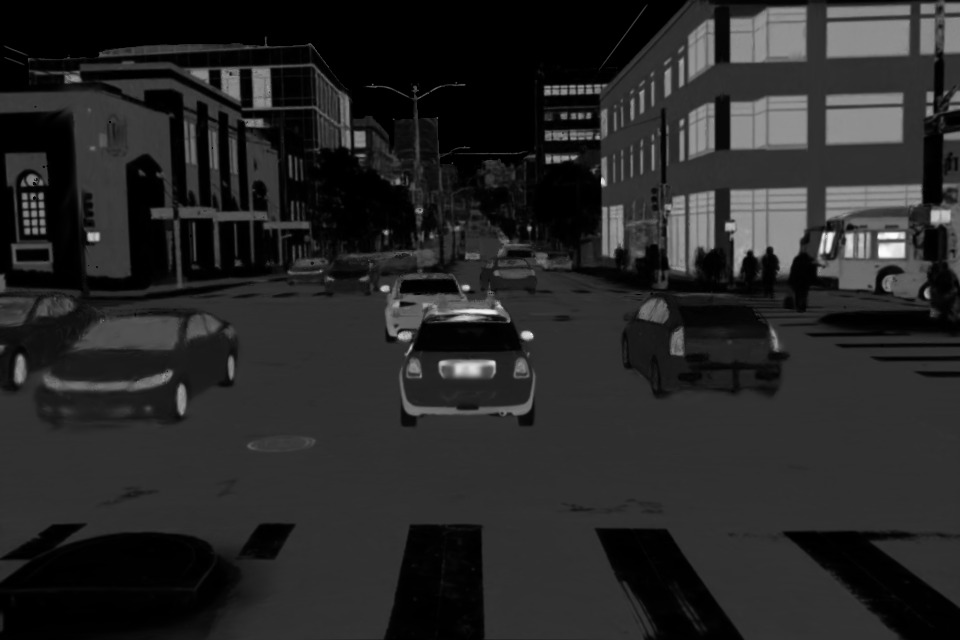}}\\
    \end{tabular}%
    }
    \caption{\textbf{Inverse rendering quality on WOD.}  
    The materials and normals predicted by InvRGB+L~\cite{invrgbl} contain artifacts and lack details; InvRGB+L does not model metallic (N/A).
    Gen3C~\cite{Gen3C}+DR~\cite{diffusionrenderer} and AW4D~\cite{autoweather4d} predict high metallic values on parts of the road.
    BRDFusion recovers detailed materials and normals.
    }
    \label{fig:qual_inverse}
    \vspace{-5mm}
\end{figure*}

\vspace{-1.0em}
\paragraph{\textbf{Volume Rendering Stage.}}
This stage recovers geometry and materials. The Gaussian means are initialized from LiDAR points~\cite{chen2025omnire}, and all Gaussian attributes are optimized through the volume rendering pass with
{\small
\begin{equation}\label{eq:loss_volume}
    \loss_{\text{vol}} = \loss_{\text{rgb}} + \weight_{O}\loss_{O} + \weight_{D}\loss_{D} + \weight_{N}\loss_{N} + \weight_{A}\loss_{A} + \weight_{R}\loss_{R} + \weight_{M}\loss_{M}.
\end{equation}
}
The photometric loss $\loss_{\text{rgb}} = \| \bC - \Cgt \| + 1 - \text{SSIM}(\bC, \Cgt)$ supervises appearance.
The opacity term $\loss_{O} = \mathrm{BCE}(\bO, \bS_{\text{non-sky}})$ aligns the opacity with the non-sky mask from SegFormer~\cite{segformer} by binary cross-entropy, suppressing sky floaters that would cast spurious shadows.
The remaining terms regularize geometry and materials toward the G-buffers $\{\bN_g, \bD_g, \bA_g, \bR_g, \bM_g\}$ that the DR inverse model $\diffusionModelI$ predicts from the input frames, with the same sliding-window averaging as in forward rendering; these generative priors are essential to reduce the ambiguity of inverse rendering (Sec.~\ref{sec:ablation}):
$\loss_{D} = \|\bD - \bD_{\text{lidar}}\| + \|(u\bD + v) - \bD_g\|$, where the first term is evaluated at pixels with LiDAR points and $u, v$ align scale and shift~\cite{MonoSDF};
$\loss_{N} = \|\bN - \bN_g\| - \bN \cdot \bN_g$;
and $\loss_{X} = \|\bX - \bX_g\|$ for $\bX \in \{\bA, \bR, \bM\}$.

\vspace{-1.0em}
\paragraph{\textbf{Generative Refinement Stage.}}

Sliding-window averaging reduces, but does not eliminate, the temporal inconsistency of the material priors $\bX_g$, $\bX \in \{\bA, \bR, \bM\}$.
This inconsistency feeds contradictory supervision into Eq.~\ref{eq:loss_volume} and produces material artifacts in the reconstruction.
BRDFusion resolves it with the same SDEdit operator, VAE residual, and sliding-window averaging as in forward rendering (Eqs.~\ref{eq:sdedit} and~\ref{eq:vae_residual}), now applied to \emph{material maps at training time} instead of PBR-rendered RGB at inference time.
After the volume rendering stage converges, each material map $\bX$ is rendered from the 3D reconstruction. These maps are temporally consistent but contain artifacts from the inconsistent priors. They serve as anchors for the DR \emph{inverse} model $\diffusionModelI$, conditioned on the input frames:
\begin{equation}\label{eq:gen_refine}
\begin{split}
    \bX_g^{'1:F} &= \bX^{1:F} + (\bX_{\text{gen}}^{1:F} - \dec(\enc(\bX^{1:F}))), \\
    \bX_{\text{gen}}^{1:F} &= \sdedit(\bX^{1:F} \mid \diffusionModelI, \enc(\Cgt^{1:F}), \diffusionTime).
\end{split}
\end{equation}
The refined maps $\bX_g^{'1:F}$ have fewer artifacts than the rendered maps and are more temporally consistent than the original priors. On our synthetic scenes, refinement reduces the temporal warping error of all three material priors (supplementary material). The refined maps replace the priors in Eq.~\ref{eq:loss_volume}, and the volume rendering stage runs once more, which reduces material artifacts in the reconstruction.
Note that SRUG~\cite{srug} also refines rendered material maps via SDEdit, but to reduce the synthetic-to-real gap of its priors rather than their temporal inconsistency.

\vspace{-1.0em}
\paragraph{\textbf{Physically-Based Inverse Rendering Stage.}}
With geometry and materials frozen, this stage optimizes the environment map by rendering the scene with the PBR pass (Eq.~\ref{eq:pbr}) and minimizing the photometric loss
{\small
\begin{equation}\label{eq:loss_pbr}
    \loss_{\text{pbr}} = \|\Cldr - \Cgt\|.
\end{equation}
}
Freezing geometry and materials reduces ambiguities such as baking shadows into the albedo.

\vspace{-1.0em}
\paragraph{\textbf{Joint Refinement Stage.}}
Finally, we jointly optimize geometry, material, and lighting with loss $\loss_{\text{vol}} + \lambda_{\text{pbr}}\loss_{\text{pbr}}$ to further improve the overall quality. With each property well initialized by the previous stages, this optimization avoids the ambiguities of optimizing all properties from scratch.
\vspace{-0.5em}
\section{Experiments}
\label{sec:experiments}
\vspace{-0.5em}

\begin{table}[t]
    \centering\setlength{\tabcolsep}{4pt}
    \caption{
        \textbf{Quantitative comparison on synthetic data.} 
        The best and second-best results are bold and underlined, respectively. Gen3C+DR and Gen3C+UR share the novel views generated by Gen3C, hence the identical novel-view synthesis metrics.
    }
    \vspace{-2mm}
    \label{tab:quant_synthetic}
    \resizebox{1.0\linewidth}{!}{%
    \begin{tabular}{lcccccccccc}
    \toprule
        & \multicolumn{4}{c}{Inverse Rendering}
        & \multicolumn{3}{c}{Novel-View Synthesis}
        & \multicolumn{3}{c}{Novel-View Relighting} \\  \cmidrule(lr){2-5} \cmidrule(lr){6-8} \cmidrule(lr){9-11}
        Method
        & \makecell[c]{Albedo \\ (si-PSNR$\uparrow$) }
        & \makecell[c]{Roughness \\ (RMSE$\downarrow$) }
        & \makecell[c]{Metallic \\ (RMSE$\downarrow$) }
        & \makecell[c]{Normal \\ (MAE ($^\circ$)$\downarrow$) }
        & PSNR$\uparrow$ & SSIM$\uparrow$ & LPIPS$\downarrow$
        & PSNR$\uparrow$ & SSIM$\uparrow$ & LPIPS$\downarrow$ \\
    \midrule
    IRGS~\cite{IRGS} & 14.01 & 0.451 & 0.223 & 19.85 & 22.66 & 0.792 & 0.389 &
    14.16 & 0.639 & 0.600 \\

    UrbanIR~\cite{urbanir} & 14.60 & - & - & 12.90 & 23.51 & \textbf{0.867} & \underline{0.215} &
    14.38 & \underline{0.735} & 0.456 \\

    InvRGB+L~\cite{invrgbl} & 16.37 & 0.302 & - & 11.62 & \underline{24.55} & \underline{0.828} & 0.320 &
    14.87 & 0.722 & 0.450 \\

    Gen3C~\cite{Gen3C}+DR~\cite{diffusionrenderer} & 17.70 & 0.220 & 0.248 & 16.30 & 22.49 & 0.781 & 0.323 & 16.85 & 0.729 & 0.462 \\

    Gen3C~\cite{Gen3C}+UR~\cite{unirelight} & 18.02 & - & - & - & 22.49 & 0.781 & 0.323 & 16.33 & 0.727 & 0.440 \\

    AutoWeather4D~\cite{autoweather4d} & \underline{19.23} & \underline{0.217} & \underline{0.176} & \underline{10.80} & 22.14 & 0.714 & \textbf{0.198} & \underline{17.91} & 0.732 & \underline{0.409} \\

    Ours & \textbf{19.33} & \textbf{0.187} & \textbf{0.144} & \textbf{8.22} & \textbf{26.40} & 0.799 & 0.250 & \textbf{20.71} & \textbf{0.775} & \textbf{0.325} \\
    \bottomrule
    \end{tabular}%
    }
    \vspace{-4mm}
\end{table}

\subsection{Evaluation Settings}
\vspace{-0.5em}
\paragraph{\textbf{Datasets.}}
We evaluate our method on real-world and synthetic datasets. For real-world evaluation, we use six scenes from the Waymo Open Dataset (WOD), four for baseline comparisons and two for applications.
From each scene, we select 51 consecutive single-camera frames, reserving every tenth frame for testing and the rest for training. This length fits within the native 57-frame setting of DR, ensuring a fair comparison with generative baselines. 

Because obtaining ground-truth materials and relighting videos is difficult, we simulate six synthetic urban scenes. Specifically, we combine city assets~\cite{blenderkit,Rocketbox} with HDR environment maps~\cite{PolyHaven2024} to build dynamic cities and render RGB images and PBR material maps using Blender Cycles~\cite{blender}. The cities contain moving vehicles, pedestrians, buildings, and traffic lights. To evaluate under diverse lighting conditions (cloudy, misty, and sunset), we render each scene with four environment maps, one for training and three for relighting evaluation. For each scene, we render a 100-frame sequence, from which we select 51 consecutive frames and hold out every tenth frame for testing.
Note that we do not use LightCity~\cite{LightCity}, as its released data lacks ground-truth albedo and the image-to-environment-map mapping.

\vspace{-1.0em}
\paragraph{\textbf{Baselines.}}
We compare our method against physically-based inverse rendering methods IRGS~\cite{IRGS}, UrbanIR~\cite{urbanir}, and InvRGB+L~\cite{invrgbl}, generative methods DR~\cite{diffusionrenderer} and UniRelight (UR)~\cite{unirelight}, and the hybrid method AW4D~\cite{autoweather4d}. 
AW4D reconstructs 4D scenes with $\pi^3$~\cite{pi3}, relights driving videos with the DR forward model, shades local lights using PBR and blends them with the DR forward result, and refines the results via SDEdit with the Canny-edge-conditioned Wan~\cite{wan} video model to improve realism. 
To ensure a fair comparison, we convert the environment map into the representations required by UrbanIR and InvRGB+L. 
Since DR and UR do not support novel-view synthesis, we first synthesize novel views with Gen3C~\cite{Gen3C} and then relight them, denoted as Gen3C+DR and Gen3C+UR.
AW4D also does not natively support novel-view synthesis, so we warp neighboring input frames to the test views using the $\pi^3$ depth.
We do not compare with FEGR~\cite{FEGR}, LightSim~\cite{LightSim}, LightCrafter~\cite{LightCrafter}, SRUG~\cite{srug}, or TRON~\cite{TRON}, as their code has not been publicly released.
The supplementary material details the environment map conversion and novel-view warping and shows qualitative comparisons with IRGS, UrbanIR, and Gen3C+UR.

\begin{figure*}[t]
\centering
\captionsetup[subfigure]{position=bottom}
\setlength{\tabcolsep}{2pt}
\resizebox{1.0\textwidth}{!}{%
\begin{tabular}{ccccc}

\subcaptionbox{Novel view}{%
\frame{\includegraphics[width=1.5in,height=1in]{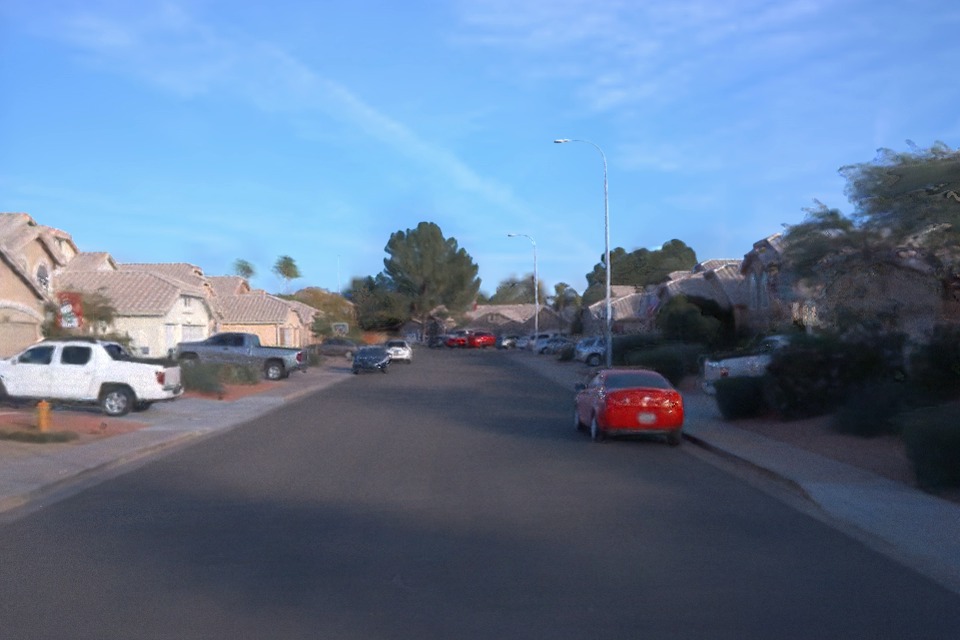}}} &
\subcaptionbox{Daylight}{%
\frame{\includegraphics[width=1.5in,height=1in]{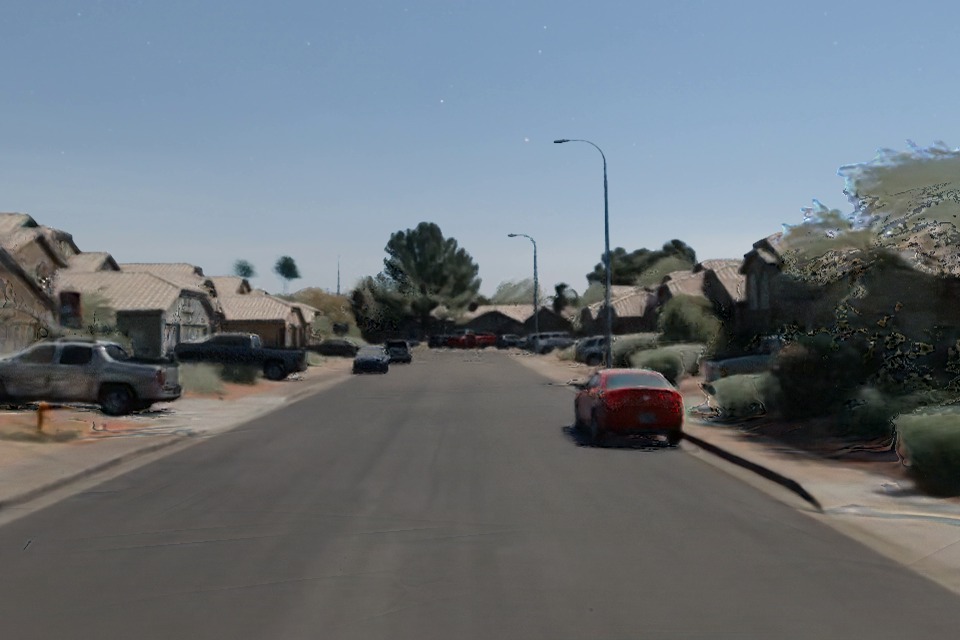}}} &
\subcaptionbox{Sunset}{%
\frame{\includegraphics[width=1.5in,height=1in]{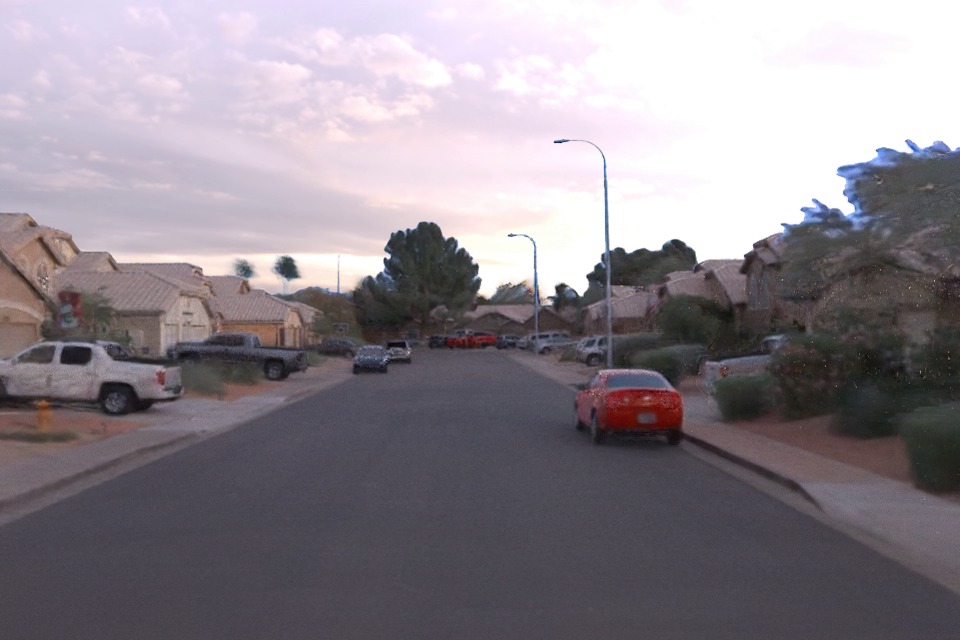}}} &
\subcaptionbox{Sunset + inserted car}{%
\frame{\includegraphics[width=1.5in,height=1in]{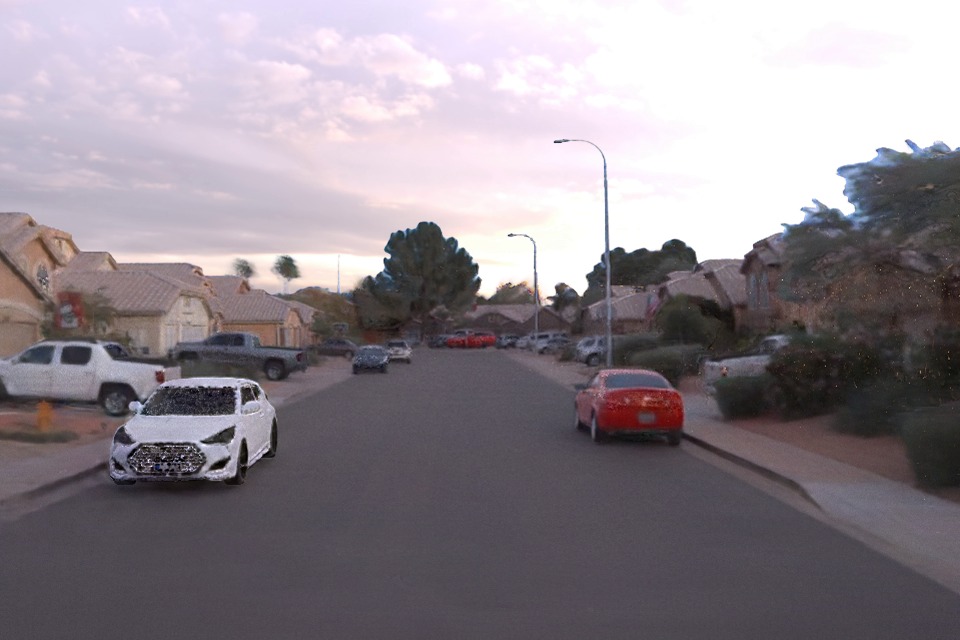}}} &
\subcaptionbox{Night + inserted car}{%
\frame{\includegraphics[width=1.5in,height=1in]{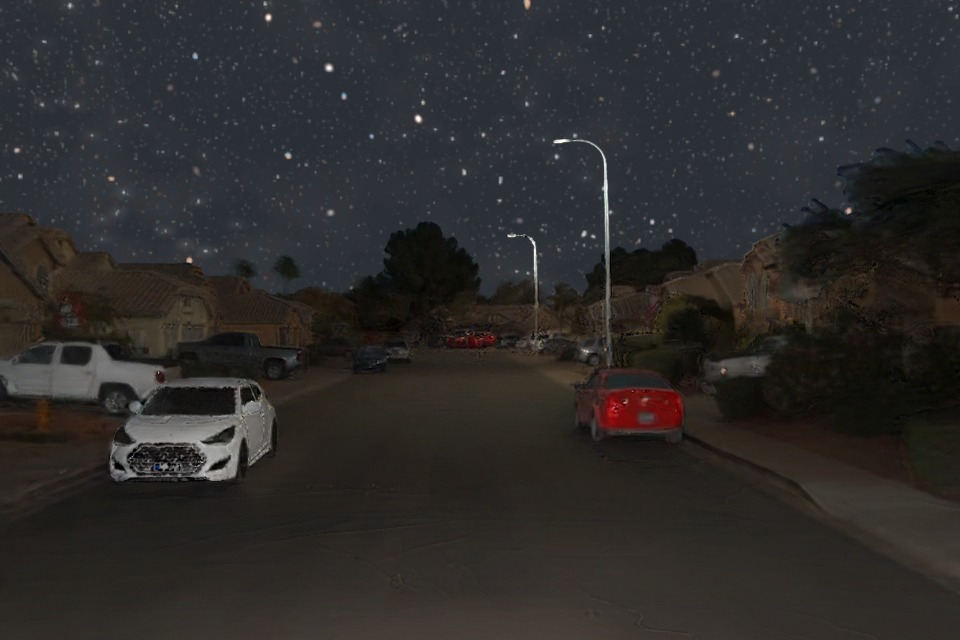}}}

\end{tabular}}
\caption{\textbf{Applications on WOD.} (a) Novel view. (b) Daylight and (c) sunset relighting. (d) An inserted car under the lighting of (c). (e) Night relighting with streetlights and headlights, including the inserted car.
}
\label{fig:qual_applications}
\vspace{-5mm}
\end{figure*}

\begin{table}[t]
    \centering\setlength{\tabcolsep}{4pt}
    \caption{
        \textbf{Ablation study on our synthetic dataset.} We use 100-frame sequences (51 frames in Tab.~\ref{tab:quant_synthetic}), so the numbers of the full method differ from Tab.~\ref{tab:quant_synthetic}. The forward-rendering variants reuse the reconstruction of the full method, so their inverse rendering metrics are identical (gray). The best and second-best results are bold and underlined, respectively. 
    }
    \vspace{-2mm}
    \label{tab:quant_ablation}
    \resizebox{1.0\linewidth}{!}{%
    \begin{tabular}{lccccccc}
    \toprule
        & \multicolumn{4}{c}{Inverse Rendering}
        & \multicolumn{3}{c}{Novel-View Relighting} \\  \cmidrule(lr){2-5} \cmidrule(lr){6-8}
        Method
        & \makecell[c]{Albedo \\ (si-PSNR$\uparrow$) }
        & \makecell[c]{Roughness \\ (RMSE$\downarrow$) }
        & \makecell[c]{Metallic \\ (RMSE$\downarrow$) }
        & \makecell[c]{Normal \\ (MAE ($^\circ$)$\downarrow$) }
        & PSNR$\uparrow$ & SSIM$\uparrow$ & LPIPS$\downarrow$ \\
    \midrule
    \multicolumn{8}{l}{\textit{Forward rendering}} \\
    No Gen. Render & \textcolor{gray}{19.27} & \textcolor{gray}{0.171} & \textcolor{gray}{0.155} & \textcolor{gray}{8.72} & 18.54 & 0.751 & 0.345 \\
    No PBR Anchor & \textcolor{gray}{19.27} & \textcolor{gray}{0.171} & \textcolor{gray}{0.155} & \textcolor{gray}{8.72} & 18.74 & 0.773 & 0.460 \\
    No VAE Residual & \textcolor{gray}{19.27} & \textcolor{gray}{0.171} & \textcolor{gray}{0.155} & \textcolor{gray}{8.72} & \textbf{20.66} & \textbf{0.787} & 0.445 \\
    \midrule
    \multicolumn{8}{l}{\textit{Inverse rendering}} \\
    No Gen. Prior & 14.73 & 0.238 & 0.736 & 115.96 & 13.48 & 0.609 & 0.564 \\
    No Multi-stage & 18.03 & 0.179 & 0.161 & 11.90 & 18.89 & 0.736 & 0.421 \\
    No Gen. Refine & \textbf{19.37} & \underline{0.178} & \underline{0.157} & \textbf{8.71} & 20.30 & 0.770 & \underline{0.338} \\
    \midrule
    Ours (Full Method) & \underline{19.27} & \textbf{0.171} & \textbf{0.155} & \underline{8.72} & \underline{20.60} & \underline{0.774} & \textbf{0.327} \\
    \bottomrule
    \end{tabular}%
    }
    \vspace{-4mm}
\end{table}

 \vspace{-1.0em}
\paragraph{\textbf{Metrics.}}
For novel-view synthesis and relighting, we report PSNR, SSIM~\cite{ssim}, and LPIPS~\cite{lpips}.
For inverse rendering, roughness and metallic are evaluated using root-mean-square error (RMSE), and normals are evaluated using mean angular error (MAE). For albedo, we follow~\cite{diffusionrenderer} and compute PSNR after a per-channel least-squares scale alignment (si-PSNR). We compute all metrics over non-sky regions to avoid ambiguity in sky material and color.

\vspace{-1.0em}
\paragraph{\textbf{Implementation Details.}}
The optimization schedule, loss weights, ray-tracing settings, hardware, training time, and SDEdit noise levels $\diffusionTime$ are in the supplementary material.

\subsection{Qualitative Evaluation}
\vspace{-0.5em}
\paragraph{\textbf{Forward Rendering.}}

Fig.~\ref{fig:qual_relighting} compares novel-view synthesis and relighting on three WOD scenes. The night scene uses streetlights and car headlights placed in 3D.
InvRGB+L follows the target lighting, but artifacts of its reconstruction degrade the relit images.
Gen3C+DR renders realistic frames, but without an explicit 3D scene it does not follow the target lighting, especially the local lights.
AW4D simulates the target lighting, but its refinement, conditioned only on Canny edges, alters scene content, and its per-pixel blending of PBR local lights with DR global lighting leaves moderately lit regions (e.g., the road under the headlights) dark.
\method renders both global and local lighting with PBR in one 3D scene and removes artifacts with a generative model anchored on the PBR result, 
so it preserves the scene content and keeps local and global lighting consistent.

\vspace{-1.0em}
\paragraph{\textbf{Inverse Rendering.}}

Fig.~\ref{fig:qual_inverse} compares inverse rendering results in novel views of a dynamic WOD scene.
InvRGB+L predicts incorrect materials (e.g., white cars with gray albedo) and produces overly smooth G-buffers without scene details.
Although Gen3C+DR and AW4D produce highly detailed outputs, their inherent randomness cannot guarantee physically plausible and consistent predictions.
In contrast, \method integrates generative modeling into physically-based optimization, yielding detailed, physically plausible, and temporally consistent geometry and material.

\subsection{Quantitative Evaluation}
\vspace{-0.5em}
Tab.~\ref{tab:quant_synthetic} reports the comparison on the synthetic dataset.
\method has the best results on all material and geometry metrics and the best relighting on all metrics, improving PSNR by 2.8 dB over AW4D, as relighting benefits from both the accurate decomposition and the light controllability.
For novel-view synthesis, \method has the highest PSNR but not the best SSIM and LPIPS. Its novel views pass through the same PBR and generative passes as relighting, and we attribute the gap mainly to residual Monte Carlo noise from the limited number of rays per pixel. AW4D attains the best LPIPS because its novel views are warped from neighboring input frames. Its lower PSNR comes mainly from its Wan SDEdit refinement, which alters scene content.

\subsection{Ablation Study}
\label{sec:ablation}
\vspace{-0.5em}

We ablate our design choices on six synthetic 100-frame scenes, which require sliding-window averaging, holding out every tenth frame for testing (Tab.~\ref{tab:quant_ablation}). Additional ablations and qualitative results are in the supplementary material.

\vspace{-1.0em}
\paragraph{\textbf{Forward Rendering.}}
Removing the generative rendering pass (\textbf{No Gen. Render}) leaves reconstruction artifacts in the PBR result and lowers relighting PSNR by 2.1 dB.
Without PBR anchoring, generating from pure noise under the same G-buffer conditioning (\textbf{No PBR Anchor}) does not reproduce the target lighting effects.
Using the SDEdit output directly without the VAE residual (\textbf{No VAE Residual}) blurs the rendered video. Smoothing the Monte Carlo noise slightly raises PSNR and SSIM, but LPIPS increases from 0.327 to 0.445. The full pipeline gives the best trade-off for relighting.

\vspace{-1.0em}
\paragraph{\textbf{Inverse Rendering.}}
Without DR priors (\textbf{No Gen. Prior}), the highly ill-posed optimization bakes lighting effects such as shadows into geometry and material and fails to recover a meaningful decomposition. On average, its normals even point away from the true normals (115.96$^\circ$ MAE).
Replacing the multi-stage strategy with a single joint optimization of the same total number of iterations (\textbf{No Multi-stage}) also fails to estimate the scene properties correctly.
Skipping generative refinement (\textbf{No Gen. Refine}) keeps the albedo and normal errors on par with the full method but increases the roughness and metallic errors, because the temporally inconsistent material priors are baked into the materials as artifacts (supplementary material).
All three variants degrade relighting, e.g., PSNR drops by 0.3 dB for \textbf{No Gen. Refine}.

\subsection{Diverse Applications}
\vspace{-0.5em}
Fig.~\ref{fig:qual_applications} shows relighting under daylight, sunset, and night conditions (b, c, e), object insertion (d, e), and added streetlights and headlights (e). Inserted vehicles are textured meshes converted to Gaussians with their PBR materials (supplementary material), so they cast and receive shadows through the same PBR and generative passes as the scene.

\vspace{-0.5em}
\section{Conclusion}
\label{sec:conclusion}
\vspace{-0.5em}
BRDFusion couples physically-based and generative rendering in both directions. PBR renderings of a 3D Gaussian scene anchor a video diffusion model in forward rendering, and the same anchoring turns per-clip generative priors into consistent supervision in inverse rendering. This coupling gives the best novel-view relighting on our synthetic benchmark and supports night simulation and object insertion on real driving videos.
Future work should address its limitations. BRDFusion does not model emissive materials, so it cannot yet decompose night videos with active lights. The PBR pass models only direct illumination and ignores interreflection. Monte Carlo ray tracing dominates the rendering time, 
and its noise lowers the perceptual quality of novel views. Floaters or failures of the generative priors can corrupt the decomposition or cast spurious shadows (supplementary material).

\section*{Acknowledgements}
This research was funded by the National Science and Technology Council, Taiwan, under Grants NSTC 112-2222-E-A49-004-MY2 and 113-2628-E-A49-023-. The authors are grateful to Google, NVIDIA, and MediaTek Inc. for their generous donations. Yu-Lun Liu acknowledges the Yushan Young Fellow Program by the MOE in Taiwan.

\section*{LLM Usage Statement}
We used OpenAI Codex, Anthropic Claude, and Claude Code to edit the wording, grammar, and clarity of all sections of this paper (abstract, Secs.~1--5, supplementary material, and the figure and table captions). We also used Claude Code to help write and debug code and to run and monitor experiments. The authors designed the method and experiments and verified all code and results.\par
{%
    \small
    \bibliographystyle{ieeenat_fullname}
    \bibliography{main}
}

\clearpage
\maketitlesupplementary
\appendix

This supplementary material provides additional details and experimental results. First, we describe our synthetic dataset in Sec.~\ref{supp:dataset}. We then present additional results on synthetic and real-world scenes in Sec.~\ref{supp:result}. Next, we provide additional ablation studies on our design choices in Sec.~\ref{supp:ablation}. We further analyze stochastic stability with PBR anchoring in Sec.~\ref{supp:stochastic}. This analysis is followed by method-level pseudocode in Sec.~\ref{supp:algorithm}, and comprehensive implementation details, including optimization formulations, SDEdit settings, and local light rendering in Sec.~\ref{supp:impl}, followed by the baseline adaptations in Sec.~\ref{supp:baseline}. Next, we detail the Cook-Torrance BRDF utilized in our rendering pipeline in Sec.~\ref{supp:brdf}. We further report the training and rendering efficiency of our method in Sec.~\ref{supp:comp}. Finally, we discuss current limitations and failure cases in Sec.~\ref{supp:fail}.

\section{Synthetic Dataset Details}
\label{supp:dataset}
\begin{figure*}[t]
    \centering\setlength{\tabcolsep}{1pt}\renewcommand{\arraystretch}{0.6}
    \resizebox{1.0\textwidth}{!}{%
    \begin{tabular}{@{}lcccccc@{}}
     & Noon (training) & Partly cloudy & Misty & Sunset & GT albedo & GT normal \\[0.3em]
    \raisebox{0.5in}[0pt][0pt]{\rotatebox[origin=c]{90}{Scene 1}} &
    \includegraphics[width=1.5in]{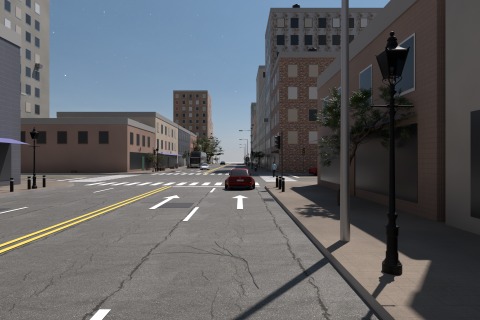} &
    \includegraphics[width=1.5in]{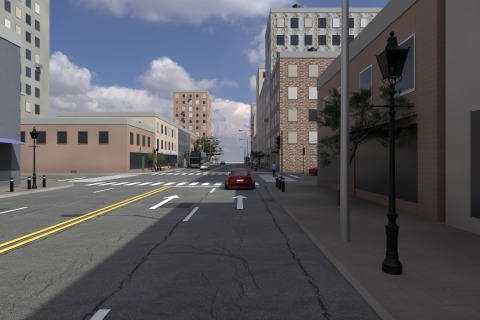} &
    \includegraphics[width=1.5in]{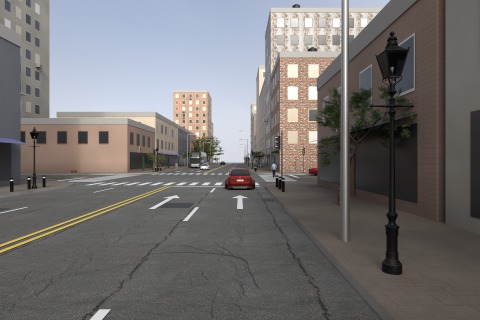} &
    \includegraphics[width=1.5in]{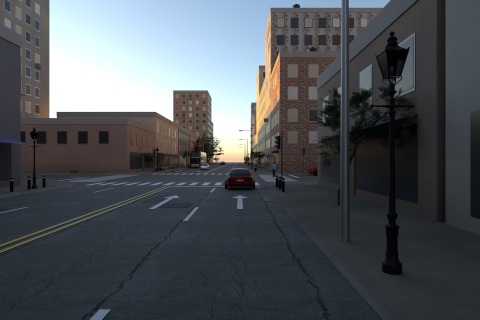} &
    \includegraphics[width=1.5in]{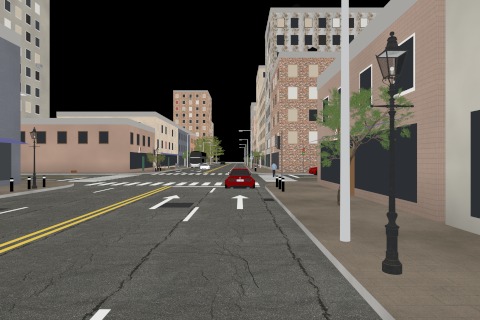} &
    \includegraphics[width=1.5in]{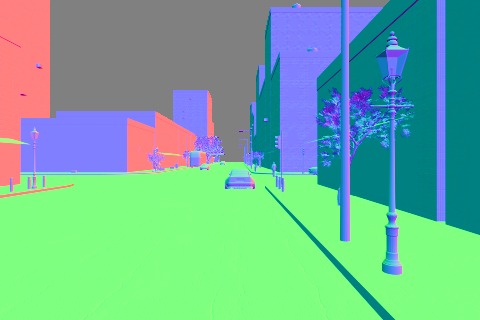} \\
    \raisebox{0.5in}[0pt][0pt]{\rotatebox[origin=c]{90}{Scene 2}} &
    \includegraphics[width=1.5in]{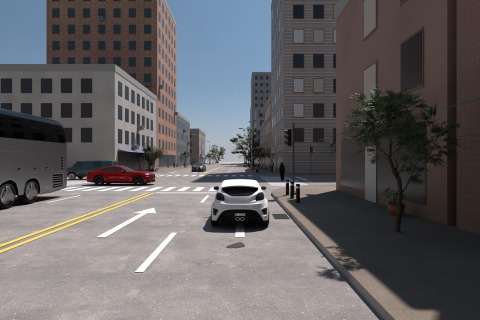} &
    \includegraphics[width=1.5in]{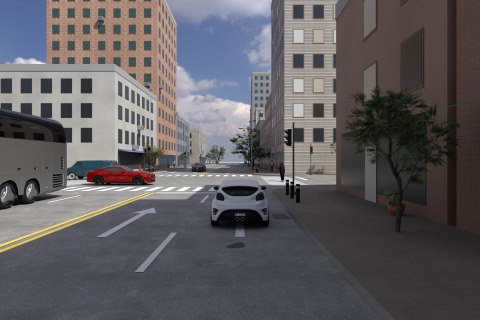} &
    \includegraphics[width=1.5in]{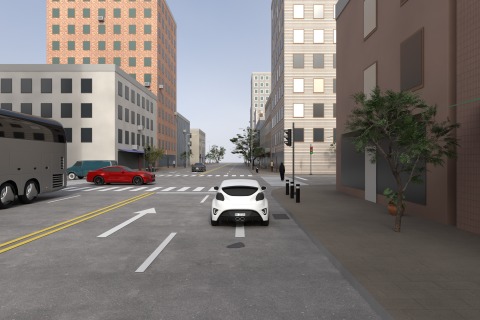} &
    \includegraphics[width=1.5in]{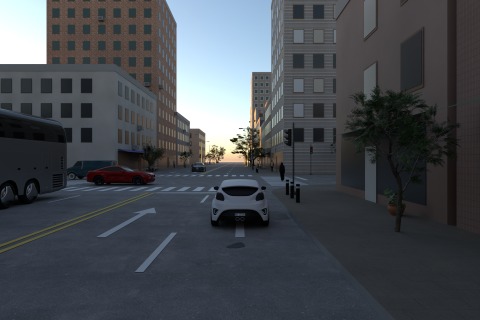} &
    \includegraphics[width=1.5in]{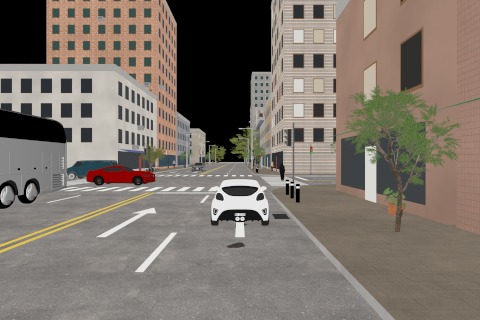} &
    \includegraphics[width=1.5in]{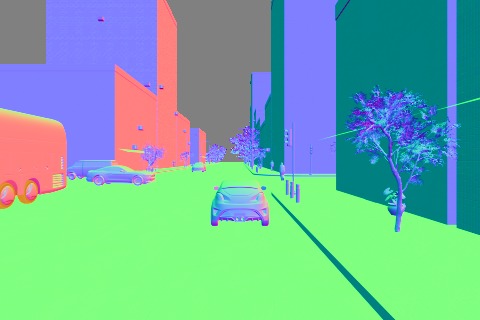} \\
    \raisebox{0.5in}[0pt][0pt]{\rotatebox[origin=c]{90}{Scene 3}} &
    \includegraphics[width=1.5in]{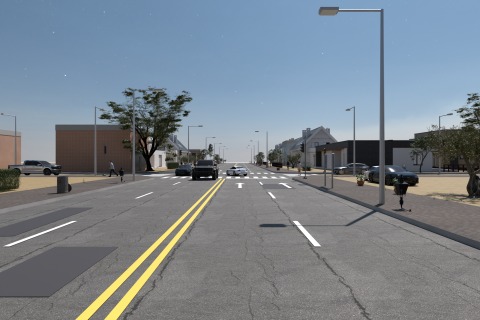} &
    \includegraphics[width=1.5in]{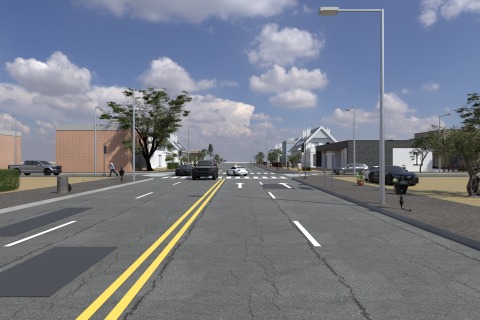} &
    \includegraphics[width=1.5in]{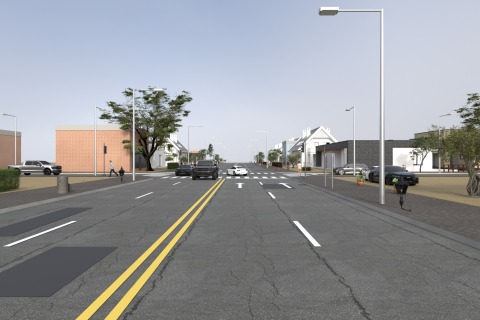} &
    \includegraphics[width=1.5in]{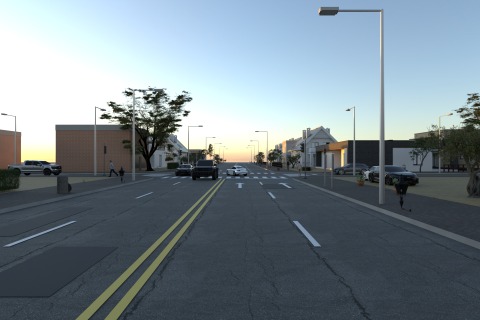} &
    \includegraphics[width=1.5in]{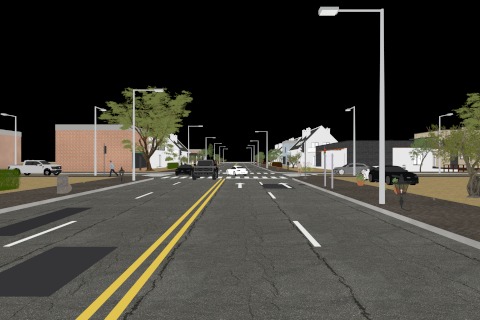} &
    \includegraphics[width=1.5in]{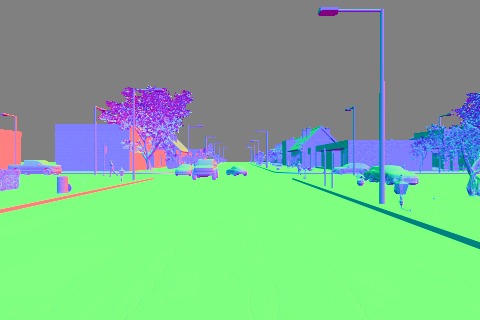} \\
    \raisebox{0.5in}[0pt][0pt]{\rotatebox[origin=c]{90}{Scene 4}} &
    \includegraphics[width=1.5in]{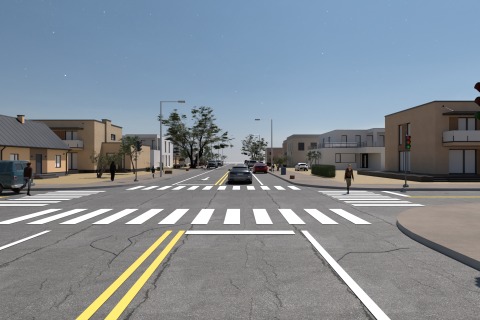} &
    \includegraphics[width=1.5in]{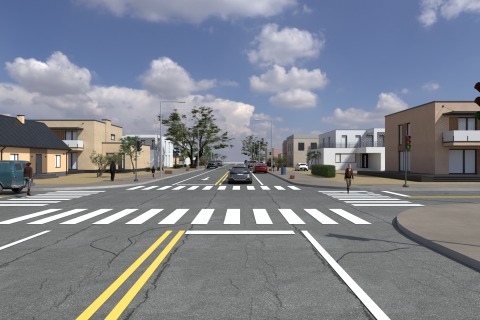} &
    \includegraphics[width=1.5in]{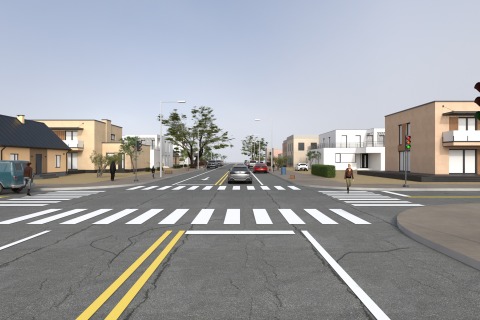} &
    \includegraphics[width=1.5in]{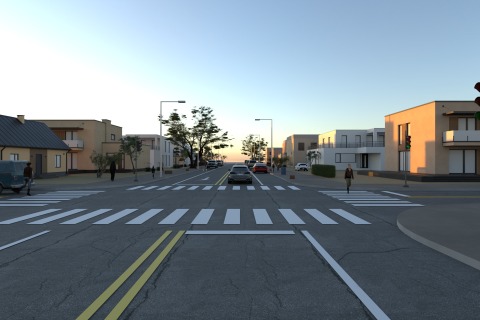} &
    \includegraphics[width=1.5in]{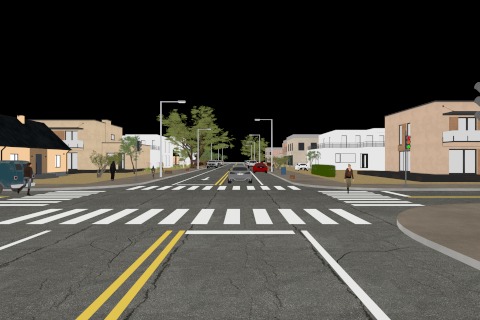} &
    \includegraphics[width=1.5in]{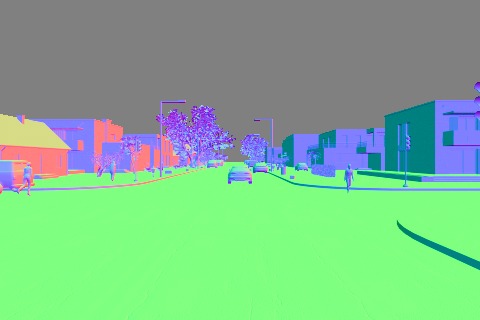} \\
    \raisebox{0.5in}[0pt][0pt]{\rotatebox[origin=c]{90}{Scene 5}} &
    \includegraphics[width=1.5in]{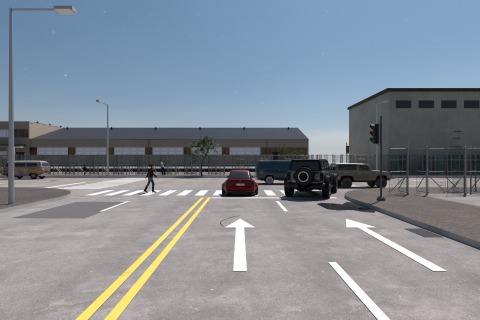} &
    \includegraphics[width=1.5in]{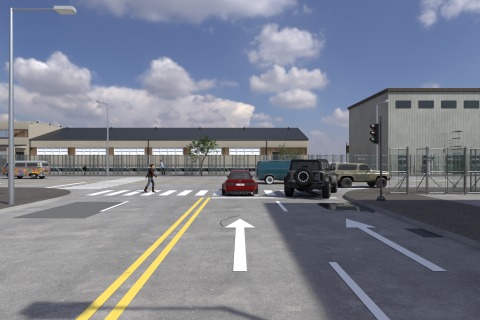} &
    \includegraphics[width=1.5in]{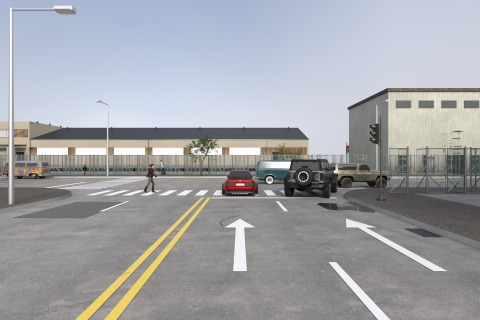} &
    \includegraphics[width=1.5in]{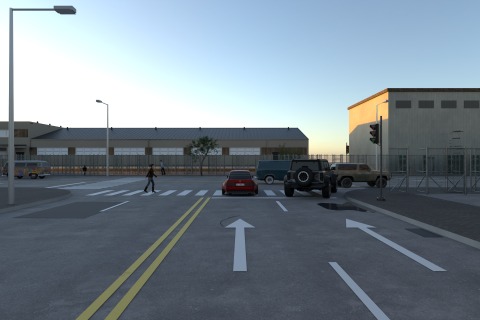} &
    \includegraphics[width=1.5in]{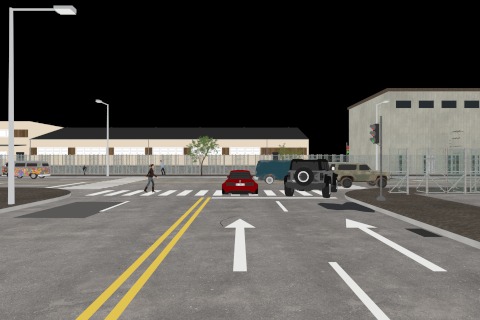} &
    \includegraphics[width=1.5in]{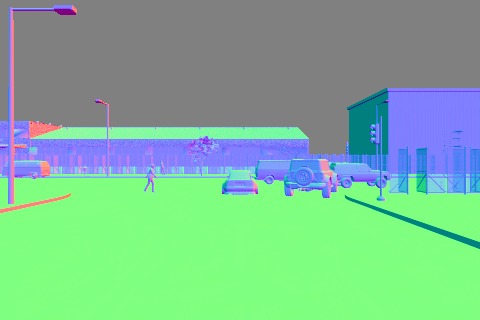} \\
    \raisebox{0.5in}[0pt][0pt]{\rotatebox[origin=c]{90}{Scene 6}} &
    \includegraphics[width=1.5in]{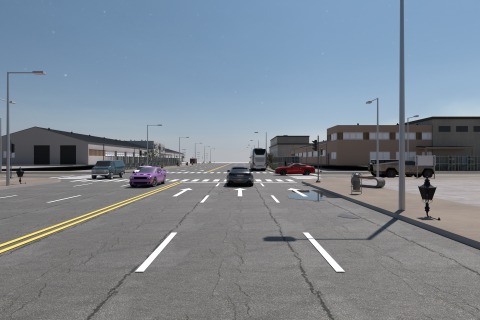} &
    \includegraphics[width=1.5in]{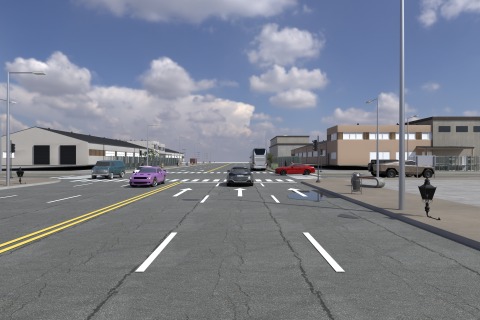} &
    \includegraphics[width=1.5in]{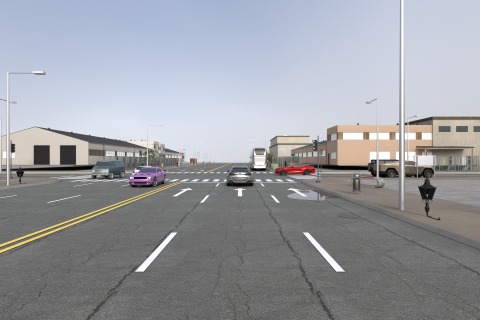} &
    \includegraphics[width=1.5in]{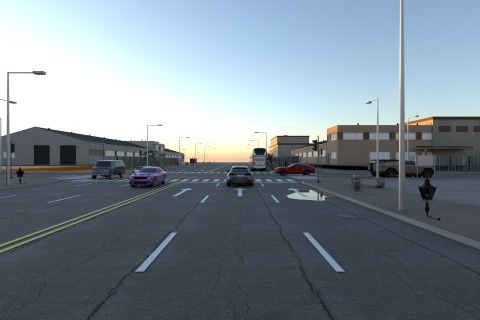} &
    \includegraphics[width=1.5in]{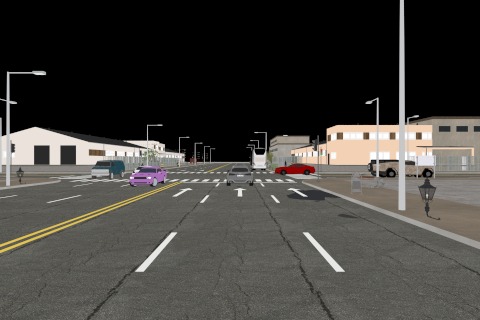} &
    \includegraphics[width=1.5in]{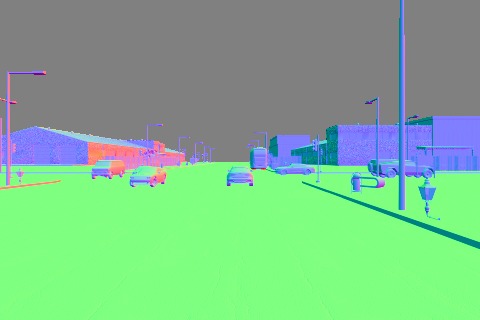} \\
    \end{tabular}%
    }
    \caption{\textbf{Overview of our synthetic dataset.} Each row shows one of the six dynamic scenes under the training environment map (noon) and the three relighting environment maps, together with the ground-truth albedo and normal. Ground-truth roughness, metallic, depth, sky masks, and dynamic-object masks are also provided.}
    \label{fig:dataset_overview}
\end{figure*}

We build six dynamic urban scenes in Blender~\cite{blender}, shown in Fig.~\ref{fig:dataset_overview}. The scenes cover different road layouts: a straight road, a right turn, a left turn, a narrow street with parked cars, a T-junction, and an avenue. They are assembled from vehicle, building, and street-furniture assets from BlenderKit~\cite{blenderkit}, road and facade textures, trees, and lamps from Poly Haven~\cite{PolyHaven2024}, and animated pedestrians from Microsoft Rocketbox~\cite{Rocketbox}. Each scene contains four to eight moving vehicles and five walking pedestrians, together with static buildings, parked cars, and traffic lights.

We render each scene with a forward-facing camera mounted on a moving ego vehicle for 100 frames at 10 fps, using Blender Cycles at $1920 \times 1280$ with a pinhole camera. All methods are trained and evaluated at $960 \times 640$. Each scene is rendered under four HDR environment maps from Poly Haven~\cite{PolyHaven2024}: a clear noon sky used for training, and a partly cloudy sky, a misty sky, and a low sun near sunset used for relighting evaluation. 

For every frame, we export the rendered RGB image under each lighting condition, together with the lighting-independent ground-truth albedo, roughness, metallic, surface normal, depth, sky mask, dynamic-object mask, and instance ID maps, as well as the camera poses and the 3D bounding-box tracks of the dynamic objects. As on the Waymo Open Dataset (WOD)~\cite{waymo_open_dataset}, the camera poses and object tracks are available to all methods. Since the synthetic scenes have no LiDAR, we use the rendered depth as pseudo-LiDAR for initialization and the depth loss.

\section{Additional Results}
\label{supp:result}

\begin{figure*}[t]
    \centering
    \setlength{\tabcolsep}{2pt}
    \resizebox{1.0\textwidth}{!}{%
    \begin{tabular}{@{}lcccccccc@{}}
    
    & IRGS~\cite{IRGS} & UrbanIR~\cite{urbanir} & InvRGB+L~\cite{invrgbl} & Gen3C~\cite{Gen3C}+DR~\cite{diffusionrenderer} & Gen3C~\cite{Gen3C}+UR~\cite{unirelight} & AW4D~\cite{autoweather4d} & Ours & GT \\[0.2em]

    \raisebox{2.5\normalbaselineskip}[0pt][0pt]{\rotatebox[origin=c]{90}{Albedo}} & 
    \frame{\includegraphics[width=1.2in]{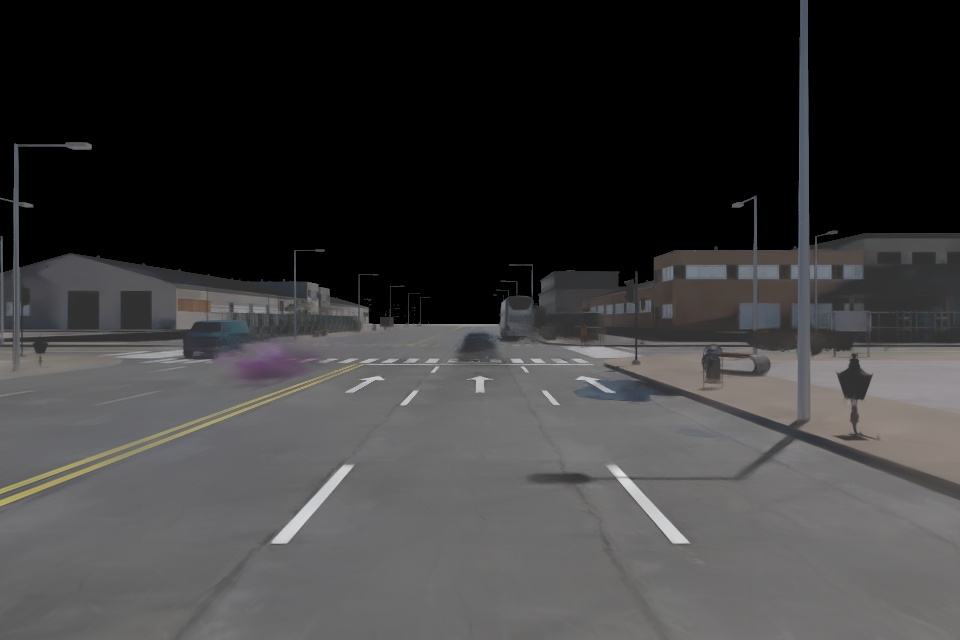}} &
    \frame{\includegraphics[width=1.2in]{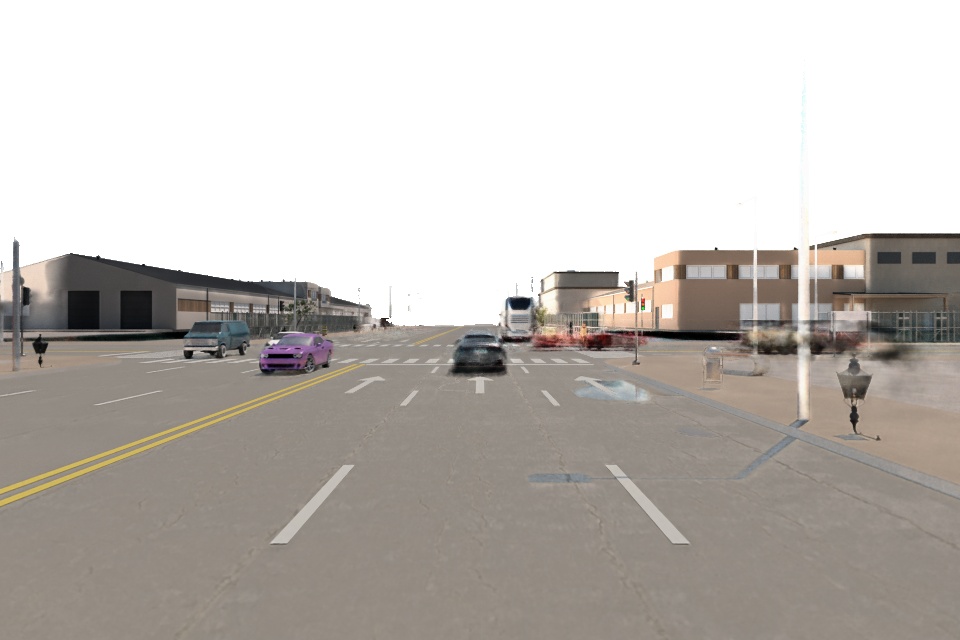}} & 
    \frame{\includegraphics[width=1.2in]{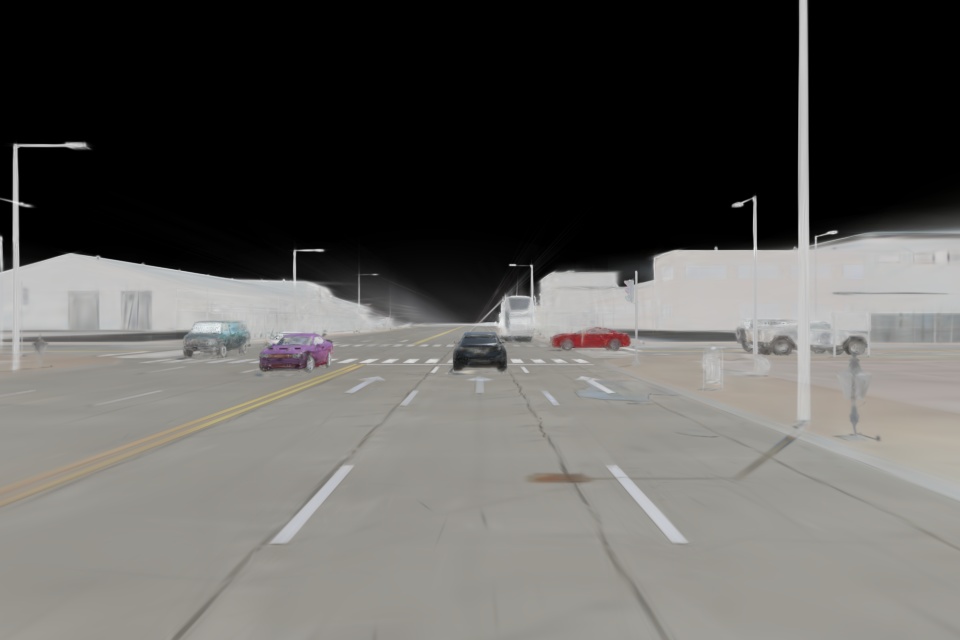}} & 
    \frame{\includegraphics[width=1.2in]{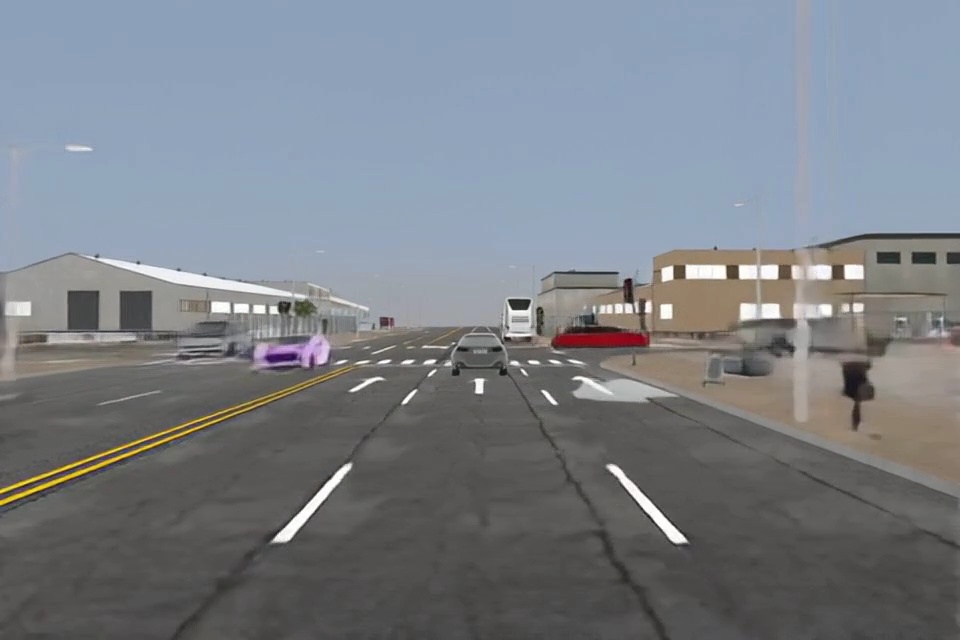}} & 
    \frame{\includegraphics[width=1.2in]{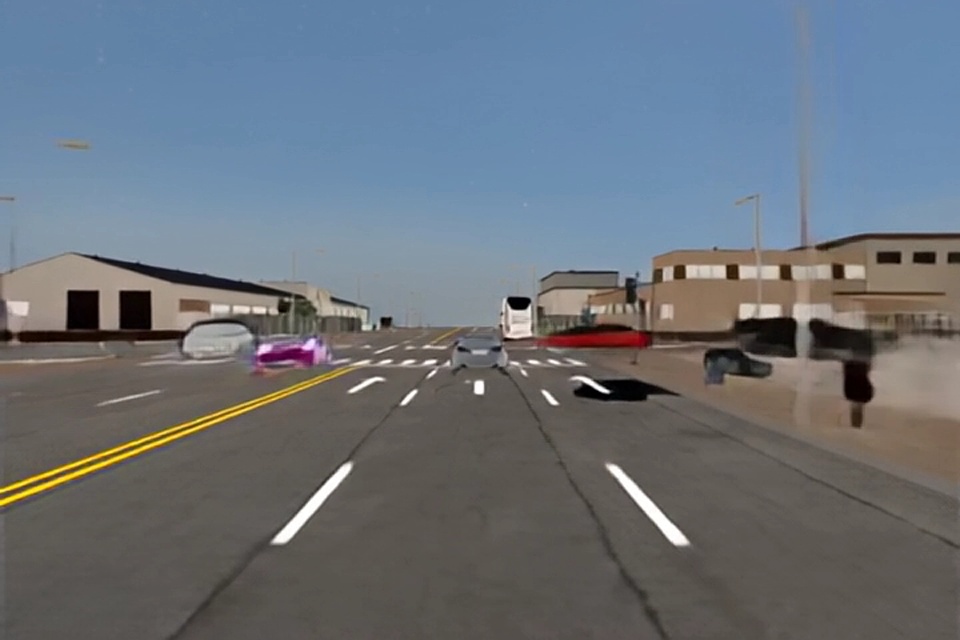}} &
    \frame{\includegraphics[width=1.2in]{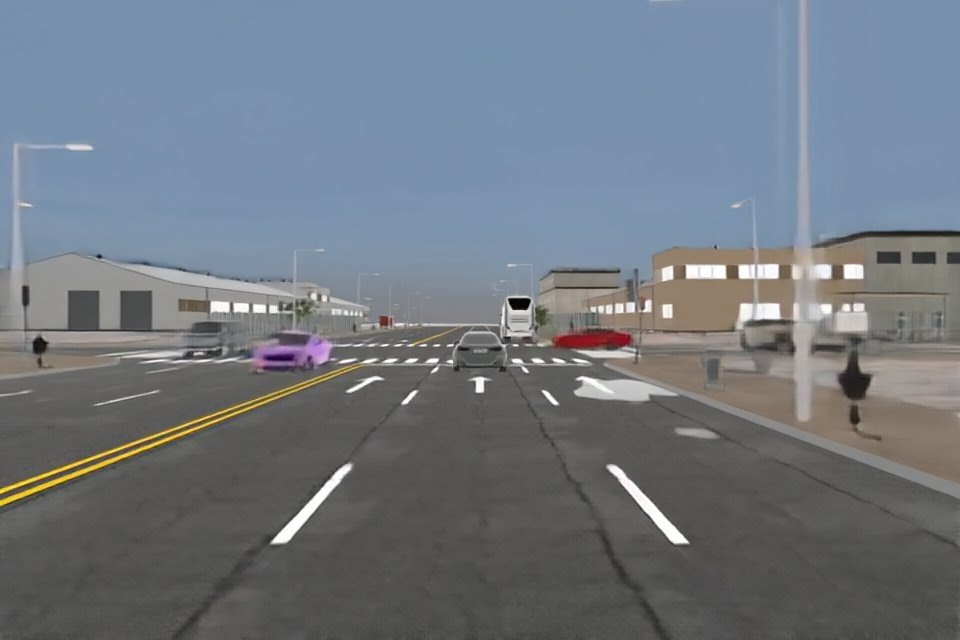}} &
    \frame{\includegraphics[width=1.2in]{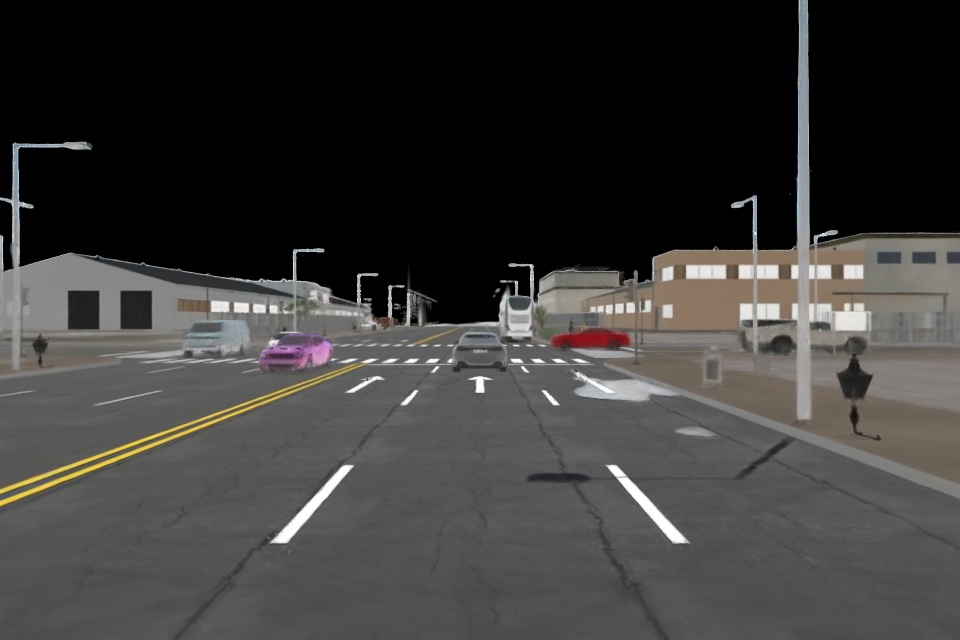}} & 
    \frame{\includegraphics[width=1.2in]{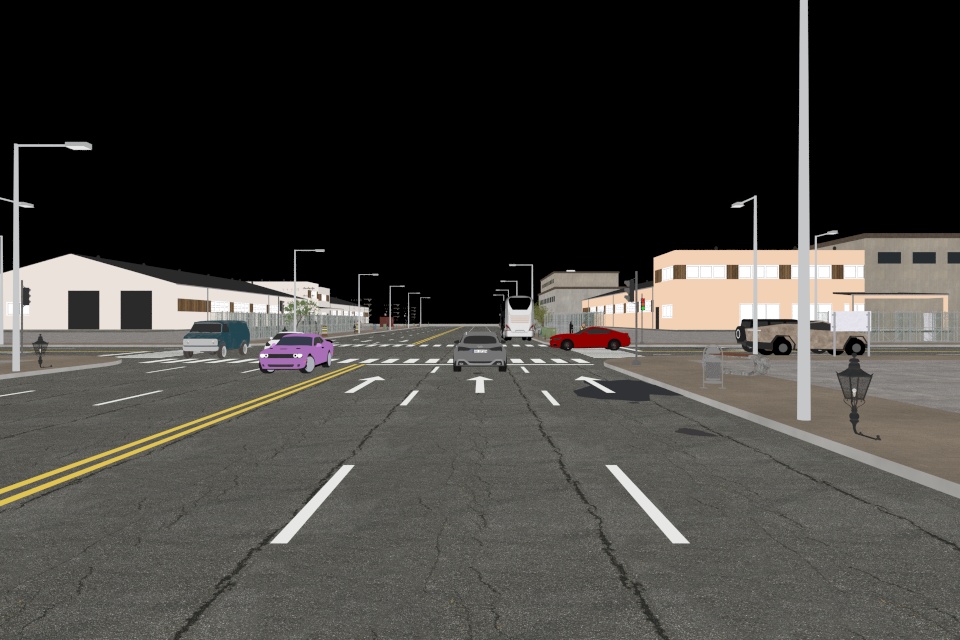}} \\

    \raisebox{2.5\normalbaselineskip}[0pt][0pt]{\rotatebox[origin=c]{90}{Roughness}} & 
    \frame{\includegraphics[width=1.2in]{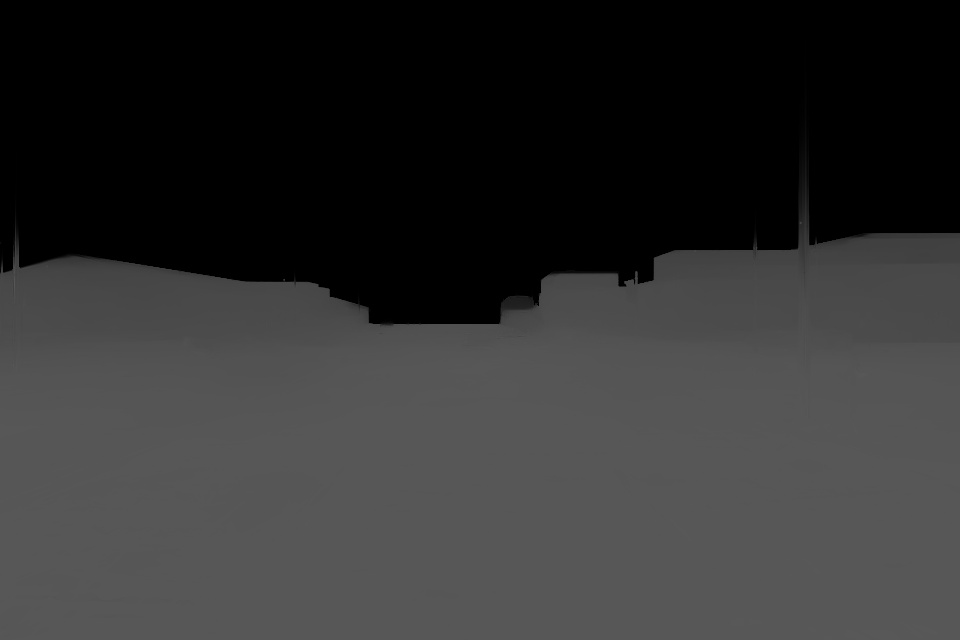}} &
    \frame{\includegraphics[width=1.2in]{figures/images/null.jpg}} & 
    \frame{\includegraphics[width=1.2in]{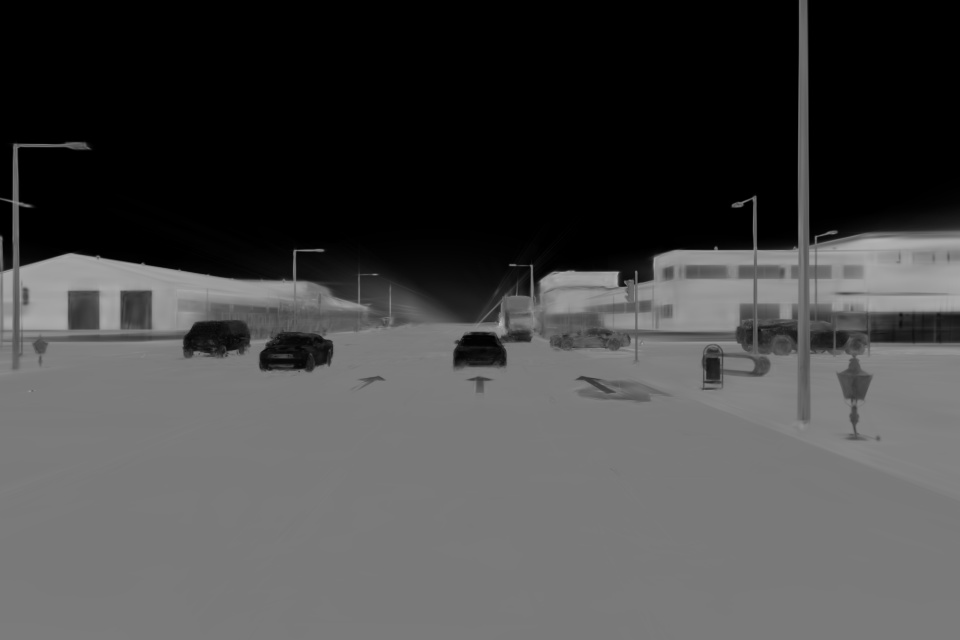}} & 
    \frame{\includegraphics[width=1.2in]{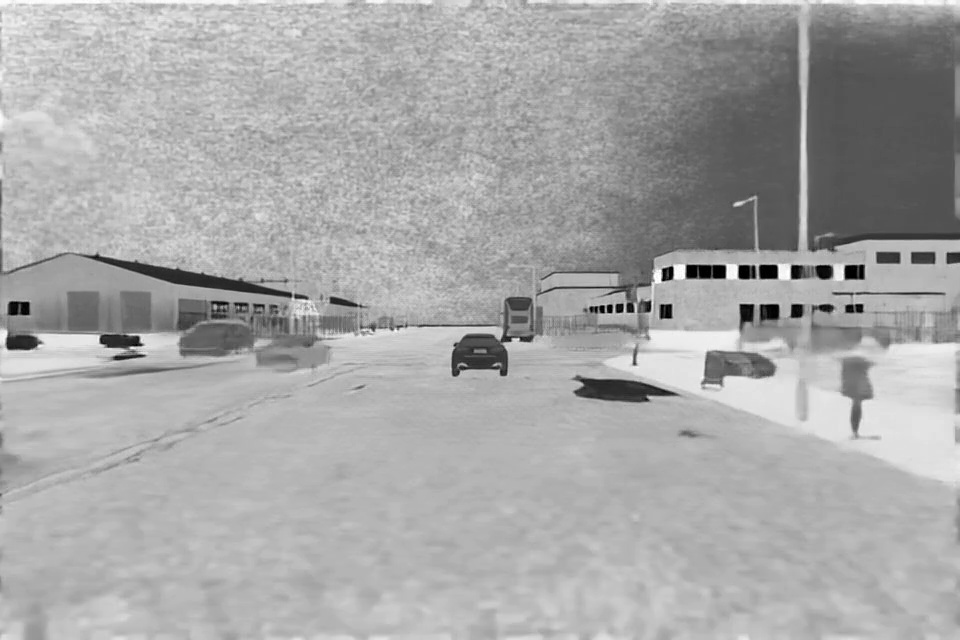}} & 
    \frame{\includegraphics[width=1.2in]{figures/images/null.jpg}} &
    \frame{\includegraphics[width=1.2in]{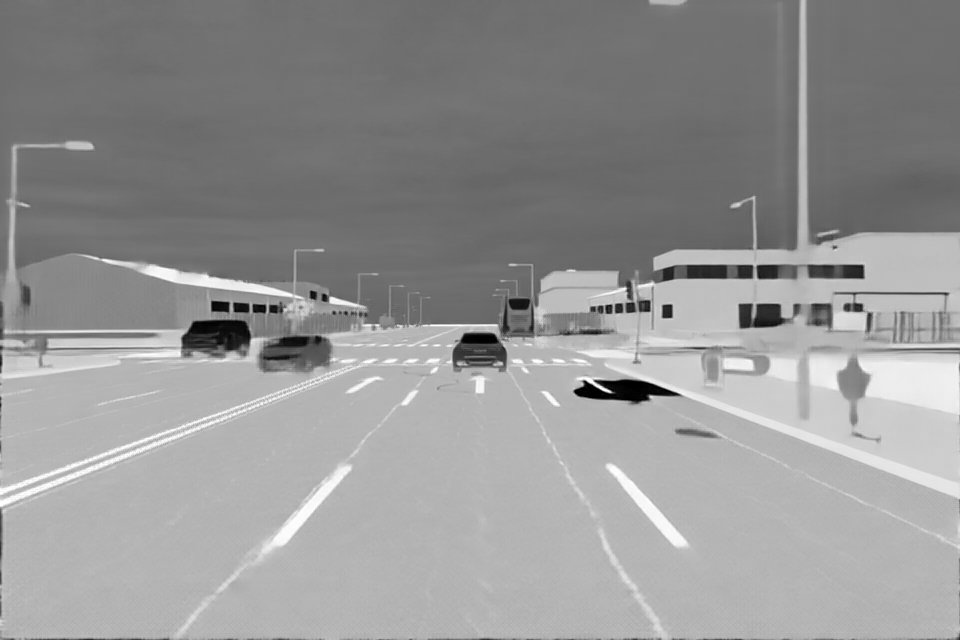}} &
    \frame{\includegraphics[width=1.2in]{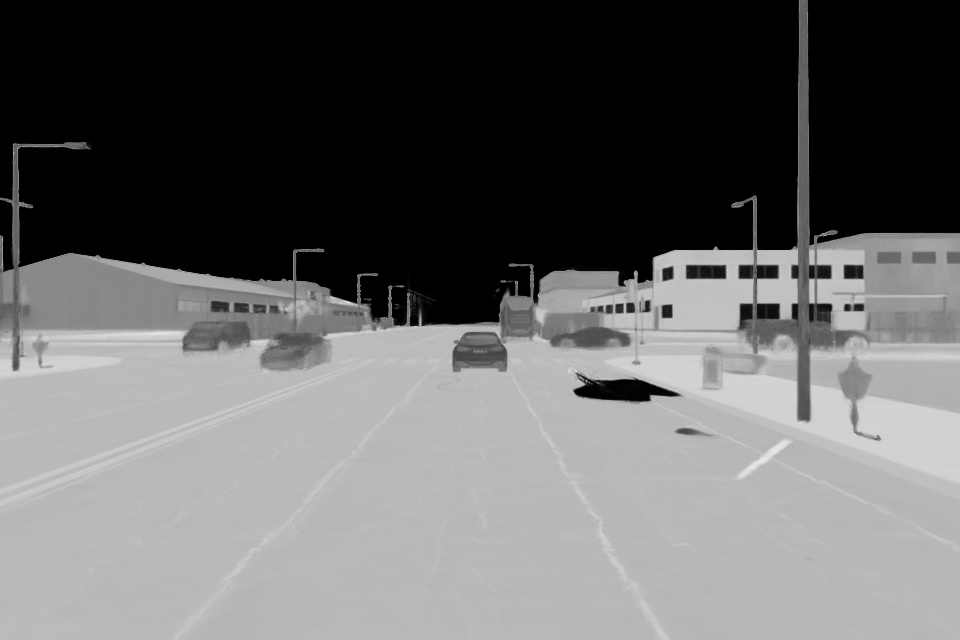}} & 
    \frame{\includegraphics[width=1.2in]{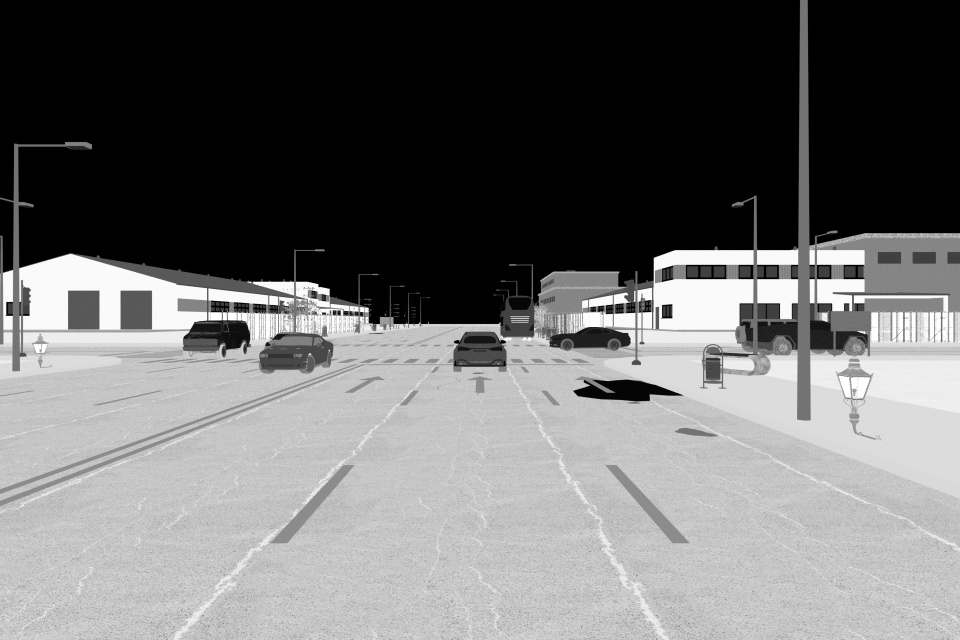}} \\

    \raisebox{2.5\normalbaselineskip}[0pt][0pt]{\rotatebox[origin=c]{90}{Metallic}} & 
    \frame{\includegraphics[width=1.2in]{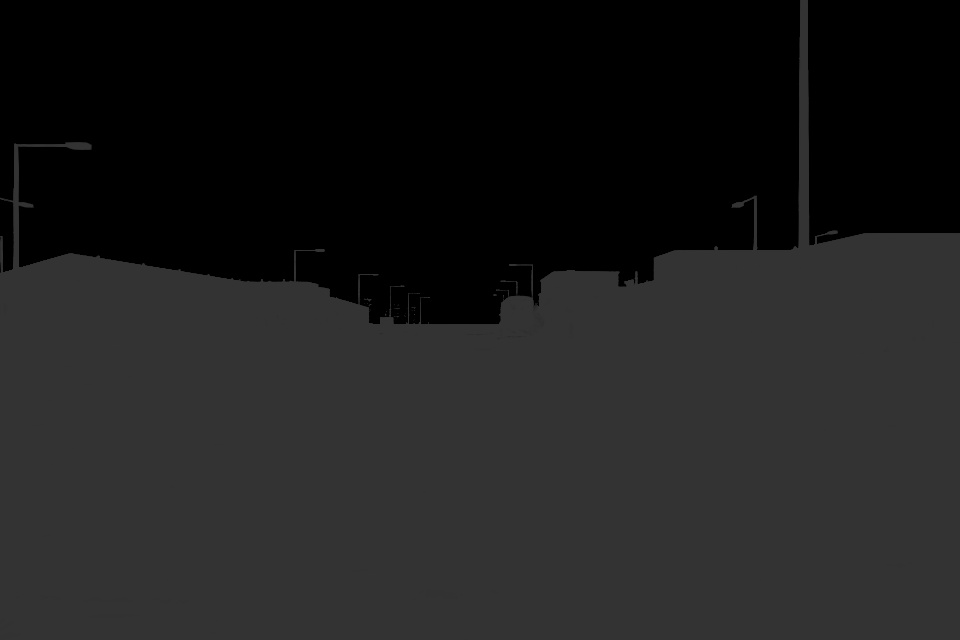}} &
    \frame{\includegraphics[width=1.2in]{figures/images/null.jpg}} & 
    \frame{\includegraphics[width=1.2in]{figures/images/null.jpg}} & 
    \frame{\includegraphics[width=1.2in]{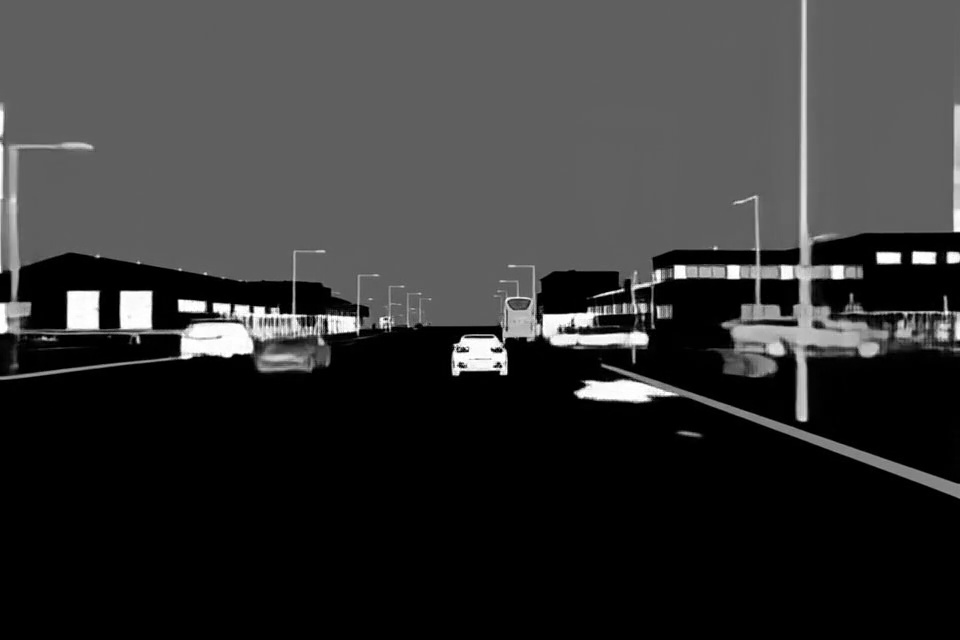}} & 
    \frame{\includegraphics[width=1.2in]{figures/images/null.jpg}} &
    \frame{\includegraphics[width=1.2in]{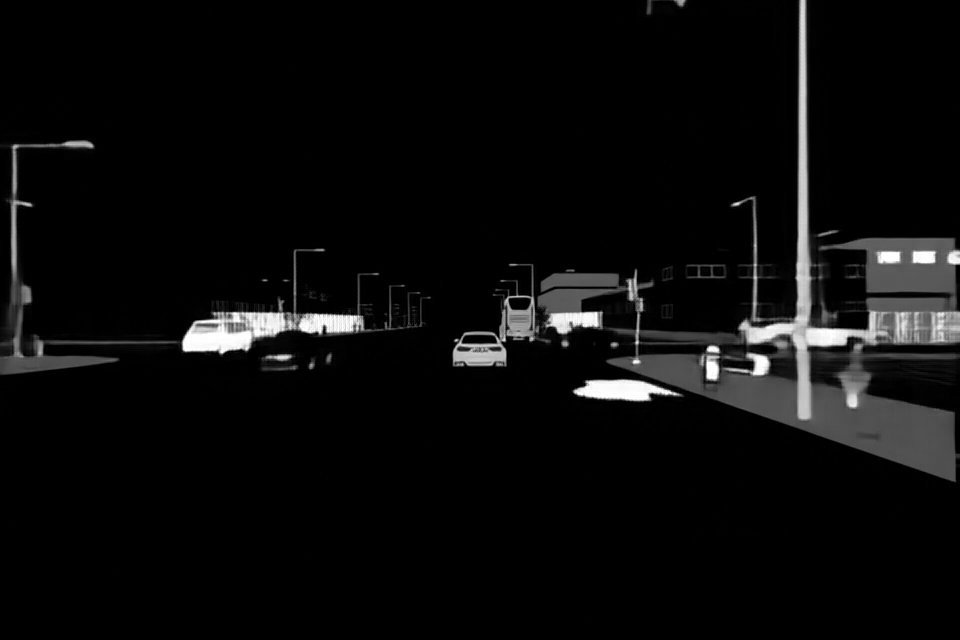}} &
    \frame{\includegraphics[width=1.2in]{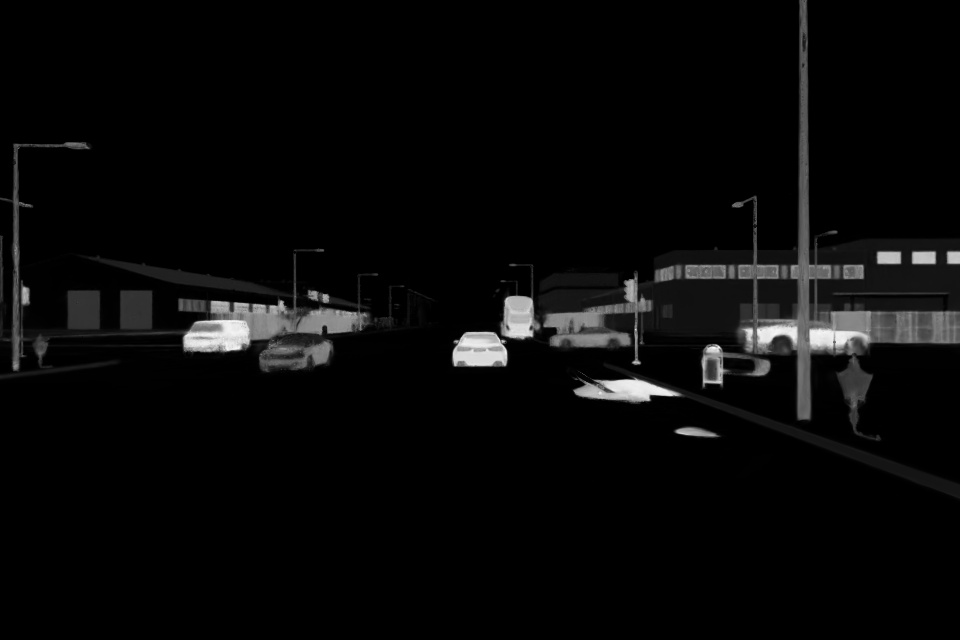}} & 
    \frame{\includegraphics[width=1.2in]{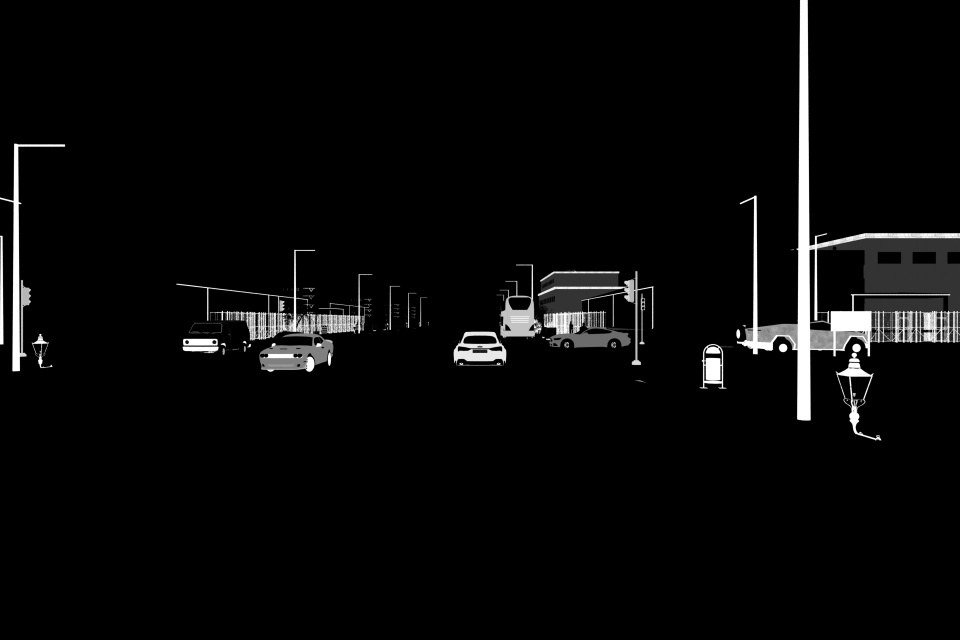}} \\

    \raisebox{2.5\normalbaselineskip}[0pt][0pt]{\rotatebox[origin=c]{90}{Normal}} & 
    \frame{\includegraphics[width=1.2in]{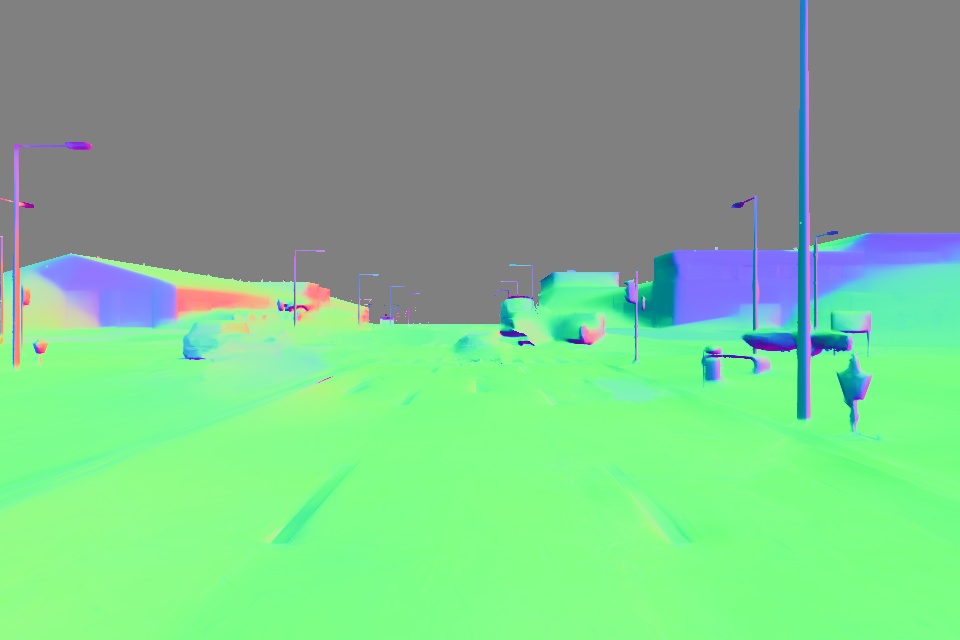}} &
    \frame{\includegraphics[width=1.2in]{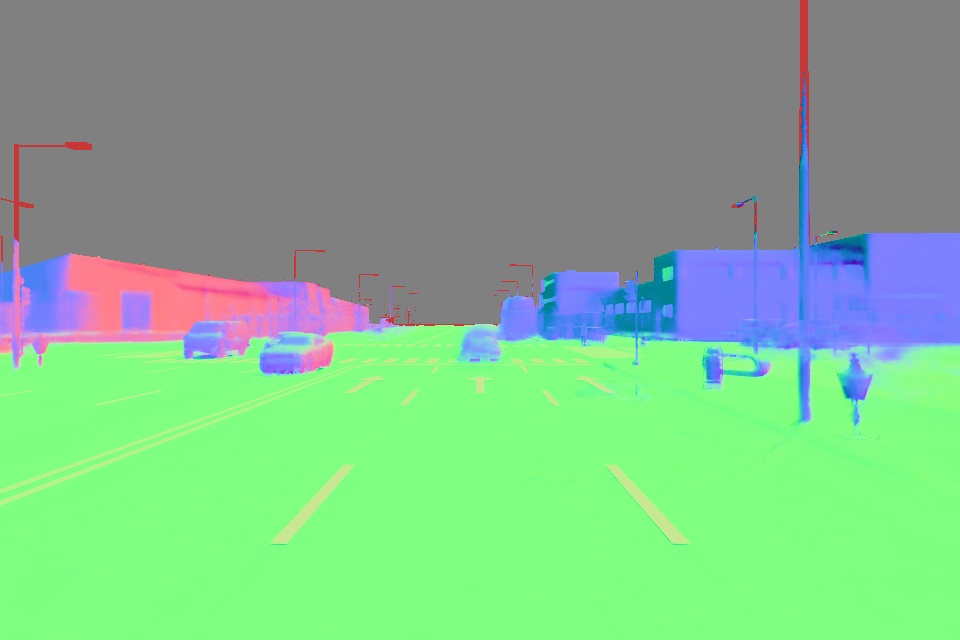}} & 
    \frame{\includegraphics[width=1.2in]{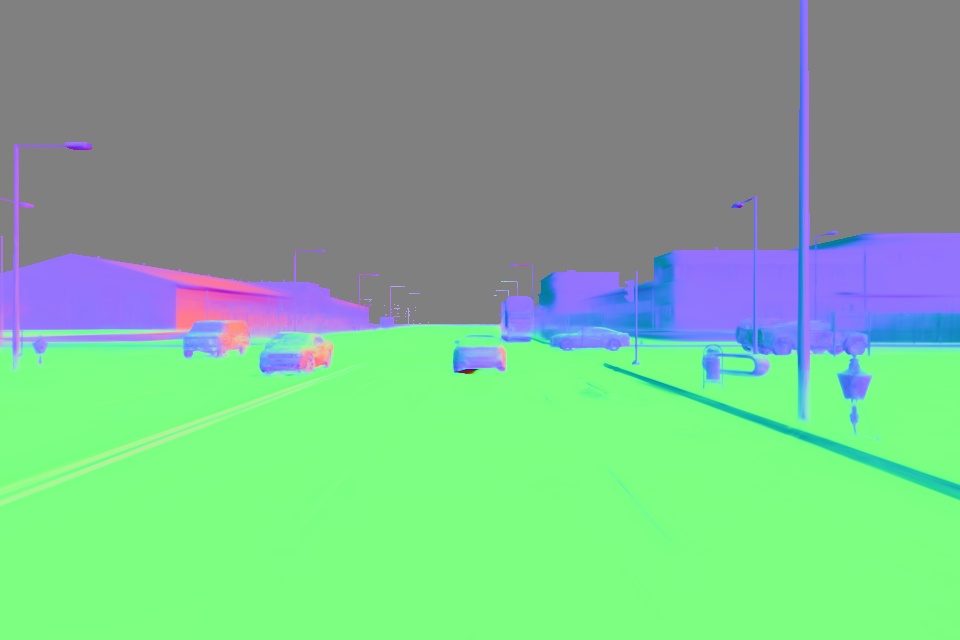}} & 
    \frame{\includegraphics[width=1.2in]{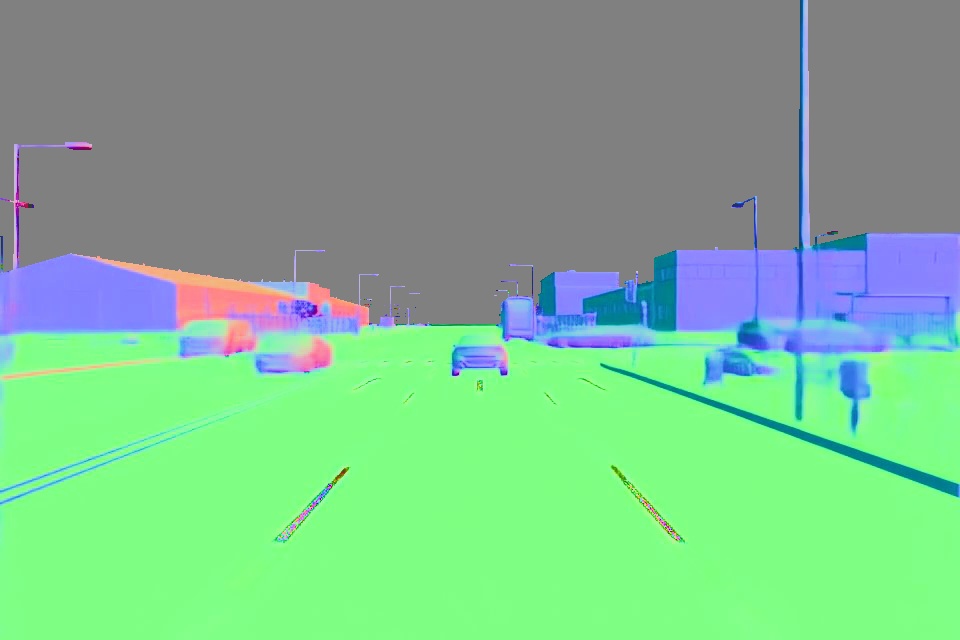}} & 
    \frame{\includegraphics[width=1.2in]{figures/images/null.jpg}} &
    \frame{\includegraphics[width=1.2in]{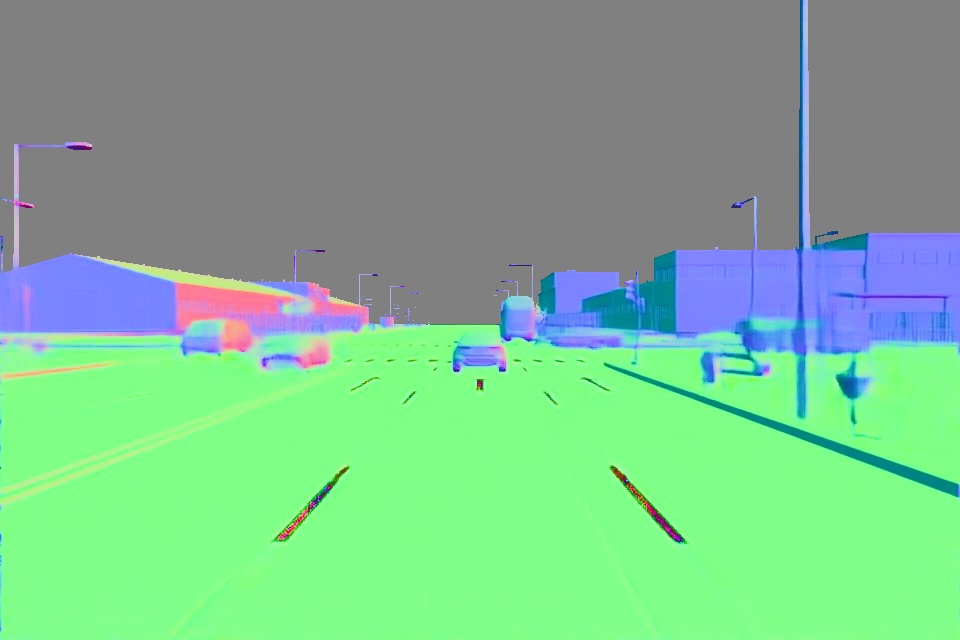}} &
    \frame{\includegraphics[width=1.2in]{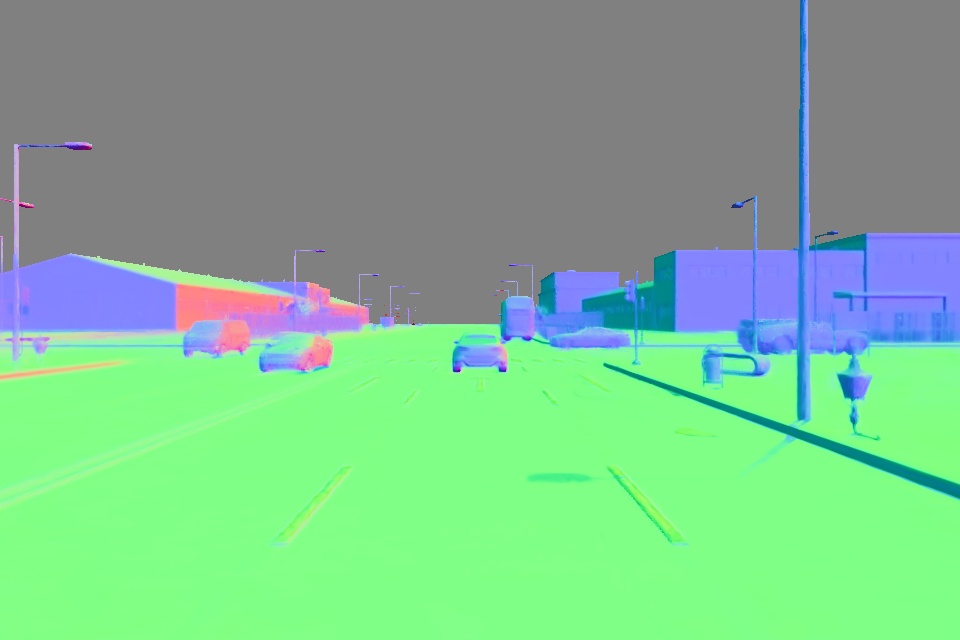}} & 
    \frame{\includegraphics[width=1.2in]{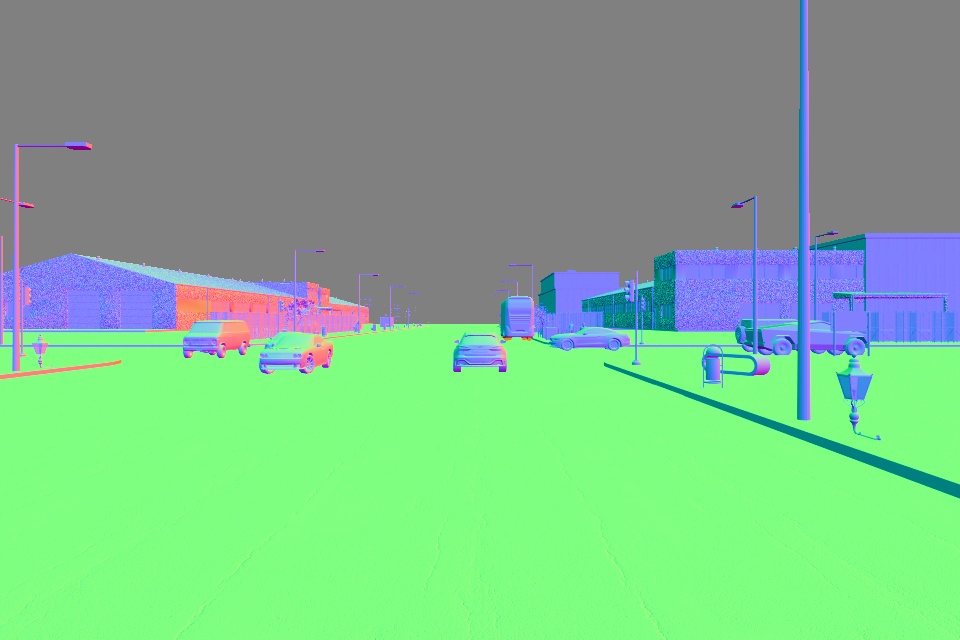}} \\

    \raisebox{2.5\normalbaselineskip}[0pt][0pt]{\rotatebox[origin=c]{90}{NVS}} & 
    \frame{\includegraphics[width=1.2in]{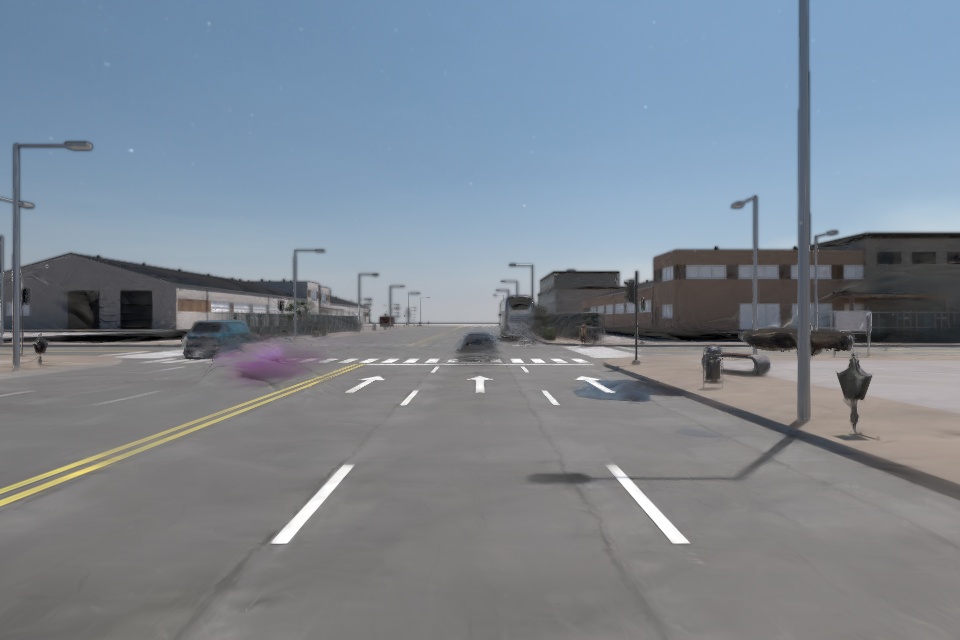}} &
    \frame{\includegraphics[width=1.2in]{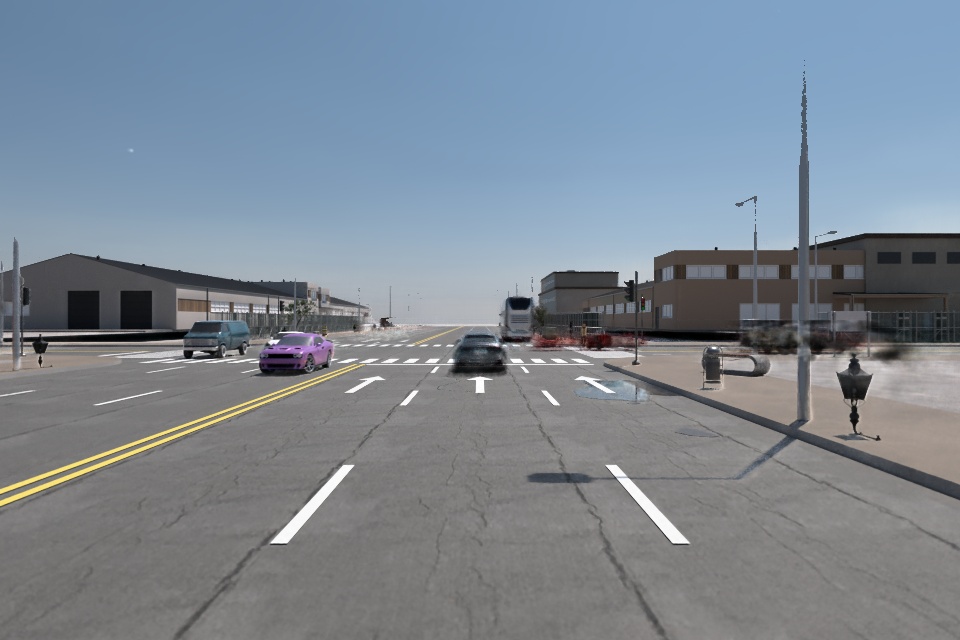}} & 
    \frame{\includegraphics[width=1.2in]{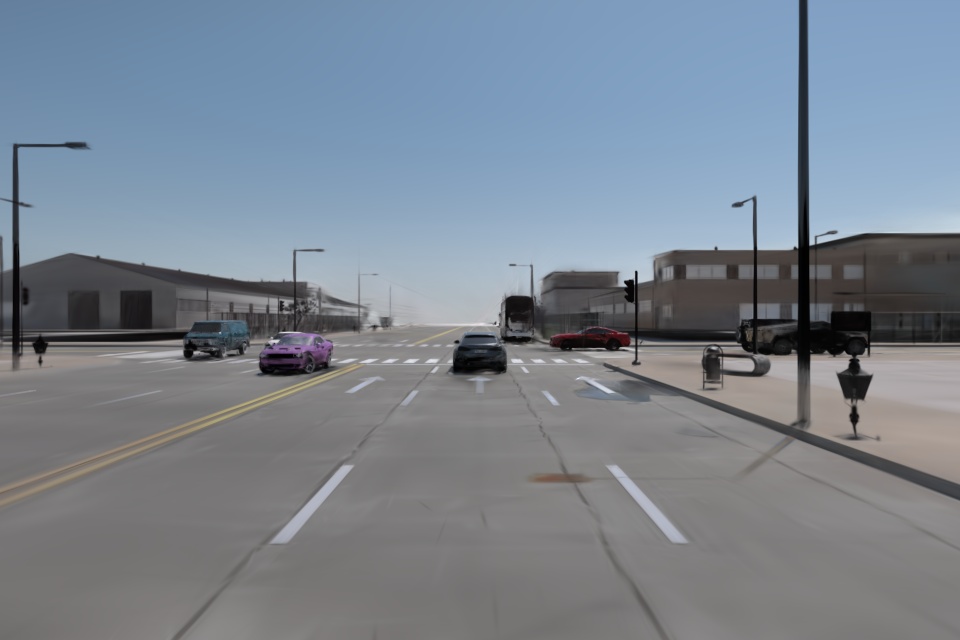}} & 
    \frame{\includegraphics[width=1.2in]{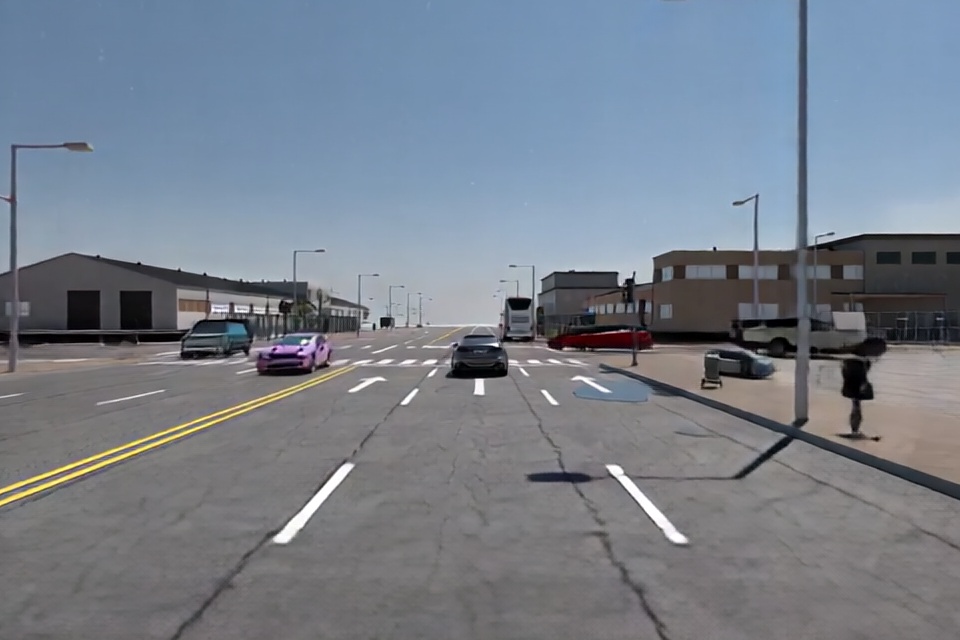}} & 
    \frame{\includegraphics[width=1.2in]{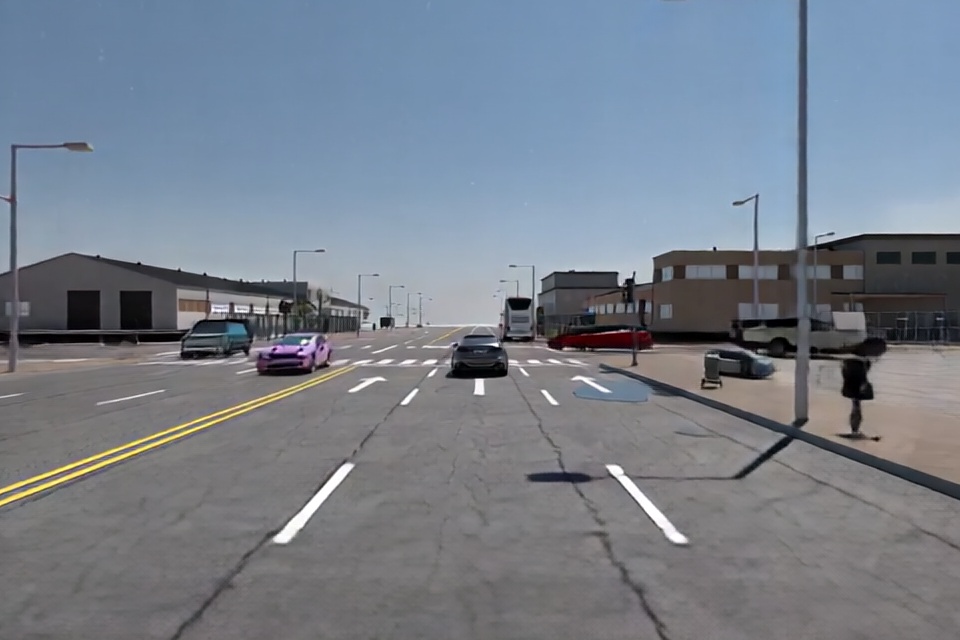}} &
    \frame{\includegraphics[width=1.2in]{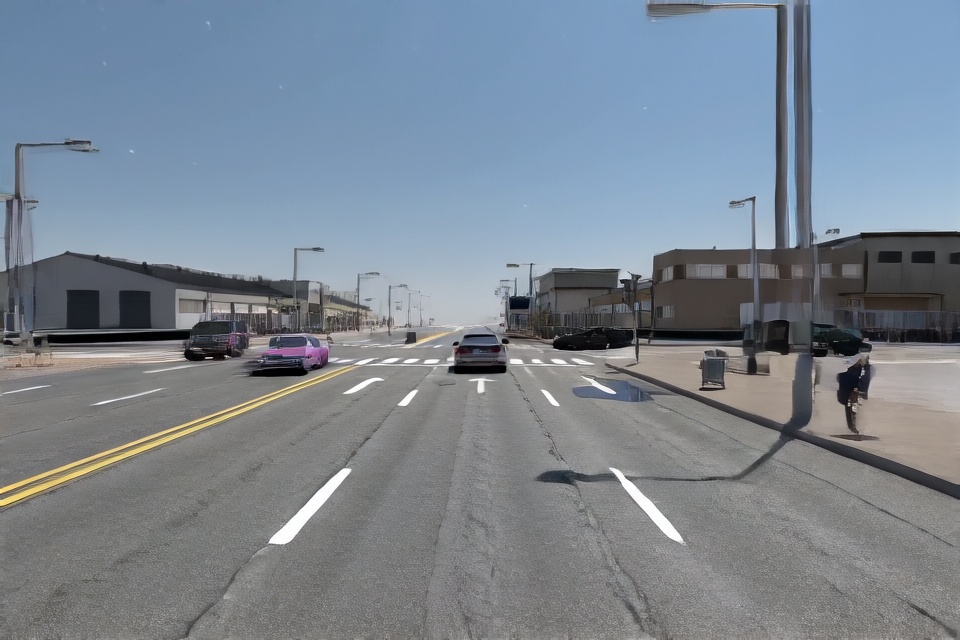}} &
    \frame{\includegraphics[width=1.2in]{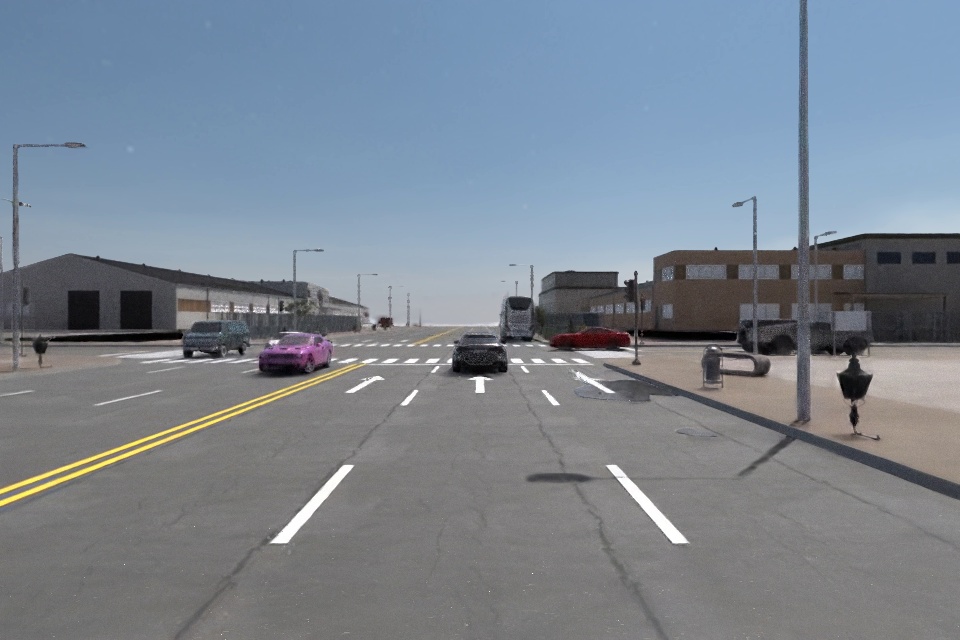}} & 
    \frame{\includegraphics[width=1.2in]{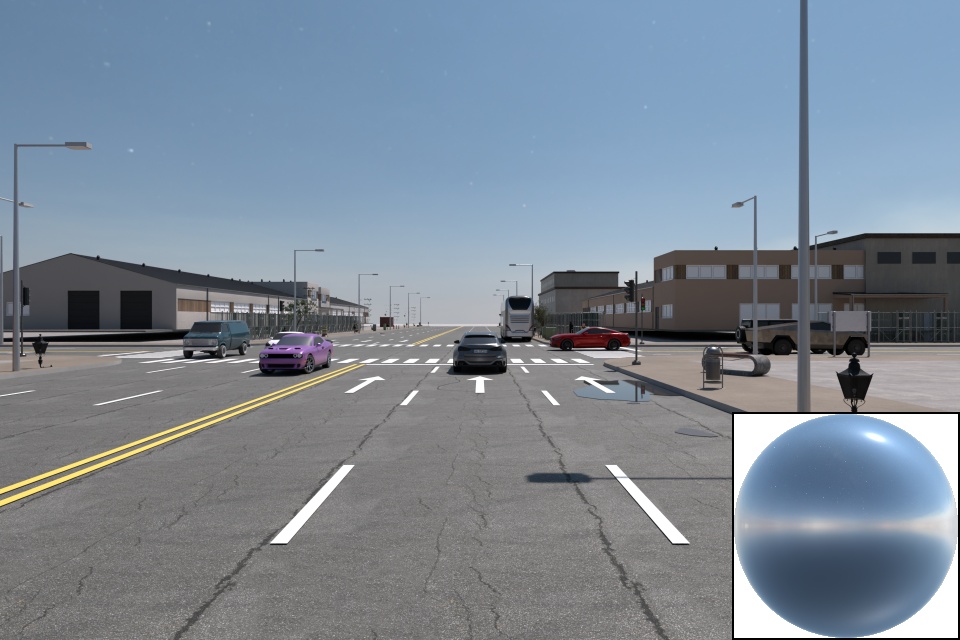}} \\

    \raisebox{2.5\normalbaselineskip}[0pt][0pt]{\rotatebox[origin=c]{90}{\makecell[c]{NVS + \\ Relighting}}} & 
    \frame{\includegraphics[width=1.2in]{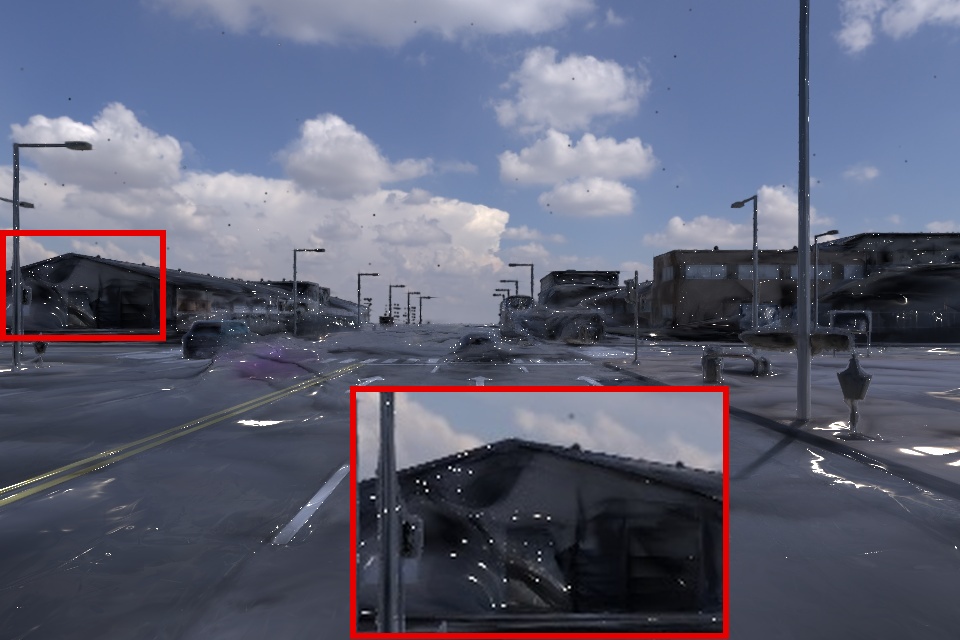}} &
    \frame{\includegraphics[width=1.2in]{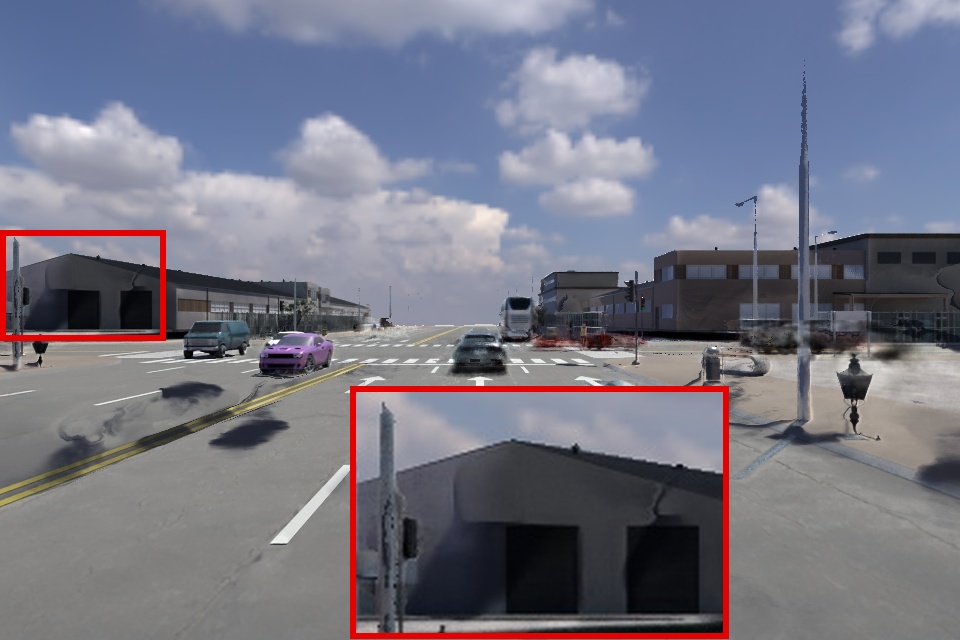}} & 
    \frame{\includegraphics[width=1.2in]{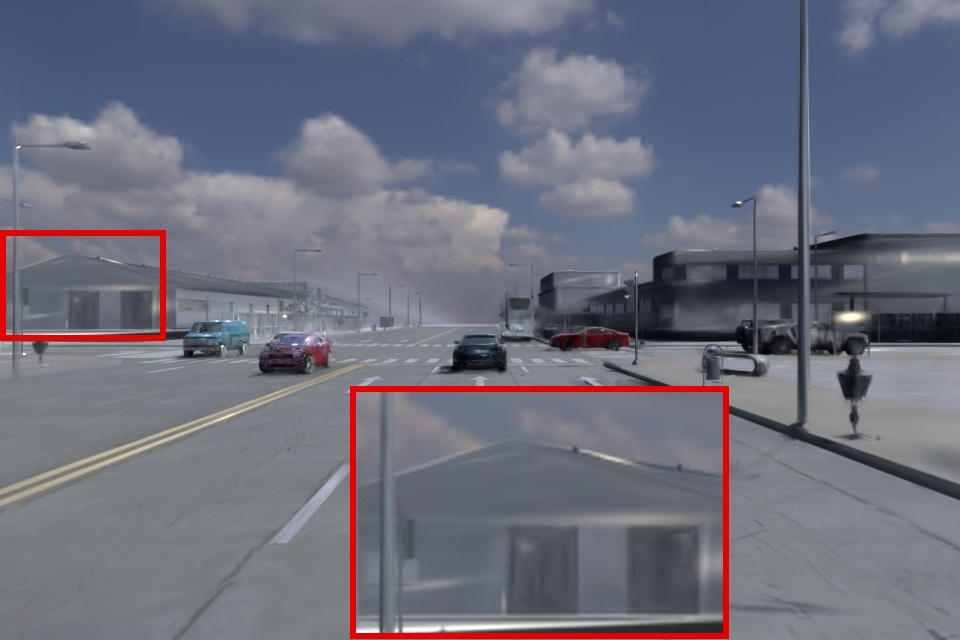}} & 
    \frame{\includegraphics[width=1.2in]{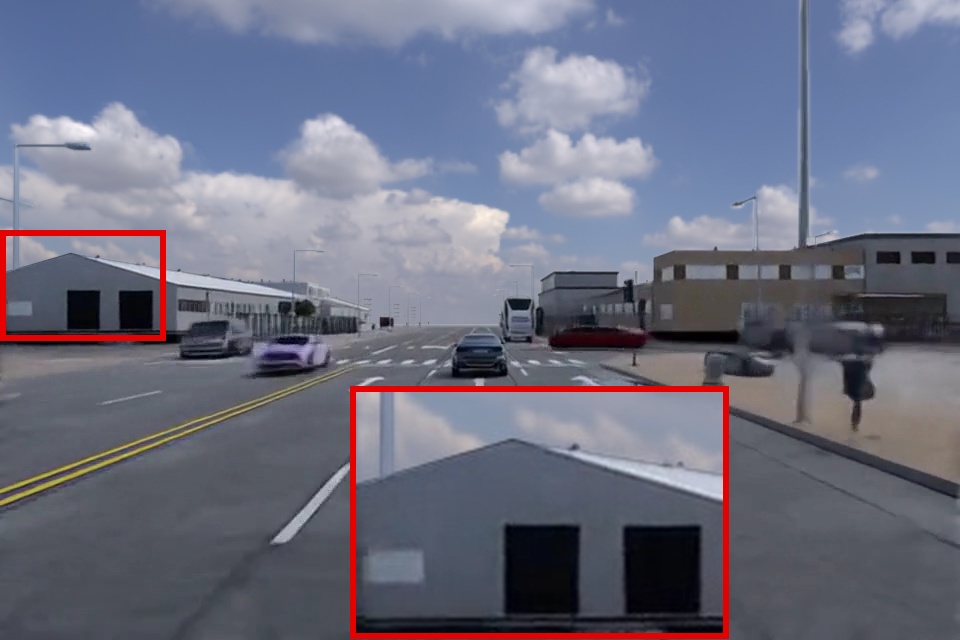}} & 
    \frame{\includegraphics[width=1.2in]{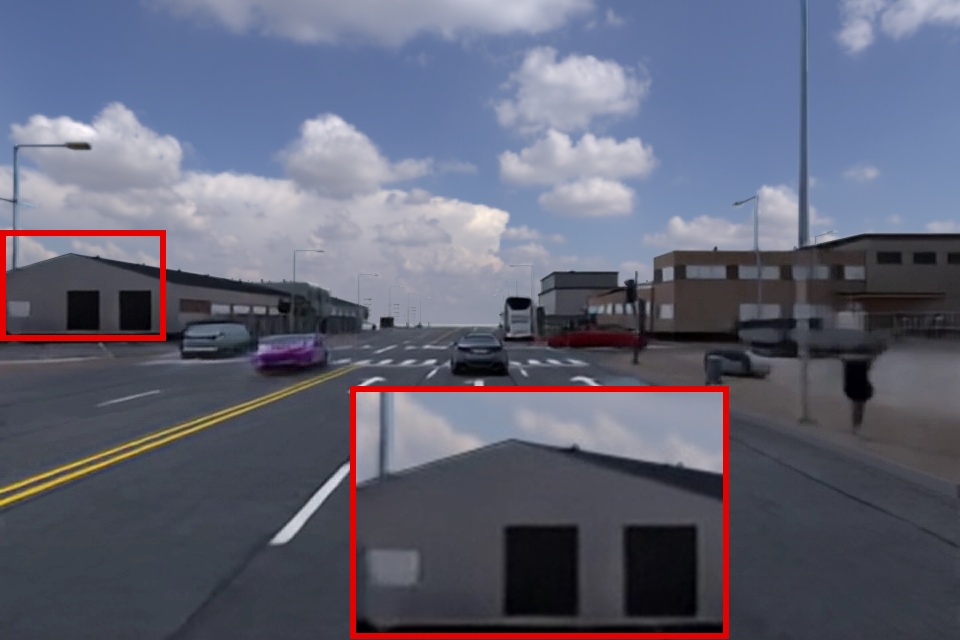}} &
    \frame{\includegraphics[width=1.2in]{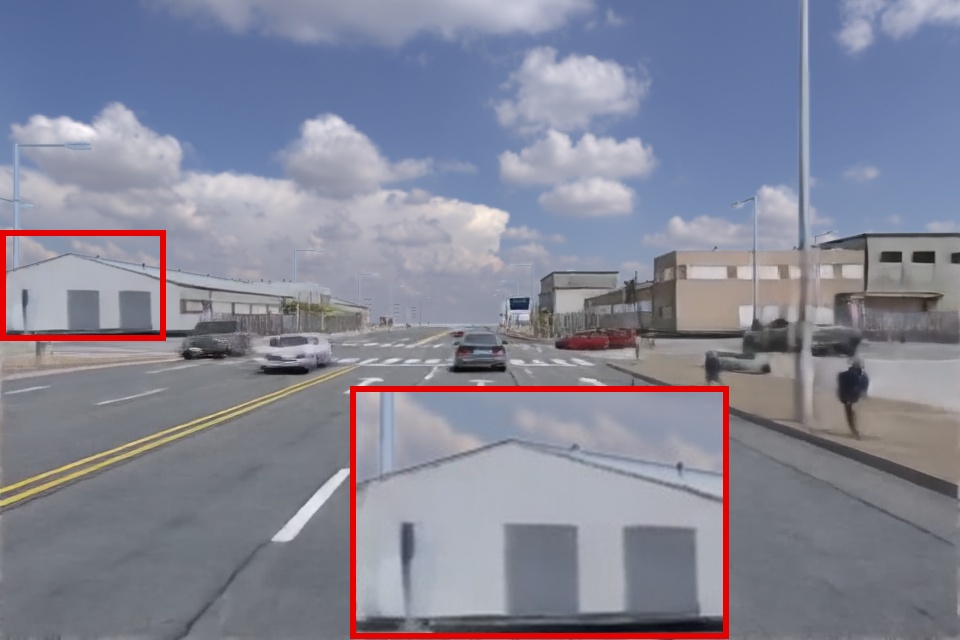}} &
    \frame{\includegraphics[width=1.2in]{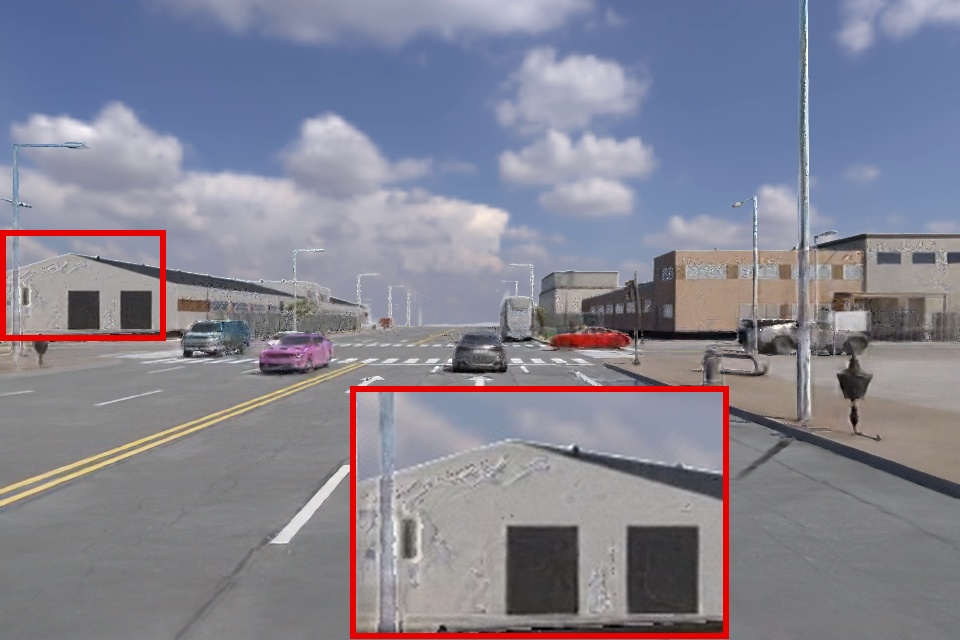}} & 
    \frame{\includegraphics[width=1.2in]{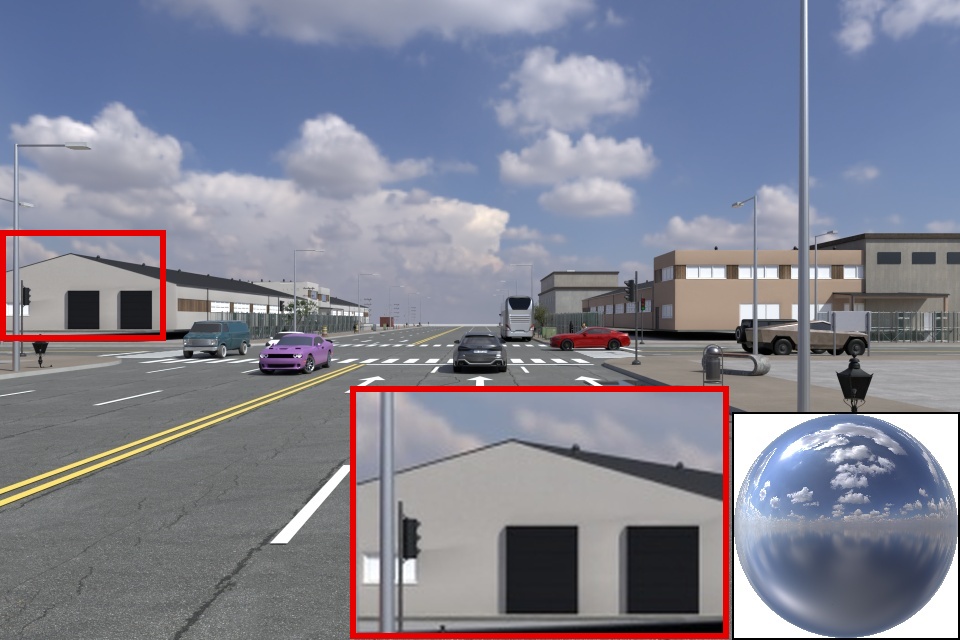}} \\

    \end{tabular}%
    }
    \caption{\textbf{Qualitative comparison of novel-view synthesis (NVS), inverse rendering, and relighting on the synthetic data.} The ground truth is provided in the rightmost column as a reference. For NVS and NVS+relighting tasks, we also provide lighting in the bottom-right corners. IRGS~\cite{IRGS}, UrbanIR~\cite{urbanir}, and InvRGB+L~\cite{invrgbl} produce rendering artifacts. Gen3C~\cite{Gen3C}+DR~\cite{diffusionrenderer} and Gen3C~\cite{Gen3C}+UR~\cite{unirelight} fail to simulate the target lighting. AW4D~\cite{autoweather4d} modifies the color of the building facade, as highlighted in the \textcolor{red}{\textbf{red}} zoom-in patches. Note that for Gen3C+DR, Gen3C+UR, and AW4D, we alpha-blend the relit results with sky color from the relighting environment map. Since IRGS does not model the sky, we also alpha-blend its NVS results with the ground-truth sky.}
    \label{fig:qual_synthetic}
\end{figure*}

\paragraph{\textbf{Qualitative Comparison on Synthetic Scenes.}}
In Fig.~\ref{fig:qual_synthetic}, we present a qualitative comparison with baseline methods using a scene from our synthetic dataset. IRGS~\cite{IRGS}, UrbanIR~\cite{urbanir}, and InvRGB+L~\cite{invrgbl} fail to reconstruct detailed geometry and materials, introducing severe artifacts under different viewpoints or illumination. Both Gen3C~\cite{Gen3C}+DR~\cite{diffusionrenderer} and Gen3C~\cite{Gen3C}+UR~\cite{unirelight} predict inaccurate shading under novel lighting (see the bottom row of Fig.~\ref{fig:qual_synthetic}). AW4D~\cite{autoweather4d} alters the scene content (e.g., the color of the building facade). Compared with the baselines, our method recovers accurate geometry and materials and produces physically plausible relighting that better aligns with the ground truth.

\paragraph{\textbf{Nighttime Relighting with Local Lights.}}
We also evaluate nighttime relighting on the six synthetic scenes. Each scene is rendered under a night-sky environment map and lit by streetlights and by the headlights and tail lights of the moving vehicles. All methods use the same night environment map. IRGS, UrbanIR, InvRGB+L, and AW4D also add the same local lights (positions, intensities, and cone angles) with their own shading (Sec.~\ref{supp:baseline}), whereas Gen3C+DR and Gen3C+UR only take the environment map.

\begin{figure*}[t]
    \centering
    \includegraphics[width=\linewidth]{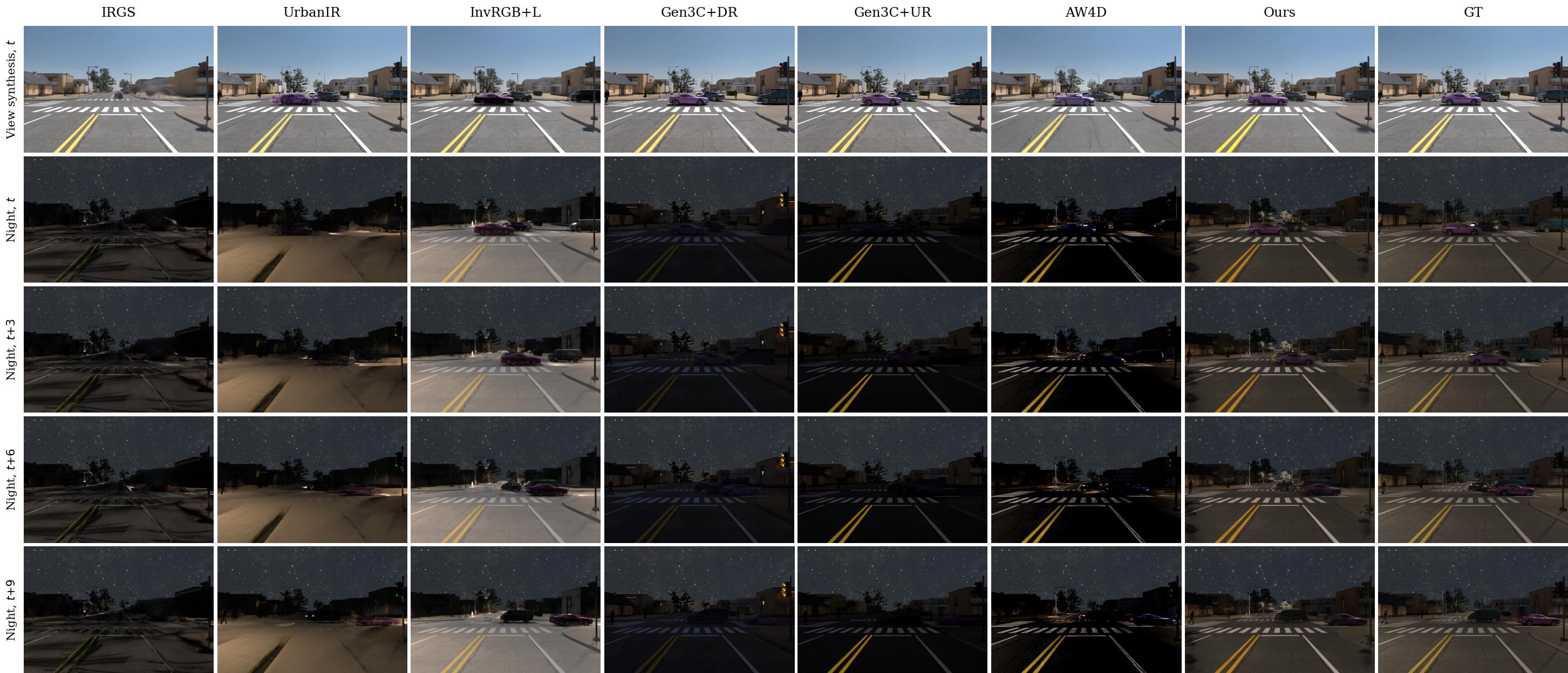}
    \vspace{-1.5em}
    \caption{\textbf{Nighttime relighting of a dynamic synthetic scene.} The top row shows the view synthesis results under the original lighting at frame $t$, and the remaining rows show the nighttime relighting results at frames $t$, $t{+}3$, $t{+}6$, and $t{+}9$. IRGS~\cite{IRGS} and UrbanIR~\cite{urbanir} cannot reconstruct the moving cars correctly. InvRGB+L~\cite{invrgbl} over-brightens the road. Gen3C~\cite{Gen3C}+DR~\cite{diffusionrenderer} and Gen3C~\cite{Gen3C}+UR~\cite{unirelight} cannot render the local lights. AW4D~\cite{autoweather4d} renders the road too dark due to its per-pixel blending of local and global lighting. Our method produces physically plausible results with correct local lighting and moving cars. For the baselines, we alpha-blend the sky of the relit results, as well as the NVS result of IRGS, with the ground-truth sky.}
    \label{fig:night_dyn}
\end{figure*}

Tab.~\ref{tab:quant_night} reports the novel-view nighttime relighting results. Our method achieves the best PSNR, SSIM, and LPIPS. Fig.~\ref{fig:night_dyn} shows a qualitative comparison on one synthetic scene. IRGS and UrbanIR fail to model the scene dynamics (e.g., moving cars). InvRGB+L over-brightens the road. Gen3C+DR and Gen3C+UR do not take the local lights as input, so the light pools cast by the local lights are missing. AW4D renders the local lights, but its per-pixel blending of the local-light rendering with the global relighting of the DR forward model makes the road too dark. In contrast, our method renders the local lights and the moving cars correctly, and its results are physically coherent because local and global lighting are rendered in the same PBR pass.

\begin{table}[t]
    \centering
    \small
    \caption{\textbf{Nighttime novel-view relighting with local lights on the synthetic dataset.} Metrics are averaged over the test frames of the six scenes and computed over non-sky regions. The best and second-best results are bold and underlined, respectively.}
    \label{tab:quant_night}
    \setlength{\tabcolsep}{5pt}
    \begin{tabular}{lccc}
    \toprule
    Method & PSNR$\uparrow$ & SSIM$\uparrow$ & LPIPS$\downarrow$ \\
    \midrule
    IRGS~\cite{IRGS} & 17.09 & 0.592 & 0.606 \\
    UrbanIR~\cite{urbanir} & \underline{19.51} & 0.688 & 0.585 \\
    InvRGB+L~\cite{invrgbl} & 15.85 & \underline{0.707} & \underline{0.454} \\
    Gen3C~\cite{Gen3C}+DR~\cite{diffusionrenderer} & 16.46 & 0.653 & 0.584 \\
    Gen3C~\cite{Gen3C}+UR~\cite{unirelight} & 15.65 & 0.574 & 0.559 \\
    AW4D~\cite{autoweather4d} & 15.74 & 0.478 & 0.554 \\
    Ours & \textbf{24.05} & \textbf{0.806} & \textbf{0.335} \\
    \bottomrule
    \end{tabular}
\end{table}
 
\paragraph{\textbf{Temporal Consistency of Material Maps.}}
We further compare the temporal consistency of the estimated materials with the hybrid baseline AW4D~\cite{autoweather4d} in Tab.~\ref{tab:quant_temporal_3d}. AW4D obtains the 2D material maps by applying the DR inverse model per clip. In contrast, we use a persistent 3D representation, which renders temporally consistent material maps. We run both methods on the 100-frame sequences of the six synthetic scenes. Since 100 frames exceed the 57-frame limit of the DR inverse model, AW4D processes the sequence with a sliding window of 57 frames and a stride of 7 frames, the same windowing we use to predict our G-buffer priors (Sec.~\ref{supp:impl}). For our method, we run the full inverse rendering pipeline on all 100 frames and render the material maps from the reconstructed scene. Our method achieves a substantially lower warping error than AW4D on all three materials over all 99 consecutive frame pairs, demonstrating the benefit of the 3D representation.
\begin{table}[tb]
\centering
\small
\caption{\textbf{Temporal consistency of material maps.} We compare the per-clip 2D material estimates of AW4D~\cite{autoweather4d} with the material maps rendered from our 3D representation. We report the temporal warping error ($\times 10^{3}$), computed as in Tab.~\ref{tab:quant_temporal} but over all 99 consecutive frame pairs of the 100-frame synthetic sequences, averaged over the six scenes. Lower is better.}
\label{tab:quant_temporal_3d}
\setlength{\tabcolsep}{4pt}
\begin{tabular}{lccc}
\toprule
Method & Albedo$\downarrow$ & Roughness$\downarrow$ & Metallic$\downarrow$ \\
\midrule
AW4D~\cite{autoweather4d} & 12.43 & 14.78 & 11.84 \\
Ours & \textbf{4.79} & \textbf{3.40} & \textbf{2.69} \\
\bottomrule
\end{tabular}
\end{table}
 
\begin{figure*}[t]
    \centering\setlength{\tabcolsep}{2pt}
    \newcommand{\figunirelightlabel}[1]{\raisebox{0.5in}[0pt][0pt]{\rotatebox[origin=c]{90}{#1}}}
    \resizebox{1.0\textwidth}{!}{%
    \begin{tabular}{@{}lcccccc@{}}

     & Novel-View Synthesis & + Relighting & Novel-View Synthesis & + Relighting & Novel-View Synthesis & + Relighting \\[0.2em]

    \figunirelightlabel{GT \& Lighting}&
    \frame{\includegraphics[width=1.5in]{figures/qual_relighting_imgs_v2/gt_nvs_049_f120.jpg}}&
    \includegraphics[width=1.5in]{figures/qual_relighting_imgs_v2/ball_049_snow_rot100.jpg}&
    \frame{\includegraphics[width=1.5in]{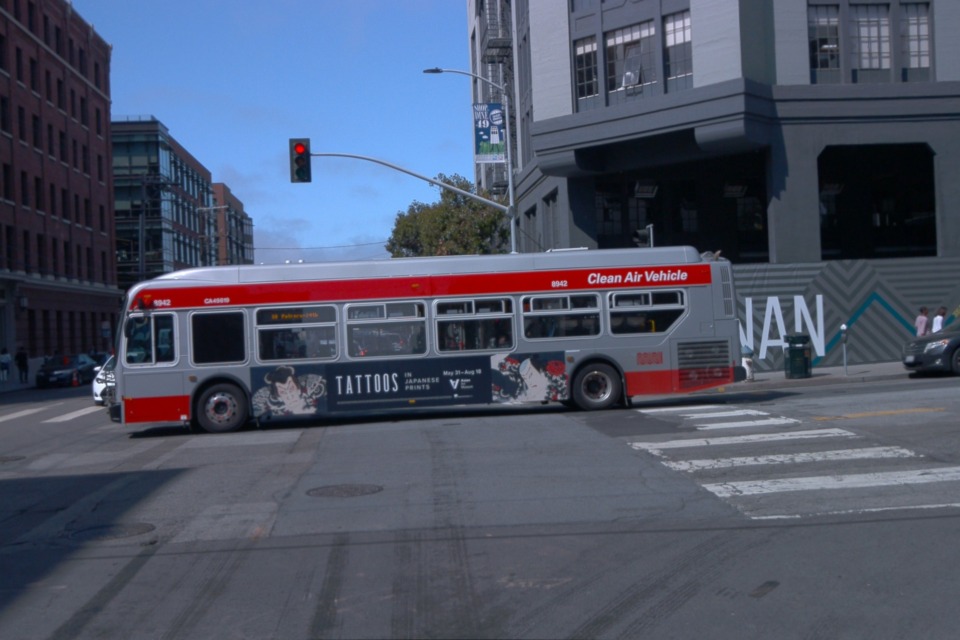}}&
    \includegraphics[width=1.5in]{figures/qual_relighting_imgs/nvs172_relight_envmap_chromeball.jpg}&
    \frame{\includegraphics[width=1.5in]{figures/qual_relighting_imgs_v2/gt_nvs_003.jpg}}&
    \includegraphics[width=1.5in]{figures/qual_relighting_imgs/night_carlight_highres.jpg} \\

    \figunirelightlabel{IRGS~\cite{IRGS}}&
    \frame{\includegraphics[width=1.5in]{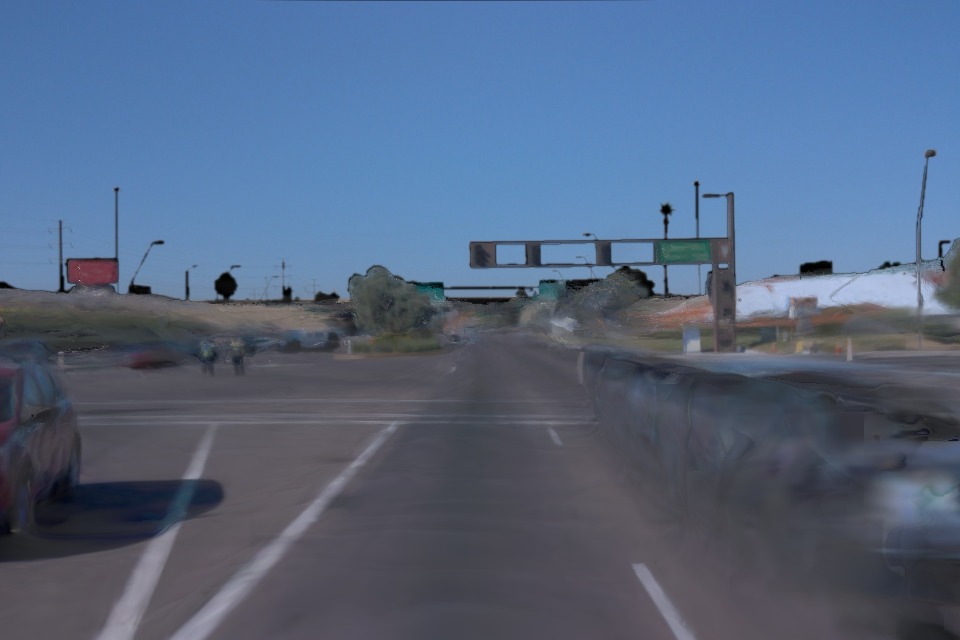}} &
    \frame{\includegraphics[width=1.5in]{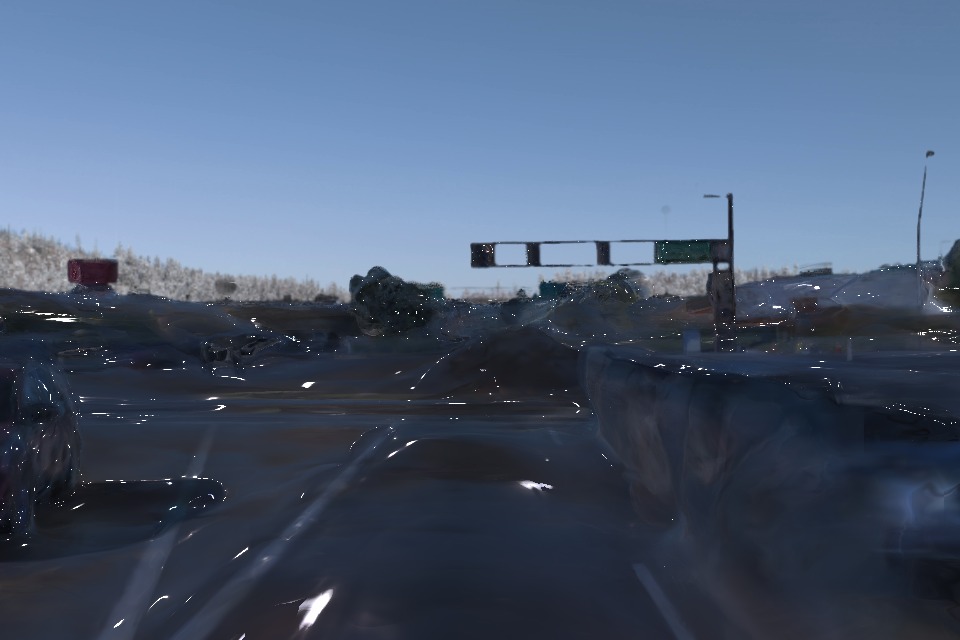}} &
    \frame{\includegraphics[width=1.5in]{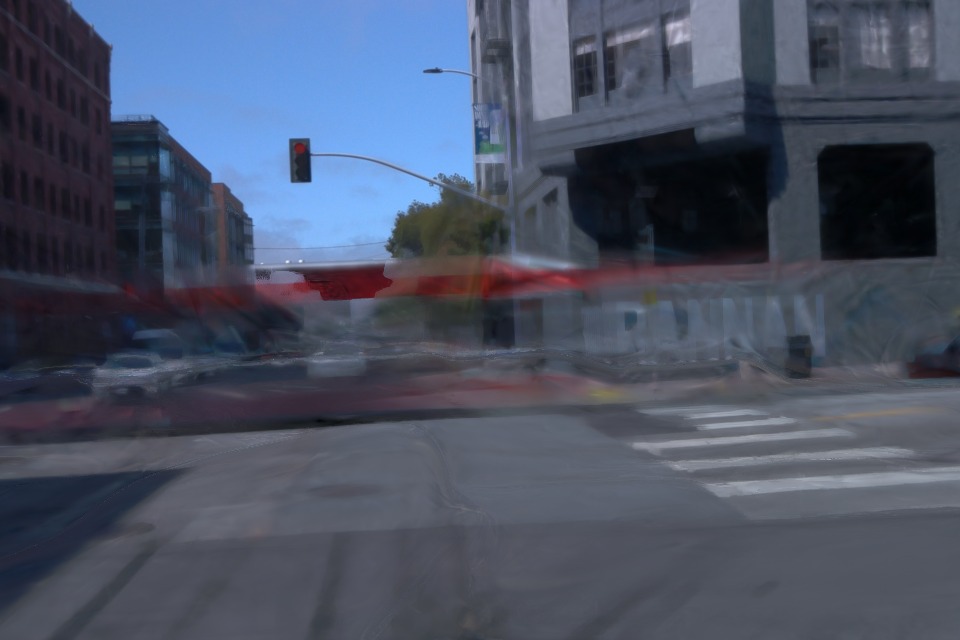}} &
    \frame{\includegraphics[width=1.5in]{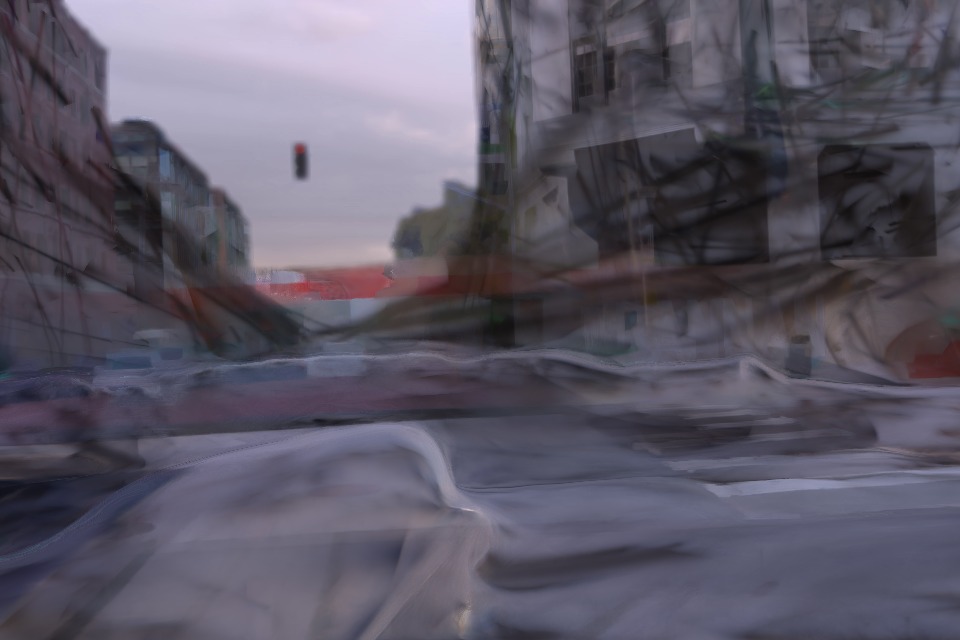}} &
    \frame{\includegraphics[width=1.5in]{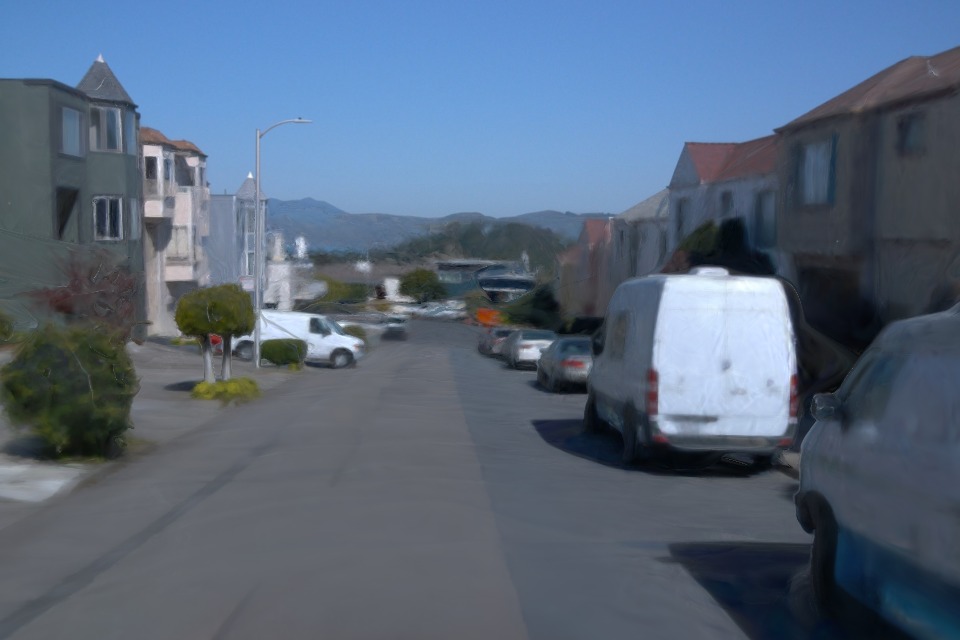}} &
    \frame{\includegraphics[width=1.5in]{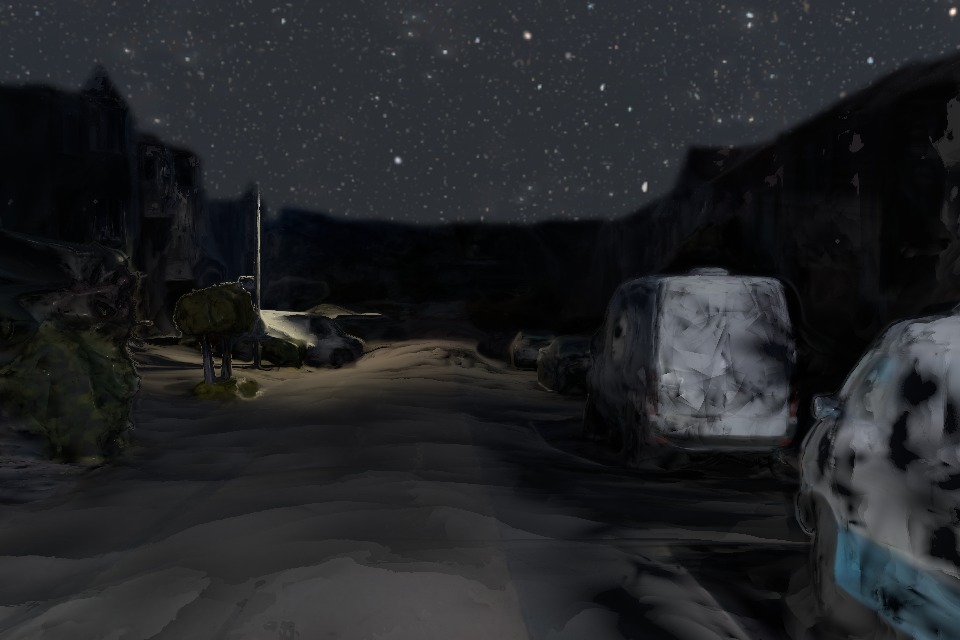}} \\

    \figunirelightlabel{UrbanIR~\cite{urbanir}}&
    \frame{\includegraphics[width=1.5in]{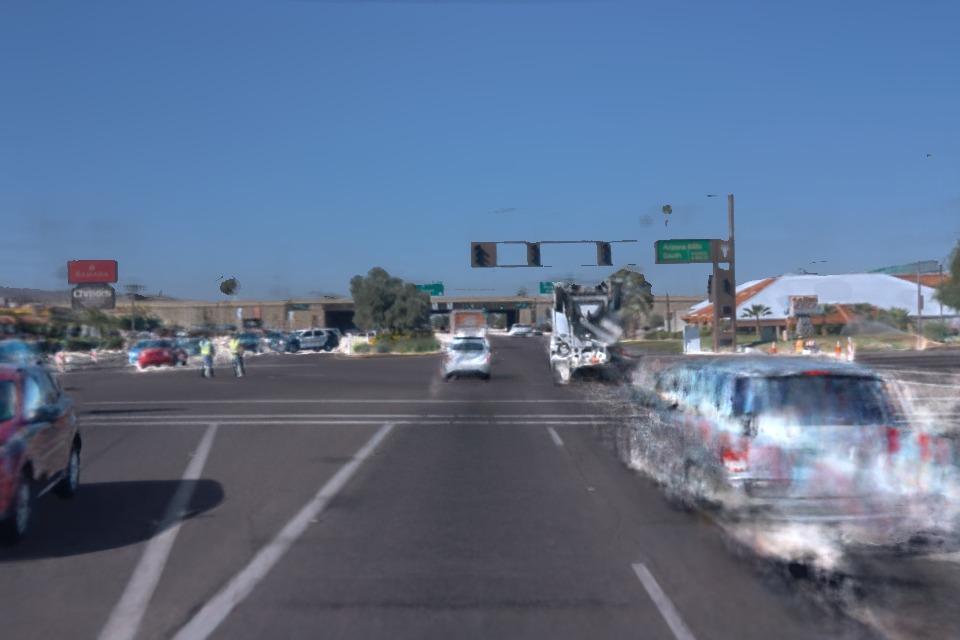}}&
    \frame{\includegraphics[width=1.5in]{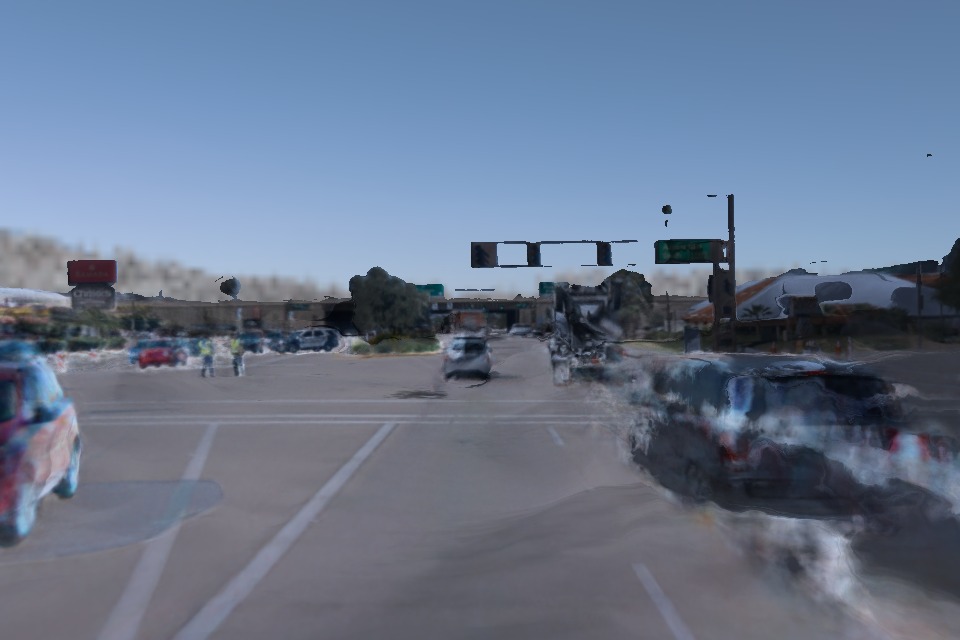}}&
    \frame{\includegraphics[width=1.5in]{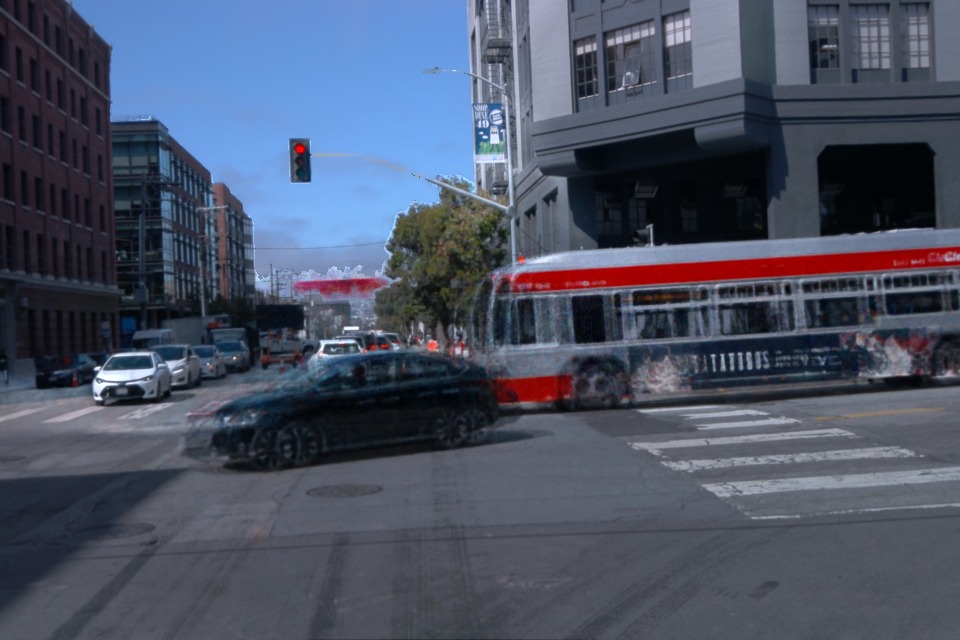}}&
    \frame{\includegraphics[width=1.5in]{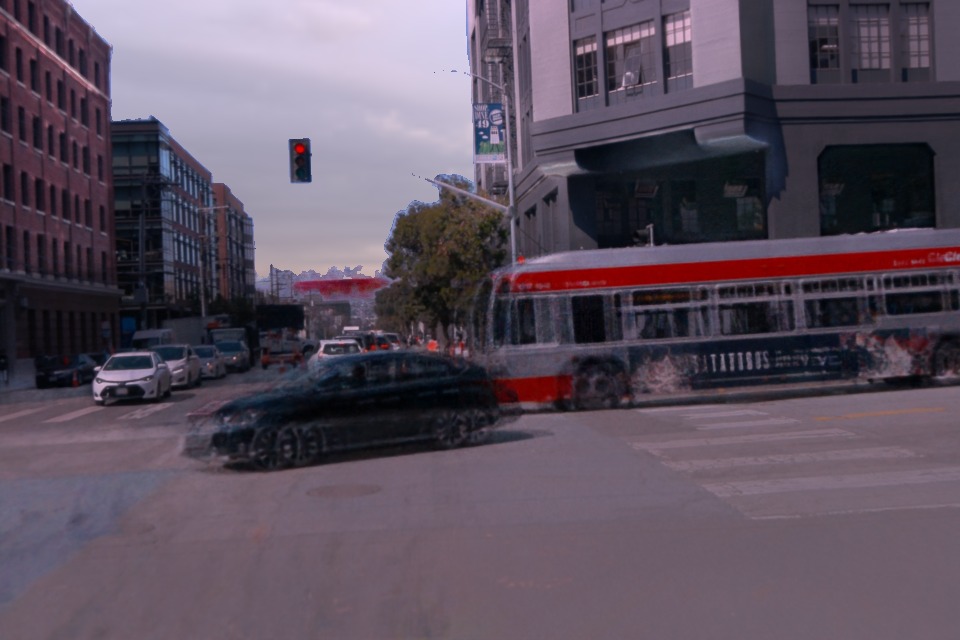}}&
    \frame{\includegraphics[width=1.5in]{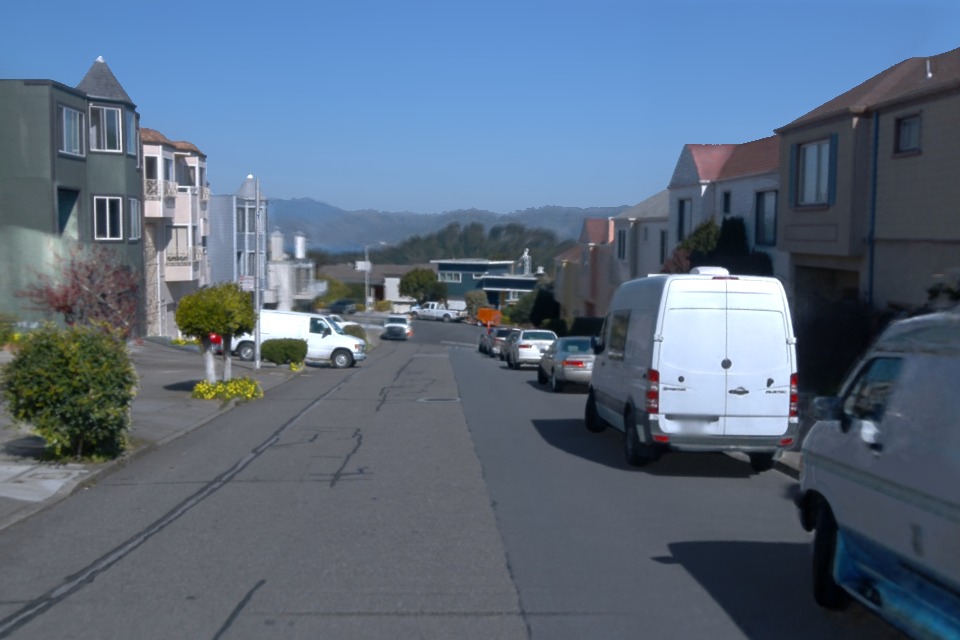}}&
    \frame{\includegraphics[width=1.5in]{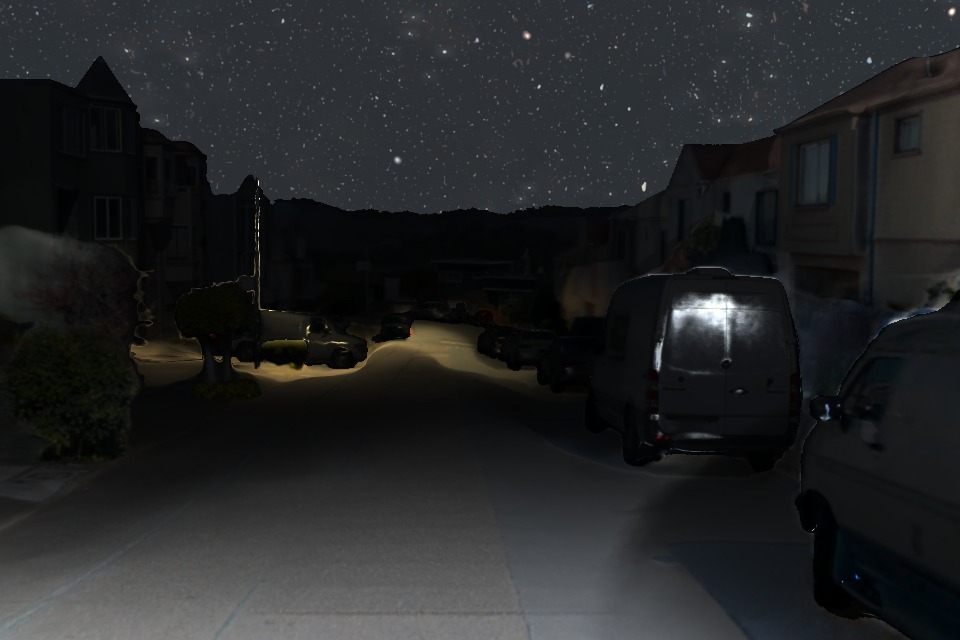}} \\

    \figunirelightlabel{Gen3C~\cite{Gen3C}+UR~\cite{unirelight}}&
    \frame{\includegraphics[width=1.5in]{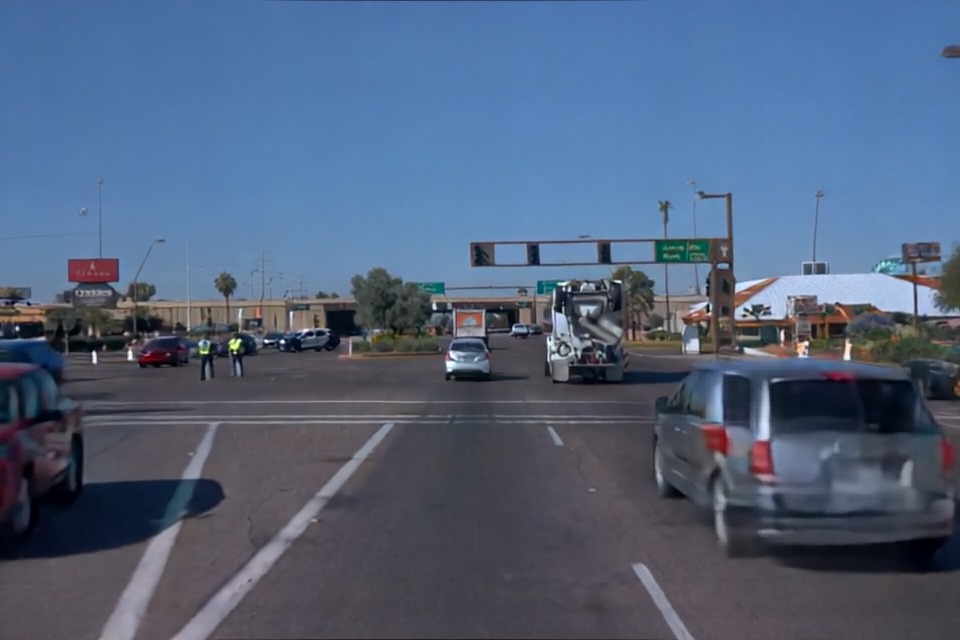}}&
    \frame{\includegraphics[width=1.5in]{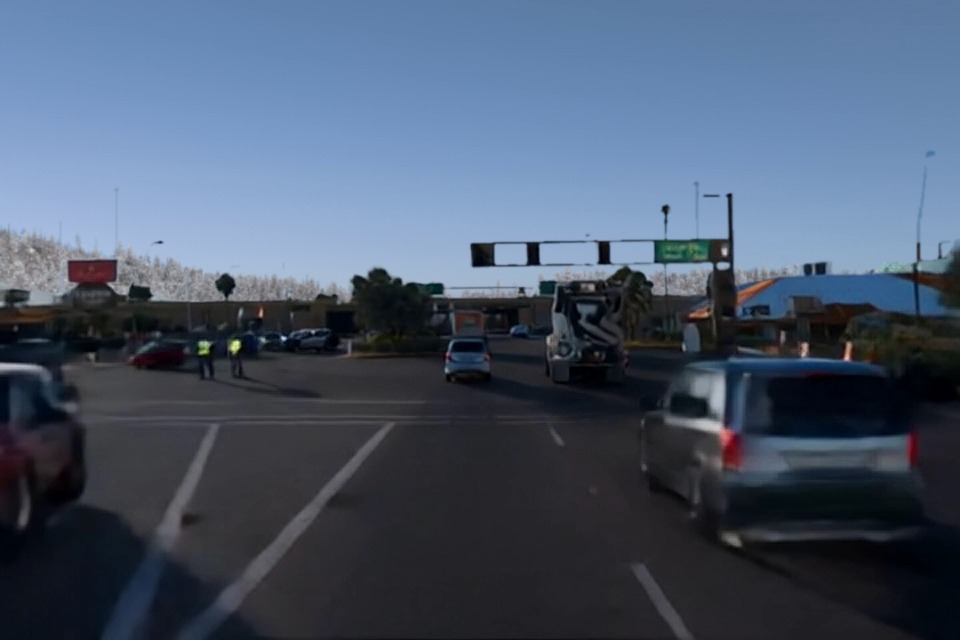}}&
    \frame{\includegraphics[width=1.5in]{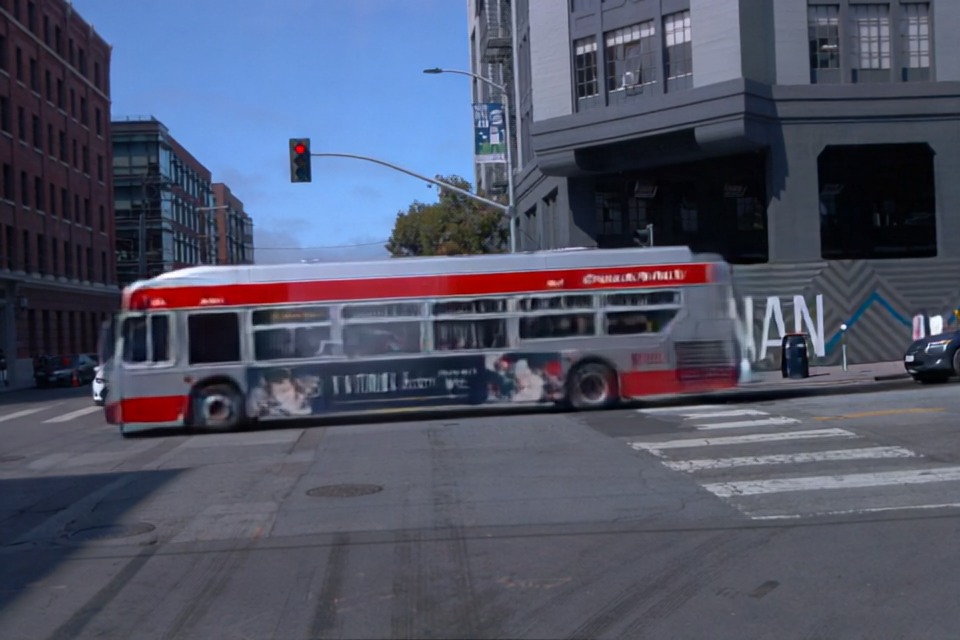}}&
    \frame{\includegraphics[width=1.5in]{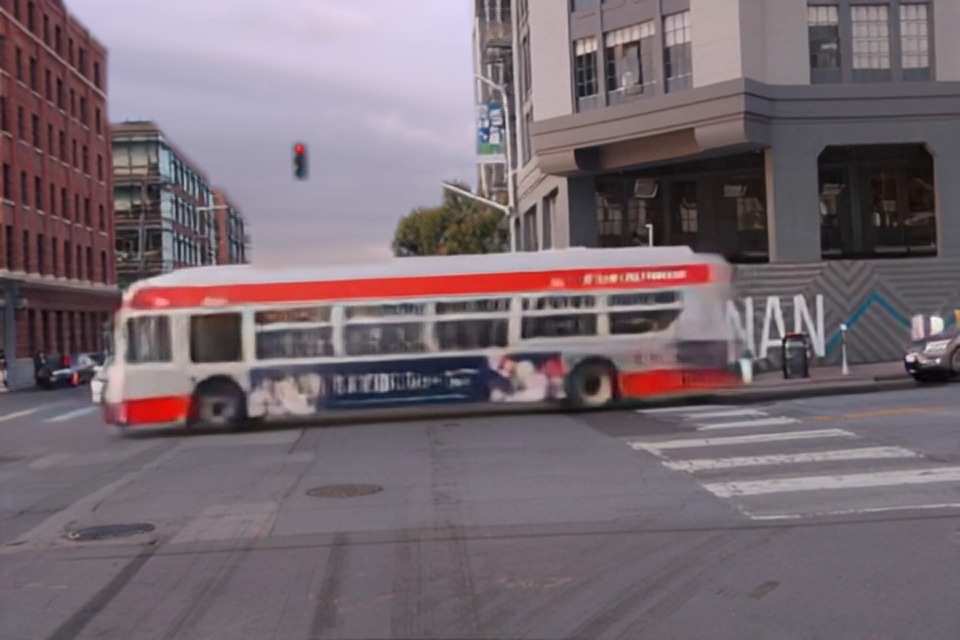}}&
    \frame{\includegraphics[width=1.5in]{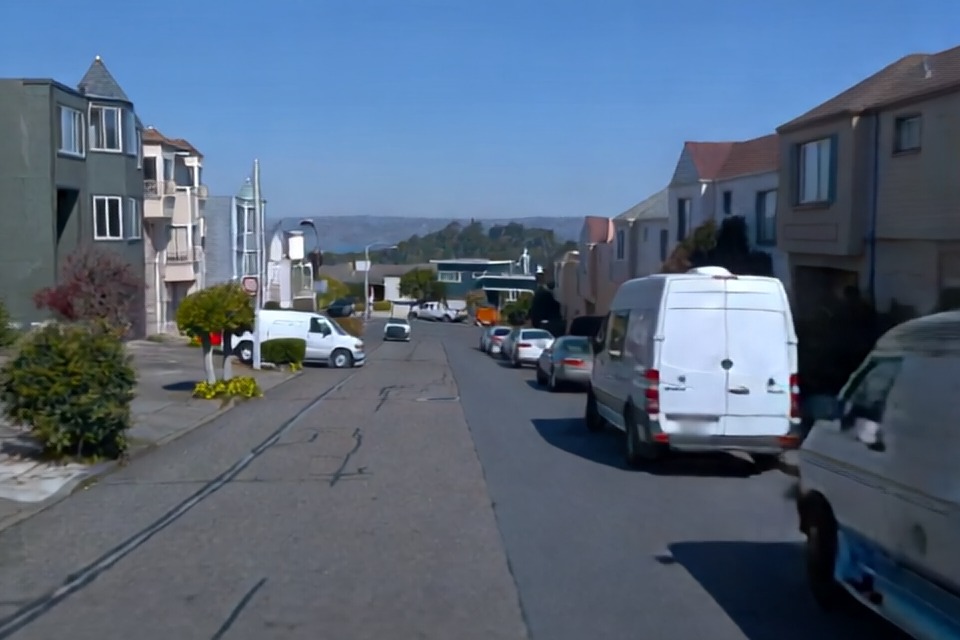}}&
    \frame{\includegraphics[width=1.5in]{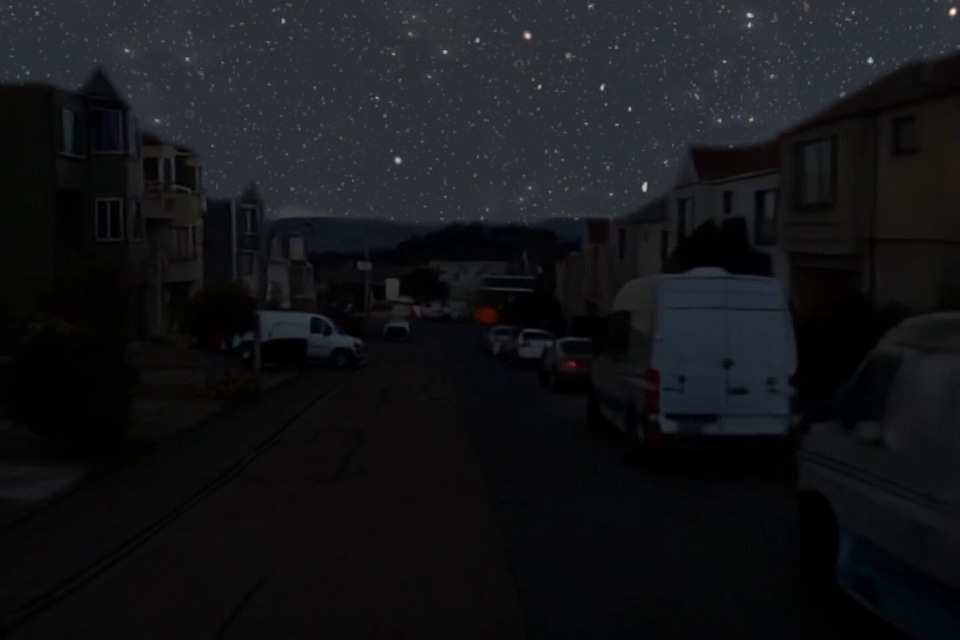}} \\

    \figunirelightlabel{Ours}&
    \frame{\includegraphics[width=1.5in]{figures/qual_relighting_imgs_v2/ours_nvs_049_f120_tau8_raster.jpg}}&
    \frame{\includegraphics[width=1.5in]{figures/qual_relighting_imgs_v2/ours_relight_049_f120_tau6_raster.jpg}}&
    \frame{\includegraphics[width=1.5in]{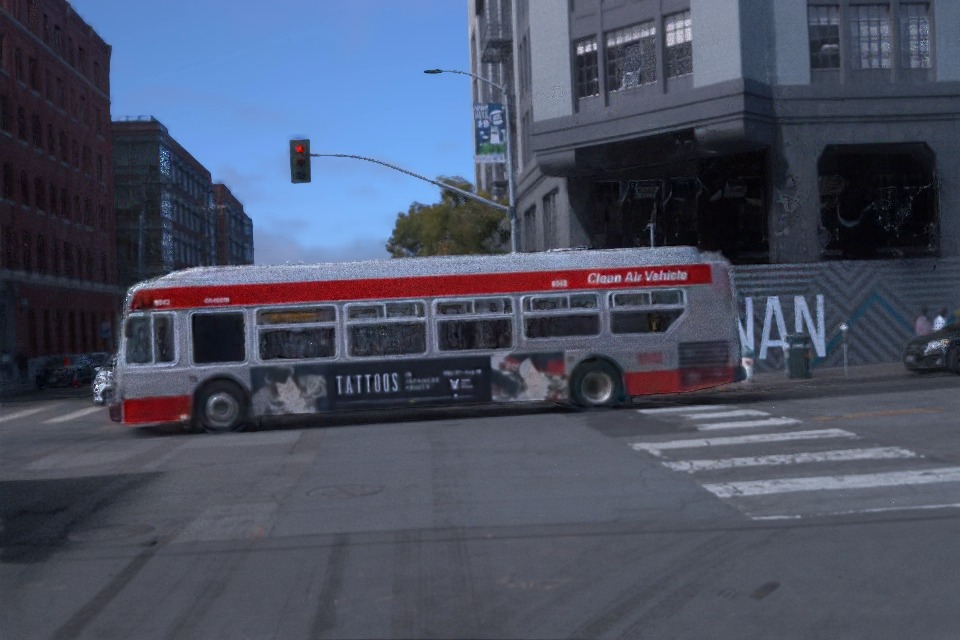}}&
    \frame{\includegraphics[width=1.5in]{figures/qual_relighting_imgs_v2/ours_relight_172_raster.jpg}}&
    \frame{\includegraphics[width=1.5in]{figures/qual_relighting_imgs_v2/ours_nvs_003_tau8_raster.jpg}}&
    \frame{\includegraphics[width=1.5in]{figures/qual_relighting_imgs_v2/ours_relight_003_raster.jpg}} \\
    \end{tabular}%
    }
    \caption{\textbf{Qualitative comparison of novel-view synthesis and relighting with IRGS, UrbanIR, and Gen3C+UR on WOD.} IRGS~\cite{IRGS} and UrbanIR~\cite{urbanir} fail to reconstruct the dynamic scenes (left and middle). Gen3C~\cite{Gen3C}+UR~\cite{unirelight} fails to simulate the correct lighting effects such as shadows. Note that for Gen3C+UR and IRGS, we alpha-blend the relit results with the sky color from the relighting environment map for fair comparison. Since IRGS does not model the sky, we also alpha-blend its NVS results with the ground-truth sky.}
    \label{fig:qual_unirelight}
\end{figure*}

\begin{figure*}[t]
    \centering\setlength{\tabcolsep}{3pt}
    \resizebox{1.0\textwidth}{!}{%
    \begin{tabular}{@{}lccccc@{}}

     & Novel-View Synthesis & Normal & Albedo & Roughness & Metallic \\[0.2em]

    \raisebox{2.5\normalbaselineskip}[0pt][0pt]{\rotatebox[origin=c]{90}{IRGS~\cite{IRGS}}}& \frame{\includegraphics[width=1.5in,height=1in]{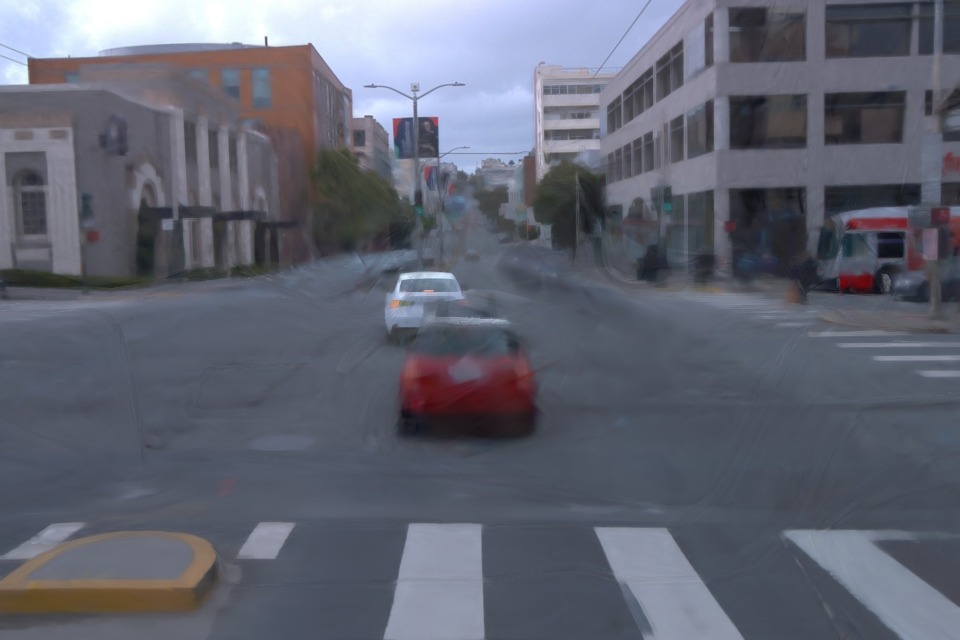}} & \frame{\includegraphics[width=1.5in,height=1in]{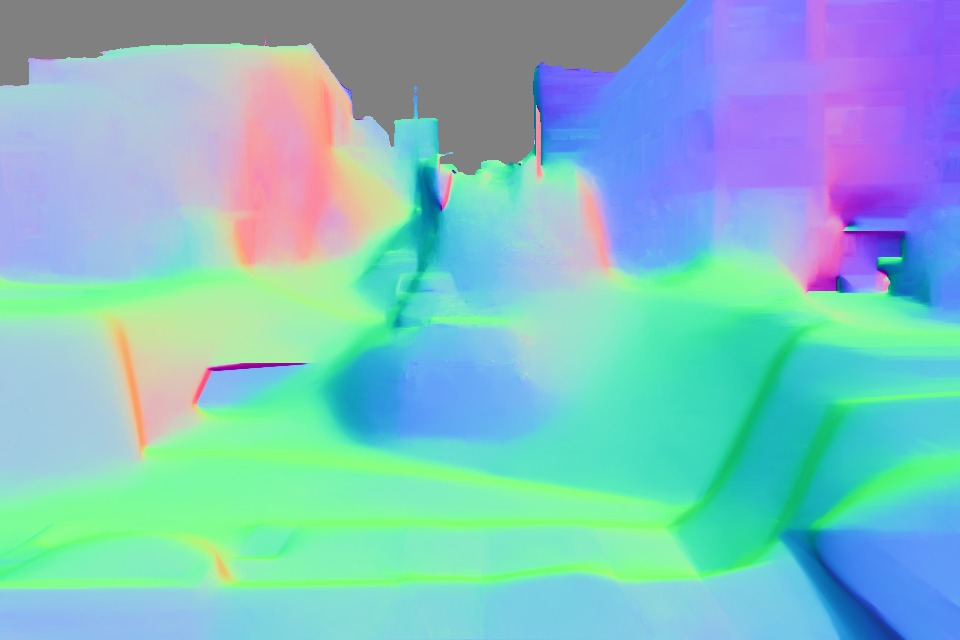}} & \frame{\includegraphics[width=1.5in,height=1in]{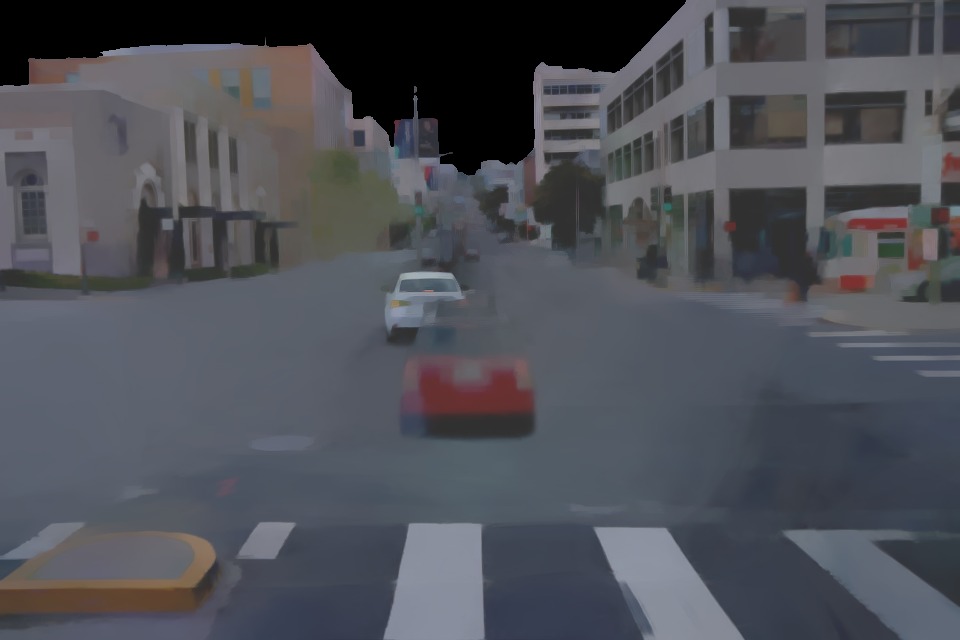}} & \frame{\includegraphics[width=1.5in,height=1in]{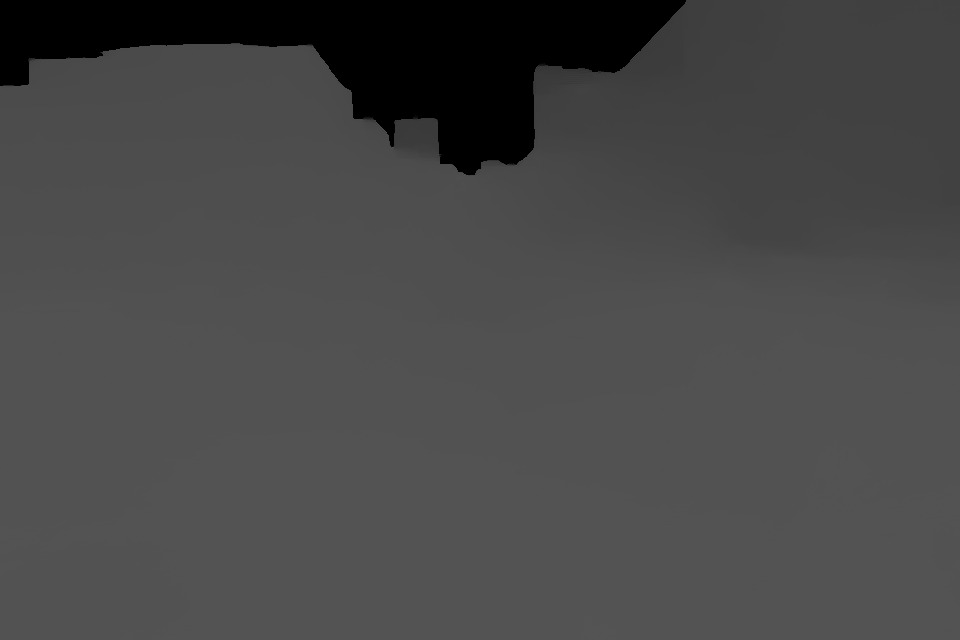}} & \frame{\includegraphics[width=1.5in,height=1in]{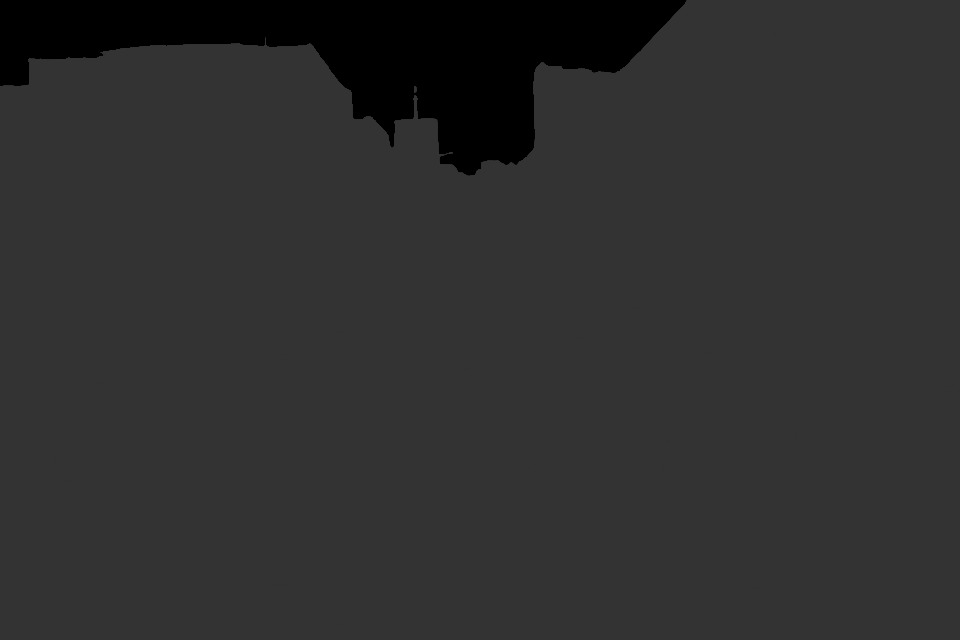}} \\

    \raisebox{2.5\normalbaselineskip}[0pt][0pt]{\rotatebox[origin=c]{90}{UrbanIR~\cite{urbanir}}}& \frame{\includegraphics[width=1.5in,height=1in]{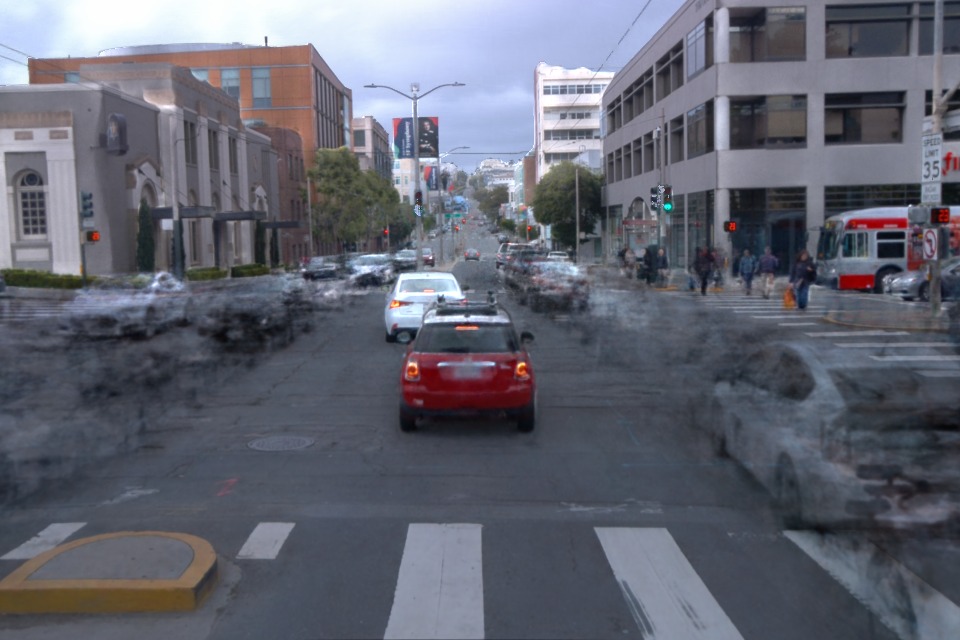}} & \frame{\includegraphics[width=1.5in,height=1in]{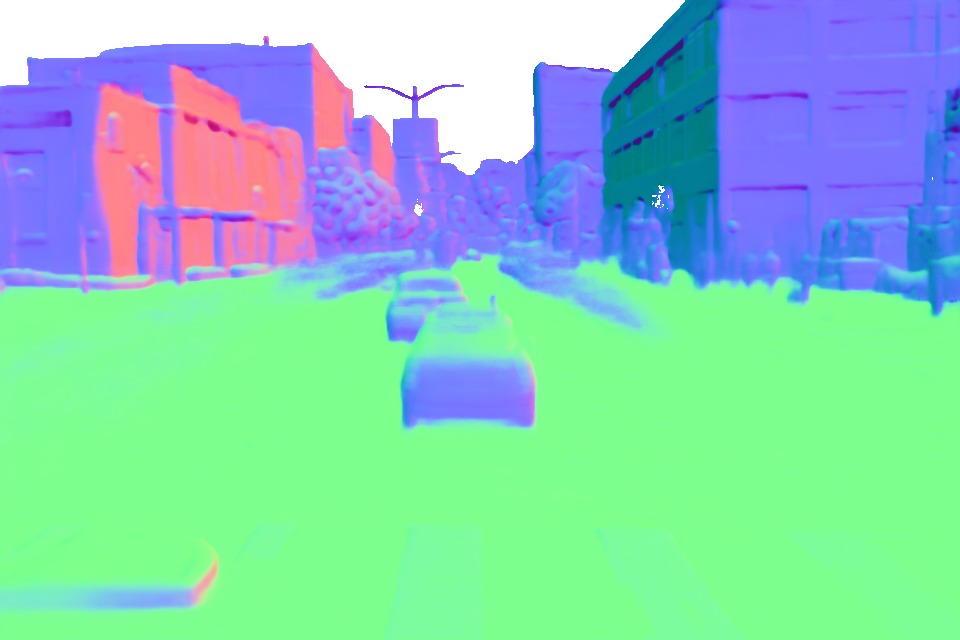}} & \frame{\includegraphics[width=1.5in,height=1in]{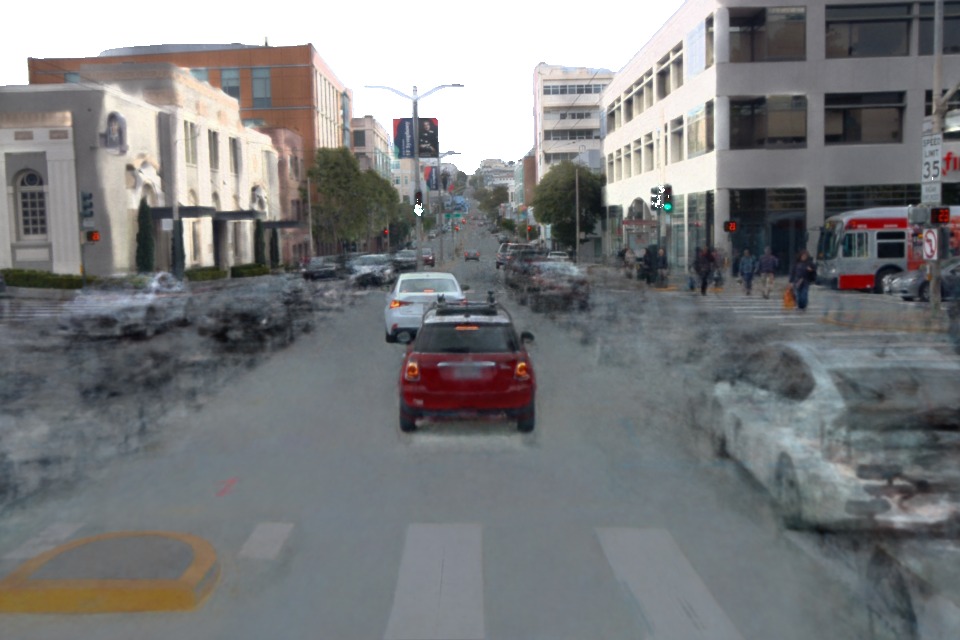}} & \frame{\includegraphics[width=1.5in,height=1in]{figures/images/null.jpg}} & \frame{\includegraphics[width=1.5in,height=1in]{figures/images/null.jpg}} \\

    \raisebox{2.5\normalbaselineskip}[0pt][0pt]{\rotatebox[origin=c]{90}{\begin{tabular}{@{}c@{}}Gen3C~\cite{Gen3C}\\+UR~\cite{unirelight}\end{tabular}}}& \frame{\includegraphics[width=1.5in,height=1in]{figures/qual_inverse_imgs_v2/dr_nvs.jpg}} & \frame{\includegraphics[width=1.5in,height=1in]{figures/images/null.jpg}} & \frame{\includegraphics[width=1.5in,height=1in]{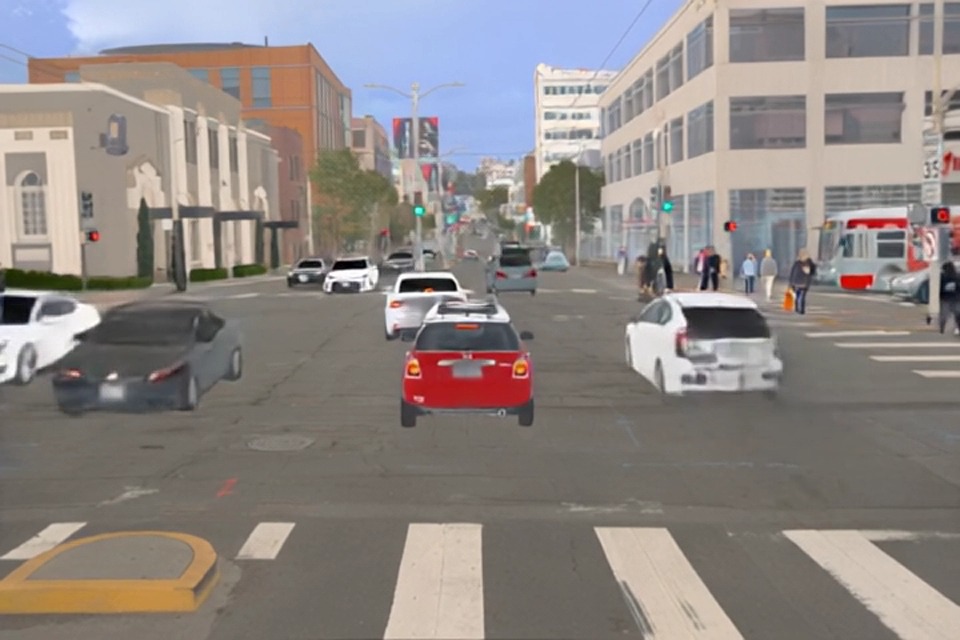}} & \frame{\includegraphics[width=1.5in,height=1in]{figures/images/null.jpg}} & \frame{\includegraphics[width=1.5in,height=1in]{figures/images/null.jpg}} \\

    \raisebox{2.5\normalbaselineskip}[0pt][0pt]{\rotatebox[origin=c]{90}{Ours}}& \frame{\includegraphics[width=1.5in,height=1in]{figures/qual_inverse_imgs_v2/ours_nvs_tau8_raster.jpg}} & \frame{\includegraphics[width=1.5in,height=1in]{figures/qual_inverse_imgs_v2/ours_normal.jpg}} & \frame{\includegraphics[width=1.5in,height=1in]{figures/qual_inverse_imgs_v2/ours_albedo.jpg}} & \frame{\includegraphics[width=1.5in,height=1in]{figures/qual_inverse_imgs_v2/ours_roughness.jpg}} & \frame{\includegraphics[width=1.5in,height=1in]{figures/qual_inverse_imgs_v2/ours_metallic.jpg}} \\
    \end{tabular}%
    }
    \caption{\textbf{Qualitative comparison of inverse rendering with IRGS, UrbanIR, and Gen3C+UR on a dynamic WOD scene.} IRGS~\cite{IRGS} and UrbanIR~\cite{urbanir} fail to reconstruct the moving vehicles.}
    \label{fig:qual_unirelight_inverse}
\end{figure*}

\paragraph{\textbf{Qualitative Comparison on Real-World Scenes.}}
We then compare our method with IRGS, UrbanIR, and Gen3C+UR on real-world scenes in Figs.~\ref{fig:qual_unirelight} and~\ref{fig:qual_unirelight_inverse}. IRGS and UrbanIR cannot handle dynamic scenes properly. Gen3C+UR fails to reproduce the correct lighting effects due to the lack of an explicit relightable 3D representation. In contrast, our method achieves controllable and realistic relighting of dynamic scenes.
 
\paragraph{\textbf{Novel-View Synthesis on WOD.}}
Tab.~\ref{tab:quant_wod_nvs} reports novel-view synthesis on the four WOD scenes. \method achieves the highest PSNR but not the best SSIM and LPIPS. As discussed in the main paper, we attribute this gap mainly to the residual Monte Carlo noise from the limited number of rays per pixel in our PBR pass.
\begin{table}[t]
    \centering
    \small
    \caption{\textbf{Novel-view synthesis on WOD.} Metrics are averaged over the four WOD scenes and computed over non-sky regions. The best and second-best results are bold and underlined, respectively. For this experiment, our method traces $N_r{=}4096$ rays per pixel.}
    \label{tab:quant_wod_nvs}
    \setlength{\tabcolsep}{5pt}
    \begin{tabular}{lccc}
    \toprule
    Method & PSNR$\uparrow$ & SSIM$\uparrow$ & LPIPS$\downarrow$ \\
    \midrule
    IRGS~\cite{IRGS} & 22.88 & 0.811 & 0.376 \\
    UrbanIR~\cite{urbanir} & 21.48 & 0.861 & 0.187 \\
    InvRGB+L~\cite{invrgbl} & \underline{24.97} & \textbf{0.893} & \underline{0.182} \\
    Gen3C~\cite{Gen3C}+DR~\cite{diffusionrenderer} & 24.79 & \underline{0.871} & \textbf{0.166} \\
    Gen3C~\cite{Gen3C}+UR~\cite{unirelight} & 24.79 & \underline{0.871} & \textbf{0.166} \\
    AW4D~\cite{autoweather4d} & 23.54 & 0.802 & 0.192 \\
    Ours & \textbf{25.62} & 0.835 & 0.235 \\
    \bottomrule
    \end{tabular}
\end{table}
 
\begin{figure*}[t]
    \centering
    \includegraphics[width=\linewidth]{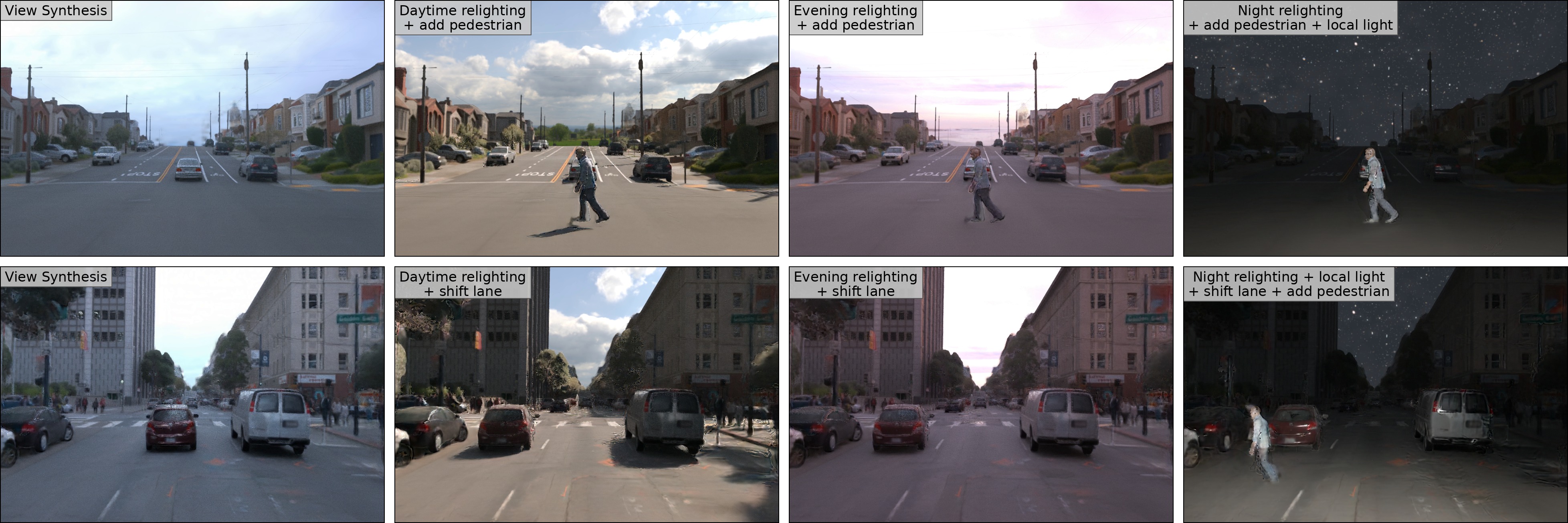}
    \vspace{-1.5em}
    \caption{\textbf{Additional qualitative results on real-world scenes.} Our method supports controllable relighting and diverse edits to 3D scenes.}
    \label{fig:qual_applications_real}
\end{figure*}

\paragraph{\textbf{Applications on WOD.}}
We further demonstrate our method's editing capabilities on two additional WOD scenes in Fig.~\ref{fig:qual_applications_real}. Our explicit 3D representation enables controllable relighting and compositional edits, including moving objects and inserting a pedestrian. These results highlight the flexibility of our representation for generating diverse driving scenarios.

\section{Additional Ablation Studies}
\label{supp:ablation}
\begin{figure*}[t]
    \centering
    \setlength{\tabcolsep}{3pt}
    \resizebox{1.0\textwidth}{!}{%
    \begin{tabular}{@{}lcccccccc@{}}
 & (a) No Gen. Render & (b) No PBR Anchor & (c) No VAE Residual & (d) No Gen. Prior & (e) No Multi-stage & (f) No Gen. Refine & (g) Ours (Full Method) & (h) GT \\[0.2em]

    \raisebox{2.5\normalbaselineskip}[0pt][0pt]{\rotatebox[origin=c]{90}{Normal}} &
    \frame{\includegraphics[width=1.5in,height=1in]{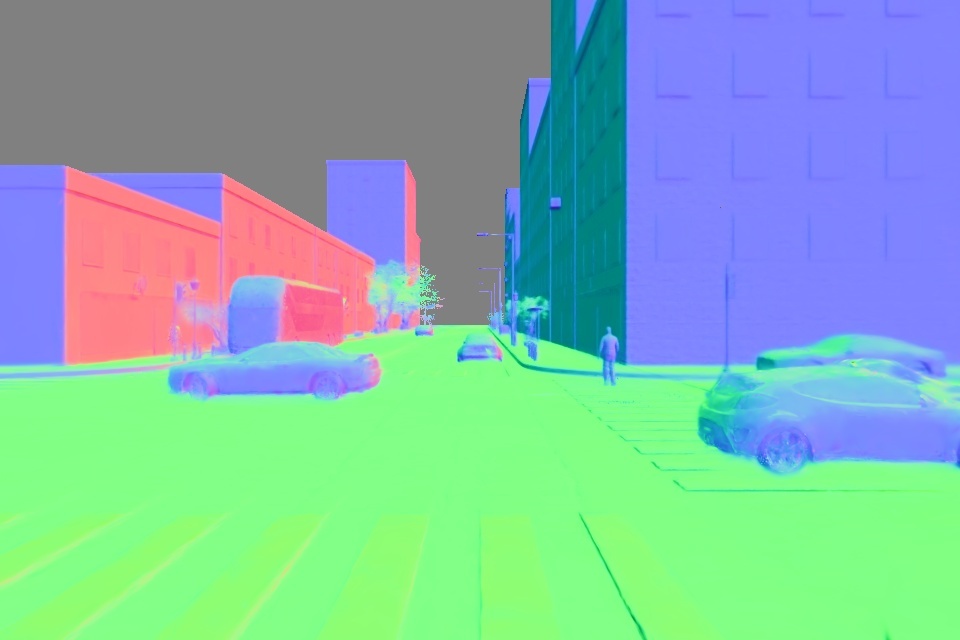}} &
    \frame{\includegraphics[width=1.5in,height=1in]{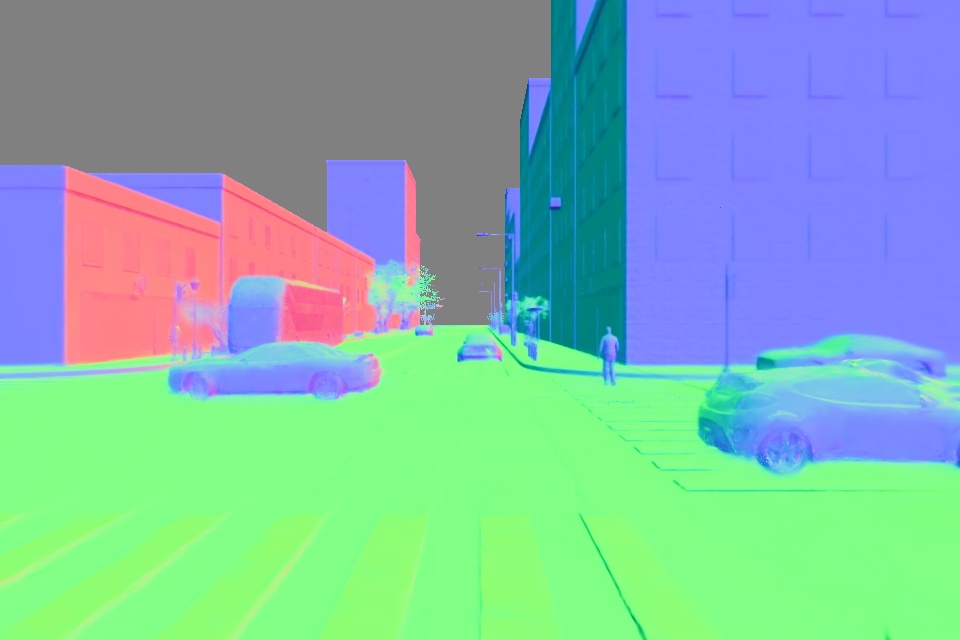}} &
    \frame{\includegraphics[width=1.5in,height=1in]{figures/supp_assets/ablation_qual_v5/ours_normal.jpg}} &
    \frame{\includegraphics[width=1.5in,height=1in]{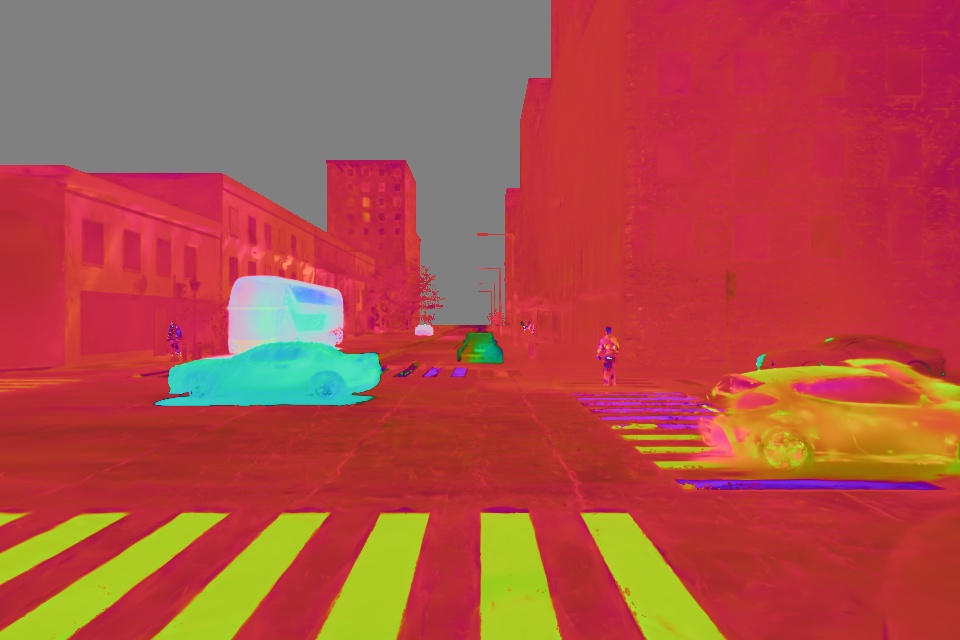}} &
    \frame{\includegraphics[width=1.5in,height=1in]{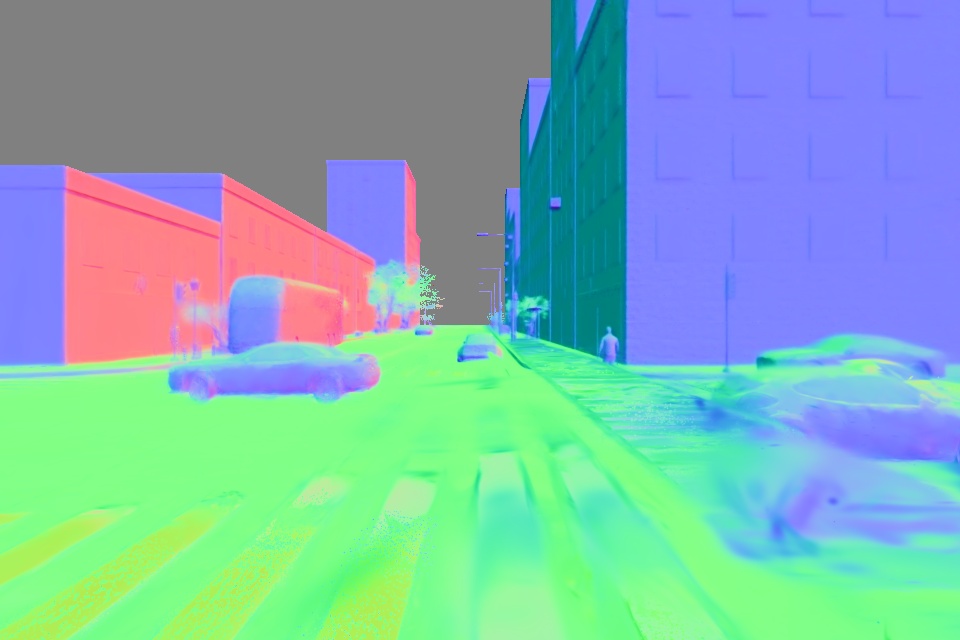}} &
    \frame{\includegraphics[width=1.5in,height=1in]{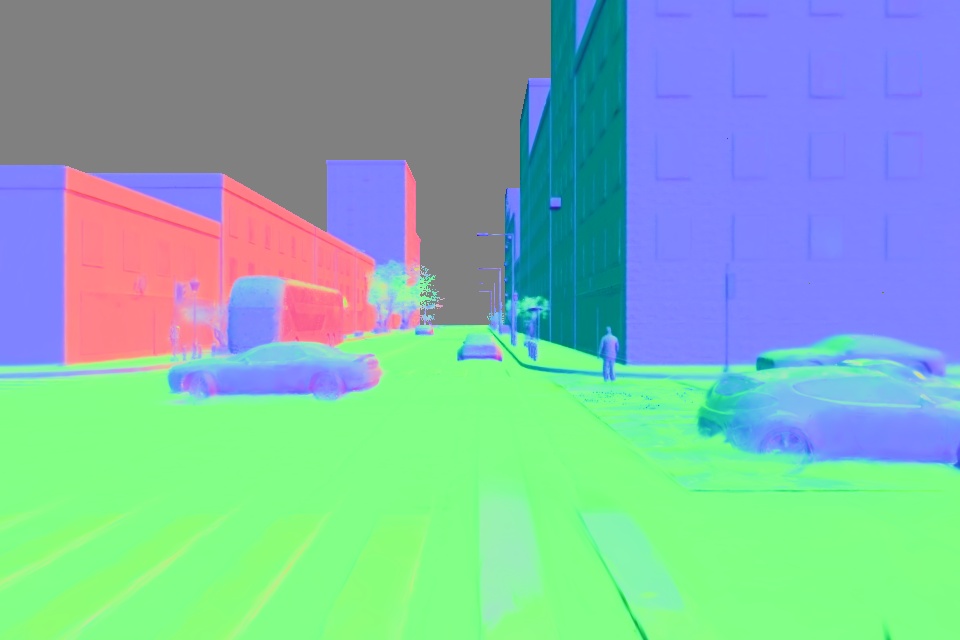}} &
    \frame{\includegraphics[width=1.5in,height=1in]{figures/supp_assets/ablation_qual_v5/ours_normal.jpg}} &
    \frame{\includegraphics[width=1.5in,height=1in]{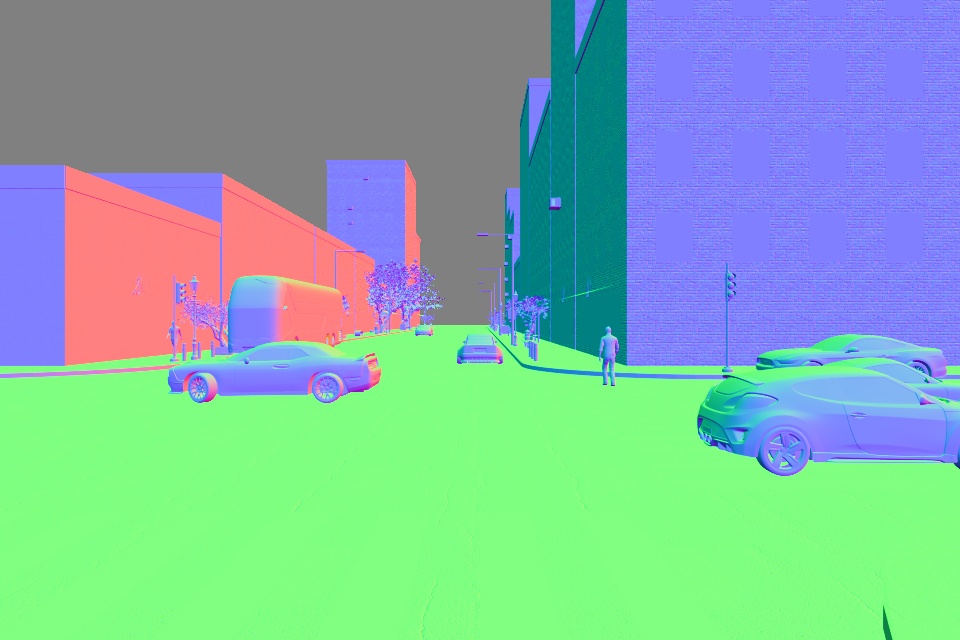}}\\

    \raisebox{2.5\normalbaselineskip}[0pt][0pt]{\rotatebox[origin=c]{90}{Albedo}} &
    \frame{\includegraphics[width=1.5in,height=1in]{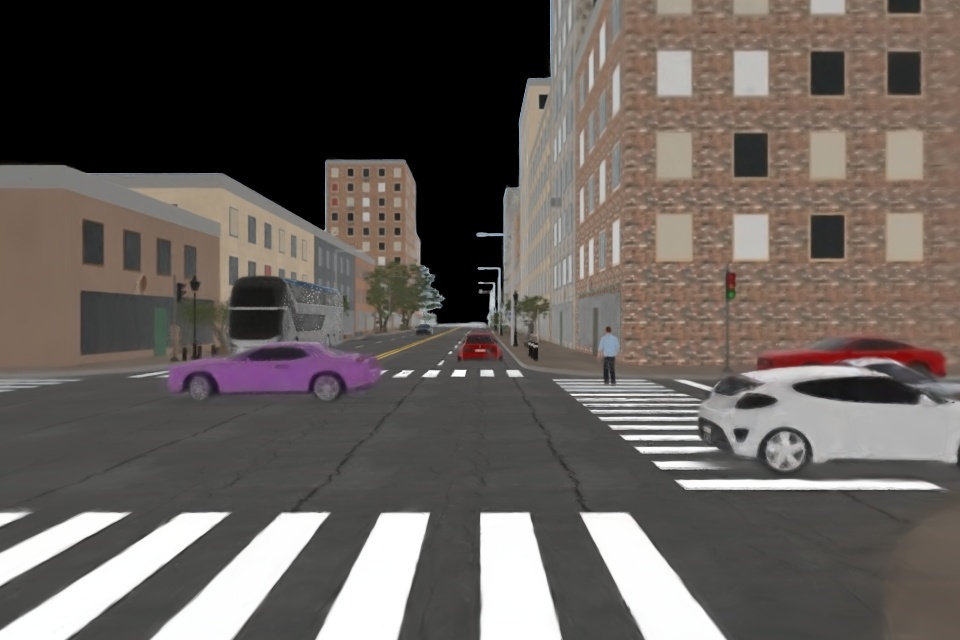}} &
    \frame{\includegraphics[width=1.5in,height=1in]{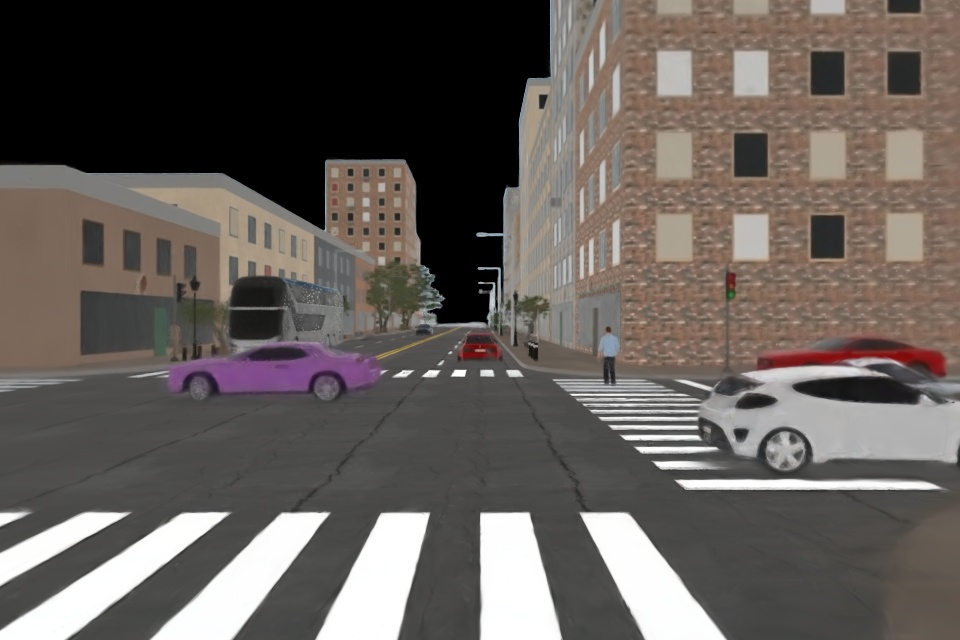}} &
    \frame{\includegraphics[width=1.5in,height=1in]{figures/supp_assets/ablation_qual_v5/ours_albedo.jpg}} &
    \frame{\includegraphics[width=1.5in,height=1in]{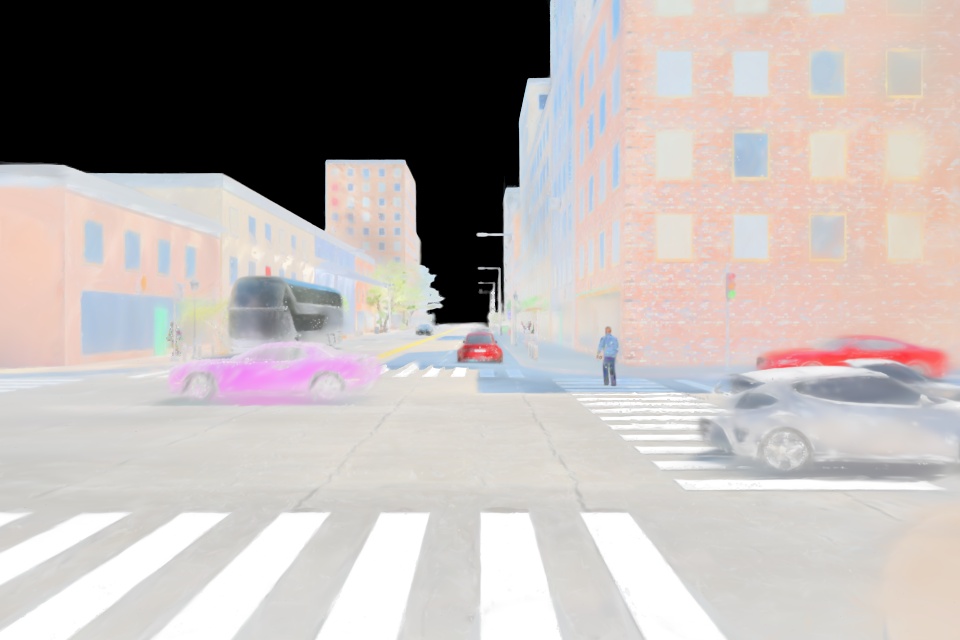}} &
    \frame{\includegraphics[width=1.5in,height=1in]{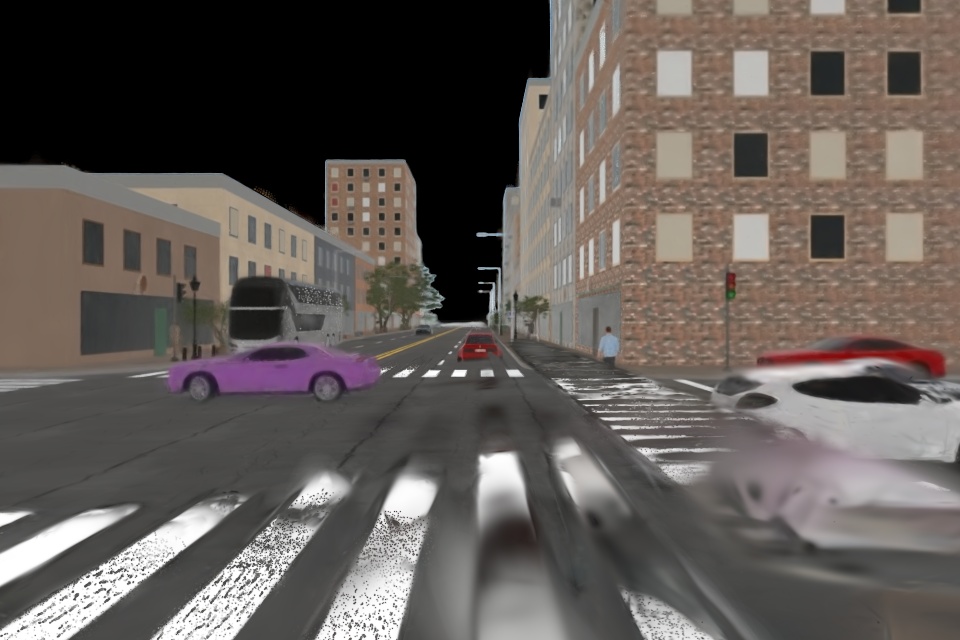}} &
    \frame{\includegraphics[width=1.5in,height=1in]{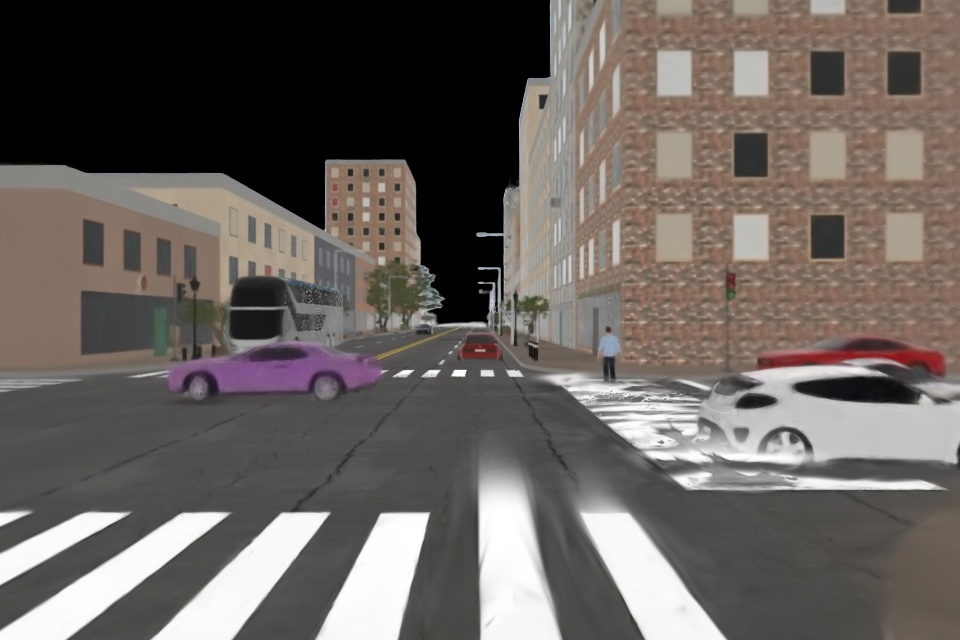}} &
    \frame{\includegraphics[width=1.5in,height=1in]{figures/supp_assets/ablation_qual_v5/ours_albedo.jpg}} &
    \frame{\includegraphics[width=1.5in,height=1in]{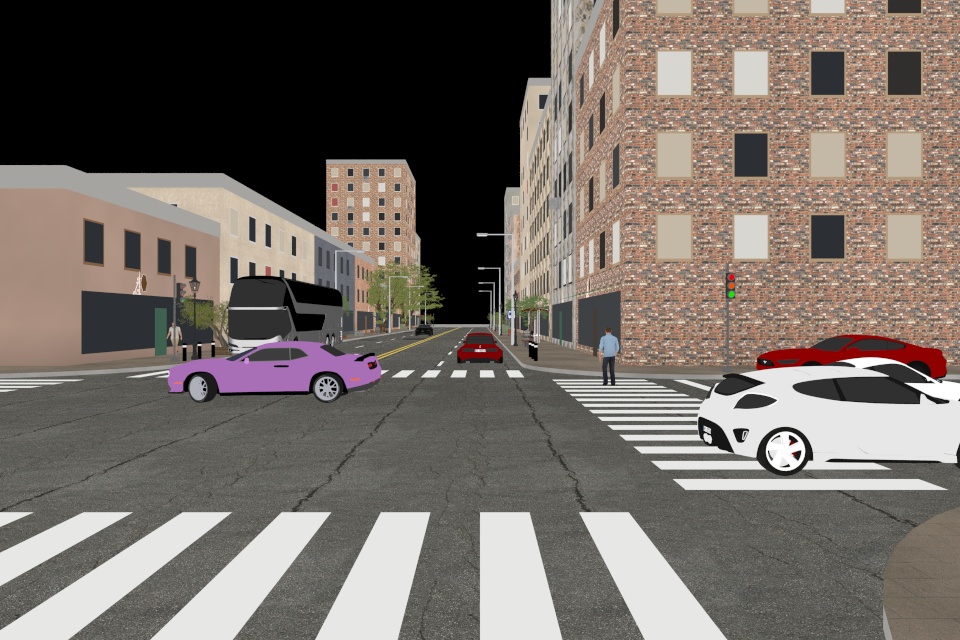}}\\

    \raisebox{2.5\normalbaselineskip}[0pt][0pt]{\rotatebox[origin=c]{90}{Roughness}} &
    \frame{\includegraphics[width=1.5in,height=1in]{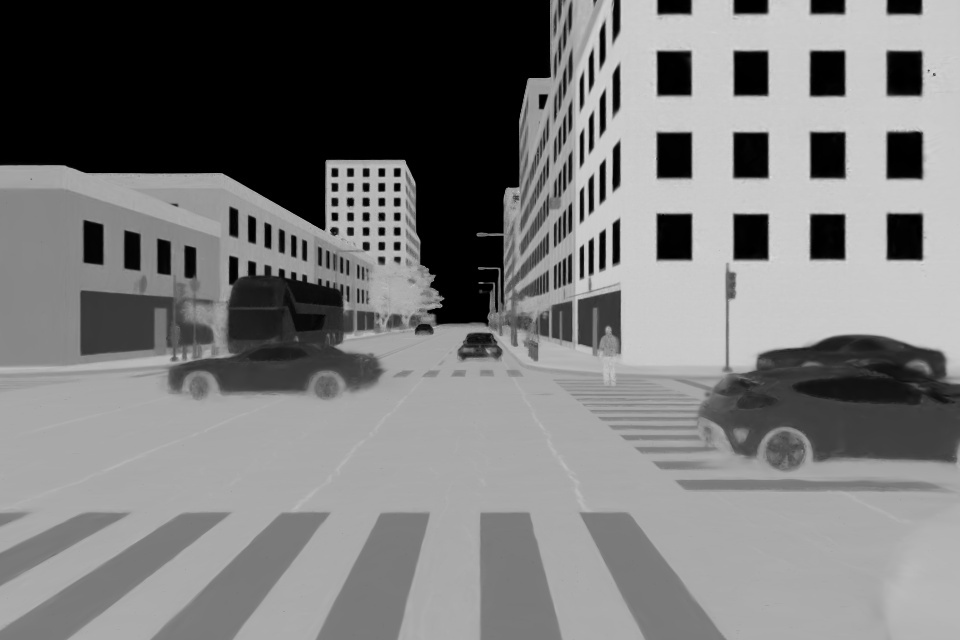}} &
    \frame{\includegraphics[width=1.5in,height=1in]{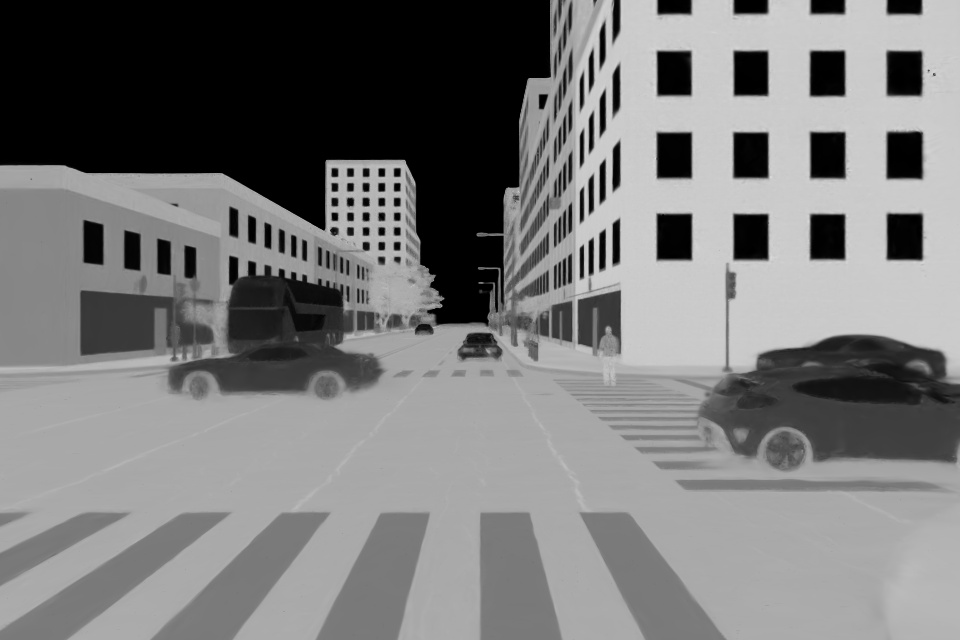}} &
    \frame{\includegraphics[width=1.5in,height=1in]{figures/supp_assets/ablation_qual_v5/ours_roughness.jpg}} &
    \frame{\includegraphics[width=1.5in,height=1in]{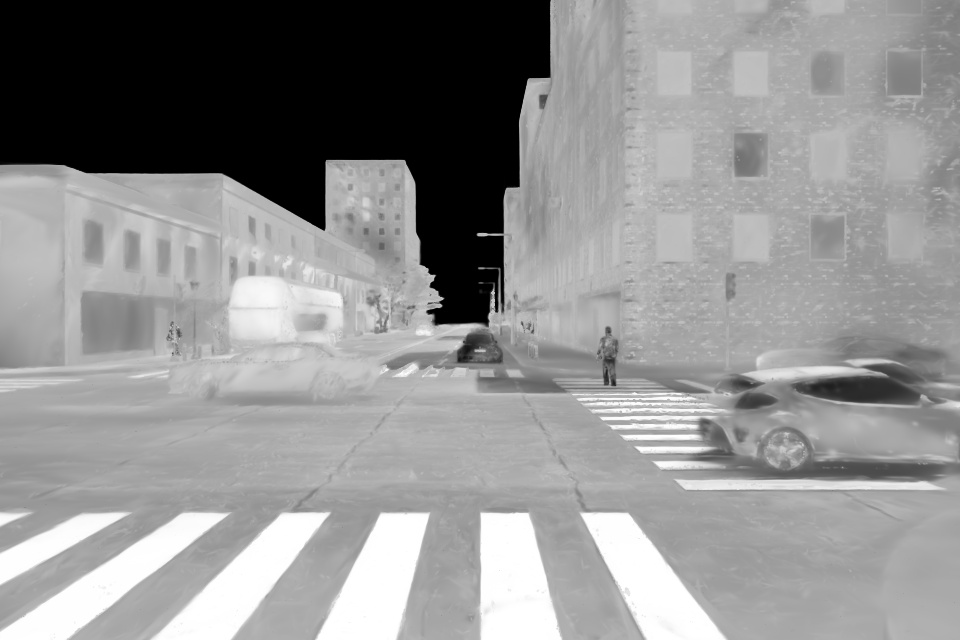}} &
    \frame{\includegraphics[width=1.5in,height=1in]{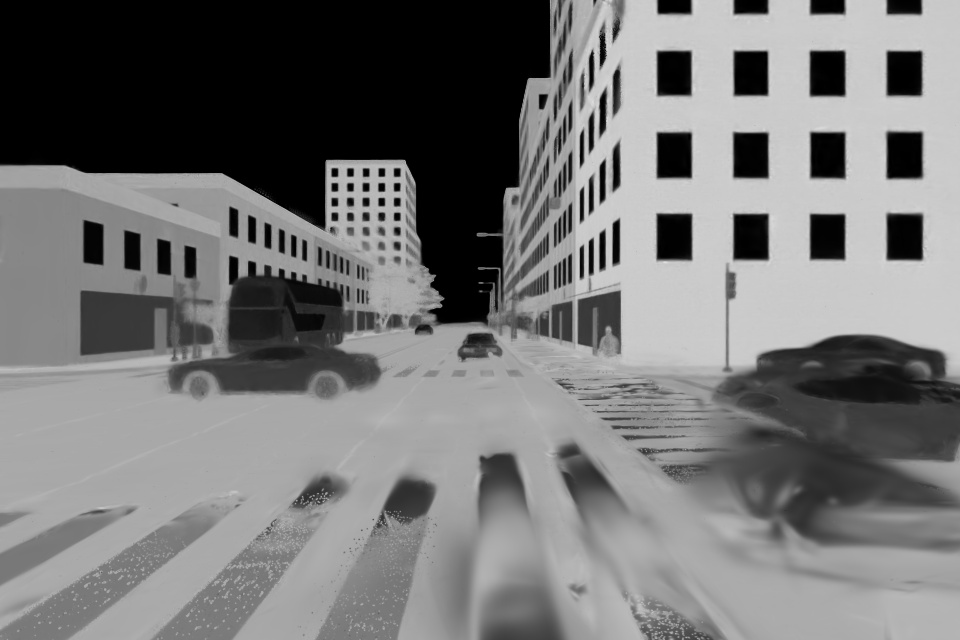}} &
    \frame{\includegraphics[width=1.5in,height=1in]{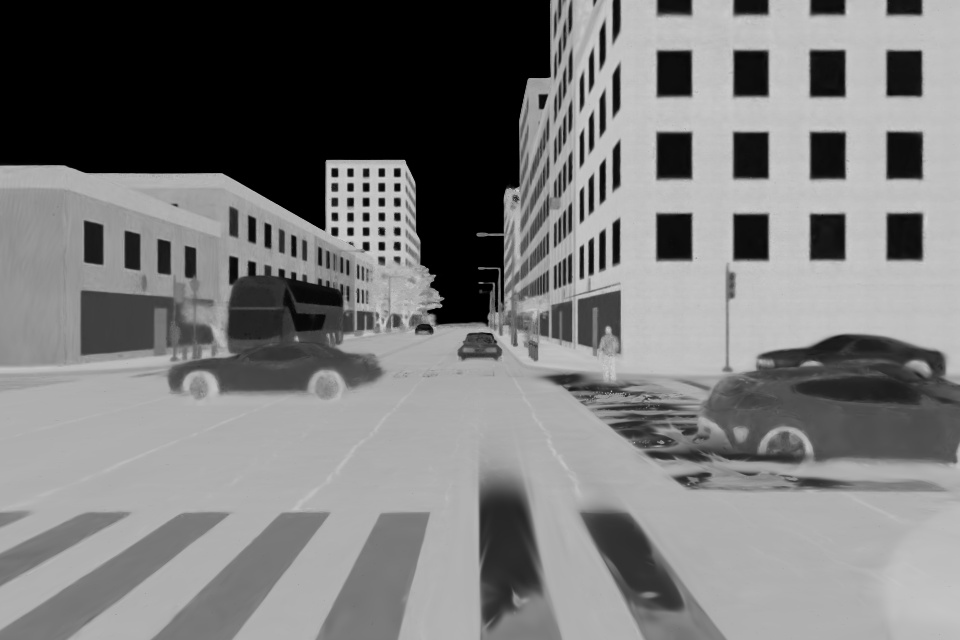}} &
    \frame{\includegraphics[width=1.5in,height=1in]{figures/supp_assets/ablation_qual_v5/ours_roughness.jpg}} &
    \frame{\includegraphics[width=1.5in,height=1in]{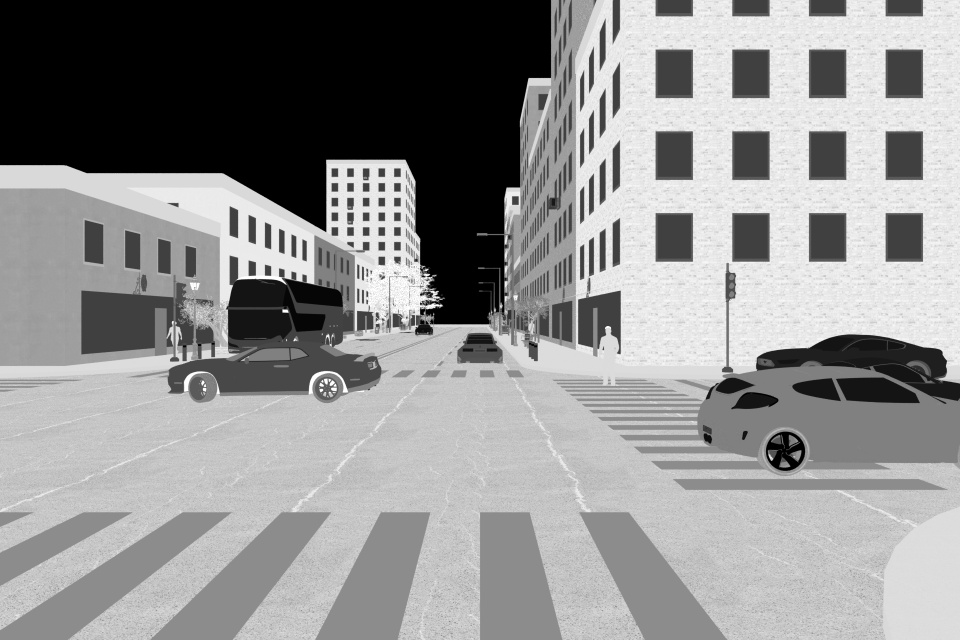}}\\

    \raisebox{2.5\normalbaselineskip}[0pt][0pt]{\rotatebox[origin=c]{90}{Metallic}} &
    \frame{\includegraphics[width=1.5in,height=1in]{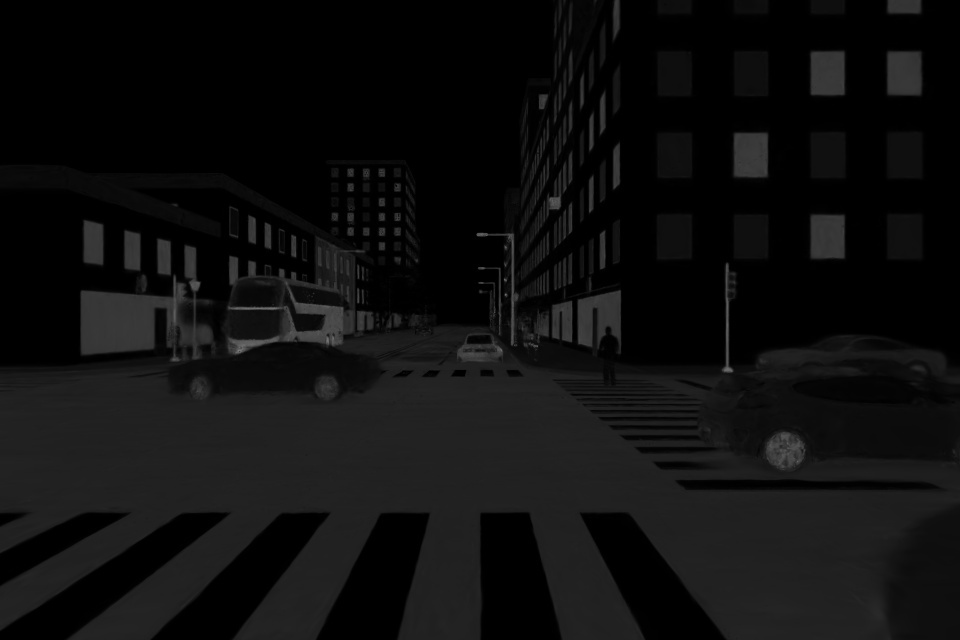}} &
    \frame{\includegraphics[width=1.5in,height=1in]{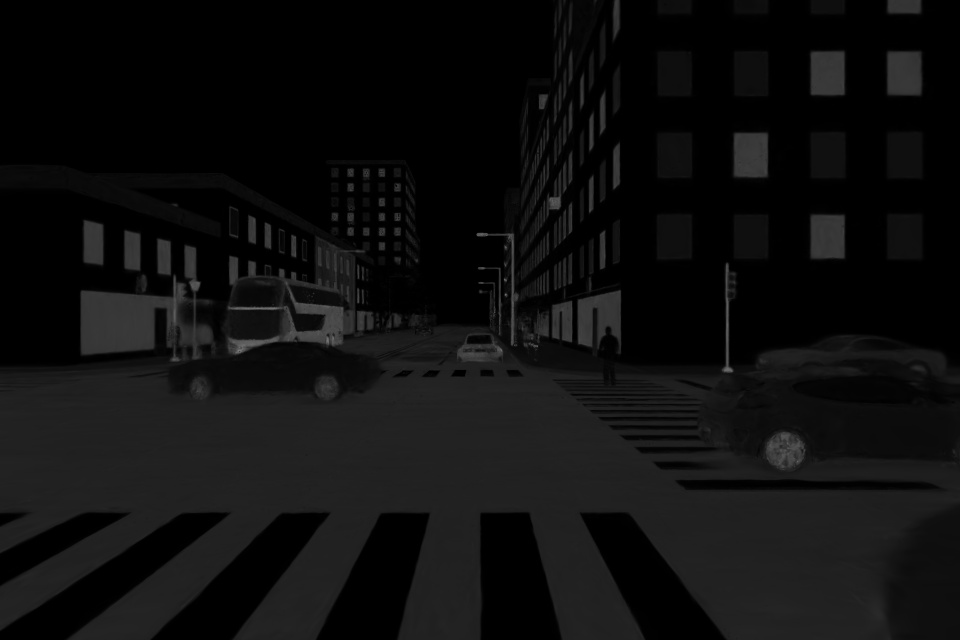}} &
    \frame{\includegraphics[width=1.5in,height=1in]{figures/supp_assets/ablation_qual_v5/ours_metallic.jpg}} &
    \frame{\includegraphics[width=1.5in,height=1in]{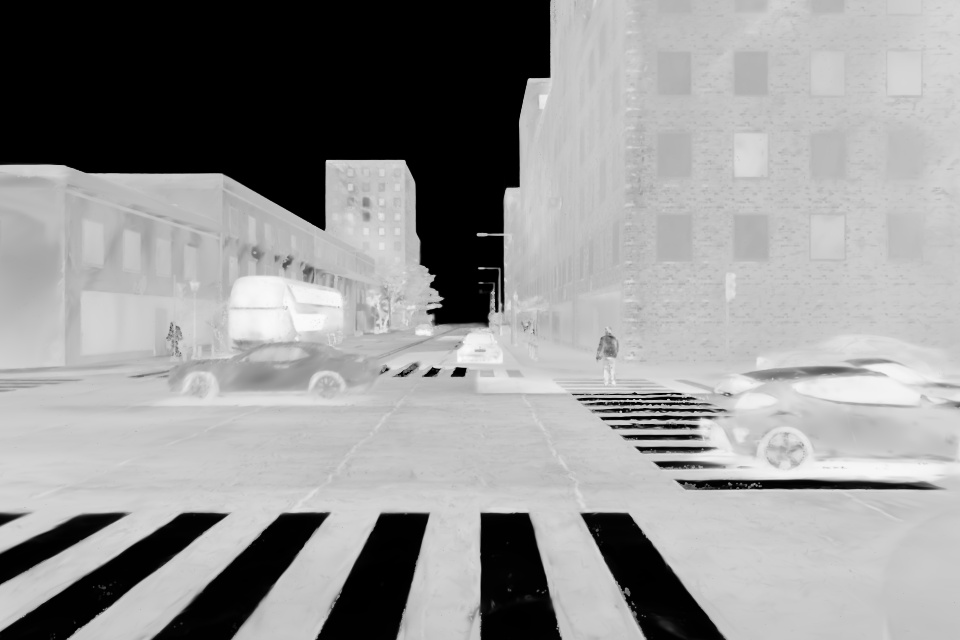}} &
    \frame{\includegraphics[width=1.5in,height=1in]{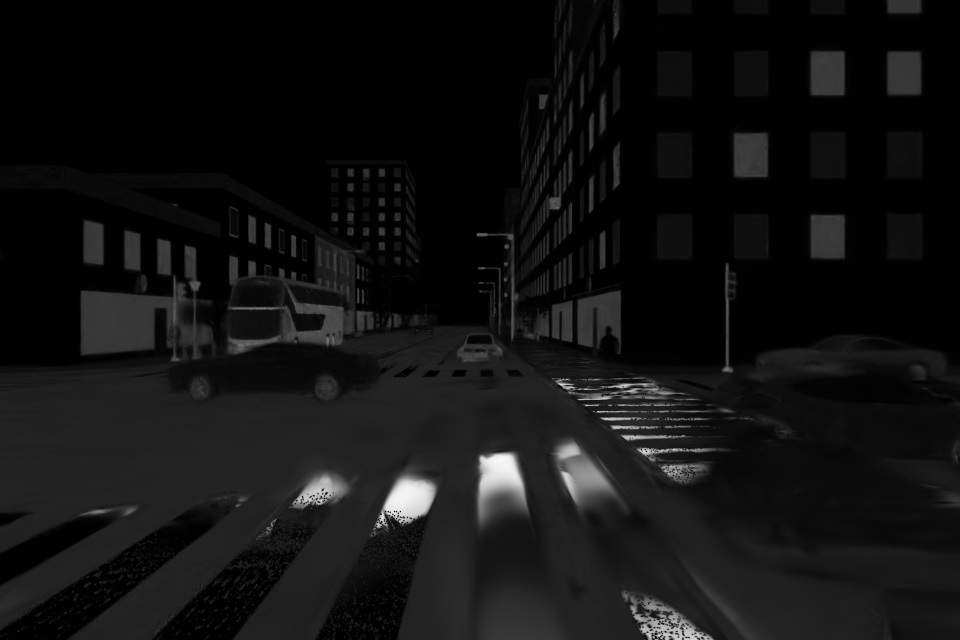}} &
    \frame{\includegraphics[width=1.5in,height=1in]{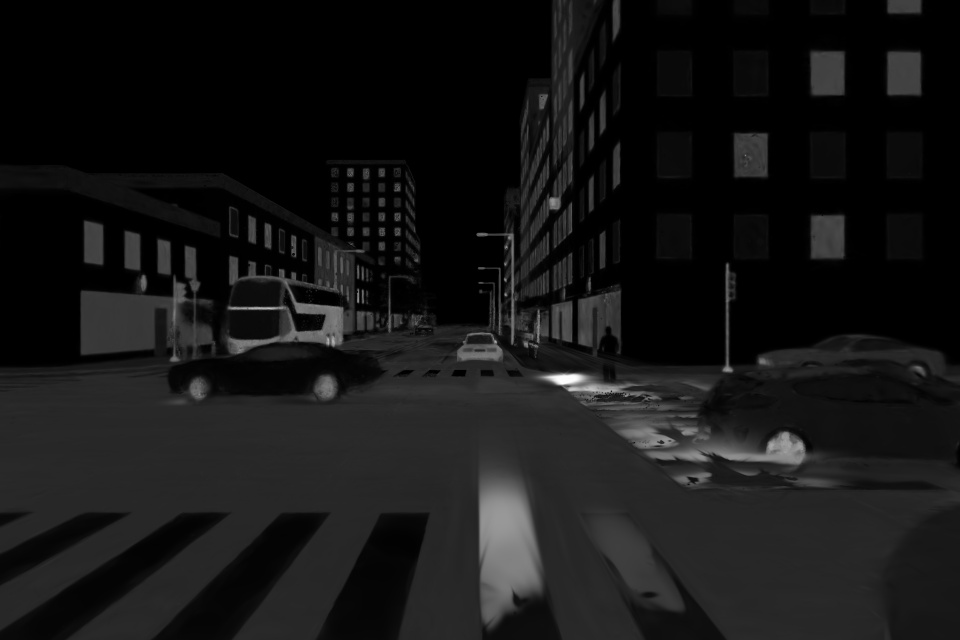}} &
    \frame{\includegraphics[width=1.5in,height=1in]{figures/supp_assets/ablation_qual_v5/ours_metallic.jpg}} &
    \frame{\includegraphics[width=1.5in,height=1in]{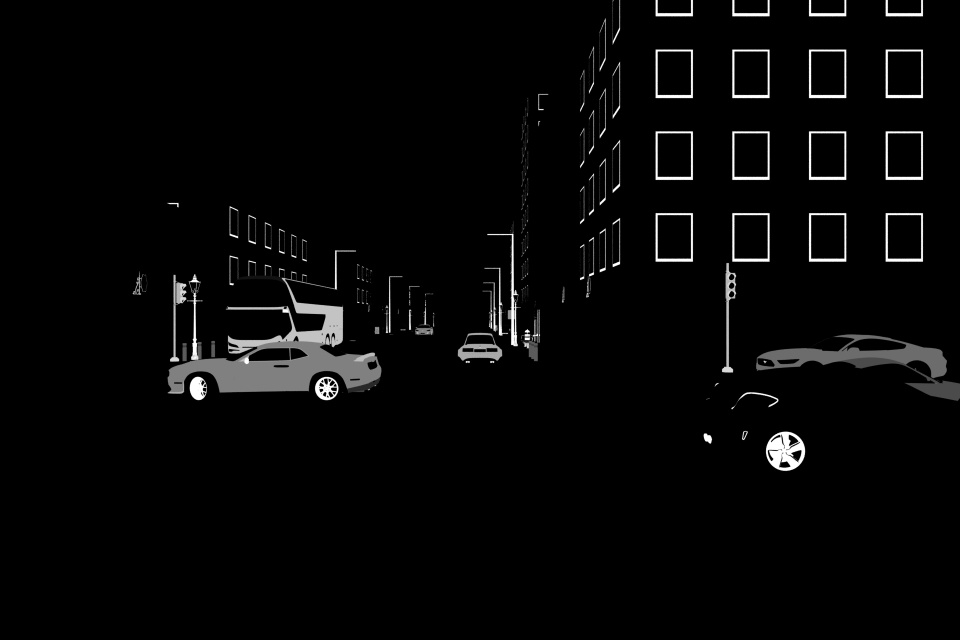}}\\

    \raisebox{2.5\normalbaselineskip}[0pt][0pt]{\rotatebox[origin=c]{90}{\makecell[c]{NVS + \\ Relighting}}} &
    \frame{\includegraphics[width=1.5in,height=1in]{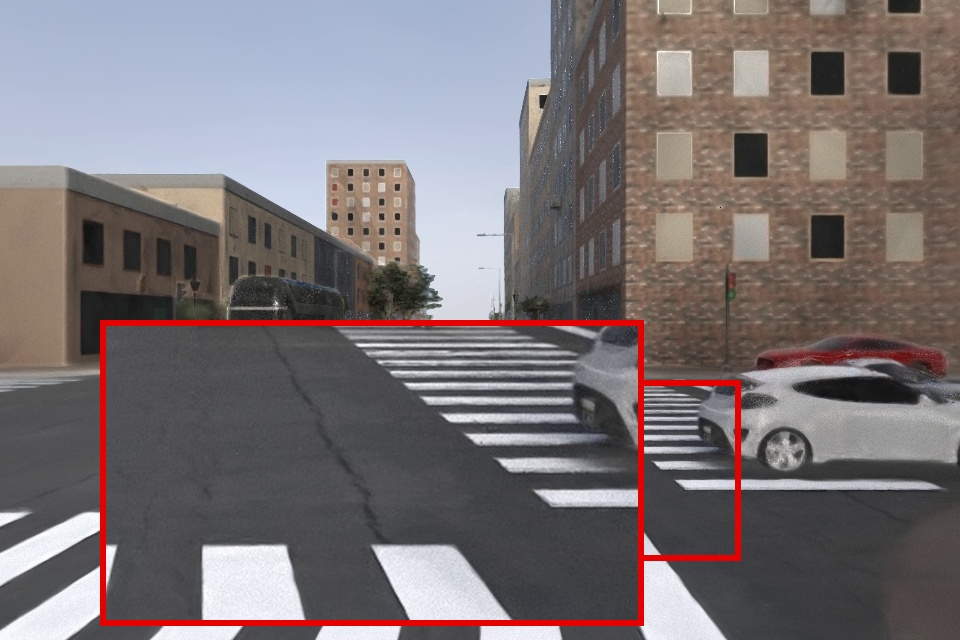}} &
    \frame{\includegraphics[width=1.5in,height=1in]{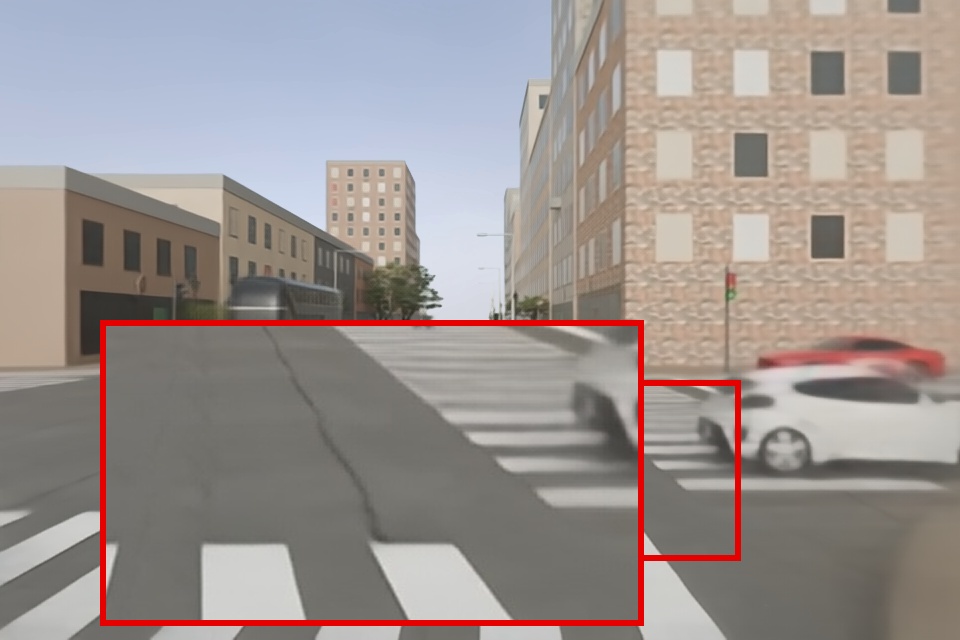}} &
    \frame{\includegraphics[width=1.5in,height=1in]{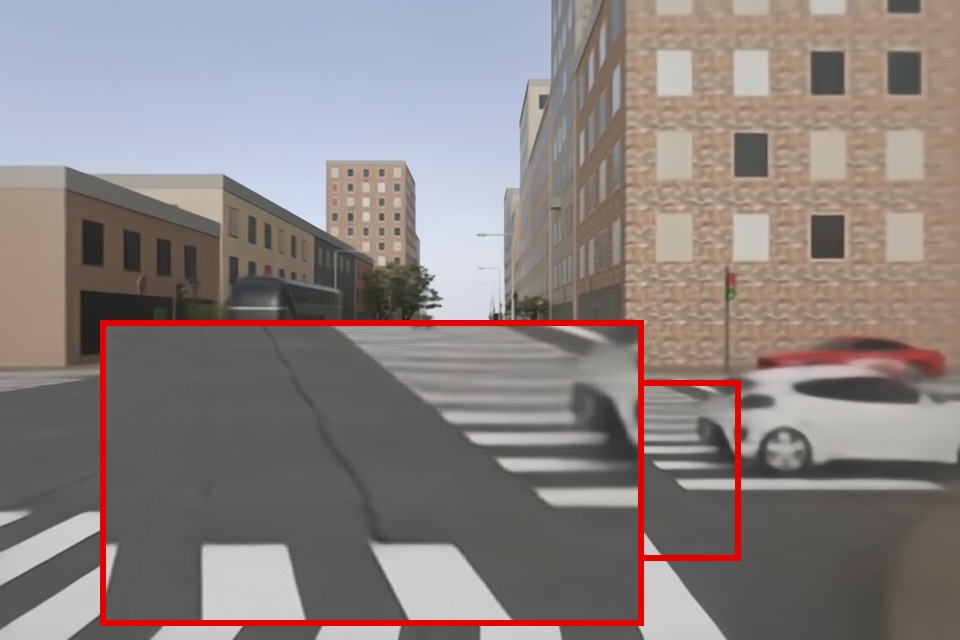}} &
    \frame{\includegraphics[width=1.5in,height=1in]{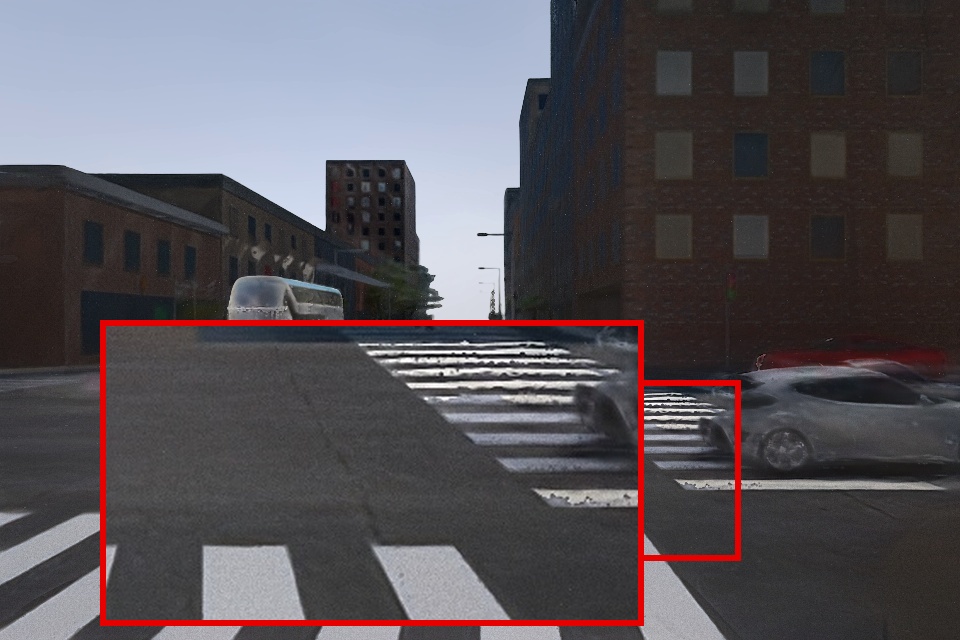}} &
    \frame{\includegraphics[width=1.5in,height=1in]{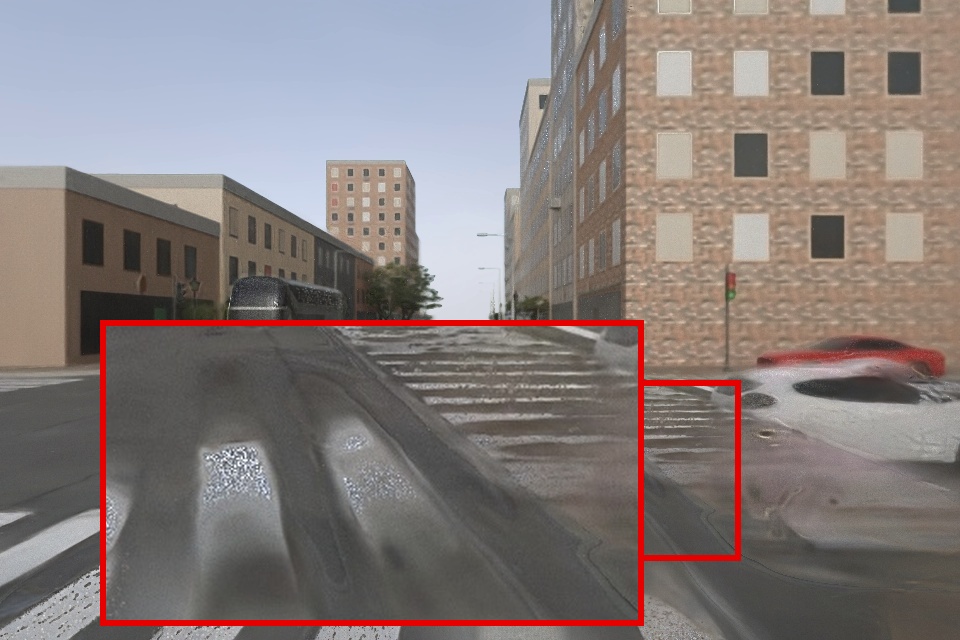}} &
    \frame{\includegraphics[width=1.5in,height=1in]{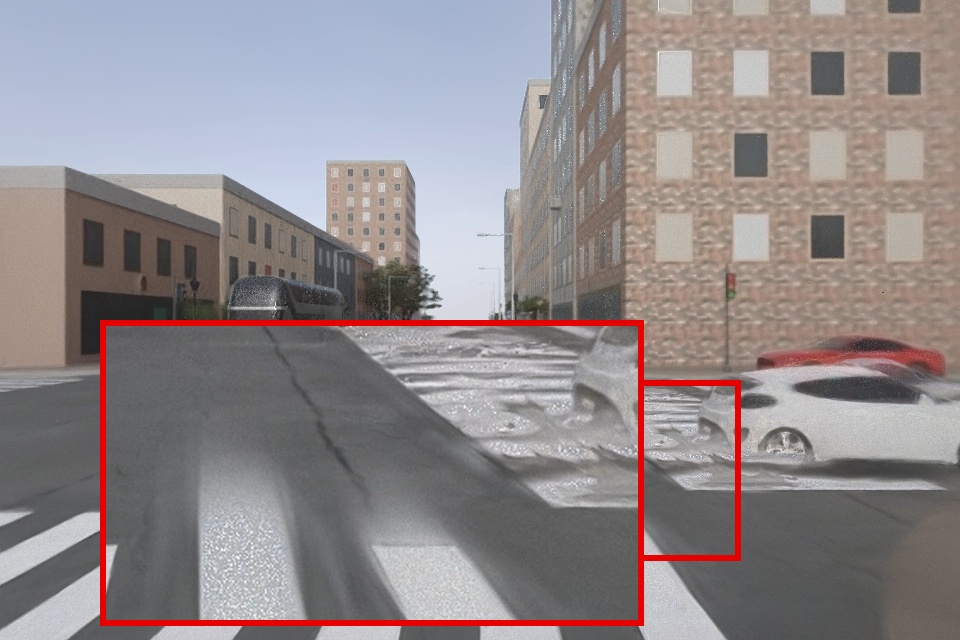}} &
    \frame{\includegraphics[width=1.5in,height=1in]{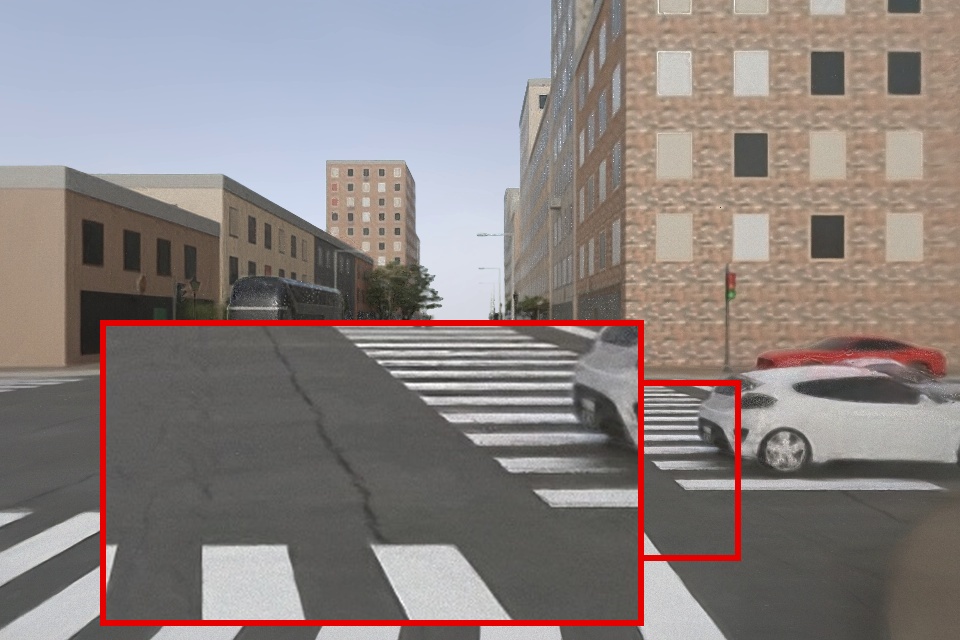}} &
    \frame{\includegraphics[width=1.5in,height=1in]{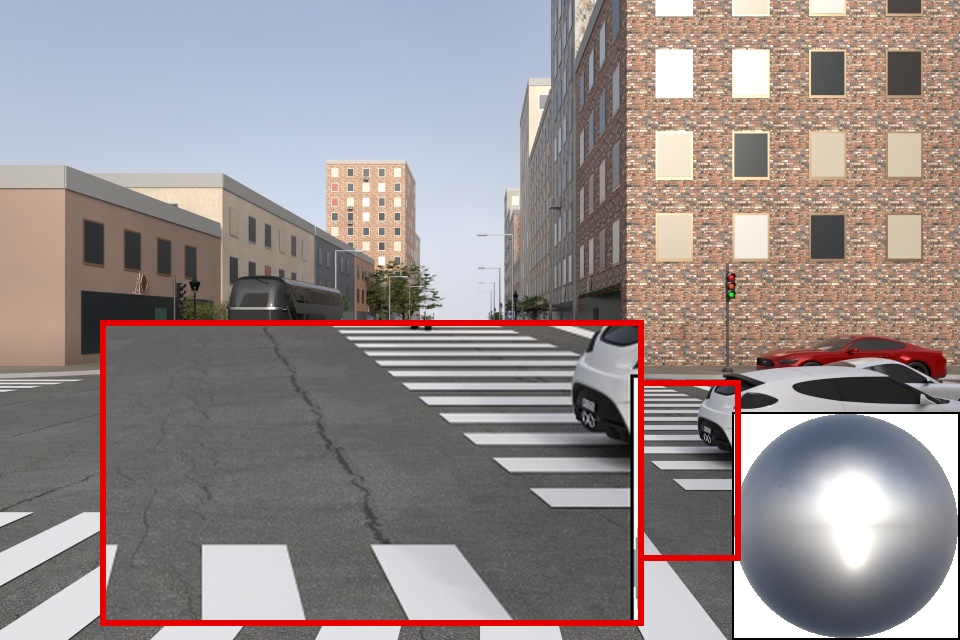}}\\

    \end{tabular}%
    }
    \vspace{-3mm}
    \caption{\textbf{Qualitative ablation study on the synthetic dataset.} Variants (a)--(c) ablate the forward rendering and share the scene properties of our full method. Variants (d)--(f) ablate the inverse rendering. The \textcolor{red}{\textbf{red}} zoom-in patches highlight the artifacts on the crosswalk.}
    \label{fig:qual_ablation}
\end{figure*}

A qualitative comparison of the ablation study is presented in Fig.~\ref{fig:qual_ablation}. Variants (a)--(c) ablate the forward rendering and share the scene properties of our full method, while variants (d)--(f) ablate the inverse rendering. 
 
\paragraph{\textbf{Ablation of Forward Rendering.}}
For forward rendering, bypassing the generative rendering pass (Fig.~\ref{fig:qual_ablation} (a) No Gen. Render) exposes rendering artifacts in the relighting results. 
Generating from pure noise without PBR anchoring (Fig.~\ref{fig:qual_ablation} (b) No PBR Anchor) fails to precisely simulate the target lighting. 
Using the SDEdit output without the VAE residual (Fig.~\ref{fig:qual_ablation} (c) No VAE Residual) blurs the rendering details (e.g., on the crosswalk). 

\paragraph{\textbf{Ablation of Inverse Rendering.}}
For inverse rendering, without DR priors (Fig.~\ref{fig:qual_ablation} (d) No Gen. Prior), the model fails to properly decompose the scene geometry, materials, and lighting due to the ill-posed nature of inverse rendering, leading to completely wrong relighting results. 
Replacing the multi-stage strategy with single-stage optimization (Fig.~\ref{fig:qual_ablation} (e) No Multi-stage) makes the optimization fragile, resulting in reconstruction artifacts.
When generative refinement is skipped (Fig.~\ref{fig:qual_ablation} (f) No Gen. Refine), artifacts caused by the temporal inconsistency of material priors are baked into the materials (e.g., the crosswalk) and propagate to the relighting results.
Our full method (Fig.~\ref{fig:qual_ablation} (g) Ours) recovers accurate scene properties and produces high-quality relighting, validating the effectiveness of each proposed design choice.

\begin{figure*}[t]
\centering
\includegraphics[width=\linewidth]{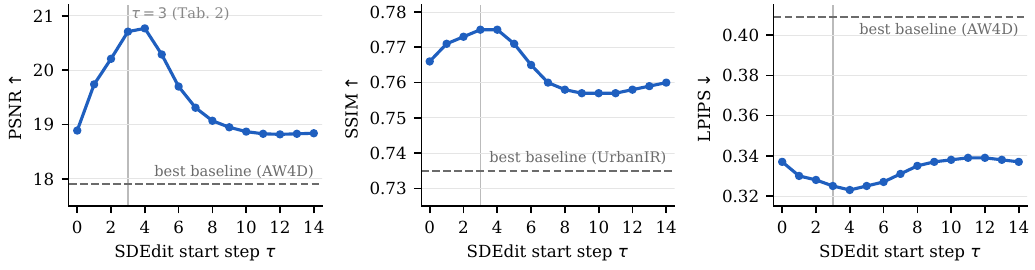}
\vspace{-1.5em}
\caption{\textbf{Ablation of the SDEdit starting step $\diffusionTime$ of generative rendering.} Novel-view relighting on the synthetic dataset for $\diffusionTime = 0, \dots, 14$. Dashed lines denote the best baseline for each metric in Tab.~\ref{tab:quant_synthetic}, and the vertical line marks $\diffusionTime=3$ used in the main paper. A smaller $\diffusionTime$ starts from a higher noise level and lets the generative model dominate, while a larger $\diffusionTime$ keeps the result close to the PBR rendering.}
\label{fig:tau_sweep}
\end{figure*}

\paragraph{\textbf{Noise Level of Generative Rendering.}}
We further ablate the SDEdit starting step $\diffusionTime$ of the generative rendering pass in Fig.~\ref{fig:tau_sweep}, using the same novel-view relighting setting as Tab.~\ref{tab:quant_synthetic}. A small $\diffusionTime$ adds strong noise to the PBR rendering, so the result is dominated by the generative model, whereas a large $\diffusionTime$ keeps the result close to the PBR rendering. An intermediate $\diffusionTime$, which balances the PBR rendering and the generative model, performs best. Moreover, every $\diffusionTime$ outperforms the best baseline on all three metrics, so our improvement does not depend on a particular choice of $\diffusionTime$.
 
\begin{figure}[t]
    \centering
    \setlength{\tabcolsep}{1pt}
    \renewcommand{\arraystretch}{1}
    \begin{tabular}{cc}
        \includegraphics[width=0.49\linewidth]{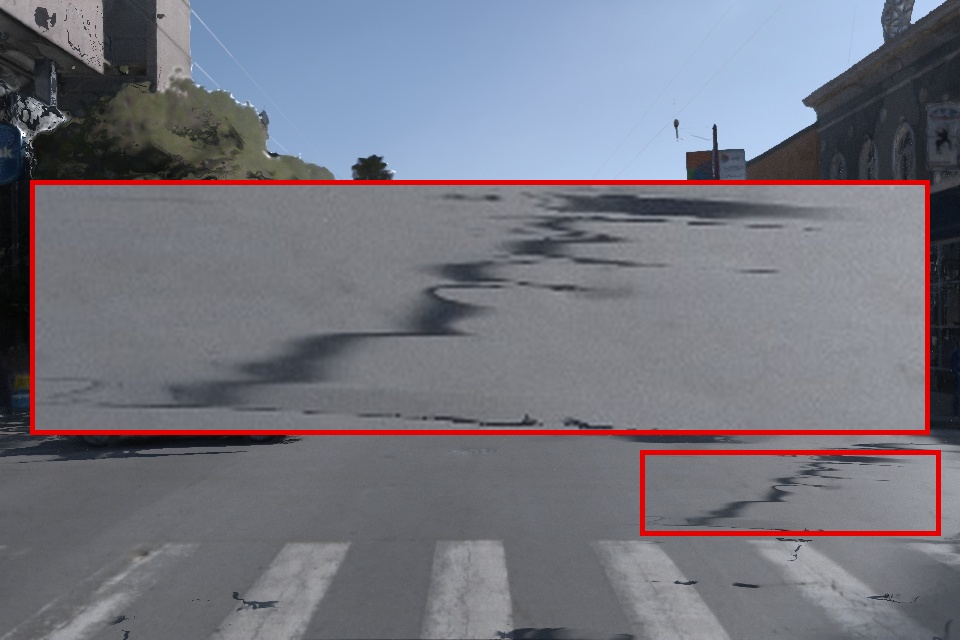} &
        \includegraphics[width=0.49\linewidth]{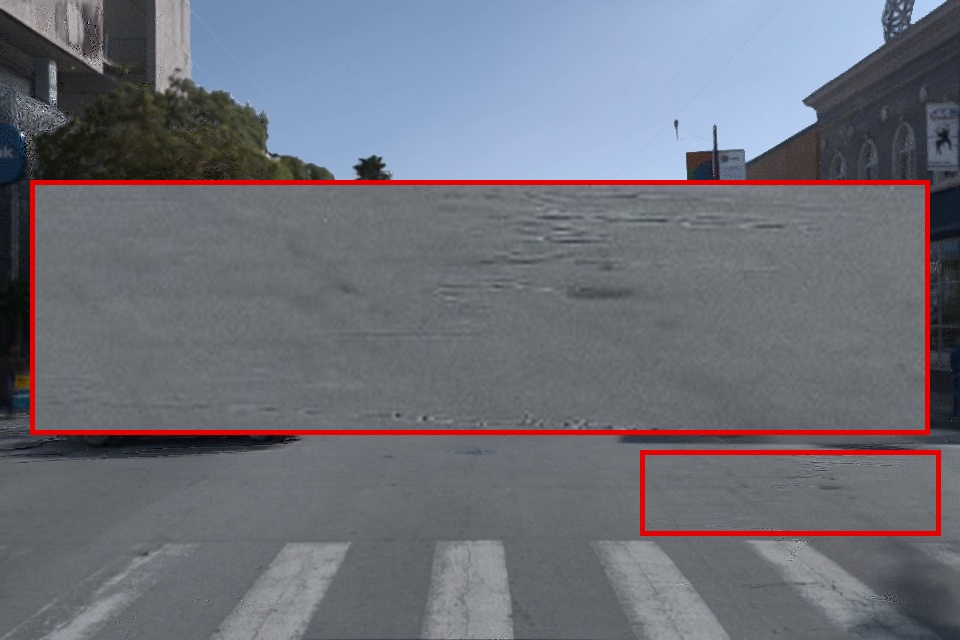} \\[1pt]
        \includegraphics[width=0.49\linewidth]{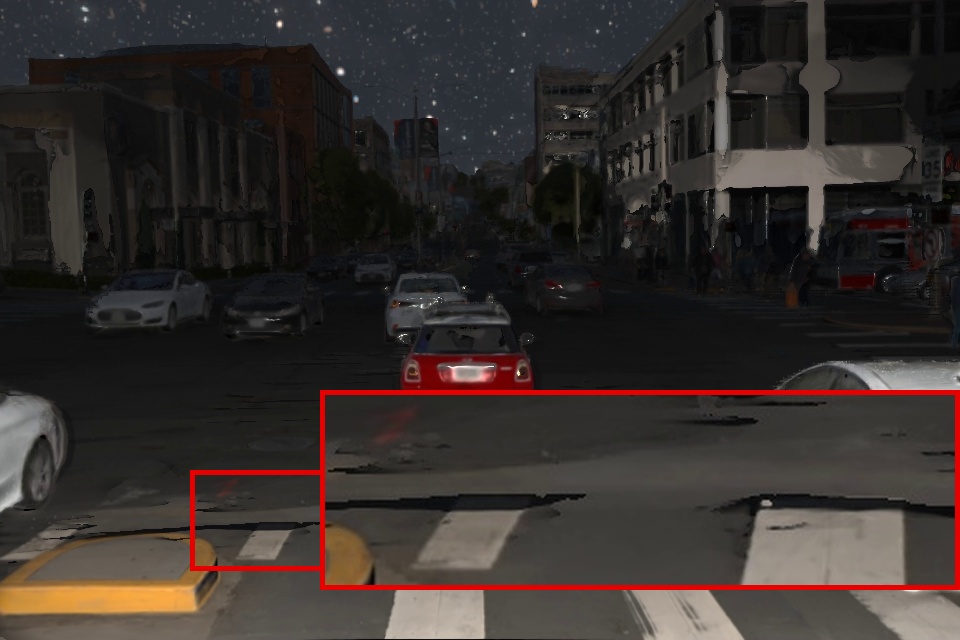} &
        \includegraphics[width=0.49\linewidth]{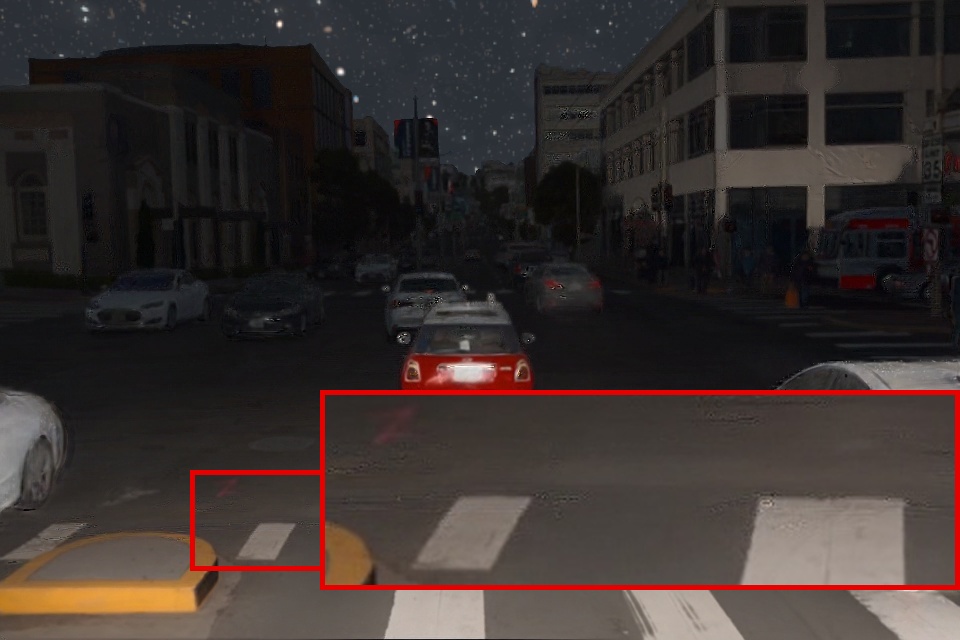} \\
    \end{tabular}
    \vspace{-0.5em}
    \caption{\textbf{Qualitative ablation of the generative rendering.} Left: PBR result. Right: after the generative rendering pass. The generative rendering pass removes reconstruction artifacts, such as crack-like shadows and holes on the road, while preserving the physically-based lighting.
    }
    \label{fig:qual_genrender}
\end{figure}

\paragraph{\textbf{Generative Rendering on Real-World Scenes.}}
We further examine the generative rendering pass on real-world scenes in Fig.~\ref{fig:qual_genrender}. Physically-based rendering alone suffers from reconstruction artifacts (e.g., on the road). The generative rendering pass removes these artifacts while preserving the physically-based lighting, substantially improving the perceptual quality of the final rendering.
 
\begin{figure}[t]
    \centering
    \setlength{\tabcolsep}{2pt} %
    \begin{tabular}{cccc}
        & \small Frame $t$ & \small Frame $t+7$ & \small Frame $t+14$ \\
        
        \smash{$\vcenter{\hbox{\rotatebox{90}{\small\begin{tabular}{@{}c@{}}(a) Init\\Prior\end{tabular}}}}$} &
        $\vcenter{\hbox{\frame{\includegraphics[width=0.280\linewidth]{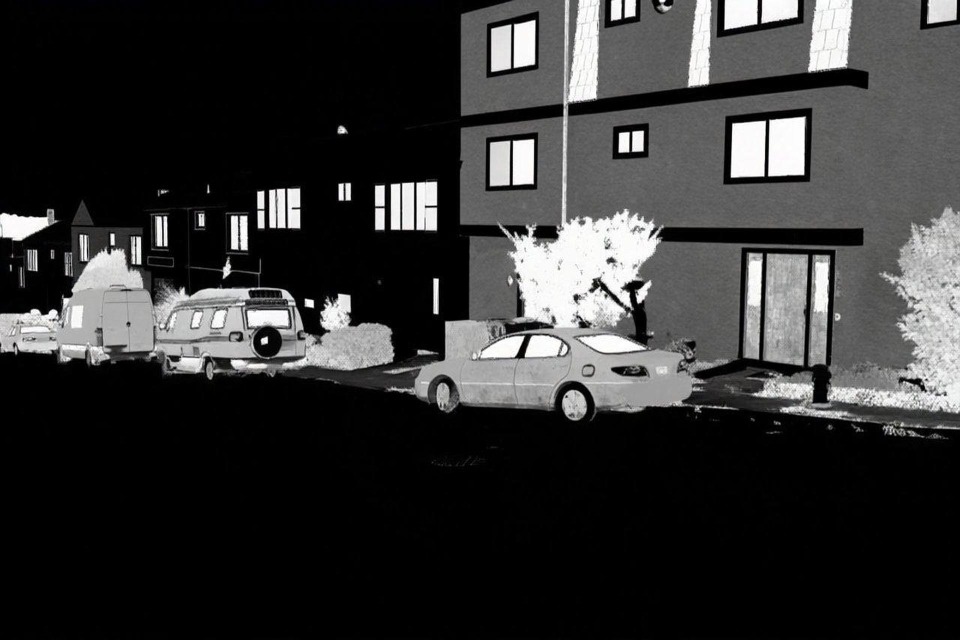}}}}$ &
        $\vcenter{\hbox{\frame{\includegraphics[width=0.280\linewidth]{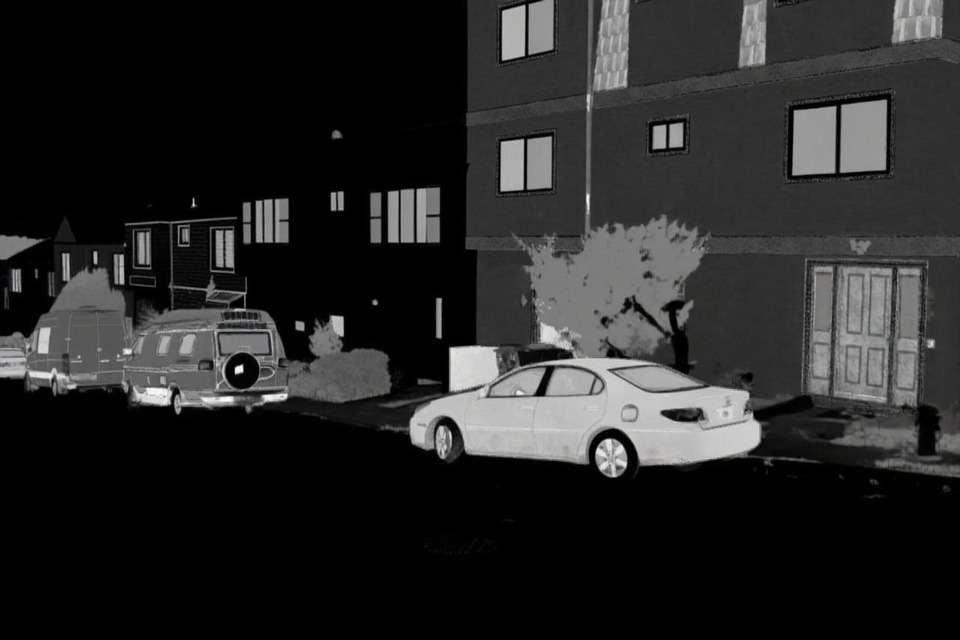}}}}$ &
        $\vcenter{\hbox{\frame{\includegraphics[width=0.280\linewidth]{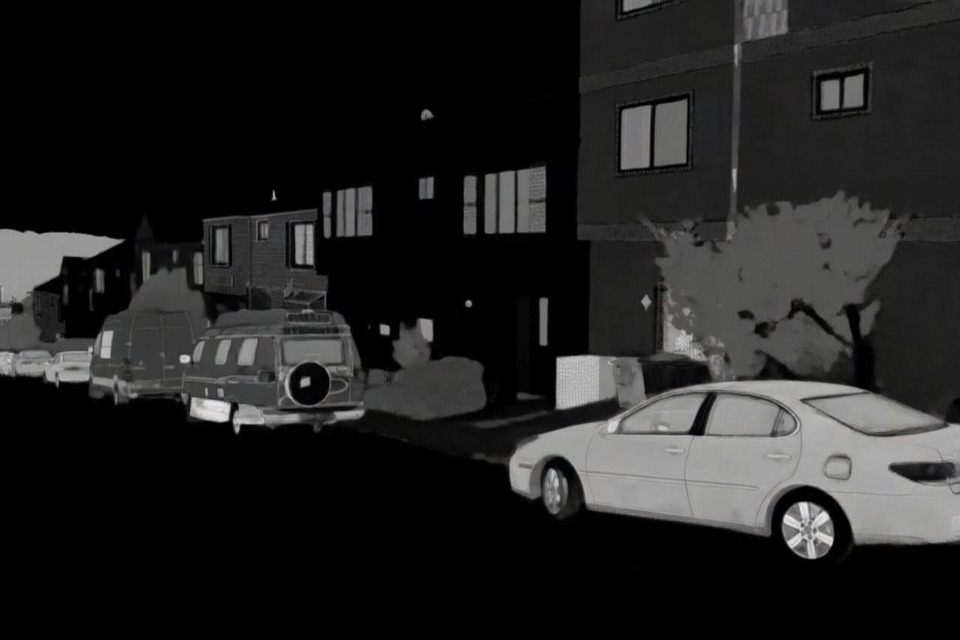}}}}$ \\
        \noalign{\vspace{4pt}} %
        
        \smash{$\vcenter{\hbox{\rotatebox{90}{\small\begin{tabular}{@{}c@{}}(b) Init\\Recon\end{tabular}}}}$} &
        $\vcenter{\hbox{\frame{\includegraphics[width=0.280\linewidth]{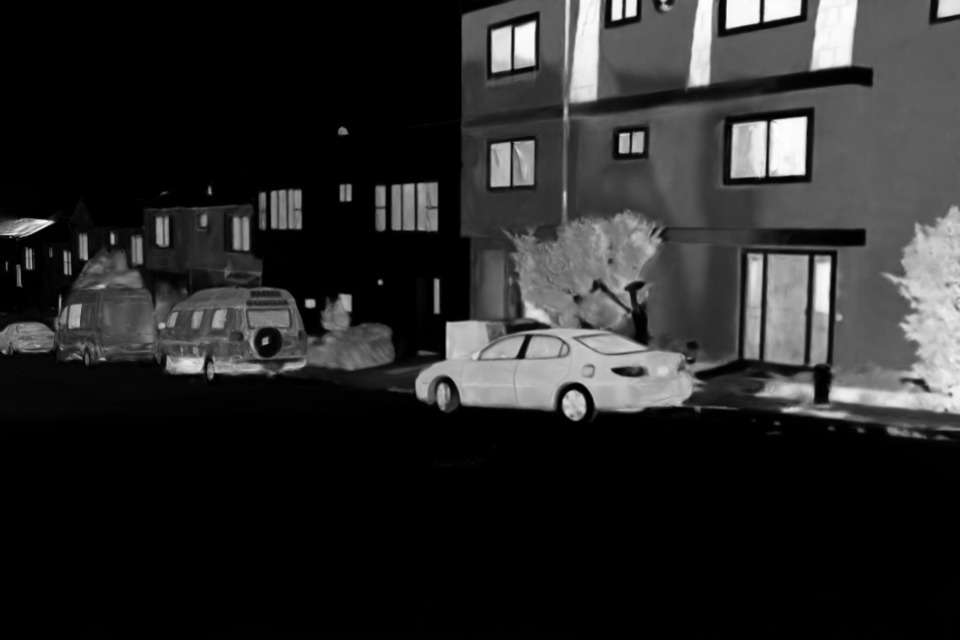}}}}$ &
        $\vcenter{\hbox{\frame{\includegraphics[width=0.280\linewidth]{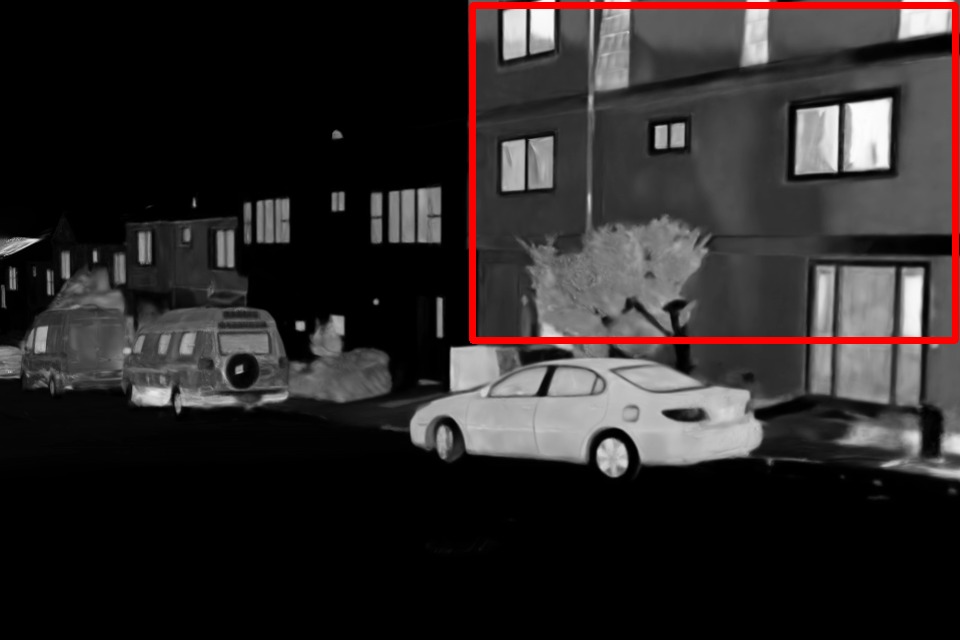}}}}$ &
        $\vcenter{\hbox{\frame{\includegraphics[width=0.280\linewidth]{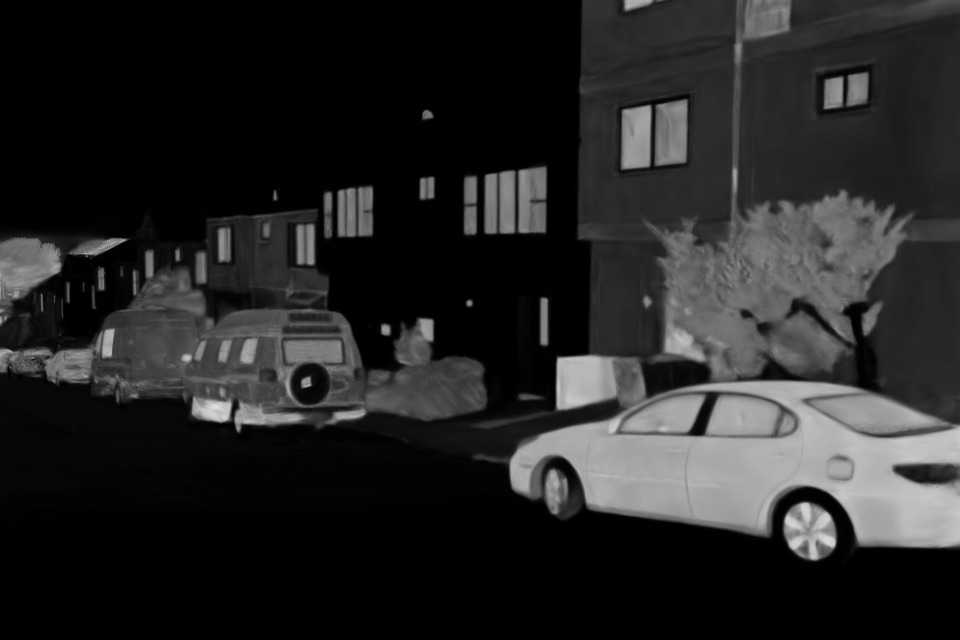}}}}$ \\
        
        & \multicolumn{3}{c}{\vspace{1mm} \Large $\downarrow$ \small \textbf{Generative Refinement} \Large $\downarrow$ \vspace{1mm}} \\
        
        \smash{$\vcenter{\hbox{\rotatebox{90}{\small\begin{tabular}{@{}c@{}}(c) Refined\\Prior\end{tabular}}}}$} &
        $\vcenter{\hbox{\frame{\includegraphics[width=0.280\linewidth]{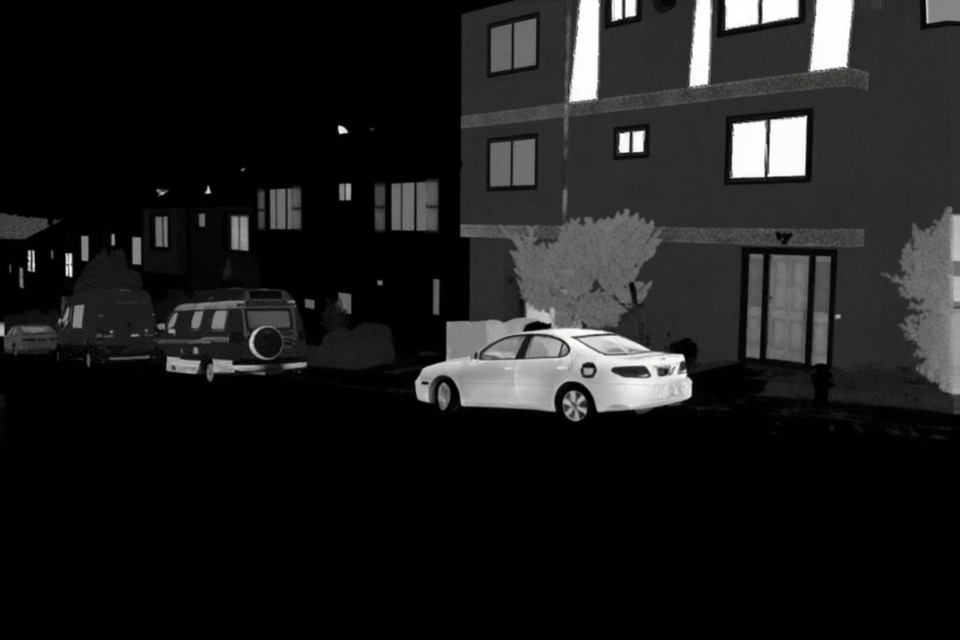}}}}$ &
        $\vcenter{\hbox{\frame{\includegraphics[width=0.280\linewidth]{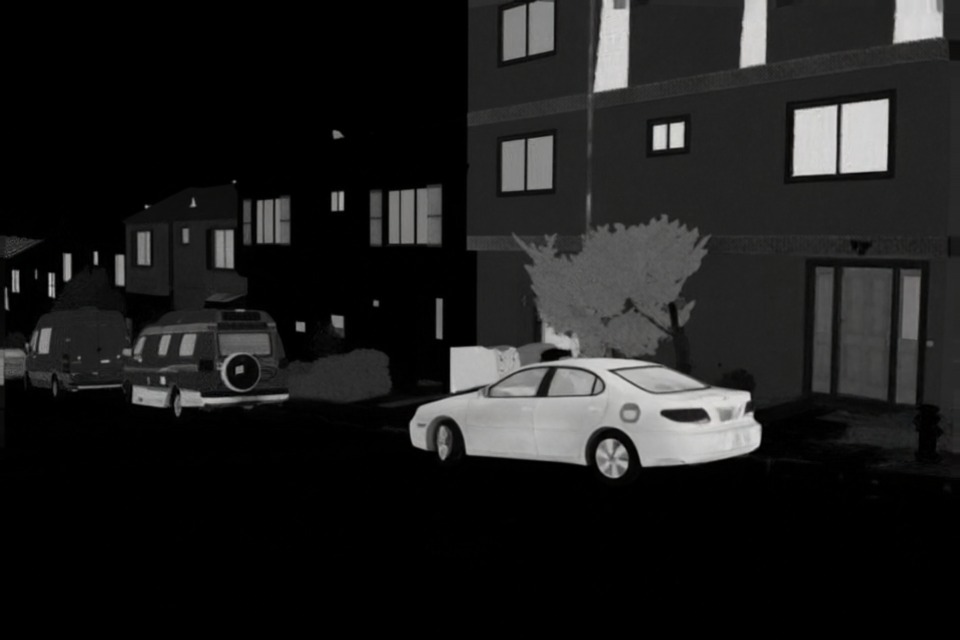}}}}$ &
        $\vcenter{\hbox{\frame{\includegraphics[width=0.280\linewidth]{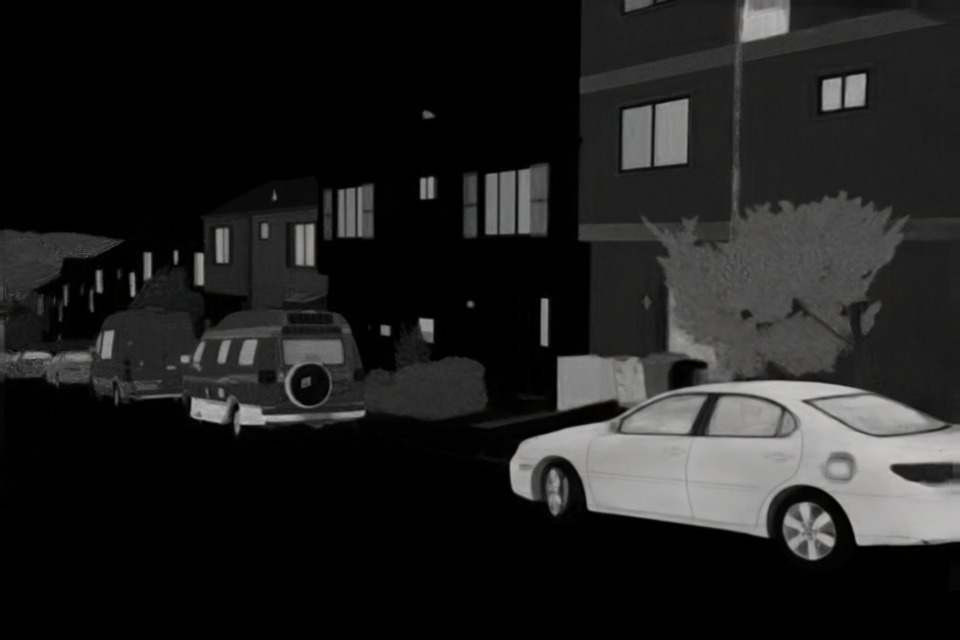}}}}$ \\
        \noalign{\vspace{4pt}} %
        
        \smash{$\vcenter{\hbox{\rotatebox{90}{\small\begin{tabular}{@{}c@{}}(d) Refined\\Recon\end{tabular}}}}$} &
        $\vcenter{\hbox{\frame{\includegraphics[width=0.280\linewidth]{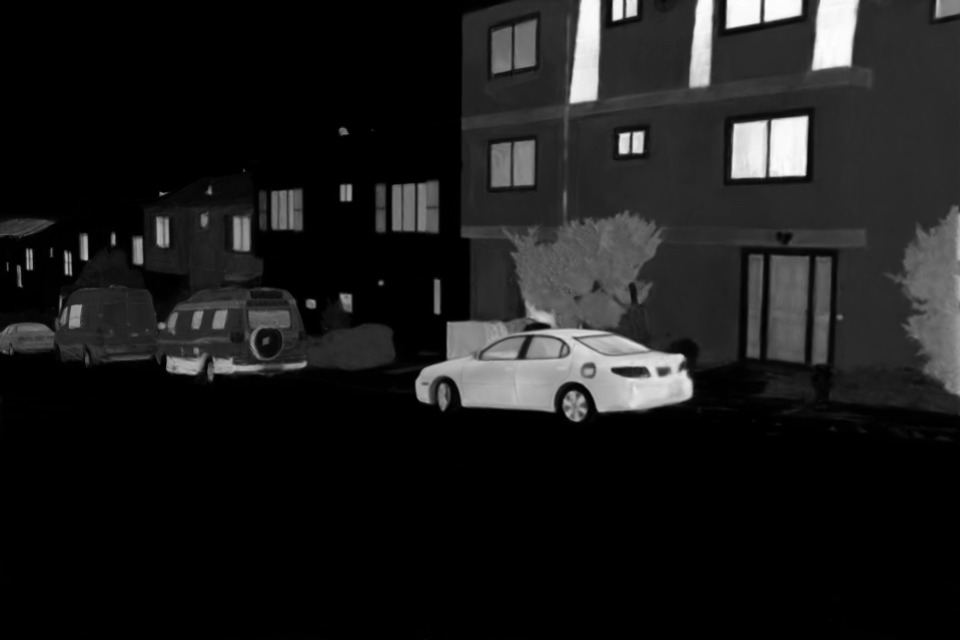}}}}$ &
        $\vcenter{\hbox{\frame{\includegraphics[width=0.280\linewidth]{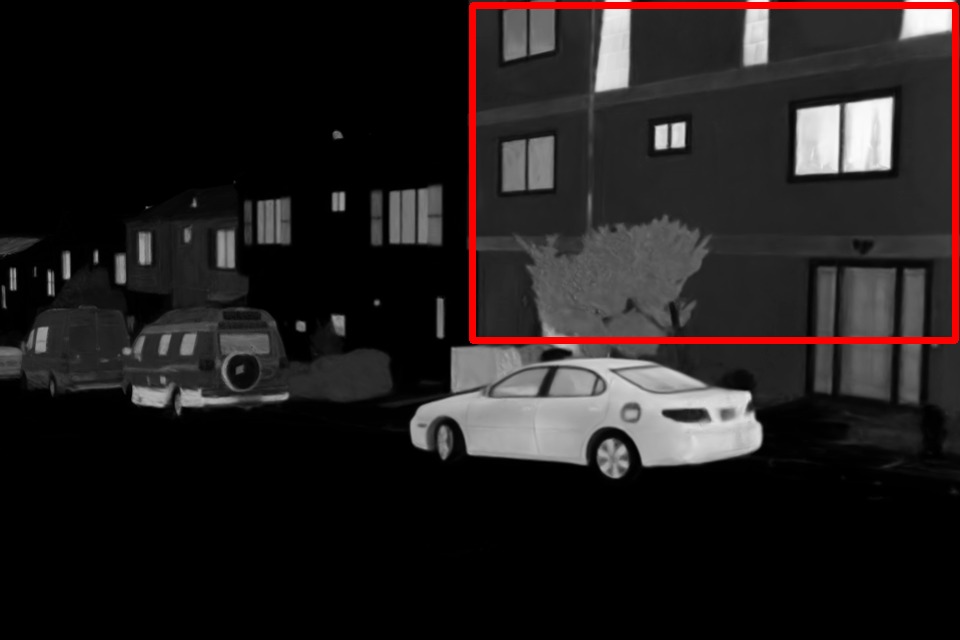}}}}$ &
        $\vcenter{\hbox{\frame{\includegraphics[width=0.280\linewidth]{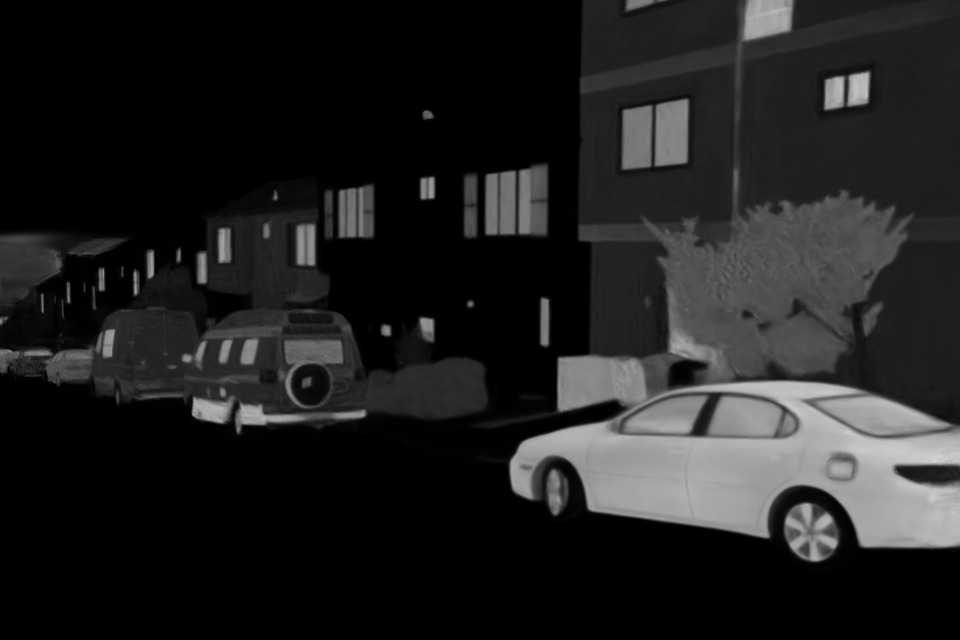}}}}$ \\
    \end{tabular}
    \caption{\textbf{Effectiveness of generative refinement.} We compare the metallic estimation across different stages in the optimization. \textbf{(a)} The initial prior predicted by DiffusionRenderer~\cite{diffusionrenderer} exhibits temporal inconsistencies, which lead to artifacts in the \textbf{(b)} initial 3DGS reconstruction (see red boxes). \textbf{(c)} Generative refinement produces a temporally consistent and clean prior, leading to fewer artifacts in the \textbf{(d)} subsequent reconstruction.}
    \label{fig:qual_ablate_genrefine}
\end{figure}

\paragraph{\textbf{Effect of Generative Refinement.}}
We additionally validate the importance of the generative refinement stage in the optimization pipeline, as shown in Fig.~\ref{fig:qual_ablate_genrefine}. The material prior generated by DR~\cite{diffusionrenderer} is temporally inconsistent (e.g., the predicted metallic value of the window changes across frames), which results in noticeable artifacts in the rendered material maps of the reconstructed scene. After applying generative refinement, the material prior becomes temporally consistent and has fewer artifacts, which in turn leads to a better reconstruction.
 We further quantify the temporal consistency of the material priors before and after the generative refinement process in Tab.~\ref{tab:quant_temporal}. We warp each prior map to the next frame using ground-truth depth, camera poses, and object motion, and compute the mean absolute difference over non-sky pixels. Generative refinement reduces the warping error for all three materials, showing that this process can effectively improve the temporal consistency of the material priors.
\begin{table}[t]
    \centering
    \small
    \caption{\textbf{Temporal consistency of material priors.} We report the temporal warping error, \ie, the mean absolute difference between each prior map and the next frame's map warped to it, with material values in $[0,1]$ (albedo in sRGB). Values are scaled by $10^{3}$ and averaged over the six synthetic scenes on consecutive training-frame pairs. Lower is better. We compare the warping error of the initial material prior from the DiffusionRenderer~\cite{diffusionrenderer} (\emph{Initial}) and the refined material prior after our generative refinement process (\emph{Refined}).}
    \label{tab:quant_temporal}
    \setlength{\tabcolsep}{5pt}
    \begin{tabular}{lccc}
    \toprule
    Prior ($\times 10^{3}$) & Albedo$\downarrow$ & Roughness$\downarrow$ & Metallic$\downarrow$ \\
    \midrule
    Initial & 10.66 & 11.37 & 8.96 \\
    Refined & \textbf{9.29} & \textbf{7.17} & \textbf{6.53} \\
    \bottomrule
    \end{tabular}
\end{table}
 
\begin{figure}[t]
    \centering
    \includegraphics[width=\linewidth]{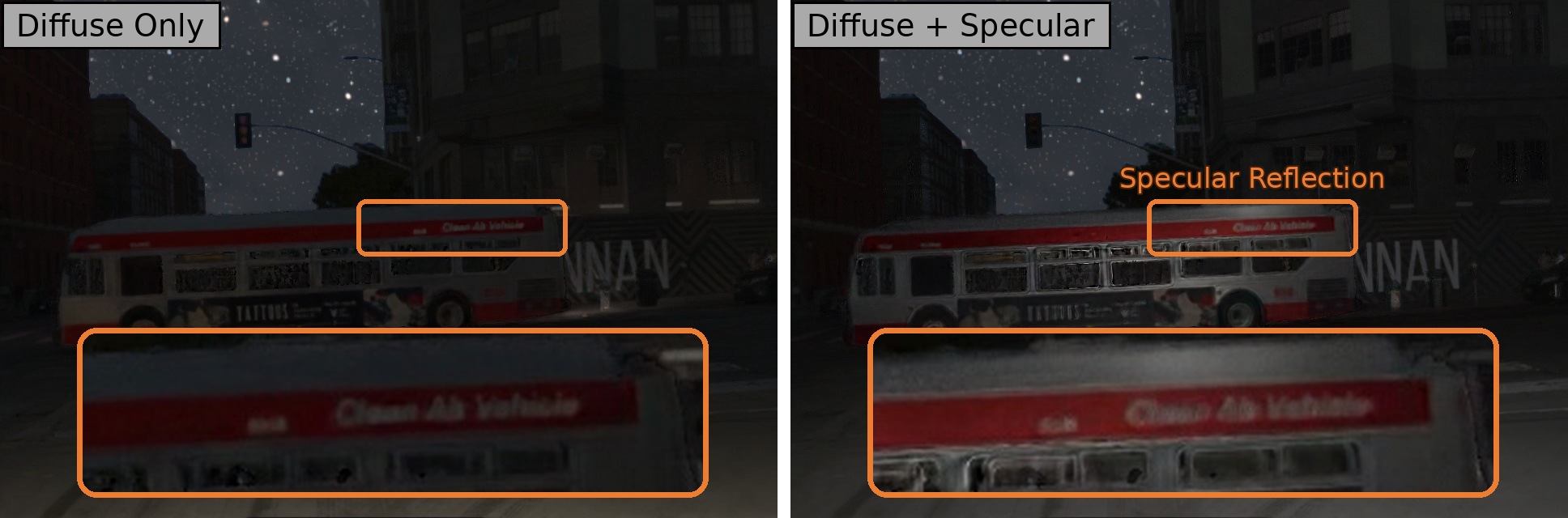}
    \vspace{-1.5em}
    \caption{\textbf{Ablation study on the necessity of the specular component of the Cook-Torrance BRDF.} Rendering with only the diffuse component of the Cook-Torrance BRDF (left) misses the specular highlights on the bus, whereas the full model with reconstructed roughness and metallic parameters (right) reproduces these reflections, which are critical for realistic relighting.}
    \label{fig:qual_roughmetal}
\end{figure}

\paragraph{\textbf{Specular Component of the BRDF.}}
Finally, we validate the necessity of using the Cook-Torrance BRDF~\cite{cooktorrancebrdf} rather than a diffuse-only Lambertian model in Fig.~\ref{fig:qual_roughmetal}. Rendering with the diffuse component alone fails to reproduce specular highlights on the bus, whereas incorporating the reconstructed roughness and metallic parameters recovers these highlights. Such specular effects are particularly important for night simulation under local and directional lighting, contributing to the realism of the relit results.
 
\section{\texorpdfstring{Stochastic Stability with PBR Anchoring}{Stochastic Stability with PBR Anchoring}}
\label{supp:stochastic}
Generative models such as DR~\cite{diffusionrenderer} and UR~\cite{unirelight}, and hybrid methods such as AW4D~\cite{autoweather4d}, are inherently stochastic, producing inconsistent relighting across random seeds. \method also uses diffusion models but anchors generation with PBR results to reduce this variation. 
We evaluate novel-view relighting on our synthetic dataset over five seeds and report the mean and standard deviation in Tab.~\ref{tab:quant_variance}. For \method, we fix the reconstruction and vary only the SDEdit seed of the generative pass. For the baselines, we vary the seeds of all their generative models. \method shows standard deviations lower than or comparable to those of these approaches across all metrics, demonstrating improved stability from PBR anchoring. 
\begin{table}[t]
    \centering
    \small
    \caption{\textbf{Stochastic stability.} Mean and standard deviation of novel-view relighting over five random seeds. Tab.~\ref{tab:quant_synthetic} reports one of these runs, so its numbers differ slightly from the means here.}
    \label{tab:quant_variance}
    \setlength{\tabcolsep}{5pt}
    \resizebox{\linewidth}{!}{%
    \begin{tabular}{lccc}
    \toprule
    Method & PSNR$\uparrow$ & SSIM$\uparrow$ & LPIPS$\downarrow$ \\
    \midrule
    Gen3C~\cite{Gen3C}+DR~\cite{diffusionrenderer}
        & $16.68 \pm 0.23$
        & $0.723 \pm 0.006$
        & $0.467 \pm 0.008$ \\
    Gen3C~\cite{Gen3C}+UR~\cite{unirelight}
        & $16.18 \pm 0.22$
        & $0.725 \pm 0.004$
        & $0.441 \pm 0.004$ \\
    AutoWeather4D~\cite{autoweather4d}
        & $17.73 \pm 0.59$
        & $0.731 \pm 0.005$
        & $0.408 \pm 0.006$ \\
    \textbf{Ours}
        & $\mathbf{20.74 \pm 0.08}$
        & $\mathbf{0.777 \pm 0.001}$
        & $\mathbf{0.319 \pm 0.004}$ \\
    \bottomrule
    \end{tabular}
    }
\end{table}

\section{\texorpdfstring{Algorithm Details}{Algorithm Details}}
\label{supp:algorithm}
We summarize the full optimization pipeline in Alg.~\ref{alg:full_optimization}, the generative refinement procedure in Alg.~\ref{alg:gen_refinement}, and the generative rendering procedure in Alg.~\ref{alg:gen_rendering}.
\algdef{SE}[PHASE]{Phase}{EndPhase}[1]{\textit{#1}}{}
\algtext*{EndPhase}
\algnewcommand\algorithmicinput{\textbf{Input:}}
\algnewcommand\Input{\item[\algorithmicinput]}
\algnewcommand\algorithmicgiven{\textbf{Given:}}
\algnewcommand\Given{\item[\algorithmicgiven]}
\algnewcommand\algorithmichyperparameter{\textbf{Hyperparameter:}}
\algnewcommand\Hyperparameter{\item[\algorithmichyperparameter]}
\algnewcommand\algorithmicoutput{\textbf{Output:}}
\algnewcommand\Output{\item[\algorithmicoutput]}
\begin{algorithm}[t]
    \caption{\textbf{Full optimization pipeline.}}
    \label{alg:full_optimization}
    \footnotesize
\begin{algorithmic}
        \Input Input video $\Cgt^{1:F}$, camera poses $\mathcal{P}$, LiDAR points $\mathcal{Q}$, and 3D bounding-box tracks $\mathcal{B}$
        \Given Inverse diffusion model $\diffusionModelI$
        \Output 3DGS scene $\mathcal{S}$ and HDR environment map $\bE$
        \Phase{Initialization}
            \State $\mathcal{X}_g \gets \Call{PredictGBufferPriors}{\Cgt^{1:F},\diffusionModelI}$
            \State $\mathcal{S} \gets \Call{Initialize3DGS}{\mathcal{Q},\mathcal{B}}$
            \State $\bE \gets \Call{InitializeEnvMap}{}$
        \EndPhase
        \Phase{Volume Rendering and Generative Refinement}
        \State $\mathcal{S} \gets \Call{VolRenderOptimize}{\mathcal{S},\Cgt^{1:F},\mathcal{X}_g, \mathcal{P}}$
        \State $\{\bA_g,\bR_g,\bM_g\} \gets \Call{GenerativeRefinement}{\mathcal{S},\Cgt^{1:F},\mathcal{P}, \diffusionModelI}$
        \State $\mathcal{S} \gets \Call{Initialize3DGS}{\mathcal{Q},\mathcal{B}}$
        \State $\mathcal{S} \gets \Call{VolRenderOptimize}{\mathcal{S},\Cgt^{1:F},\mathcal{X}_g, \mathcal{P}}$
        \EndPhase
        \Phase{Lighting and Joint Optimization}
            \State $\bE \gets \Call{PBROptimize}{\mathcal{S},\bE,\Cgt^{1:F},\mathcal{P}}$
            \State $(\mathcal{S},\bE) \gets \Call{JointRefine}{\mathcal{S},\bE,\Cgt^{1:F},\mathcal{X}_g,\mathcal{P}}$
        \EndPhase
        \State \Return $(\mathcal{S},\bE)$
    \end{algorithmic}
\end{algorithm}
\begin{algorithm}[t]
    \caption{\textbf{Generative refinement stage.}}
    \label{alg:gen_refinement}
    \footnotesize
\begin{algorithmic}
        \Input 3DGS scene $\mathcal{S}$, input video $\Cgt^{1:F}$, and camera poses $\mathcal{P}$
        \Given Inverse diffusion model $\diffusionModelI$, encoder $\enc$, and decoder $\dec$
        \Hyperparameter SDEdit timestep $\diffusionTime$
        \Output Refined material priors $\mathcal{X}'_g = \{\bA'_g,\bR'_g,\bM'_g\}$
    
        \State $\mathcal{X}'_g \gets \varnothing$
        \ForAll{$\bX \in \{\bA,\bR,\bM\}$}
            \State $\bX^{1:F} \gets \Call{RenderGBuffer}{\mathcal{S},\mathcal{P},\bX}$
            \State $\noise \sim \mathcal{N}(0,\mathbf{I})$
            \State $\latent_{\diffusionTime} \gets \beta_{\diffusionTime}\enc(\bX^{1:F})+\sigma_{\diffusionTime}\noise$
            \State $\latent_0 \gets \Call{Denoise}{\latent_{\diffusionTime},\diffusionModelI,\enc(\Cgt^{1:F}),\diffusionTime}$
            \State $\bX_{\text{gen}}^{1:F} \gets \dec(\latent_0)$
            \State $\bX'_g{}^{1:F} \gets \bX^{1:F} + \bX_{\text{gen}}^{1:F} - \dec(\enc(\bX^{1:F}))$
            \State $\mathcal{X}'_g \gets \mathcal{X}'_g \cup \{\bX'_g{}^{1:F}\}$
        \EndFor
        \State \Return $\mathcal{X}'_g$
    \end{algorithmic}
\end{algorithm}
\begin{algorithm}[t]
    \caption{\textbf{Generative rendering pass.}}
    \label{alg:gen_rendering}
    \footnotesize
\begin{algorithmic}
        \Input LDR PBR video $\Cldr^{1:F}$, rendered G-buffer videos $\mathcal{X}^{1:F}$ and target environment map $\bE$
        \Given Forward diffusion model $\diffusionModelF$, encoder $\enc$, and decoder $\dec$
        \Hyperparameter SDEdit timestep $\diffusionTime$
        \Output Refined video $\Cfin^{1:F}$
        \State $\cond \gets \Call{EncodeConditions}{\mathcal{X}^{1:F},\bE}$
        \State $\noise \sim \mathcal{N}(0,\mathbf{I})$
        \State $\latent_{\diffusionTime} \gets \beta_{\diffusionTime}\enc(\Cldr^{1:F})+\sigma_{\diffusionTime}\noise$
        \State $\latent_0 \gets \Call{Denoise}{\latent_{\diffusionTime},\diffusionModelF,\cond,\diffusionTime}$
        \State $\Cgen^{1:F} \gets \dec(\latent_0)$
        \State $\Cfin^{1:F} \gets \Cldr^{1:F} + \Cgen^{1:F} - \dec(\enc(\Cldr^{1:F}))$
        \State \Return $\Cfin^{1:F}$
    \end{algorithmic}
\end{algorithm}
 
\section{Implementation Details}
\label{supp:impl}

\paragraph{\textbf{General.}}
Our method is built upon OmniRe~\cite{chen2025omnire} and 3DGRT~\cite{3dgrt}.
Our pipeline executes sequentially: an initial volume rendering stage for 30,000 iterations, generative refinement, a second volume rendering stage for 30,000 iterations, a physically-based inverse rendering stage for 5,000 iterations, and a final joint refinement stage for 20,000 iterations. All training is conducted on a single RTX 4090 GPU. Generative rendering is run on a single RTX A6000 GPU or a single RTX PRO 6000 GPU. During training, we trace $N_r = 512$ rays and sample 1024 pixels per iteration~\cite{IRGS} to accelerate the process. For testing, we use $N_r = 1024$. Total training takes approximately 5 hours.

\paragraph{\textbf{Optimization.}}
For the volume rendering stage, the total loss is defined as:
\begin{equation*}
\begin{aligned}
    \loss_{\text{vol}} = {}& \loss_{\text{rgb}} + \weight_{O}\loss_{O} + \weight_{D}\loss_{D} + \weight_{N}\loss_{N} \\
    & + \weight_{A}\loss_{A} + \weight_{R}\loss_{R} + \weight_{M}\loss_{M}.
\end{aligned}
\end{equation*}
We empirically set the loss weights to $\lambda_O = 0.05$, $\lambda_D = 0.01$, $\lambda_N = 0.3$, $\lambda_A = 0.5$, and $\lambda_R = \lambda_M = 0.3$.
For the physically-based inverse rendering stage, the loss objective is:
\begin{equation*}
\begin{aligned}
    \loss_{\text{pbr}} &= \|\Cldr - \Cgt\|.
\end{aligned}
\end{equation*}
In the final joint refinement stage, both losses are combined:
\begin{equation*}
    \loss_{\text{all}} = \loss_{\text{vol}} + \lambda_{\text{pbr}}\loss_{\text{pbr}},
\end{equation*}
with $\lambda_{\text{pbr}} = 0.15$. To balance the joint optimization, the volume rendering weights are adjusted to $\lambda_N = 0.2$, $\lambda_A = 0.65$, $\lambda_R = 0.2$, and $\lambda_M = 0.15$.
These optimization hyperparameters are fixed for all scenes.

\paragraph{\textbf{SDEdit Process.}}
We use the released Cosmos-based checkpoints of DR, which process 57-frame clips, and sample with an EDM~\cite{EDM} sampler with $N{=}15$ steps ($\sigma_{\max}{=}80$, $\sigma_{\min}{=}0.02$, $\rho{=}7$). Steps are zero-indexed, so starting at step $\diffusionTime$ runs $N-\diffusionTime$ denoising steps. Generative refinement starts at $\diffusionTime=4$ ($\sigma_{\diffusionTime} \approx 17.0$). For generative rendering, we use the following noise levels:
\begin{itemize}
    \item Synthetic dataset: $\diffusionTime=3$ ($\sigma_{\diffusionTime} \approx 25.9$) for daytime relighting, $\diffusionTime=8$ ($\sigma_{\diffusionTime} \approx 2.33$) for nighttime relighting, and $\diffusionTime=11$ ($\sigma_{\diffusionTime} \approx 0.32$) for novel-view synthesis.
    \item WOD: for relighting, $\diffusionTime=6$ ($\sigma_{\diffusionTime} \approx 6.75$) in the baseline comparisons and $\diffusionTime \in [3, 8]$ in the other relighting figures; for novel-view synthesis, $\diffusionTime=8$ ($\sigma_{\diffusionTime} \approx 2.33$) in all figures and Tab.~\ref{tab:quant_wod_nvs}.
\end{itemize}
Sequences longer than 57 frames are processed with sliding windows of 57 frames and a stride of 7 frames, and overlapping frames are averaged in pixel space. The G-buffer priors, generative refinement, and generative rendering all use this windowing.

\paragraph{\textbf{Local Light Rendering.}}
To support localized illumination (e.g., car lights and streetlights), we extend our rendering to evaluate the direct contributions of point and spot lights. Specifically, for each surface point $\bx$, we compute its visibility to a local point light at position $\bp_l$ by casting a deterministic shadow ray, yielding visibility $V(\bx, \bomega_l)$, where $\bomega_l = \frac{\bp_l - \bx}{\|\bp_l - \bx\|}$ is the normalized direction to the light. The incident local radiance $\bE_l$ is the light's base intensity attenuated by a distance falloff $1/\|\bp_l - \bx\|^2$. For spot lights, $\bE_l$ is additionally modulated by a cone attenuation factor smoothly interpolating between predefined inner and outer cone angles. This local contribution is evaluated via the Cook-Torrance BRDF and added to the environment-mapped reflectance to compute the total outgoing radiance $\bL_\text{r}$:

\begin{equation*}
\begin{aligned}
\bL_\text{r} &= \sum_{k=1}^{N_r}\frac{\bE(\bomega_k)V(\bx, \bomega_k)f(\bx, \bomega_k, \bomega_\text{o})(\bomega_k \cdot \bN(\bx))}{N_r\,p_k} \\
&\quad + \sum_{l} \bE_l V(\bx, \bomega_l) f(\bx, \bomega_l, \bomega_\text{o}) (\bomega_l \cdot \bN(\bx))
\end{aligned}
\end{equation*}

\paragraph{\textbf{Object Insertion.}}
We insert vehicles from BlenderKit~\cite{blenderkit} and animated pedestrians from Microsoft Rocketbox~\cite{Rocketbox} into the reconstructed scenes. For vehicles, we convert the textured mesh directly into 3D Gaussians by sampling points on each triangle proportionally to its area and placing a flat Gaussian at each sample, whose shortest axis and normal follow the face normal. The albedo, roughness, and metallic values are read from the physically-based textures of the mesh, and the converted vehicle is added as a rigid node. For pedestrians, we fit SMPL~\cite{smpl} poses to the skeleton of the animated avatar in every frame, sample Gaussians on the posed mesh, and map them to the canonical SMPL space by inverting the fitted skinning, so that they are driven by the deformable human nodes of our scene graph~\cite{chen2025omnire}. The insertion position, orientation, and trajectory are chosen manually. Since the inserted Gaussians are added to the scene before ray tracing, they are rendered by the same physically-based rendering pass, casting and receiving shadows under both the environment map and local lights, followed by the generative rendering pass.

\section{Baseline Details}
\label{supp:baseline}
\paragraph{\textbf{Light Representation Conversion.}}
We specify all target lighting conditions as HDR environment maps. However, the baselines UrbanIR~\cite{urbanir} and InvRGB+L~\cite{invrgbl} do not take environment maps as input. Instead, they rely on parametric sun-sky models. To ensure a fair comparison, we convert each target environment map into their respective light representations.

UrbanIR~\cite{urbanir} represents lighting as a sun-sky model:
\begin{equation*}
    \mathbf{L} = \{(\mathbf{L}_{\text{sun}}, \boldsymbol{\omega}_{\text{sun}}),\; \mathbf{L}_{\text{amb}},\; \mathbf{C}_{\text{sky}}\},
\end{equation*}
where $\mathbf{L}_{\text{sun}}$ and $\boldsymbol{\omega}_{\text{sun}}$ denote the intensity and direction of a directional sun light, $\mathbf{L}_{\text{amb}}$ is a constant ambient term that approximates the illumination from the sky, and $\mathbf{C}_{\text{sky}}$ is the sky color, predicted by an MLP. InvRGB+L~\cite{invrgbl} similarly models the sun as a directional light with direction $\boldsymbol{\omega}_{\text{sun}}$ and intensity $I_{\text{sun}}$, and represents the sky illumination with third-order spherical harmonics $\mathbf{L}_{\text{sky}} \in \mathbb{R}^{16\times3}$. For both methods, we bypass the learned sky models when rendering sky pixels and directly query the target environment map instead.

We convert each target environment map $\bE$ into the sun-sky representations of both methods with the same three steps.
\textit{(1) Sun extraction.} Since both methods model the sun as a separate directional light, we first separate the sun from the sky. The sun region $\mathcal{S}$ consists of the pixels around the brightest pixel $p^{*}$ of the upper hemisphere (luminance above $2\%$ of the peak, within $3^{\circ}$). The sun direction is $\boldsymbol{\omega}(p^{*})$, and its irradiance is the total radiance of this region, $\mathbf{E}_{\text{sun}} = \sum_{p \in \mathcal{S}} \bE(p)\, \Delta\omega_p$, where $\Delta\omega_p$ is the solid angle of pixel $p$. The remaining pixels form the sky.
\textit{(2) Sky conversion.} For InvRGB+L, whose renderer can shade with an environment map, we replace the spherical harmonics with the sky map $\bE_{\text{sky}}$, \ie, the target map with the sun region removed. For UrbanIR, whose sky illumination is a single constant ambient term, we use the average irradiance of the sky over all surface orientations, $\mathbf{K}_{\text{sky}}/4$, where $\mathbf{K}_{\text{sky}} = \sum_{p \notin \mathcal{S}} \bE(p)\, \Delta\omega_p$ is the integrated sky radiance and the factor $1/4$ is the average of the clamped cosine over all orientations. 
\textit{(3) Brightness alignment.} Both methods are trained on LDR images, so their lighting has no absolute scale and is only meaningful relative to the lighting learned on the training sequence. We therefore rescale the converted sun and sky by a single factor $k$, which preserves their relative strength in the target map. We choose $k$ by auto-exposure, such that the relit test frames are, on average, as bright as the method's own renderings under the learned training lighting:
\begin{equation*}
    \frac{1}{|\Omega|}\sum_{(t,\mathbf{x})\in\Omega} \hat{I}_t(\mathbf{x}; k)
    = \frac{1}{|\Omega|}\sum_{(t,\mathbf{x})\in\Omega} I^{\text{tr}}_t(\mathbf{x}),
\end{equation*}
where $\Omega$ is the set of non-sky pixels of the test frames, $\hat{I}_t(\cdot\,; k)$ is the method's rendering of frame $t$ under the converted lighting scaled by $k$, and $I^{\text{tr}}_t$ is its rendering under the learned training lighting. We solve for $k$ by bisection, once per scene and target lighting. Without this scaling, the results would be overexposed or underexposed, unfairly penalizing these methods.
Finally, for UrbanIR we set $\mathbf{L}_{\text{sun}} = k\,\mathbf{E}_{\text{sun}}$ and $\mathbf{L}_{\text{amb}} = k\,\mathbf{K}_{\text{sky}}/4$. For InvRGB+L, we shade the environment term with $k\,\bE_{\text{sky}}$ and set $I_{\text{sun}}$ such that its sun term equals $k\,\mathbf{E}_{\text{sun}} \cos\theta$ for a surface at incident angle $\theta$.
This conversion assumes a dominant sun and is not applied to night scenes. For night scenes, we use the night environment map without rescaling for the ambient and sky terms, and add the same local lights as in our method.

\paragraph{\textbf{IRGS.}}
IRGS does not natively support local lights, so for night scenes we add the same local lights as for the other baselines at render time. Each light contributes $f\, I\, d^{-2} \cos\theta\, s(\theta)\, V$ to a surface point, where $f$ is the BRDF of IRGS, $I$ is the light intensity, $d$ is the distance to the light, $s(\theta)$ is a smooth cone falloff for spot lights, and $V$ is the visibility traced with the 2D Gaussian ray tracer of IRGS. The result is converted to sRGB in the same way as its environment-map relighting.

\paragraph{\textbf{AutoWeather4D.}}
We reimplement AW4D~\cite{autoweather4d} following its official codebase and paper wherever possible, using the same inputs as our method (the front-camera clip, ground-truth camera poses, and ground-truth instance tracks). Following AW4D, the $\pi^3$~\cite{pi3} depth is converted to metric scale with ground-truth depth or LiDAR.

\textit{Novel-view synthesis.} Since AW4D does not support novel views, we synthesize each test view $t$ by forward-warping the two neighboring input training frames $t-1$ and $t+1$, captured under the original lighting, using the $\pi^3$ depth, the camera poses, and the instance tracks. The warping is applied to the input frames before any relighting. Sky pixels are not warped, and the remaining holes (the sky and disoccluded regions) are filled with the pixels of frame $t-1$ at the same image locations, so the warped views contain no empty pixels. The warped test frames and the input training frames form the input video of AW4D, which is then refined by the official Wan~\cite{wan} refiner (Canny-edge-conditioned SDEdit) with its default setting. As a result, AW4D achieves a low LPIPS, since the test views are warped from neighboring input frames, but low PSNR and SSIM, since the Wan refinement hallucinates textures that are not aligned with the ground truth (Fig.~\ref{fig:qual_synthetic}).

\textit{Global relighting.} For novel-view relighting under a target HDR environment map (the same map used by our method), the input video above is passed through $\pi^3$, the DR inverse model, and the DR forward model, followed by the same Wan refinement. For relighting figures, we alpha-blend the sky with the target environment map, as for Gen3C+DR.

\textit{Night relighting with local lights.} AW4D renders local lights in 2D. For each frame, it shades the point and spot lights with a Cook-Torrance BRDF using the per-pixel positions from $\pi^3$ and the albedo, normal, roughness, and metallic maps predicted by the DR inverse model, without casting shadows, yielding the local-light image $\mathbf{C}_{\text{local}}$. In parallel, the DR forward model relights the video, and the relit frames are darkened by a night color lookup table (including the sky), yielding $\mathbf{C}_{\text{relit}}$. The two images are converted to linear RGB and blended at each pixel $\mathbf{p}$:
\begin{equation*}
\begin{aligned}
    \mathbf{C}(\mathbf{p}) &= \big(1-w(\mathbf{p})\big)\,\mathbf{C}_{\text{relit}}(\mathbf{p}) + w(\mathbf{p})\,\mathbf{C}_{\text{local}}(\mathbf{p}), \\
    w(\mathbf{p}) &= \lambda \cdot \mathrm{clamp}\big(\bar{C}_{\text{local}}(\mathbf{p}) - 0.05,\, 0,\, 1\big),
\end{aligned}
\end{equation*}
where $\bar{C}_{\text{local}}(\mathbf{p})$ is the mean of the three channels of $\mathbf{C}_{\text{local}}(\mathbf{p})$ and $\lambda$ is a blending strength. Thus, the local-light shading replaces the relit frame only where it is bright. The blended video is finally refined by the Wan refiner. Note that this is the process implemented in the official AW4D codebase, which we do not modify. We use the default values of the official configuration, including $\lambda = 3.5$, and apply the same light types, intensities, positions, and directions as for the other baselines, leaving all other settings unchanged.

\section{Cook-Torrance BRDF Details}
\label{supp:brdf}

Our rendering pipeline employs a standard Cook-Torrance BRDF~\cite{cooktorrancebrdf}. The surface material response $f(\bx, \bomega_\text{i}, \bomega_\text{o})$ is modeled as a linear combination of a Lambertian diffuse component and a Cook-Torrance microfacet specular component~\cite{cooktorrancebrdf}:
\begin{equation*}
f(\bx, \bomega_\text{i}, \bomega_\text{o}) = (1 - \bM) \frac{\bA}{\pi} + \frac{D(\bH) G(\bomega_\text{i}, \bomega_\text{o}) F(\bomega_\text{o}, \bH)}{4 (\bN \cdot \bomega_\text{i}) (\bN \cdot \bomega_\text{o})},
\end{equation*}
where $\bA$ is the albedo, $\bM \in [0, 1]$ is metallic, $\bN$ is the surface normal, and $\bH = \frac{\bomega_\text{i} + \bomega_\text{o}}{\|\bomega_\text{i} + \bomega_\text{o}\|}$ is the half-vector between the incident direction $\bomega_\text{i}$ and outgoing direction $\bomega_\text{o}$.
The specular component is evaluated using the following standard approximations~\cite{karis2013real}:

\textbf{Normal Distribution Function} ($D$): We use the GGX distribution~\cite{microfacet_refraction} parameterized by $\alpha = \bR^2$, where $\bR$ is surface roughness:
\begin{equation*}
D(\bH) = \frac{\alpha^2}{\pi((\bN \cdot \bH)^2 (\alpha^2 - 1) + 1)^2}
\end{equation*}

\textbf{Geometry Function} ($G$): We apply Schlick's approximation of the Smith geometry term, $G(\bomega_\text{i}, \bomega_\text{o}) = G_1(\bomega_\text{o}) G_1(\bomega_\text{i})$, where:
\begin{equation*}
G_1(\bomega) = \frac{\bN \cdot \bomega}{(\bN \cdot \bomega)(1 - k) + k}, \quad k = \frac{(\bR + 1)^2}{8}
\end{equation*}

\textbf{Fresnel Function} ($F$): We utilize Schlick's approximation~\cite{schlick_approx}:
\begin{equation*}
F(\bomega_\text{o}, \bH) = F_0 + (1 - F_0)(1 - \bomega_\text{o} \cdot \bH)^5
\end{equation*}
where the normal-incidence reflectance $F_0$ is linearly interpolated between a baseline dielectric value of $0.04$ and the surface albedo $\bA$:

\[F_0 = (1 - \bM) 0.04 + \bM \bA\]

\section{Computation Efficiency}
\label{supp:comp}
We report training time of a single 51-frame scene in Tab.~\ref{tab:training_time} and rendering speed in Tab.~\ref{tab:rendering_speed}, and compare with the physically-based baselines IRGS~\cite{IRGS}, UrbanIR~\cite{urbanir}, and InvRGB+L~\cite{invrgbl}. All measurements are conducted at a resolution of $960 \times 640$. For \method, IRGS, and UrbanIR, we measure the training time on a single RTX 4090 GPU and the rendering speed on a single RTX A6000 GPU. InvRGB+L requires about 75 GB of GPU memory, so we measure it on a single RTX PRO 6000 GPU. Generative refinement and generative rendering process videos in 57-frame chunks, so their time is amortized over each chunk. \method trains faster than UrbanIR and InvRGB+L but slower than IRGS, which does not use generative priors or model dynamic objects. Our rendering speed is dominated by Monte Carlo ray tracing in the PBR pass. Since the InvRGB+L paper reports efficiency on an NVIDIA A100 GPU, our measurements may differ from theirs.

\begin{table}[t]
    \centering
    \small
    \caption{\textbf{Training time.} Vol. Render, Gen. Ref., PBR, and Joint-refine denote volume rendering, generative refinement, physically-based rendering, and joint refinement, respectively.
    }
    \label{tab:training_time}
    \vspace{-3mm}
    \setlength{\tabcolsep}{3pt}
    \begin{tabular}{lc}
    \toprule
    Method & Training (hrs) $\downarrow$ \\
    \midrule
    IRGS~\cite{IRGS} & 1.08 \\
    UrbanIR~\cite{urbanir} & 15.80 \\
    InvRGB+L~\cite{invrgbl} & 9.00 \\
    \midrule
    Ours (Total) & 5.00 \\
    \quad \textit{- Vol. Render ($\times 2$)} & \textit{2.75} \\
    \quad \textit{- Gen. Ref.} & \textit{0.25} \\
    \quad \textit{- PBR} & \textit{0.35} \\
    \quad \textit{- Joint-refine} & \textit{1.65} \\
    \bottomrule
    \end{tabular}
\end{table}

\begin{table}[t]
    \centering
    \small
    \caption{\textbf{Rendering speed.} We report frames per second (FPS) for physically-based rendering (PBR), generative rendering (Gen. Render), and the full rendering pipeline (Full).}
    \label{tab:rendering_speed}
    \vspace{-3mm}
    \setlength{\tabcolsep}{3pt}
    \begin{tabular}{lc}
    \toprule
    Method & Rendering (FPS) $\uparrow$ \\
    \midrule
    IRGS~\cite{IRGS} & 0.060 \\
    UrbanIR~\cite{urbanir} & 0.080 \\
    InvRGB+L~\cite{invrgbl} & 0.190 \\
    \midrule
    Ours (PBR) & 0.015 \\
    Ours (Gen. Render) & 0.200 \\
    Ours (Full) & 0.014 \\
    \bottomrule
    \end{tabular}
\end{table}
 
\begin{figure}[t]
    \centering
    \setlength{\tabcolsep}{3pt}
    \begin{tabular}{cc}
        \begin{tabular}{c}
            \includegraphics[width=0.3\linewidth]{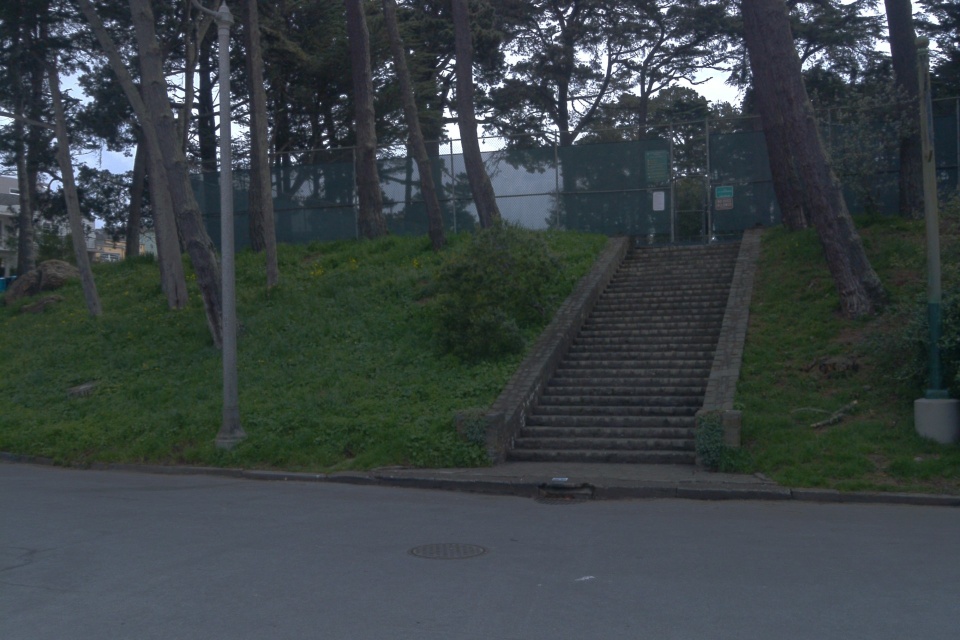} \\
            \small Input
        \end{tabular} &
        \begin{tabular}{c}
            \includegraphics[width=0.3\linewidth]{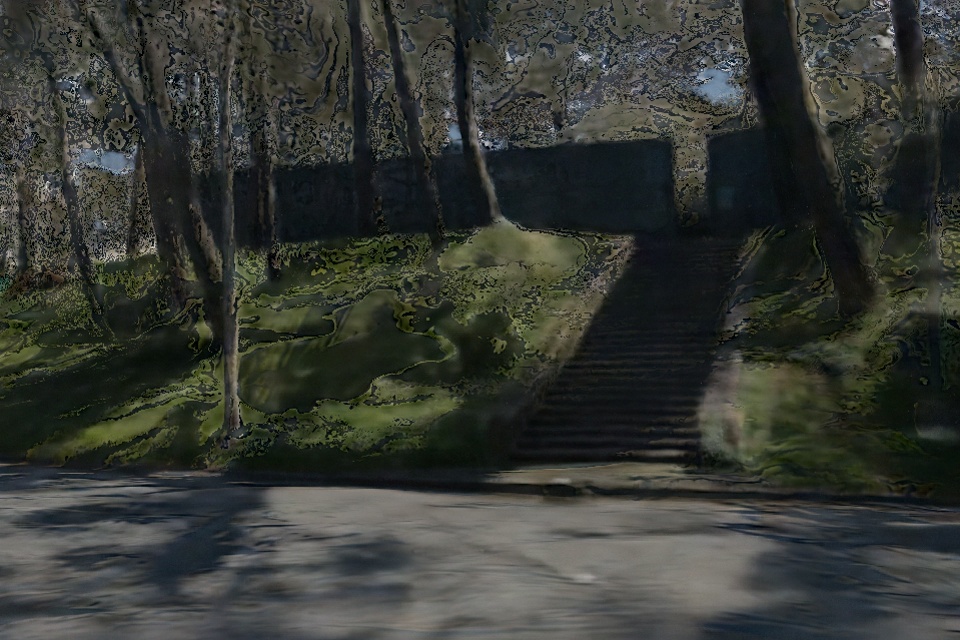} \\
            \small Relighting
        \end{tabular}
    \end{tabular}
    \caption{\textbf{Failure case.} Floaters in the unobserved regions cast unexpected shadows on the ground when relighting.} 
    \label{fig:limitation}
\end{figure} 
\section{Limitations and Failure Cases}
\label{supp:fail}
\method has five main limitations.
First, the inverse rendering pipeline does not model emissive materials. Local lights can be inserted at render time, but nighttime sequences with active headlights or streetlights cannot yet be decomposed.
Second, the PBR pass relies on Monte Carlo ray tracing with a limited number of rays per pixel. The residual noise lowers the perceptual quality (SSIM and LPIPS) of novel views, and ray tracing dominates the rendering time (0.015 FPS for the PBR pass vs.\ 0.200 FPS for generative rendering, Tab.~\ref{tab:rendering_speed}).
Third, the PBR pass models only direct illumination. Interreflections are ignored and must be absorbed by the environment map and the albedo during inverse rendering, or synthesized by the generative pass during forward rendering.
Fourth, as in other 3D reconstruction methods, floaters may appear in unobserved regions due to limited training views. They produce artifacts and cast spurious shadows during relighting (Fig.~\ref{fig:limitation}). 
Finally, while generative refinement mitigates minor inconsistencies of the priors, catastrophic failures of the generative models can still corrupt the decomposition.

\end{document}